\documentclass{osa-article}

\journal{osajournal}

\articletype{Research Article}

\usepackage{lineno}
\usepackage{enumitem}
\usepackage{longtable,booktabs}
\usepackage{subcaption}
\usepackage{tikz,pgfplots}
\usetikzlibrary{patterns}
\newcommand{\comment}[1]{}

\newcommand*\egor[1]{}
\usepackage{titlesec}

\def\lf{\left\lfloor}   
\def\rf{\right\rfloor}

\titleformat{\paragraph}
{\normalfont\normalsize\bfseries}{\theparagraph}{1em}{}
\titlespacing*{\paragraph}
{0pt}{3.25ex plus 1ex minus .2ex}{1.5ex plus .2ex}
\pgfplotsset{compat=1.18}

\begin{document}

\title{Artificial Neural Networks for Photonic Applications: From Algorithms to Implementation }

\author{Pedro Freire,\authormark{*} Egor Manuylovich, Jaroslaw E. Prilepsky and Sergei K. Turitsyn}
\address{Aston Institute of Photonic Technologies, Aston University, Birmingham B4 7ET, UK}

\email{\authormark{*} p.freiredecarvalhosourza@aston.ac.uk} 



\begin{abstract}
This tutorial-review on applications of artificial neural networks in photonics targets a broad audience, ranging from optical research and engineering communities to computer science and applied mathematics. We focus here on the research areas at the interface between these disciplines, attempting to find the right balance between technical details specific to each domain and overall clarity. First, we briefly recall key properties and peculiarities of some core neural network types, which we believe are the most relevant to photonics, also linking the layer's theoretical design to some photonics hardware realizations. After that, we elucidate the question of how to fine-tune the selected model's design to perform the required task with optimized accuracy.  Then, in the review part, we discuss recent developments and progress for several selected applications of neural networks in photonics, including multiple aspects relevant to optical communications, imaging, sensing, and the design of new materials and lasers. In the following section, we put a special emphasis on how to accurately evaluate the complexity of neural networks in the context of the transition from algorithms to hardware implementation. The introduced complexity characteristics are used to analyze the applications of neural networks in optical communications, as a specific, albeit highly important example, comparing those with some benchmark signal processing methods. We combine the description of the well-known model compression strategies used in machine learning, with some novel techniques introduced recently in optical applications of neural networks. It is important to stress that although our focus in this tutorial-review is on photonics, we believe that the methods and techniques presented here can be handy in a much wider range of scientific and engineering applications.
\end{abstract}


\section{Introduction}
Machine learning has a tremendous number of definitions, which often reflect the specific interests of the researchers who formulate them. Here, we use the definition of machine learning as a bevy of algorithms that ``... allows computer programs to automatically improve through experience and that automatically infer some general laws from specific data'', taken from the classical Tom Mitchell’s monograph \cite{TM1997}.  In this tutorial-review, we will discuss blending machine learning with various photonics technologies and applications. The mixture of these two complementary disciplines enables the development of new scientific and engineering techniques that benefit both from the speed and parallelism inherent to optical systems and the ability of machine learning to infer from data and automatically improve system performance. Nonlinear photonics often features complex light dynamics and deals with systems that cannot be easily comprehended or controlled. Therefore, another attractive feature of machine learning in photonics applications is its capability to deal with complex nonlinear systems, whilst staying flexible and re-adaptable. Additionally, photonic devices and systems operating at high speed can quickly generate a vast amount of data. This makes them well-suited to the application of various data-based machine learning algorithms that improve performance with increasing available data sets. Therefore, photonics and machine learning look like a perfect fit for each other, and their combination can naturally bring forth new ideas, theories, and devices, as well as novel concepts for understanding the description of light-related phenomena.

Artificial neural networks, which we will henceforth call simple neural network (NNs), are computational machine learning frameworks that attempt to mimic some brain operations. The attractive features of biological NNs, which we would like to keep when using their artificial analogs, are robustness and fault tolerance; flexibility and easiness of re-adaptation to the changing conditions; ability to deal with a variety of data, meaning that the network can make do with information that is fuzzy, probabilistic, noisy, and even
inconsistent; collective computation, i.e., the network can process the data in parallel and in a distributed manner \cite{yegnanarayana2009artificial}. Whilst the NNs are frequently attributed to supervised learning thanks to numerous widely-known successful examples in, e.g., image recognition  \cite{pakkim,gu2018recent}, they are also applicable to unsupervised learning \cite{karhunen2015unsupervised}, semi-supervised learning \cite{van2020survey,NEURIPS2018_c1fea270}, and reinforcement learning \cite{mnih2015humanlevel,Silver_2016,li2017deep,henderson2018deep}, to mention the most noticeable directions. Of course, in this tutorial-review, we cannot address each specific item from the list above. Instead, we will focus on some particular examples of using the NNs in photonics, trying to explain why the particular combination of a machine learning method with a photonics application has turned out to be successful.

Here, it is pertinent to note that ultra-fast photonic applications can bring about conditions and requirements (in terms of accuracy, speed, and complexity), which differ from those in more ``traditional'' use cases of NNs. For example, in optical communications, the typical bit-error-rate (the probability of error occurrence in the dataset, speaking in ``machine learning'' language) before forward error correction, is of the order $10^{-2}$, which is, for instance, much lower than we have in typical image recognition tasks\cite{freire2022pitfalls}. Therefore, the solutions developed in deep learning applications for image recognition and language processing often require adaptation and/or substantial modifications when we deal with, e.g., an equalization task in optical communications. We specifically notice that the real-time operation of NNs in ultra-fast photonics inevitably sets a limit on the acceptable level of NN's complexity and processing latency (inference). Thus, in this review, we pay special attention to the NNs with reduced complexity, and this, in turn, emanates into the reduction of the energy consumption used for signal processing, the sought-for feature in almost every application nowadays.

There are numerous recently-emerged and still developing areas at the interface of machine learning and photonics: general neuromorphic photonics, unconventional optical computing, photonic neural networks,  optical communications, imaging, and sensing, to mention a few important examples where the cross-fertilization of the fields has already proven to be fruitful. Typically, the NNs' application in photonics is related to the processing of large data sets, which is the case in optical communications, ultra-fast photonics, optical imaging and sensing, lasers, optical metrology, design of new photonic materials, and so on. However, we would like to stress that this tutorial-review is not aimed to be a comprehensive overview of all applications of NNs (or, in more general terms, of the machine learning methods) in photonics, as this goal would be too large and general to fit into any review paper or even in a monograph. More information, details, methods, and examples of merging the photonics and artificial intelligence solutions can be found in other recent review papers covering different aspects of the subject and presenting various view-points\cite{R01,R02,R03Darko,R04,R05,R06,R07,R08,R09,R10,R11,NNMaterial01,R13,R14,Saad02,wettlin2020complexity, wei2020special}. How is then this tutorial-review different from numerous other review papers in the field? In this paper, we aim to improve some photonic techniques and technologies by using NNs for signal or data processing, providing analysis of the complexity and hardware implementation. We do not provide a comprehensive survey of optical reservoir computing or photonic NNs, which form a huge, rapidly expanding, and utterly fascinating area; we refer the reader to recent works and reviews on the subject \cite{shen2017deep,sunada2021photonic,huang2021silicon,shastri2022silicon,huang2022prospects,Prucnal02,Prucnal03,Prucnal05,Fan01,Fan05,pai2023experimentally,Volker01,Volker02,Volker03,VOlker04,christensen20222022,wei2020special}, including critical opinions \cite{peserico2022emerging}. In particular, a good exposition of the known and potential benefits of using neuromorphic devices in place of their ``von Neumann'' counterparts, including estimates of energy consumption, is given in Ref.~\cite{mehonic2022brain}.

Now, we emphasize that signal processing (inference) speed and energy efficiency are the two factors that quite often (virtually always) emerge when we talk about the practical implementation of a particular model or method. Both of these factors relate to the complexity of the NNs. Therefore, the tutorial part of our work is focused on a rather specific challenge: how to pick a correct NN structure fitting the task in hand and how to manipulate (typically reduce) the complexity of the NN to make them practically implementable and power-/cost-efficient, while not losing much in the efficiency/functionality of the initially-developed unrestricted (typically complex) NN solutions. Below, we will try to follow the whole path, from the NN algorithms explanations and development stage down to notes on the existing approaches for the hardware implementation of NNs. Thus, the tutorial part systematically describes the tools that can be used when we already have some NN model performing the desired task ``well enough'', and when the next step refers to how to match the model with the constraints (imposed by, say, the limited available resources) for the practical implementation. Evidently, there is some trade-off between performance and complexity. The performance of the compressed and/or quantized models, in general, degrades, and the important goal of complexity optimization is to identify the acceptable balance between complexity reduction and performance degradation. We hope that this tutorial-review can provide the necessary assistance along the ``thorny way'' of modifying your model toward a much simpler, but still efficient structure that would not flabbergast hardware designers.

The plan of our tutorial-review is as follows. Firstly, in Sec.~\ref{sec:NN-basic}, we briefly overview/remind the optical community of the basics of NNs and discuss their key photonic applications, trying to stay as close to the layman level but providing a viewpoint from a traditional digital signal processing perspective. In Sec.~\ref{sec:choose}, we describe how to select the NN architecture appropriate for the task at hand. In the review part, Sec.~\ref{sec:optcom}, we overview different applications of NN in various branches of photonics, discuss the open problems and challenges, and outline future research directions in the field. In the tutorial part, Sec. \ref{sec:reduce}, we describe versatile directions and methods that can be used to reduce the complexity of NNs (i.e., the model compression techniques) in photonics applications, also paying attention to the different metrics that we can employ to quantify our complexity.  We also append our work with a considerable number of references that can add more particular details to the questions considered.

\section{Basics of artificial neural networks for photonics community}\label{sec:NN-basic}

In an artificial NN, several neurons are connected together in a nonlinear manner. The network learns by adjusting the weights and biases, according to the feedback (typically provided by the so-called back-propagation technique) based on the evaluation of the accuracy of the NN's prediction, which is quantified by a cost (loss) function.
The number of neurons in the input layer corresponds to the input characteristics, whereas the number of output neurons is linked to the batch of classes of interest for classification; or it can be just a single neuron when we do with a single-class regression. In the deep NN structures, the layers between the input and output layer are referred to as hidden layers; the number of neurons per layer is arbitrary, and the choice of NN's hyperparameters (the number of neurons in the hidden layers and the number of hidden layers) requires designer's expertise in adjusting the NN structure to the task in hand; the choice of hyperparameters also depends on the complexity of the system to be modeled, as these parameters ultimately define the representation capability of an NN. For convenience of presentation, in this section, we briefly revisit some basic types and features of artificial NNs that will be discussed throughout the paper. 

However, we note that in spite of the (deceptive) simplicity of the short description of NNs given above, there are a plethora of unresolved puzzles and problems in the theory of deep NN, which typically relate to the three fundamental deep NN challenges: expressibility, optimizability, and generalizability. At the moment, we do not seem to have a good universal theory that would give us persuasive answers to all the problems itemized above, while the works shedding light on some of the NNs' properties, features, and peculiarities emerge continuously.
 
\subsection{Dense Layer}
We start from the basic feedforward NN, the so-called multi-layer perceptron (MLP). The simplest variant of the perceptron idea was first developed in 1943 by McCulloch and Pitts \cite{mcculloch1943logical}, but this concept drew the essential attention of scientific society only after Frank Rosenblatt's implementing it as a machine built in 1958 \cite{rosenblatt1958perceptron}. While Rosenblatt used just a single layer of neurons for binary prediction, nowadays, the perceptron's original idea has been largely generalized, such that it evolved into a (deep) feed-forward densely-connected multilayer structure that we call the MLP. 
 
\begin{figure}[ht!]
\centering
\includegraphics[width=.6\linewidth]{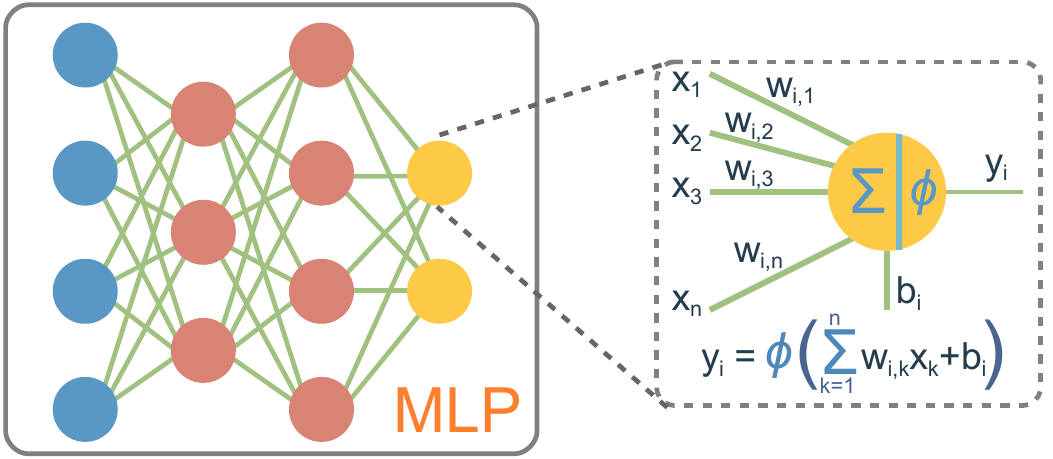}
\caption{Schematics of a McCulloch-Pitts Neuron.}
\label{fig: single MLP neuron schematics}
\end{figure}

 
A dense layer, also known as a fully connected layer, is a layer in which each neuron (labeled as $i$) is connected with all the neurons (labeled as $j$) from the previous layer with a specific weight $w_{{i}{j}}$. The input vector is mapped to the output vector in a nonlinear manner by the dense layer, due to the presence of a non-linear activation function. Dense layers can be combined to form an MLP, which is a class of a feed-forward deep NN. Fig. \ref{fig: single MLP neuron schematics} illustrates the working operation of a single neuron in such a dense layer.

The output vector $y$ of a dense layer, given $x$ as an input vector, is written as:
\begin{equation}\label{eq.dense}
 y =\phi (Wx+ b ),
\end{equation}
where $y$ is the output vector, $\phi$ is a nonlinear activation function, $W$ is the weight matrix, and $b$ is the bias vector.

Now, let us turn to the hardware implementation aspect of this most prolific NN structure, where we first mention the electronic implementation. The traditional matrix multiplier-and-accumulator (MAC) is used for the implementation of such layers in the digital domain \cite{hong2020low}. More recently, the electrical analog implementation of a dense layer was demonstrated using a CMOS with transistors and resistors \cite{heidari2019analog,geng2020analog}, or using an operational transconductance amplifier \cite{abden2020multilayer}. As a drawback, the analog NNs' implementation typically renders a lower accuracy and is more sensitive to noise compared to their digital counterparts\cite{sarpeshkar1998analog}. 

 \begin{figure*}[ht!] 
\begin{subfigure}{\textwidth}
    \centering
 \includegraphics[width=0.7\linewidth]{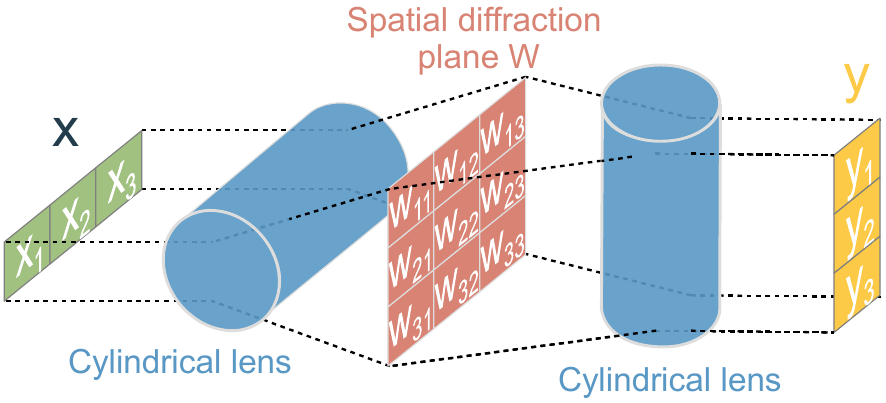}
    \caption{Optical implementation of PLC-MVM  using an array of light sources, cylindrical lenses for wavefront deformation, transmissive spatial light modulator $W$, and an array of photodiodes.}
    \label{fig:MVM} 
\end{subfigure}\hfill
\begin{subfigure}{.49\textwidth}
    \centering
 \includegraphics[width=\linewidth]{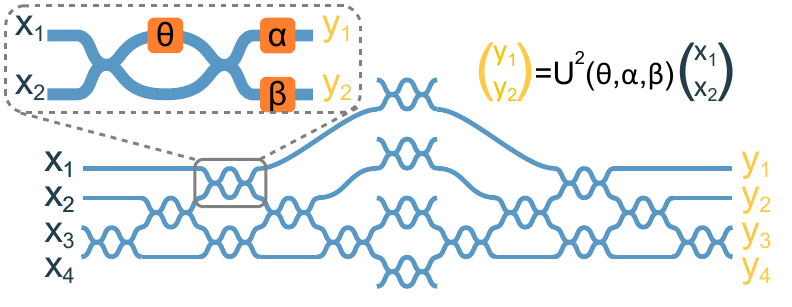}
    \caption{Optical MVM realized using Mach-Zehnder interferometers and tunable phase shifts as a weight matrix. Redrawn from Ref.~\cite{zhou2022photonic}.}    
    \label{fig:MZI} 
\end{subfigure}\hfill
\begin{subfigure}{.49\textwidth}
    \centering
 \includegraphics[width=\linewidth]{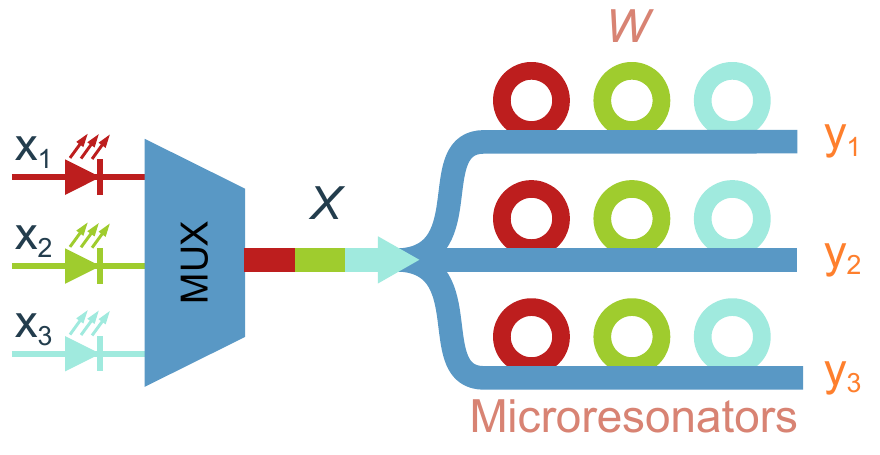}
    \caption{Wave division multiplexer (WDM) based MVM with microresonators and tunable filters as a weight matrix. Redrawn from Ref.~\cite{zhou2022photonic}.}
    \label{fig:WDM} 
\end{subfigure}
  \caption{Optical implementations of vector-matrix multipliers. }
  \label{fig:multiplications_optics} 
\end{figure*}

Now, we mention that there are two different elements of the NN processing that are addressed in the photonic feed-forward NN implementation: the matrix-vector multiplications, and the activation function. First, we address the differences in the activation function. The first widely adopted approach for the activation of photonic NNs, which can be called a ``fully-analog'' implementation,  entails utilizing silicon photonic meshes comprising the networks of Mach-Zehnder interferometers and programmable phase shifters (electro-optic activations). However, lately, a novel approach for the activations coined ``hybrid'' photonic programmable NNs has emerged, demonstrating remarkable features in terms of low latency and energy efficiency for inference. These hybrid photonic NNs combine programmable photonic linear optical elements, such as meshes, with digital nonlinear activation functions\cite{harris2018linear,bogaerts2020programmable,pai2023experimentally}. In comparison to existing fully analog photonic NNs that employ electro-optic nonlinear activation functions, hybrid designs can overcome the significant challenge of photonic loss and provide improved flexibility in performing logical operations between layers as compared to the fully-analog counterparts. Quite importantly, the hybrid design has been shown to be able to learn online \cite{Fan02,pai2023experimentally}, which gives immense opportunities for the prompt reconfiguration of photonic NNs and, so, for their usage for real-life problems, see also the explanatory note \cite{roques2023learning}.

Let us consider probably the most resource-consuming NN part: matrix-vector multiplication (MVM). There are three main ways to implement the MVM in the optical domain. The first kind of optical MVM (the plain light conversion, PLC-MVM) is based on the diffraction of light in free space. Fig.~\ref{fig:MVM} shows a typical MVM configuration. First, the incident vector of $X$ distributed along the $x$ direction can be expanded and replicated along the $y$ direction through a cylindrical lens or other optical elements. Then, the spatial diffraction plane is used to adjust each element independently, and its transmission matrix is $W$. Finally, the $x$-direction beams are combined and summed similarly, and the final output vector of $Y$ along the $y$-direction is the product of the matrix of W and the vector X, that is, $Y =WX$. The second MVM exploits a Mach–Zehnder interferometer (MZI) network (MZI-MVM). Fig.~\ref{fig:MZI} depicts the configuration diagram, which is based mainly on rotation submatrix decomposition and singular value decomposition. The calibration of the transmission matrix is more difficult since every matrix element is affected by multiple dependent parameters. For a simple 2x2 MZI multiplier, considering the inputs $x_1$ and $x_2$, the matrix multiplication with a 2x2 weight $W$ results in an output that follows the formula\cite{cheng2021photonic}:
\begin{equation}
      Y = U^2(\theta, \alpha,\beta) X,
\end{equation}
\begin{equation}
  W = U^2(\theta, \alpha,\beta) = \begin{bmatrix}
e^{-j\alpha}(e^{-j\theta}-1) & je^{-j\alpha}(1+e^{-j\theta}) \\
je^{-j\beta}(e^{-j\theta}+1) & e^{-j\beta}(1-e^{-j\theta}) 
\end{bmatrix},
\end{equation}
where to set the weight values to the desired ones, the phase shifter $\theta, \alpha,$ and $\beta$ need to be properly adjusted.

The third MVM (i.e., WDM-MVM) is an incoherent matrix computation method based on WDM technology. Fig. \ref{fig:WDM} shows a typical diagram based on microring resonators (MRRs). The input vector of X is loaded onto beams with different wavelengths, which pass through the micro-rings with one-on-one adjustment of the transmission coefficients of W. Then, the total output power vector is given by $Y=WX$. 

In Ref.~\cite{zhang2021optical}, an optical neural chip was designed in which matrix multiplications were performed using the MZI network, and a simple nonlinear activation function was based on intensity detection $f(x) =||x||$. A good survey and comparison of the different MVM realizations and photonic chip architectures are given in recent reviews \cite{peserico2022photonic,thomaschewski2023high}.

\subsection{Convolutional Neural Networks}
\begin{figure}[ht!]
\centering
\includegraphics[width=.8\linewidth]{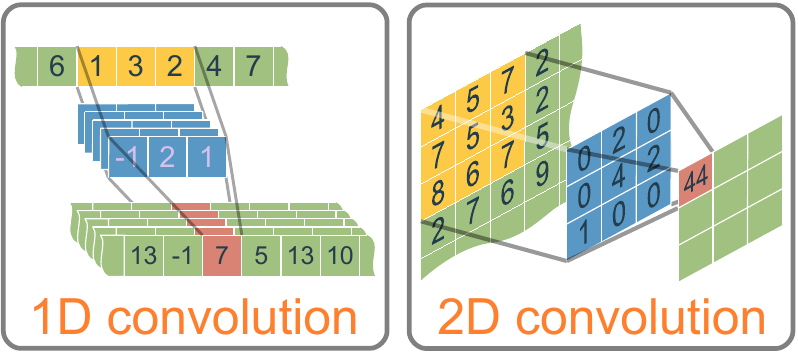}
\caption{Schematics of 1D and 2D convolutional filters. A series of arrays of such filters constitute a convolutional neural network.}
\label{fig: CNN filter}
\end{figure}

In a convolutional NN (CNN), we apply the convolutions with different filters to extract the features and convert them into a lower-dimensional feature set, \egor{as can be seen in Fig.~\ref{fig: CNN filter}}. The CNNs can be used in 1D, 2D, or 3D network arrangements depending on the applications. Here we focus on 1D-CNNs, which are applicable to, e.g., processing sequential data \cite{kiranyaz20211d}. 
The 1D-CNN processing with padding equal to 0, dilation equal to 1, and stride equal to 1, can be summarized as the following transformation:
\begin{equation}\label{eq.cnn}
  y^{f}_{i} = \phi \left(\sum_{n=1}^{n_i}\sum_{j=1}^{n_k}x^{in}_{i+j-1,n} \cdot k^{f}_{j,n} + b^{f} \right),
\end{equation}
where $y^{f}_{i}$ denotes the output, known as a feature map, of a convolutional layer built by the filter $f$ in the $i$-th input element, $n_k$ is the kernel size, $n_i$ is the size of the input vector, $x^{in}$ represents the raw input data, $k^{f}_{j}$ denotes the $j$-th trainable convolution kernel of the filter $f$ and $b^{f}$ is the bias of the filter $f$.    

In the general case, the additional parameters, such as padding, dilation, and stride, also affect the output size of the CNN. The padding adds information (often zeros) to the empty points around the edges of an input signal so that its size stays the same after the convolution operation. The dilation and stride affect how the kernel operation will behave in the convolution. The dilation ``inflates'' the kernel by adding holes between the kernel elements, and the stride controls how the filter convolves the input signal by setting the number of shifting units at a time that the kernel will do in the convolution. The generalized output shape of the 1D-CNN can be formalized as:
\begin{equation}\label{outputwidth.cnn}
Output Size = \! \left \lfloor \frac{n_s +2  \, padding \! - \! dilation (n_k-1) \! - \! 1 }{stride} \! + \! 1 \right\rceil \! ,
\end{equation}
where $\left\lfloor...\right\rceil$ is the nearest integer operation,  $n_s$ is the input time sequence size and $n_k$ is, again, the respective kernel size.

To understand the relation of the CNN to the ordinary digital signal processing (DSP) filtering, recall that the output of the 1D finite impulse response (FIR) filter can be presented as follows (see, e.g., \cite[p.~58]{woods2017fpga}):
\begin{equation}\label{eq:fir}
y^{\text{FIR}}_i = \sum_{m=0}^{n_{\text{FIR}}-1} x_{i-m} \cdot \kappa_m, 
\end{equation}
where $\kappa_m$ is the set of coefficients (time-reversed impulse response) that generate the required filter response (e.g., low-pass, high-pass, baseband, etc.);  $n_{\text{FIR}}$ is the number of filter taps in the output, i.e., the FIR filter order.  Comparing ~(\ref{eq:fir}) with (\ref{eq.cnn}), we can put: $m=1-j$, and designate $n_k = 1- n_{\text{FIR}}$, $\kappa_{1-j} = k_j$, to obtain: $y_i = \sum_{j=1}^{n_k} x_{i+j-1} \cdot k_j$. We can readily see that the action of a CNN layer before the activation is tantamount to the convolution of the several FIR filters' outputs, and the whole CNN layer adds the nonlinearity to the convolution of FIR filters via the activation function; if, otherwise, $\phi$ in Eq.~(\ref{eq.cnn}) is a linear function, $\phi(x)=x$, the CNN transforms into the direct FIRs convolution. 

Two-dimensional convolutional filters used in the vast majority of modern image processing CNNs are similar to their 1D counterparts \egor{(see Fig. \ref{fig: CNN filter})}, but with the increased dimensionality the additional summation over the second axis is added:
\begin{equation}\label{eq. 2D cnn}
  y^{f}_{i,j} = \phi \left(\sum_{i=1}^{n_i}\sum_{j=1}^{n_j}\sum_{m=\lf-n_m/2\rf}^{\lf n_m/2 \rf}\sum_{l=\lf-n_l/2\rf}^{\lf n_l/2 \rf}x_{i+m,j+l} \cdot K^{f}_{m,l} + b^{f} \right).
\end{equation}

\begin{figure}[ht!]
\centering
\includegraphics[width=.8\linewidth]{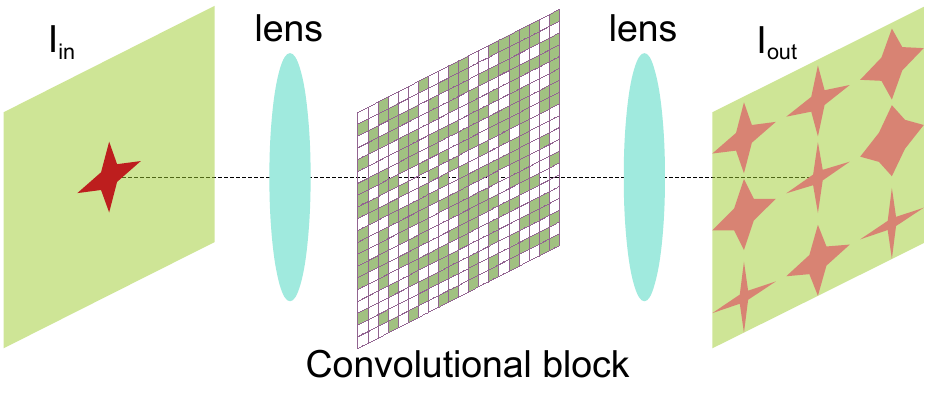}
\caption{Optical 2D convolution using scattering matrix in a Fourier plane of a 4-f imaging system.}
\label{fig: 2D CNN experimental}
\end{figure}

These 2D convolutions are very similar to the optical concept of a point spread function (PSF) that is used for the description of the response of a focused optical imaging system to a point source or point object. In free space optical systems, the image behind a scattering medium can be described as a convolution of the original image with a PSF \egor{remove extra colon :} \cite{chang2018hybrid}:
 \begin{equation}\label{eq. PSF convolution}
  I_{out}(x,y) = I_{in}(x,y)\ast PSF(x,y),
\end{equation}
where $\ast$ denotes a 2D convolution. Thus, a 2D convolution can be implemented in free-space optics with a diffraction mask in the Fourier plane of a 4-f imaging system, as shown in Fig.~\ref{fig: 2D CNN experimental}

 \subsection{Vanilla Recurrent Neural Networks} \label{sec.rnn}
\begin{figure}[ht!]
\centering
\includegraphics[width=\linewidth]{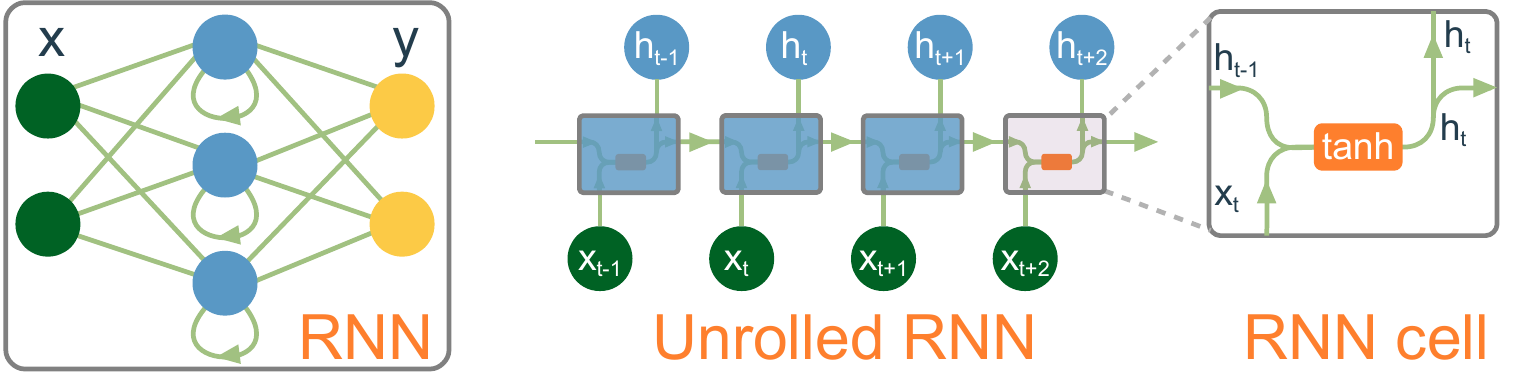}
\caption{Schematics of a recurrent neural network. Hidden layer neurons with closed-loop connections underlie the memory effect.}
\label{fig: RNN schematics}
\end{figure}

Vanilla RNN is different from MLP and CNN in terms of its ability to handle memory, which is quite beneficial for time series analysis and prediction. Here, we note that the feedforward models (e.g., those described above) can be reckoned, according to J. L. Elman \cite{elman1990finding}, as an ``... attempt to ``parallelize time'' by giving it a spatial representation...  However, there are problems with this approach,  and it is ultimately not a good solution. A better approach would be to represent  time  implicitly rather than explicitly.'' The recurrent structures described in the following subsections do that implicit representation, Fig.~\ref{fig: RNN schematics}: RNNs take into account the current input and the output that the network has learned from the prior input. The propagation step for the vanilla RNN at the time step $t$, can be described as follows: 
\begin{equation}\label{eq.rnn}
    \begin{gathered}
    h_{t} = \phi(W{x}_{t} + Uh_{t-1} + b), \\
    \end{gathered}
\end{equation}
where $\phi$ is again the nonlinear activation function, $x_{t}\in \mathbb{R}^{n_i}$ is the $n_i$-dimensional input vector at time $t$, $h_{t} \in \mathbb{R}^{n_h}$ is a hidden layer vector of the current state with size $n_h$,  $W \in \mathbb{R}^{n_h \times n_i}$  and $U\in \mathbb{R}^{n_h \times n_h}$ represent the trainable weight matrices, and $b$ is the bias vector. For more explanations on the vanilla RNN operation, see, e.g., Ref.~\cite{lipton2015critical}. Even though the RNNs were tailored for efficient memory handling, they still suffer from the inability to capture the long-term dependencies because of the infamous vanishing gradient issue \cite{bengio1994vanishgrad}. 

\begin{figure}
    \centering
    \includegraphics[width=\linewidth]{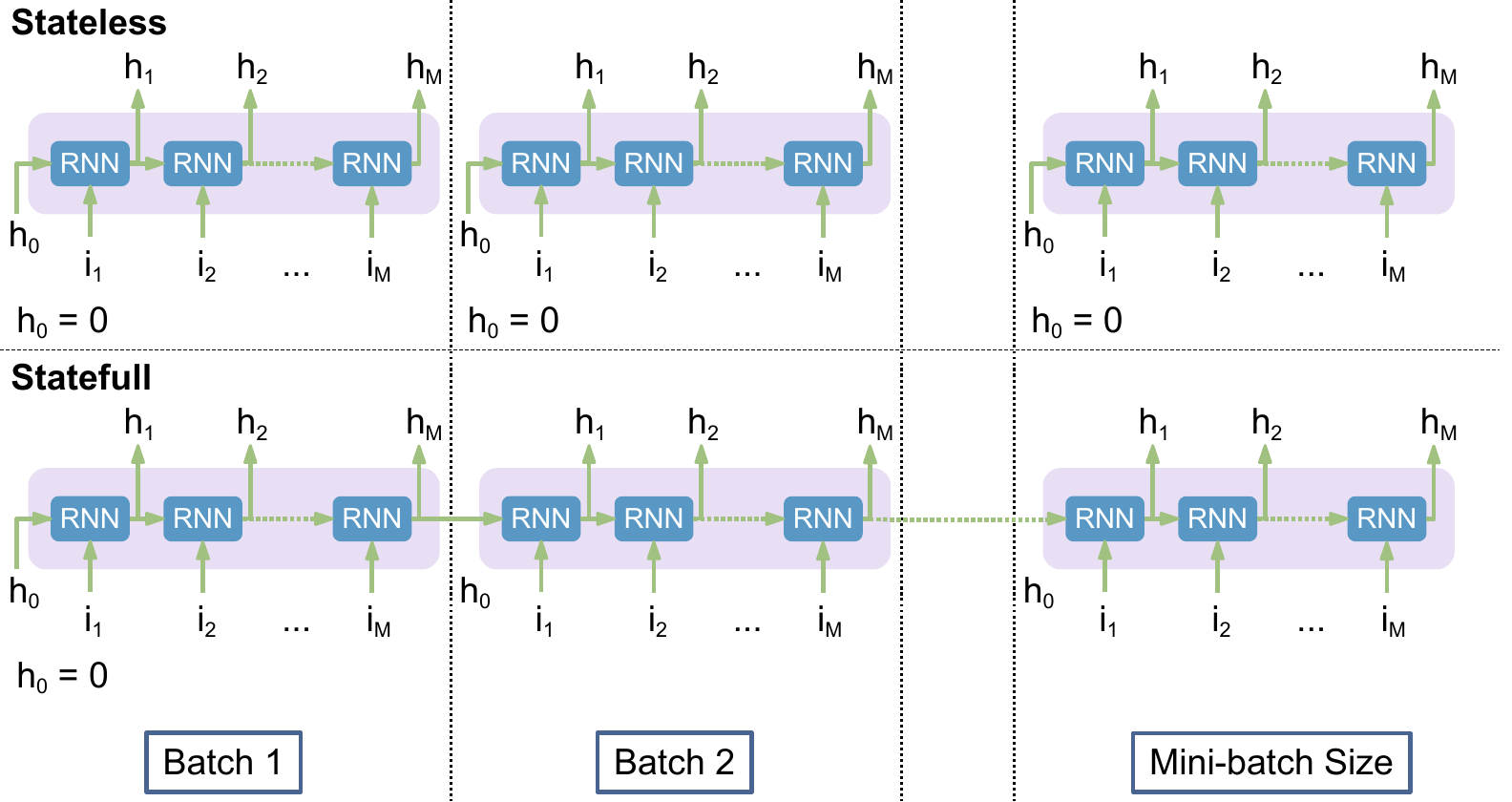}
    \caption{Stateful and stateless RNN training architectures.}
    \label{fig:stateful}
\end{figure}

In addition to the mathematical description of such a layer, when designing sequence modeling algorithms (i.e., the algorithms involving recurrent layers), it is crucial to consider whether the training architecture is stateless or stateful \cite{sanchez2020exploiting, saha2022comprehensive,pham2019recurrent}. Fig.~\ref{fig:stateful} schematically illustrates how both architecture types work. The primary difference between these two architectures is how the first state ($h_0$) of the model (corresponding to each batch) is initialized as the training advances from one batch to the following one. Considering that the input share of these sequential data is $input=[batch_{size}, time_{length}=M, input_{features}]$,  in both architectures, for each batch of data, we utilize $M$ recurrent cells in forward propagation. However, in the stateless architecture, every batch initializes the first state as $h_0 = 0$. This causes the model to forget the prior batch's learning. This design is utilized when the i.i.d. assumption for the data distribution is true\footnote{This is similar to other supervised learning methods where we assume that each batch of the dataset you pass, is i.i.d. with respect to each other.}. This means that when building the training batches, there is no interdependence between the batches, and each batch is independent. This is not to be confused with the parameters/weights, which have already propagated through the entire training process, which is the goal of training.
 
Nevertheless, not all sequential data, such as time series, contain non-i.i.d. samples; hence, it is not reasonable to always presume that the divided batches are completely independent. Therefore, it is natural to propagate the learned states across successive batches in such a way that the model not only reflects the temporal dependence inside each sample sequence but also does this across batches. The second type, called stateful architecture, is the solution to this problem. In the stateful architecture, the cell and the hidden states of the recurrent cell for each batch are initialized using the states learned from the previous batch, allowing the model to learn the dependence between batches for each sample in the batch. As indicated in Fig.~\ref{fig:stateful}, the states are reset to zero only at the start of each epoch. 

Here, we highlight that the default implementation of recurrent cells in the most popular machine learning libraries (TensorFlow and PyTorch) uses the stateless setup. In order to transit to a stateful architecture, in addition to connecting the hidden states across batches, it must be ensured that the batches cannot be shuffled internally (which, otherwise, is the default step in the case of stateless architecture, and in the case of stateful architectures it would break the learning process)\footnote{Most applications in practice use the stateless RNN, because if we use the stateful RNN, then in production, the network is forced to deal with infinitely long sequences, and this property can be quite difficult to handle.}.

In the stateless configuration, the linear part that describes the vanilla RNN resembles the well-known infinite impulse response (IIR) filter. The IIR filter is characterized by its theoretically infinite impulse response as \cite{van2001reference,storn1996differential}:
\begin{equation}
    y(t) = \sum_{k=0}^{\infty}r(k)x(t-k),
\end{equation}
where $r(k)$ is the linear time-invariant filter’s impulse response. 

Practically speaking, it is not possible to compute the output of the IIR using this equation. Therefore, the equation may be rewritten in terms of a finite number of poles $p$ and zeros $q$ of the IIR filter, as defined by the linear constant coefficient difference equation: 
\begin{equation} \label{eq:IIR}
        y(t) = \sum_{k=0}^{q}a_k x(t-k) + \sum_{k=1}^{p}b_k y(t-k),
\end{equation}
where $a_k$ and $b_k$ are the filter’s denominator and numerator polynomial coefficients, the roots of which are equal to the filter’s poles and zeros, respectively. In this sense, if $p=q=1$ and if we consider that the number of hidden units in the recurrent cell is equal to 1, Eq.~(\ref{eq:IIR}) and the linear part of (\ref{eq.rnn}) become the same.

Now, it is interesting to analyze the interrelation of the RNN models with a Kalman filters theory\cite{kalman12,brookner1998tracking}. To do this, we first consider the Elman variant of RNN \cite{elman1990finding,cruse2006neural}, a relatively simple 3-layer recurrent structure, where the ``hop'' of the variable $h$ from $t-1$ to $t$, is given by Eq.~(~\ref{eq.rnn}), and the output $y_t$ (the prediction associated with the input $x_t$) is defined by:
\[ y_{t} = \phi_y ( H h_t + d), \] 
where $H$ is the matrix of parameters to optimize for our getting the best prediction, and $d$ is the bias vector; $\phi_y$ is the activation function, whose subindex $y$ signifies that it can be different from the activation function in (\ref{eq.rnn}). Index $t$ can be understood as the number of the pairs $\{x_t,\tilde{y}_t\}$ in the overall dataset used for the training, where $\tilde{y}_t$ is the true value, while $y_t$ marks the prediction given by the RNN; $y_t$ produced by the RNN run number $t$, can also be understood as a ``measurement'' rendered by our RNN model at the $t$-th step, whereas $\tilde{y}_t$ can be reckoned as the true result of the ``measurement''.

Applied to the regression task, the goal of the RNN training is to identify the optimal NN structure, namely the particular values of the parameters (matrices and vectors) $W$, $b$, $U$, $H$, and $d$, which are further used in the inference stage. Optimization is performed by minimizing the loss function, i.e., typically some characteristic function of the difference of $y_t$ (predicted by the RNN) and the ``true value'', $\tilde{y}_t$. Quite often, the regression loss function to minimize is the mean-squared error (MSE), such that we minimize $\mathrm{MSE} (y_t - \tilde{y}_t) = |y_t - \tilde{y}_t|^2$ (where $| \ldots |$ means the norm); the goal is to minimize this function across all allowed values of $W$, $b$, $U$, $H$, and $d$\footnote{The word ``allowed'' in this statement means that we can impose some specific borders on the range of each parameter's change, based on our experience, a priori information, desired solution properties, etc.}. 

Now, turning to the ordinary two-step Kalman filter, it deals with the estimates (predictions) attributed to the linear systems of discrete equations \cite{brookner1998tracking}:
\begin{equation}\label{eq:kalman} 
\begin{split}
\eta_t &= W^K \varkappa_t + U^K \eta_{t-1} + \beta_t, \\
\psi_t &= H^K \eta_t + \delta_t,
\end{split}
\end{equation}
where $\eta$  variables represent the ``hidden'' system state to which we do not have direct access (cf. the internal $h$-variables in the RNN), the matrix $U^K$ defines the state's discrete transition model which is applied to the previous state at step $t-1$, $\eta_{t-1}$ in our case; $W^K$ is the control-input model which is applied to the control vector $\varkappa_t$; $\beta_t$ is the process noise, which is, for the correctness of the Kalman filter theory, assumed to be drawn from a zero mean multivariate normal distribution with known covariance independent of any other variables\footnote{Sometimes, in the literature, the variables $\kappa_t$ and $\beta_t$ in (\ref{eq:kalman}) are marked with index $t-1$, but not $t$; of course, this change of notations does not affect the physical meaning of the result. The matrices $W^K$, $U^K$, and $H^K$, can also alter with the index $t$, but we omit this dependence here for simplicity.}. In the lower equation from Kalman filter set (\ref{eq:kalman}), $H^K$ is the observation model, which maps the true state space $\eta$ into the observations space $\psi$, and $\delta_t$ is the observation noise, which is, again, assumed to be drawn from a zero mean multivariate normal distribution with known covariance independent of any other variables. 

The Kalman filter algorithm can be conceptualized in two steps (that we do not detail here mathematically): i) a prediction step and ii) an update step. Initially, we assume that we have the \textit{a priori} estimate of $ \eta_{t-1}$ (the prior), say $\hat{\eta}_{t-1}^-$, obtained at the previous step of the algorithm, and we know the (diagonal) error covariance matrix associated with $\eta_{t-1}^-$; the latter is also computed at the previous step. For the prediction at step $t$, we now use the value $\psi_t$ and calculate the optimal Kalman gain matrix. The Kalman gain allows us to update the prior and calculate \textit{a posteriori} (the posterior) value for the hidden variable estimate, $\hat{\eta}_{t-1}$. For the optimal Kalman gain, the value of the expectation for $\mathrm{MSE}(\eta_{t-1} - \hat{\eta}_{t-1})$ (the variance of the posterior) is minimal, and it is used to calculate the prior of the covariance matrix associated with $\eta_t$, see Refs.~\cite{kalman12,brookner1998tracking} for detailed equations and explanations.

We can notice the difference in the outputs for the RNN and for the Kalman algorithm: while the former attempts to minimize the MSE for the difference of the RNN result $y_t$ with the observed value, $\tilde{y}_t$, the Kalman algorithm ensures the minimization of the errors' posterior estimates for the hidden states $\eta$, and the latter is the `analogs`'' of the hidden RNN variables $h$. Thus, we ought to understand how the Kalman filter handles the estimation of $\psi$ errors (the so-called \textit{innovations}), the measurement, or the ``observer'', i.e., to deal with the so-called pre-fit and post-fit residuals. However, it is known that the observer for the optimal Kalman gain is also optimal in the MSE sense\cite{juang1993estimation}, and, so, the Kalman filter also minimizes the observation error; therefore, the tasks for the regression Elman RNN and Kalman filter are, indeed, similar, and it is possible to compare the results of two approaches. As the answer for the ``ideal'' Kalman system is obvious, and it has been rigorously proven that the Kalman filter for such a system is an optimal linear estimator, the two approaches are often compared for systems and conditions different from ~(\ref{eq:kalman}): it was found that the RNN can give good results in conditions where the classical Kalman filter fails, e.g., when the system to estimate is nonlinear \cite{decruyenaere1992comparison}.  

Now, let us turn to the distinctions between the two approaches. Firstly, obviously, the RNN contains nonlinear activation functions, while the Kalman system is linear. Secondly, the Kalman system contains random variables, the process, and measurement noises, and the optimal Kalman gain is expressed through the (known) variances of the two noises. In contrast, the ``ordinary'' RNN's parameters are deterministic\footnote{We do not consider here the case of stochastic NNs \cite{jospin2022bayesian}.}. However, the important difference is that the Kalman filter in its original positioning cannot learn, it just gives the estimate based on the known system's parameters, and this estimate is optimal in the MSE sense if the specific conditions (system's linearity and the white additive Gaussian character of participating noises) are fulfilled. We can, of course, state the problem differently: find some (or all) of the parameters of the system using the given input (control vector) and measurement pairs; the latter are supposed to be associated with the Kalman system\cite{chung1991thesis}. The latter problem statement is already closer to the learning phase of the RNN\footnote{As noted in \cite{juang1993estimation}, when we have the deviations of the true system from the ideal Kalman case, the resulting filter identified through the input-measurement pairs is not the Kalman filter. In such a case, the identified filter is simply an
observer that is computed from input-output data that minimizes the filter residual in an MSE sense.}. More details on the comparison of different Kalman filtering-based techniques and RNNs, as well as the interpenetration of these two techniques, can be found in \cite{chenna04,parlos01,haykin2001kalman,mandic2009complex}. Finally, we also mention that the Kalman filter theory and its extensions can be efficient in the training of NNs\cite{shao2021training}, yet another important application relating the two concepts.

Concerning the photonic implementation of recurrent structures, we notice that these are rarer compared to the feed-forward counterparts. First, we notice Ref.~\cite{tait2017neuromorphic}, where the authors proposed a photonic architecture enabling all-to-all continuous-time RNN. We also mention Ref.~\cite{bueno2018reinforcement}, where a free-space network of up to 2025 diffractively coupled photonic nodes, forming a large-scale RNN, was demonstrated, and Ref.~\cite{zhou2022photonic}, where the experimental realization of diffractive RNN was also evaluated. Some further analysis of recurrent topology implementation is given in review \cite{shastri2021photonics}. another RNN (coupled with CNN) realization is considered in \cite{Prucnal04}. An interesting generalized look at the realization of RNN in hardware is presented in Ref.~\cite{Fan03}.


\subsection{Long Short-Term Memory Neural Networks} \label{sec.lstm}
\begin{figure}[ht!]
\centering
\includegraphics[width=.4\linewidth]{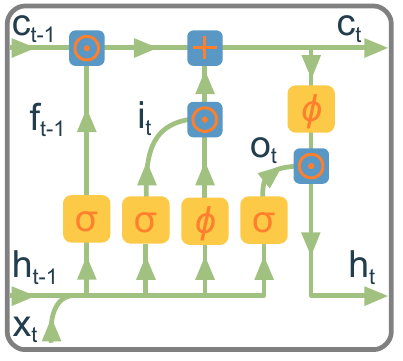}
\caption{Schematics of an LSTM cell that constitutes the backbone of LSTM neural networks. Weight matrices are omitted.}
\label{fig: LSTM cell}
\end{figure}

LSTM is an advanced type of RNN. While RNNs suffer from short-term memory issues, the LSTM network has the ability to learn long-term dependencies between time steps ($t$), insofar as it was specifically designed to address the gradient problems encountered in RNNs~\cite{hochreiter1997long, gers2000learning}. LSTM networks are made up of LSTM cells, which are units that contain a series of gates that can control the flow of information into and out of the cell, as shown in Fig. \ref{fig: LSTM cell}. The gates can learn to keep relevant information and discard irrelevant information, allowing the LSTM cell to remember important information for long periods of time. More specifically, there are three types of gates in an LSTM cell: an input gate ($i_t$), a forget gate ($f_t$), and an output gate ($o_t$). More importantly, the cell state vector ($C_t$) was proposed as a long-term memory to aggregate relevant information throughout the time steps.

The LSTM equation, as shown in Eq.~(\ref{eq.lstm}), describes the computations involved in a single time step of an LSTM model. The input at time step $t$, $x_t\in \mathbb{R}^{n_i}$, is processed by the LSTM model to produce an output at time step $t$, $h_t\in (-1,1)^{n_h}$. The subscript $t$ denotes the current time step, while $t-1$ denotes the previous time step.

\begin{equation}\label{eq.lstm}
    \begin{gathered}
i_{t} = \sigma(W^{i}{x}_{t} + U^{i}{h}_{t-1} + b^{i} ),  \\
    f_{t} = \sigma(W^{f}{x}_{t} + U^{f}{h}_{t-1} + b^{f}), \\
o_{t} = \sigma(W^{o}{x}_{t} + U^{o}{h}_{t-1} + b^{o}),\\
    C_{t} = f_{t}\odot C_{t-1} + i_{t}\odot \phi(W^{c}{x}_{t} + U^{c}{h}_{t-1}+ b^{c}), \\
    h_{t} = o_{t} \odot \phi(C_{t}),
    \end{gathered}
\end{equation}
with $\odot$ being the element-wise (Hadamard) multiplication, where $\phi$ is usually the ``tanh'' activation functions, $\sigma$ is usually the sigmoid activation function, the sizes of each variable are  $x_{t}\in \mathbb{R}^{n_i}$,  $f_{t}, i_{t}, o_{t}\in (0,1)^{n_h}$, $C_{t}\in \mathbb{R}^{n_h}$ and $h_{t}\in (-1,1)^{n_h}$. 

To explain further, the LSTM equation above is divided into 5 stages. First, the input gate controls the flow of information into the memory cell. It takes the input $x_t$ and the previous hidden state $h_{t-1}$ as inputs, and produces an output $i_t\in (0,1)^{n_h}$ that represents the degree to which the input should be written to the memory cell.  Second, 
the forget gate controls the flow of information out of the memory cell. It takes the input $x_t$ and the previous hidden state $h_{t-1}$ as inputs, and produces an output $f_t\in (0,1)^{n_h}$ that represents the degree to which the previous cell state $C_{t-1}$ should be retained. Next, the output gate controls the flow of information out of the memory cell. It takes the input $x_t$ and the previous hidden state $h_{t-1}$ as inputs, and produces an output $o_t\in (0,1)^{n_h}$ that represents the degree to which the current cell state $C_{t}$ should be outputted.  Then, the memory cell $C_t$ is responsible for storing and updating information over time. It takes the input $x_t$, the previous hidden state $h_{t-1}$, and the previous cell state $C_{t-1}$ as inputs, and produces a new cell state $C_t\in \mathbb{R}^{n_h}$ that integrates the current input and the previous memory. Finally, the hidden state $h_t$ is the output of the LSTM model at time step $t$. It takes the current cell state $C_t$ and the output gate $o_t$ as inputs, and produces an output $h_t\in (-1,1)^{n_h}$ that represents the current hidden state of the LSTM model. 

Note that LSTM networks are trained using backpropagation through time, where the error is propagated back through the network over multiple timesteps. This allows the LSTM network to learn how to use information from earlier timesteps to make predictions at later timesteps.

Here, it is also important to mention the existence of another structure called bidirectional LSTM (BiLSTM). The BiLSTM is a type of LSTM that processes the input sequence in both forward and backward directions and concatenates the output of both directions at each time step. This allows the model to have access to information from both past and future contexts of the input sequence, making it more effective in capturing both past and future contexts of the input sequence, which is important for analyzing temporal patterns in optical signals. In particular, in optical fiber communications, BiLSTMs have been shown to be effective in analyzing the temporal patterns of optical signals for detecting and mitigating various impairments, such as polarization mode dispersion and chromatic dispersion.  Similarly, in optical sensing,  BiLSTMs can be more effective than regular LSTMs in capturing the temporal patterns of optical signals. By processing the optical signal in both forward and backward directions, BiLSTMs can capture the context of the signal from both the past and the future, leading to more accurate detection and measurement of physical parameters.

The equations for the forward and backward LSTM layers are similar to the ones for the regular LSTM, except that they are computed in opposite directions. The forward LSTM layer processes the input sequence from the first time step to the last, while the backward LSTM layer processes it from the last time step to the first. The output of the forward LSTM layer at time step $t$ is denoted by $h_{t}^{f}$, and the output of the backward LSTM layer at the same time step is denoted by $h_{t}^{b}$. 

\subsection{Gated Recurrent Units}
\begin{figure}[ht!]
\centering
\includegraphics[width=.4\linewidth]{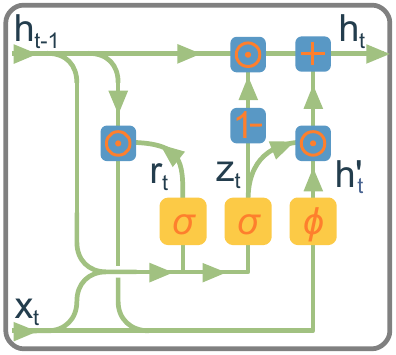}
\caption{Schematics of a GRU cell which is a less computationally complex alternative to the LSTM cell. Weight matrices are omitted.}
\label{fig: GRU cell}
\end{figure}

Introduced in 2014 \cite{cho2014learning}, the GRU network, similar to the LSTM,  was designed to overcome the short-term memory issues of RNNs. However, the GRU is less complex than the LSTM\footnote{The difference in the LSTM and GRU functioning is studied in detail in Ref.~\cite{chung2014empirical}.}, as it has only two types of gates: the reset ($r_t$) and update ($z_t$) gates, as shown in Fig. \ref{fig: GRU cell}. The reset gate is used to handle short-term memory, whereas the update gate is responsible for long-term memory \cite{dey2017gate}. In addition, the candidate hidden state ($h'_{t}$) is also introduced to measure how relevant the previous hidden state is to the candidate state.
The GRU for a time step $t$ can be formalized as: 
\begin{equation}\label{eq.gru}
    \begin{gathered}
    z_{t} = \sigma(W^{z}{x}_{t} + U^{z}{h}_{t-1} + b^{z}),  \\
    r_{t} = \sigma(W^{r}{x}_{t} + U^{r}{h}_{t-1} + b^{r}), \\
    h'_{t} = \phi(W^{h}{x}_{t} + r_{t} \odot U^{h}{h}_{t-1} + b^{h}), \\
    h_{t} = z_{t} \odot {h}_{t-1} + (1 - z_{t}) \odot h'_{t},
    \end{gathered}
\end{equation}
where $\phi$ is typically the ``tanh'' activation function and the rest of the designations are the same as in Eq.~(\ref{eq.lstm}).

In addition to ~(\ref{eq.gru}), defining the so-called fully gated unit, the simpler GRU architecture variants called minimal gated unit are also sometimes used\cite{heck2017simplified}: in these types, the reset, and update gates are merged. Some other GRU variants are described and compared in Ref.~\cite{dey2017gate}.

\subsection{Echo State Networks} \label{sec.esn}

Echo state networks (ESNs) belong to the class of recurrent structures, more specifically, to the reservoir computing category\cite{jaeger2004harnessing}. The ESN was proposed to simplify the training process while staying efficient and simple to implement. The ESN comprises three layers: an input layer, a recurrent layer, known as a reservoir, and an output layer, which is the only layer that is trainable. The reservoir with random weights assignment is used to replace back-propagation in traditional NNs to reduce the computational complexity of training~\cite{wu2018statistical}. We notice that the reservoir of the ESNs can be implemented in two domains: digital and optical~\cite{sorokina2019fiber}. With the optical implementation of the reservoir, the computational complexity dramatically falls; however, the degradation of the performance due to the change of domain can be non-negligible~\cite{brain-inspired}. In this work, we only examine the digital domain implementation. Moreover, we focus on the leaky-ESN, as it is believed to often outperform the ``standard'' ESNs and is more flexible due to time-scale phenomena \cite{sun2020review,jaeger2007optimization}. 
The equations of the leaky-ESN for a certain time step $t$ are given as:
\begin{equation}\label{eq.esn1}
 a_t = \phi \left( W^{r} s_{t-1} + W^{\text{in}} x_t + W^{\text{back}} y_{t-1} \right),
\end{equation}
\begin{equation}\label{eq.esn2}
  s_t = (1- \mu) s_{t-1} + \mu a_t,
\end{equation}
\begin{equation}\label{eq.esn3}
y_t = W^{o}s_{t} + b^{o},
\end{equation}
where $s_t$ represents the state of the reservoir at time $t$, $W^r$ denotes the weight of the reservoir with the sparsity parameter $s_p$, $W^{in}$ is the weight matrix that shows the connection between the input layer and the hidden layer, $\mu$ is the leaky rate, $W^{o}$ denotes the trained output weight matrix, and $y_t$ is the output vector.

\begin{figure}[ht!]
\centering
\includegraphics[width=.5\linewidth]{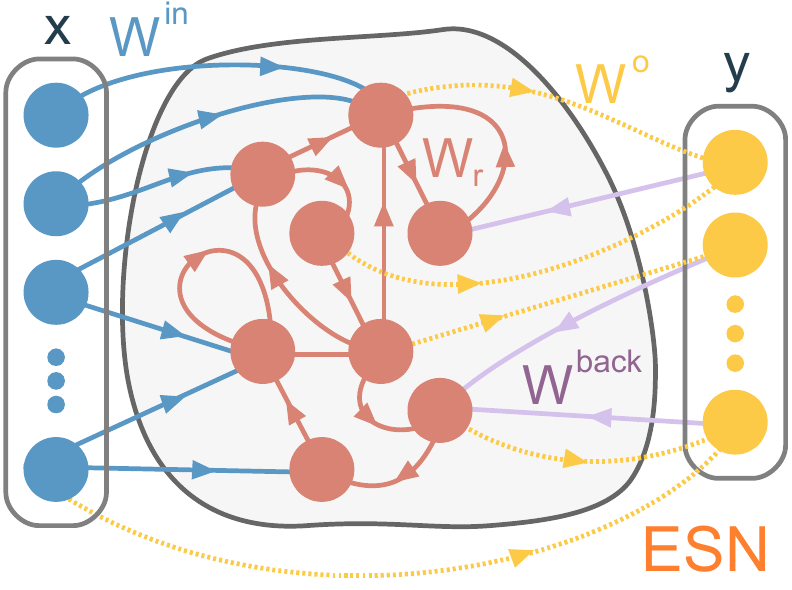}
\caption{Schematics of an echo state network is a recurrent neural network with only output weights trainable.}
\label{fig: ESN schematics}
\end{figure}

\begin{figure*}[ht!] 

\begin{subfigure}{.48\textwidth}
    \centering
\includegraphics[width=\linewidth]{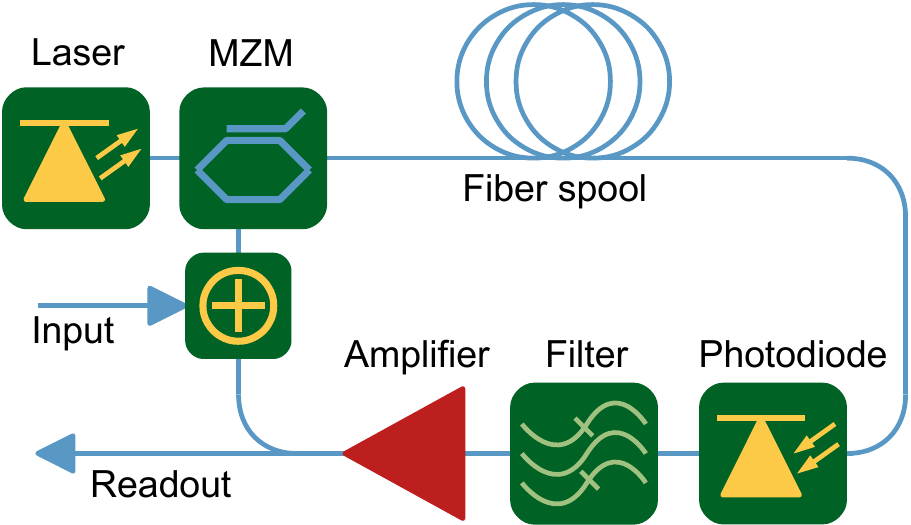}
    \caption{Optoelectronic ESN based on fiber cavity. Redrawn from Ref. \cite{van2017advances}}
    \label{fig:RC1 experimental} 
\end{subfigure}\hfill
\begin{subfigure}{.48\textwidth}
    \centering
\includegraphics[width=\linewidth]{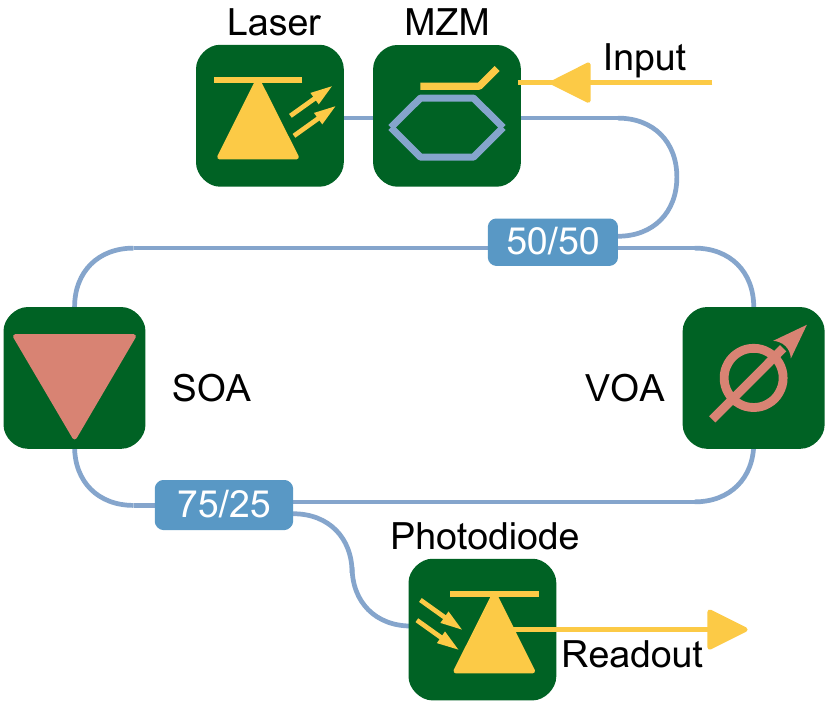}
    \caption{SOA-based reservoir computer. Redrawn from Ref. \cite{van2017advances}}    \label{fig:RC2 experimental} 
\end{subfigure}\hfill
\caption{Examples of optical reservoir computers or ESNs.}
\label{fig: experimental implementations of optical reservoir} 
\end{figure*}

The schematics of an ESN are shown in Fig. \ref{fig: ESN schematics}. The crucial point in the  ESN or reservoir computing concept is that despite the complex structure of these networks, only the weights of the output (readout) layer are trainable. One can see that the multiple interconnections described by matrices $W^{in}$, $W^{r}$, and $W^{back}$, constitute a complex recurrent structure with rich internal dynamics. Training of a classical MLP or RNN with a comparable number of neurons would be time-consuming. However, the concept of ESN speeds up the training process drastically and reduces it to linear regression on the output layer. 
The important feature of this type of NNs is that it can be easily implemented in the physical domain. Many dynamical systems with large internal phase space and exhibiting nonlinear properties can be employed as a reservoir. There are various experimental implementations of ESNs, including fiber-cavity-based schemes \cite{van2017advances}. Fig.~\ref{fig: experimental implementations of optical reservoir} shows fiber optic implementations of this concept: the fiber cavity with a circulating modulated signal serves as an optical reservoir.

Finally, we would like to highlight some potential drawbacks of using ESN which include: i)
Difficulty in training: ESN can be difficult to train, as they require careful tuning of the network's hyperparameters in order to achieve a good performance; ii) Limited ability to model long-term dependencies: An ESN is not able to effectively model long-term dependencies in the data, as they have a fixed-size reservoir and do not allow information to flow through the network over many timesteps.
\subsection{Attention Layers} 

\begin{figure}[ht!]
\centering
\includegraphics[width=.4\linewidth]{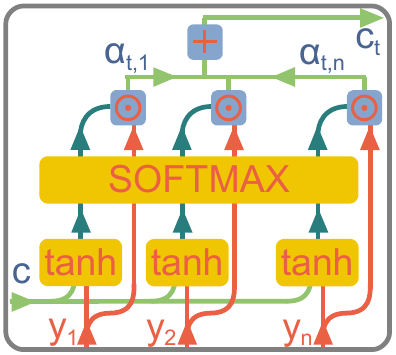}
\caption{Schematics of an attention unit that constitutes the main element of any attention NN. Trainable weights are omitted.}
\label{fig: attention cell}
\end{figure}

Attention is an NN mechanism that observes a whole collection of data and selectively focuses on a subset of the collection. In other words, attention mechanisms are a way to allow a model to focus on specific parts of its input when processing it, rather than using the entire input equally. The attention unit is schematically represented in Fig. \ref{fig: attention cell}. It was first applied to sequence-to-sequence learning in \cite{bahdanau2014neural} and was used mostly to further exploit the importance of each subset among the input data. In other words, attention is one add-on component of a network’s architecture, in charge of managing and quantifying the interdependence between the data of interest. General attention investigates the interdependence between input and output elements, whilst self-attention deals with finding correlations among input elements~\cite{quinn2019dive, luong2015effective, kim2017structured}. 

Let us turn to the case of general attention to account for the interdependence between the final predicted symbol and both the input symbols and the output hidden states.  By adding such an attention mechanism, we expect to find the contribution of the input symbols and their hidden representations to the final received symbol prediction. Therefore, we can identify the essential part of the input sequence for training that could lower the computational complexity.

The attention is generally a single- or multi-layer feed-forward NN with trainable weights and biases, which are applied to the output hidden states of the RNN layer. 

In the original attention mechanism \cite{bahdanau2014neural}, an input sequence $\{x_{1},...,x_{T_x}\}$ targets an output sequence $\{y_{1},...,y_{T_y}\}$. The conditional probability for a certain target output $y_i$, is defined as:
\begin{equation}
\label{eq:nmt_att_cond_prob}
    p(y_i|y_1,...y_{i-1},\mathbf{x}) = g(y_{i-1},s_i,c_i),
\end{equation}
where $g$ is a nonlinear, potentially multi-layered, function that outputs the probability of $y_i$; $s_i$ is an RNN's hidden state for time $i$ computed through $s_i = f(s_{i-1},y_{i-1},c_i)$. $c_i$ is a context vector conditioned for each target $y_i$, i.e.,  a vector generated from the sequence of the hidden states for predicting the current target output $y_i$; it is computed as a weighted sum of the hidden states $\{h_{1},\;...,\;h_{T_x}\}$:
\begin{equation}
\label{eq:nmt_c_i}
    c_i = \sum_{j=1}^{T_{x}} \alpha_{i,j}h_{j},
\end{equation}
where the weight $\alpha_{i,j}$ of each $h_j$ is computed by
\begin{equation}
\label{eq:nmt_alpha}
    \alpha_{i,j} =\frac{\exp{e_{ij}}}{\sum_{k=1}^{T_x} \exp{e_{ik}}},
\end{equation}
where $e_{ij}=a(s_{i-1}, h_j)$
is an alignment model which scores how well the inputs around position $j$ and the output at position $i$ match.

Instead of predicting the conditional probability of each target $y_i$ from a sequence of targets, we focus only on the received symbol $y_i$:
\begin{equation}
    y_i = g(c), \quad \text{where} \, \,    c = \alpha \ast h = [\alpha_{1}h_{1}, \; ...\;\alpha_{T_x}h_{T_x}].
\end{equation}
The weight $\alpha_{i}$ of each $h_i$ is calculated by
\begin{equation}
    \alpha_{i} =\frac{\exp{e_{i}}}{\sum_{j=1}^{2k+1} \exp{e_{j}}},
\label{eq:att_softmax}
\end{equation}
where $e_{i}=a(h_i)$ is the adapted alignment model and indicates the matching score between the output symbol $y_i$ and the hidden representations $\mathbf{h}$ of the input sequence $\mathbf{x}$.
According to \cite{bahdanau2014neural}, we can define the activation function $f$ of the RNN and the alignment model $a$ by choice. A single-layer perceptron (SLP) is selected as our alignment model.
 Matrix multiplication is first performed between the hidden input states and a trainable weight matrix $W_a$ $\in$ ${\rm I\!R}^{1\times n_h}$ with bias $b_a$ $\in$ ${\rm I\!R}^{1\times n_s}$, where $n_h$ is the number of hidden units, and $n_s$ is the input sequence length, after which a $\tanh$ function is applied as the activation function of the SLP:
\begin{equation}
    a(h_j) = \tanh(W_ah_j + b_{a_j}).
\label{eq:alignment}
\end{equation}
The softmax activation function is then applied to the alignment model to compute a probability, i.e., the attention score of the hidden states with respect to the final output symbol. The context vector $c$ is then obtained by an element-wise matrix multiplication between the attention score $\alpha$ and the hidden states. The attention score specifies the amount of attention given to each element of the hidden state sequence that corresponds to that of the input symbol sequence. 

Finally, we conclude by highlighting some  potential drawbacks and benefits of using attention mechanisms in machine learning models, including:
\begin{itemize}
    \item Increased complexity: Attention mechanisms can add additional complexity to a model, which can make the model more difficult to understand and debug.
    \item Increased training time: Attention mechanisms can also require more computation to train, which can increase the training time for a model.

    \item Improved performance: Attention mechanisms can allow a model to focus on the most relevant parts of the input, which can improve the model's performance on a variety of tasks.
    \item Better handling of long input sequences: Attention mechanisms can be particularly useful for tasks that involve long input sequences, such as machine translation, as they allow the model to focus on the most relevant parts of the input rather than processing the entire sequence equally.
    \item Improved generalization: Attention mechanisms can also improve the generalization of a model, as they allow the model to adapt to different input patterns and focus on the most important features.
\end{itemize}

\subsection{Transformers} 
\begin{figure}[ht!]
    \centering
    \includegraphics[width=0.9\linewidth]{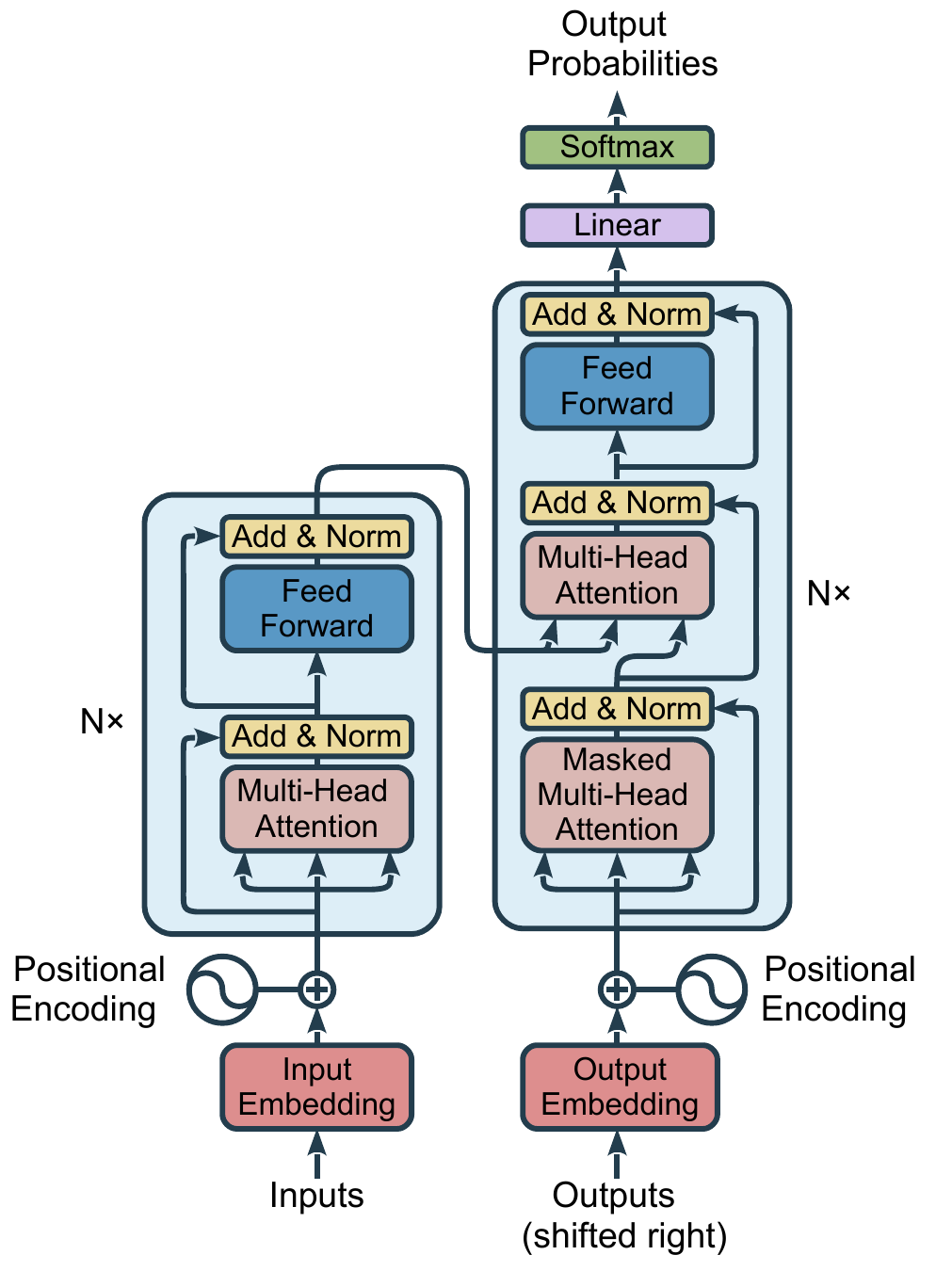}
    \caption{Transformer architecture reproduced from the original paper ~\cite{vaswani2017attention}}
    \label{fig: transformer}
\end{figure}

The vanilla transformer is a deep learning architecture that was introduced in Ref.\cite{vaswani2017attention}. Its architecture is shown in Fig.~\ref{fig: transformer}. The transformer is a sequence-to-sequence model that operates on sequences of vectors, where the goal is to learn a mapping from one sequence to another. The key innovation of the transformer is the use of the previously mentioned self-attention mechanism, which allows the model to weigh the importance of different parts of the input sequence when generating the output sequence. In a nutshell, the transformer consists of an encoder and a decoder. The encoder takes the input sequence and produces a sequence of hidden representations, which are then used by the decoder to generate the output sequence. The self-attention mechanism is used in both the encoder and the decoder, allowing the model to attend to different parts of the input sequence when generating each element of the output sequence.  The vanilla transformer can be expressed mathematically as follows:

Let $X = \{x_1, x_2, ..., x_n\}$ be the input sequence, where $x_i$ is a vector of dimension $d_{model}$, so the shape of $X$ is  [$n \times d_{model}$]. Similarly, let $Y = \{y_1, y_2, ..., y_m\}$ be the output sequence, where $y_j$ is a vector of dimension $m \times d_{model}$.

The encoder consists of $N$ identical layers, where each layer has three sub-layers: a multi-head self-attention mechanism, an Add\&Norm layer, and a position-wise fully connected feed-forward network. The output of the $i$th layer of the encoder is denoted as $H_i = \{h_{i,1}, h_{i,2}, ..., h_{i,n}\}$, where $h_{i,j}$ is a vector of dimension $n \times d_{model}$.

The multi-head self-attention mechanism can be expressed as:

\begin{align*}
\mathrm{MultiHead}(Q,K,V) &= \mathrm{Concat}(head_1, head_2, ..., head_h)W^O\\
\mathrm{where}\;\;\;\;\;\;\;\;\; head_i &= \mathrm{Attention}(QW_i^Q,KW_i^K,VW_i^V)\\
\mathrm{Attention}(Q,K,V) &= \mathrm{softmax}\left(\frac{QK^T}{\sqrt{d_k}}\right)V
\end{align*}

Here, $Q$, $K$, and $V$ are the query, key, and value matrices, respectively, with dimensions $n\times d_{model}$. The matrices $W_i^Q$, $W_i^K$, and $W_i^V$ \footnote{An important aspect of this setup is that each attention head has its own $W_V$, $W_Q$, and $W_K$ transforms. That means that each head can zoom in and expand the parts of the embedded space that it wants to focus on, and it can be different from what each of the other heads is focusing on.} are learned projection matrices with dimensions $d_{model} \times d_k$,$d_{model} \times d_k$, and $d_{model} \times d_v$, respectively. In this case, $d_k$ is the dimensions in the embedding space used for keys and queries and $d_v$ is the dimensions in the embedding space used for values.\footnote{Usually, $d_v$ is considered to be equal to $d_k$, but in reality they don't have to be.}. Also, note that because the input data as well as the linear layer weights are uniformly partitioned across the attention heads, the $d_k$ dimension is usually equal to $d_{model}/h$ and $h$ is the number of attention heads. $W^O$ is a learned projection matrix that concatenates the outputs of all the attention heads.

In the context of optical communications for denoising, one can interpret that the "key" represents the information that the model uses to look up relevant parts of the input signal (i.e, representations of the noisy signal at different positions/times), the "query" represents the information that the model is trying to find or pay attention to denoise the signal (i.e, representations of the noisy signal after some initial processing or encoding), and the "value" represents the actual content or information at each position in the signal sequence (i.e, these could be the noisy signal representations or even the same as the "key" in some cases). 

Additionally, it is important to note that masked multi-head attention can also be present in the transformer structure. In such case, the inputs of the softmax function are masked out by adding the matrix $M$ which contains 0's and $-\infty$'s. The $-\infty$'s correspond to invalid connections. The equation then is modified to
\begin{equation}\label{eq.maskedattention}
   \mathrm{Attention_{Masked}}(Q,K,V) = \mathrm{softmax}\left(\frac{QK^T}{\sqrt{d_k}} + M\right)V
\end{equation}

Next, the position-wise fully connected feed-forward network can be expressed as:

\begin{align*}
\mathrm{FFN}(x) &= \mathrm{ReLU}(xW_1 + b_1)W_2 + b_2
\end{align*}

Here, $x$ is a vector of dimension $d_{model}$, $W_1$ and $W_2$ are learned weight matrices, and $b_1$ and $b_2$ are learned bias vectors. The decoder also consists of $N$ identical layers, where each layer has three sub-layers: a multi-head self-attention mechanism, a multi-head attention mechanism between the encoder output and the decoder input, and a position-wise fully connected feed-forward network.

Besides the multi-headed and feed-forward layers, we also have the Add\&Norm layer. Such  Add\&Norm layer involves a residual connection around each of the two sub-layers \cite{he2016deep} followed by layer normalization \cite{ba2016layer}. The output of this Add\&Norm layer is:
\begin{equation}\label{eq.addnorm}
   y_\text{{Add\&Norm}} = \mathrm{LayerNorm}(x+\text{Sublayer}(x)),
\end{equation}
where $\text{Sublayer}(x))$ refers to the function implemented by the sub-layer itself, for instance, FFN or multi-head. The LayerNorm is then  defined as:
\begin{equation}\label{c.attention}
   \mathrm{LayerNorm}(z) = \frac{g(z-\mu_z)}{\sigma_z}+b,
\end{equation}
where $g$ and $b$ denote gain and bias, respectively, $\mu$ and $\sigma$ are the mean and variance of the summed inputs within each layer, respectively.

Finally, the transformer output can be computed as:

\begin{align*}
\mathrm{Transformer}(X,Y) &= \mathrm{softmax}(Z_N W^V)
\end{align*}

Here, $Z_N$ is the output of the last layer of the decoder, with dimensions $m \times d_{model}$, and $W^V$ is a learned projection matrix with dimensions $d_{model} \times |V|$, where $|V|$ is the size of the output vocabulary.

In optical communications, the transformer can potentially be used to perform various tasks, such as equalization, modulation classification, and channel estimation. In particular, the self-attention mechanism of the transformer can be used to model the complex interactions between the different components of the optical communication system, such as the transmitter, the channel, and the receiver. Additionally, the parallel processing nature of Transformers, coupled with the direct interactions among symbols in an input sequence, allows them to capture memories more efficiently than LSTM models, where memory is handled sequentially. This feature makes Transformers well-suited for hardware developments in ultra-high-speed optical transmissions\cite{hamgini2023application}.

\subsection{Residual Neural Networks} 

\begin{figure}[ht!]
\centering
\includegraphics[width=.4\linewidth]{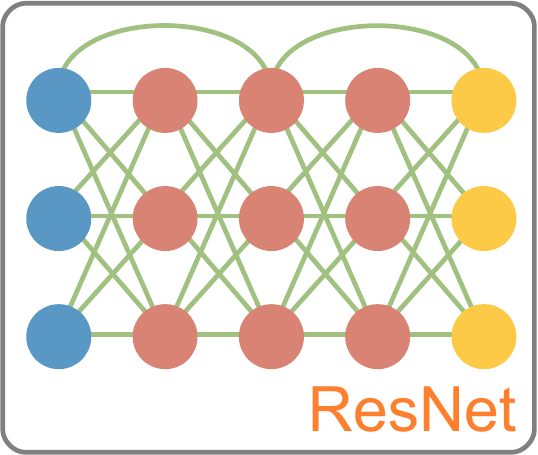}
\caption{Schematics of a residual neural network with double layer skips.}
\label{fig: ResNet schematics}
\end{figure}

An artificial neural network becomes a Residual Neural Network (ResNet)~\cite{he2016deep} if the input of a specific layer is also passed (or skipped) to another deeper layer in the network; this connection is called a residual connection. 
The utilization of skip connections or shortcuts, visually illustrated in Fig. \ref{fig: ResNet schematics}, is a distinctive feature of ResNets. These connections facilitate the bypassing of specific layers, thereby addressing challenges like vanishing gradients and promoting more efficient training within deep architectures.

Another famous architecture that uses residual connections is the HighwayNet~\cite{Srivastava2015}, The HighwayNet preserves the shortcuts introduced in the ResNet, but augments them with a learnable parameter to determine to what extent each layer should be a skip connection or a nonlinear connection. It is noteworthy that HighwayNets possess the capacity to autonomously learn the skip weights through an additional weight matrix governing their gates. In contrast, ResNet models are conventionally characterized by double or triple-layer skips, incorporating non-linear activation functions such as ReLU and batch normalization, which enhance the expressiveness and convergence capabilities of the models.

Additionally, DenseNets~\cite{huang2017densely} serve as a relevant descriptive reference for models incorporating multiple parallel skip connections, underscoring the adaptability and versatility of residual connections in contemporary neural network designs. 

Let us now define what is the feed-forward equations for such a type of NN layer. Given the weight matrix $W^{\ell-1, \ell}$ for the connection weights from layer $\ell-1$ to $\ell$, and the weight matrix $W^{\ell-2, \ell}$ for the connection weights from layer $\ell-2$ to $\ell$, then the forward propagation through the activation function would be (aka HighwayNets)
$$
\begin{aligned}
a^{\ell} &:=\mathbf{g}\left(W^{\ell-1, \ell} \cdot a^{\ell-1}+b^{\ell}+W^{\ell-2, \ell} \cdot a^{\ell-2}\right) \\
&:=\mathbf{g}\left(Z^{\ell}+W^{\ell-2, \ell} \cdot a^{\ell-2}\right),
\end{aligned}
$$
where
$a^{\ell}$ the activations (outputs) of neurons in layer $\ell$,
$\mathbf{g}$ the activation function for layer $\ell$,
$W^{\ell-1, \ell}$ the weight matrix for neurons between layer $\ell-1$ and $\ell$, and
$Z^{\ell}=W^{\ell-1, \ell} \cdot a^{\ell-1}+b^{\ell}$
Absent an explicit matrix $W^{\ell-2, \ell}$ (aka ResNets), forward propagation through the activation function simplifies to
$$
a^{\ell}:=\mathbf{g}\left(Z^{\ell}+a^{\ell-2}\right),
$$ activations from layer $\ell-2$ are passed to layer $\ell$ without weighting (aka DenseNets):
$$
a^{\ell}:=\mathbf{g}\left(Z^{\ell}+\sum_{k=2}^{K} W^{\ell-k, \ell} \cdot a^{\ell-k}\right).
$$

\begin{figure}[ht!]
\centering
\includegraphics[width=.95\linewidth]{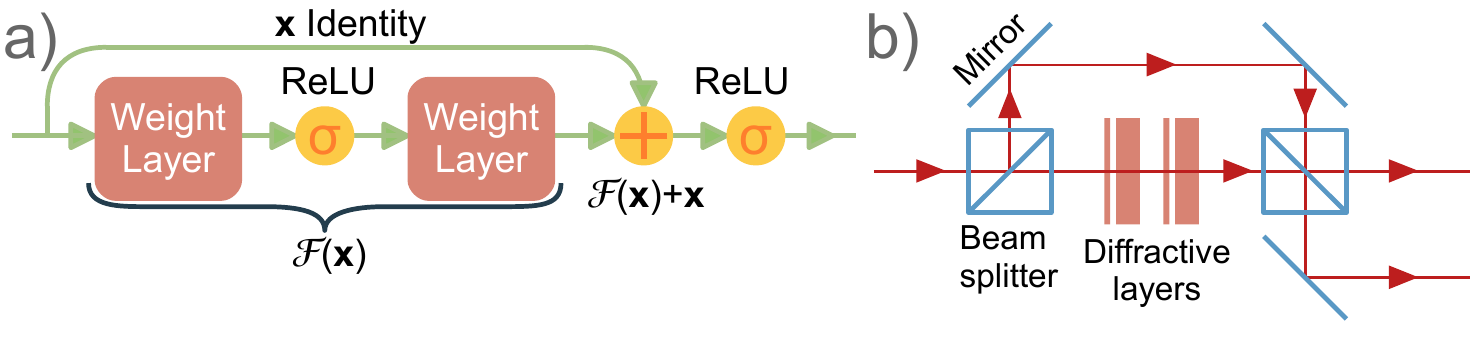}
\caption{a) Schematics of a residual neural network and b) corresponding optical implementation. Redrawn from \cite{dou2020residual}.}
\label{fig: ResNET schematics and experimental}
\end{figure}

The all-optical realization of the ResNet structure can be implemented using the scheme shown in Fig. \ref{fig: ResNET schematics and experimental}. In diffractive optical neural networks, the skip layer  can be easily realized by shortcutting the part of diffractive layers by using semi-transparent mirrors or beam splitters. Both the shortcutted beam and the signal propagated through additional diffractive layers can be spatially combined by using another beam splitter, as shown in the right-hand side of Fig. \ref{fig: ResNET schematics and experimental}.

It is pertinent to comment here on why the residual structures are needed: a very good and comprehensive example on the subject is given in Ref.~\cite{gin2021deep}. The authors of that Ref. consider the seemingly elementary problem of representing the identity function, $f(x)=x$, via a small seven-parameter NN with a one-node input layer, a two-node hidden layer with a ReLU activation, and a one-node linear output layer (see Fig. 2 of the aforementioned Ref.). When training this NN to approximate the identity function, but taking the data from the $[-1,1]$ region, the authors observed that the NN representation of the identity function diverged outside the training domain. Further (see Fig. 3 of the aforementioned Ref.), if the NN is trained repeatedly with different random initialization of the parameters, different results were observed: in some cases, the training loss plateaued and the NN failed to
accurately fit the training data at all. This observation was attributed to the tendency of deep and narrow ReLU
networks to collapse to the mean value of the function. The following conclusions were drawn in Ref.~\cite{gin2021deep}: i) it is yet another manifestation of the fact that the NNs cannot be relied upon to extrapolate outside the training domain; ii) even though it is possible to represent the identity function with this non-linear NN, it is non-trivial for the training algorithm to fit the data. Therefore, it is recommended that we use the residual NNs to avoid the aforementioned problem. 

\subsection{Radial basis function neural network} 

\begin{figure}[ht!]
\centering
\includegraphics[width=.75\linewidth]{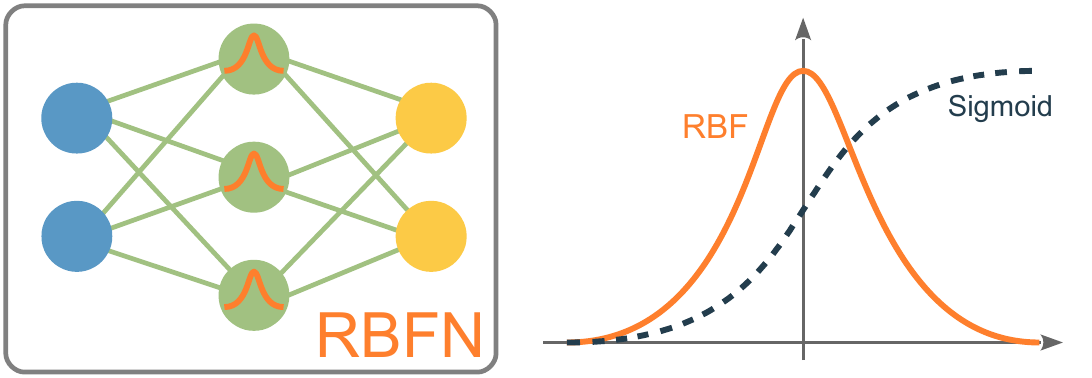}
\caption{Schematics of a radial basis neural network and RBF neuron.}
\label{fig: RBFN schematics}
\end{figure}





A radial basis function (RBF) network is an artificial NN that uses the RBFs as activation functions. Its schematics and a comparison of RBF function and sigmoid function are given in Fig. \ref{fig: RBFN schematics}. The network output is a linear combination of input RBFs and neuron parameters. The concept itself was introduced by Broomhead and Lowe in 1988 \cite{broomhead1988radial}. There are numerous applications for RBF networks, including function approximation, time series prediction, classification, and system control. Even though the RBF concept is considerably old and familiar, and, often, the other NN types are preferred nowadays, it still attracts the attention of data scientists~\cite{que2016back}.

The RBF networks typically have three layers: an input layer, a hidden layer with a non-linear RBF activation function and a linear output layer \cite{beheim2004new}. The input can be modeled as a vector of real numbers $\mathbf{x} \in \mathbb{R}^{n}$. The output of the network is then a scalar function of the input vector, $\varphi: \mathbb{R}^{n} \rightarrow \mathbb{R}$, and is given by:
$$
\varphi(\mathbf{x})=\sum_{i=1}^{N} a_{i} \rho\left(\left\|\mathbf{x}-\mathbf{c}_{i}\right\|\right),
$$
where $N$ is the number of neurons in the hidden layer, $\mathbf{c}_{i}$ is the center vector for neuron $i$, and $a_{i}$ is the weight of neuron $i$ in the linear output neuron. Functions that depend only on the distance from a center vector are radially symmetric about that vector, hence the name radial basis function. In the basic form, all inputs are connected to each hidden neuron. The norm is typically taken to be the Euclidean distance (although the Mahalanobis distance \cite{de2000mahalanobis} appears to perform better with pattern recognition) and the radial basis function is commonly taken to be a Gaussian function:
$$
\rho\left(\left\|\mathbf{x}-\mathbf{c}_{i}\right\|\right)=\exp \left[-\beta_{i}\left\|\mathbf{x}-\mathbf{c}_{i}\right\|^{2}\right].
$$
The Gaussian basis functions are local to the center vector in the sense that
$$
\lim _{\|x\| \rightarrow \infty} \rho\left(\left\|\mathbf{x}-\mathbf{c}_{i}\right\|\right)=0,
$$
i.e., changing the parameters of one neuron has only a small effect on input values that are far away from the center of that neuron.

The RBF networks are the universal approximators on a compact subset of $R^n$ under certain modest restrictions regarding the activation function shape. This implies that an RBF network with sufficient hidden neurons can approximate any continuous function on a closed, constrained set with arbitrary accuracy \cite{park1991universal}.

In addition to the unnormalized architecture mentioned, the RBF networks can be normalized. In this case, the mapping is
$$
\varphi(\mathbf{x}) \stackrel{\text { def }}{=} \frac{\sum_{i=1}^{N} a_{i} \rho\left(\left\|\mathbf{x}-\mathbf{c}_{i}\right\|\right)}{\sum_{i=1}^{N} \rho\left(\left\|\mathbf{x}-\mathbf{c}_{i}\right\|\right)}=\sum_{i=1}^{N} a_{i} u\left(\left\|\mathbf{x}-\mathbf{c}_{i}\right\|\right),
$$
where
$$
u\left(\left\|\mathbf{x}-\mathbf{c}_{i}\right\|\right) \stackrel{\text { def }}{=} \frac{\rho\left(\left\|\mathbf{x}-\mathbf{c}_{i}\right\|\right)}{\sum_{j=1}^{N} \rho\left(\left\|\mathbf{x}-\mathbf{c}_{j}\right\|\right)}
$$
is known as a normalized radial basis function.

Here, we outline the primary reason why the RBFs application failed to gain traction. RBFs are fundamentally flawed because they are a) too nonlinear, b) do not perform dimension reduction, and c) RBFs were always trained using k-means as opposed to gradient descent. In contrast, the deep NN have their nonlinearity under control, are able to reduce their dimensionality proportionately, and are learning by means of gradient descent. In spite of the fact that one might make the RBF's covariance matrix adaptive and hence achieve dimensionality reduction, this makes it even more challenging to train the RBF networks.


\subsection{Autoencoders} \label{subsec:auto}

Autoencoders are the NN architectures that are trained to reconstruct their inputs. A basic architecture of an autoencoder is shown in Fig.~\ref{fig:autoencoder}. In a more formal description, consider that the data $X$ is encoded by $\phi$ to a latent representation $Z$, which is passed through a ``bottleneck''. In this sense, this first part can be summarized by an encoder NN $g_{\phi}(.)$.  The bottleneck output is decoded terminating with an output layer with the same dimensionality as the encoder's input layer ($\hat{X}$), a reconstruction of $X$. Here, the decoder part can be described as $f_{\theta}(.)$. It is important to highlight that without the bottleneck, the encoder, and decoder would copy their input to the output, and by having a bottleneck, the encoder compresses the data to a latent representation that is more robust. The bottleneck appears in many autoencoder variations\cite{Georg2021book, bank2020autoencoders, Sensor05}. Regarding the training, both encoder and decoder NNs are simultaneously trained by minimizing a reconstruction loss function. The loss function will depend on the nature of $X$ and the task at hand. If $X$ has $n$ features with continuous values, this problem can be understood as a regression problem and one of the possible loss functions is the MSE:
\begin{figure}[ht!]
    \centering
    \includegraphics[width=0.55\linewidth]{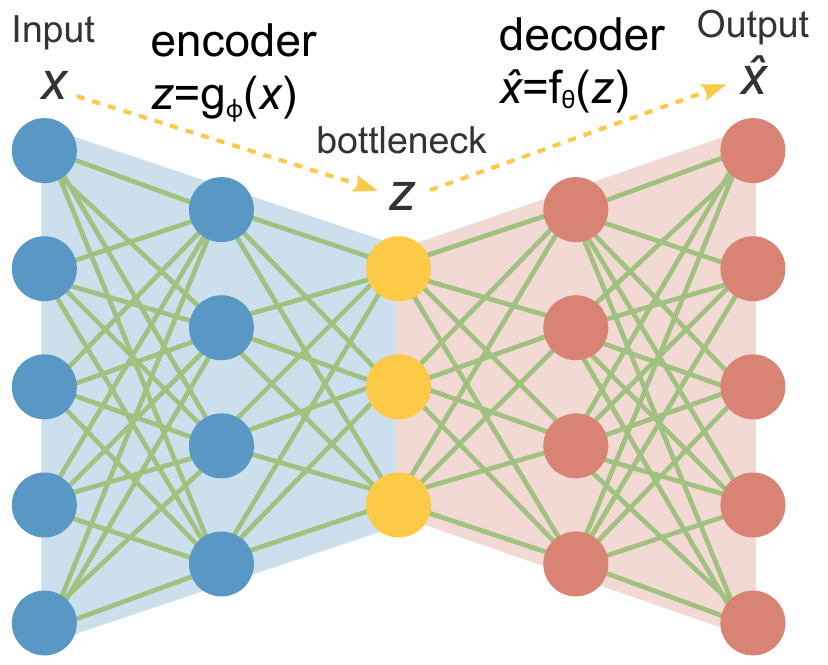}
    \caption{Schematic of autoencoders.}
    \label{fig:autoencoder}
\end{figure}
\begin{equation}
    L_{MSE}(x,\hat{x}) = \frac{1}{n}\sum_{i=1}^{n}\left[(x_i - f_{\theta}(g_{\phi}(x_i)))^2\right].
\end{equation}
However, if the data $X$ is discrete, two possible tasks can be performed: multi-class classification or multi-label classification. In the first case, $X$ is categorical in nature and is described by a one-hot-vector with $n$ possible classes in which just one of those features is equal to one at a time. For this case, the decoder's output needs to have a softmax activation function and the loss function can be the categorical cross-entropy loss: 
\begin{equation}
    L_{CCEL}(x,\hat{x}) = -\frac{1}{n}\sum_{i=1}^{n}\left[x_i\log(f_{\theta}(g_{\phi}(x_i)))\right].
\end{equation}

In the second case, $X$ also has $n$ features, but this time multiple features can be assigned to one. This case is known as the multi-label classification, and the decoder needs to have a sigmoid activation function; the loss function to use can be the binary cross-entropy loss: 
\begin{equation}
    L_{BCEL}(x,\hat{x}) = -\frac{1}{n}\sum_{i=1}^{n}\left[x_i\log(f_{\theta}(g_{\phi}(x_i))) + (1-x_i)\log(1-f_{\theta}(g_{\phi}(x_i))) \right].
\end{equation}
Moreover, the classical applications of autoencoders are dimensionality reduction (acting in the same way as a principal component analysis), data denoising, data generation, anomaly detection, and clustering. In the next section, we will show a few more examples of how this architecture is used in photonics.

Variational Autoencoders (VAEs) are a type of autoencoder that adds a probabilistic spin to the model \cite{doersch2016tutorial}. Unlike traditional autoencoders that learn deterministic functions for encoding and decoding, VAEs learn probability distributions for both. They encode the input data into a mean and variance of a probability distribution, from which a sample is drawn and then decoded to generate the output. VAEs employ a unique training strategy called the "reparameterization trick" \cite{kingma2013auto}, which allows for the optimization of the model using standard backpropagation. This additional complexity of modeling the input data as probability distributions, however, tends to increase the computational complexity of training, relative to standard autoencoders. VAE has been applied for predictive control and hidden parameters' retrieval \cite{baumeister2018deep}. There is also a photonic realization of VAE for high-throughput and low-latency image transmission \cite{chen2023photonic}.

Adversarial Autoencoders (AAEs) are another variant of autoencoders, and they leverage the power of GANs for training \cite{makhzani2015adversarial}. They consist of an autoencoder paired with a discriminator network that is trained to differentiate between the encoded representations of real and generated data. The autoencoder is trained to fool the discriminator, thereby encouraging it to produce encoded representations that resemble the distribution of real data. The training of AAEs is more complex than that of standard autoencoders and VAEs, as it involves a min-max game between the autoencoder and the discriminator, which makes it computationally more intensive. AAEs are used for machine-learning-assisted metasurface design \cite{kudyshev2020machine}, global optimization of photonic devices \cite{kudyshev2020machine2} and hyperspectral anomaly detection \cite{creswell2018denoising, xie2020autoencoder}.

Finally, we note that the serious challenge when dealing with the autoencoders is to understand the variables that are relevant to the problem under investigation. In this sense, it is important to highlight that the decoder part is never perfect and is highly dependent on the bottleneck used, since we can miss out on important dimensions of the problem.

\subsection{Generative Adversarial Network}\label{subsec:gan}

The idea behind the generative adversarial networks (GAN) was based on the concept of zero-sum game theory \cite{goodfellow2020generative}. 

\begin{figure}[ht!]
    \centering
    \includegraphics[width=0.9\linewidth]{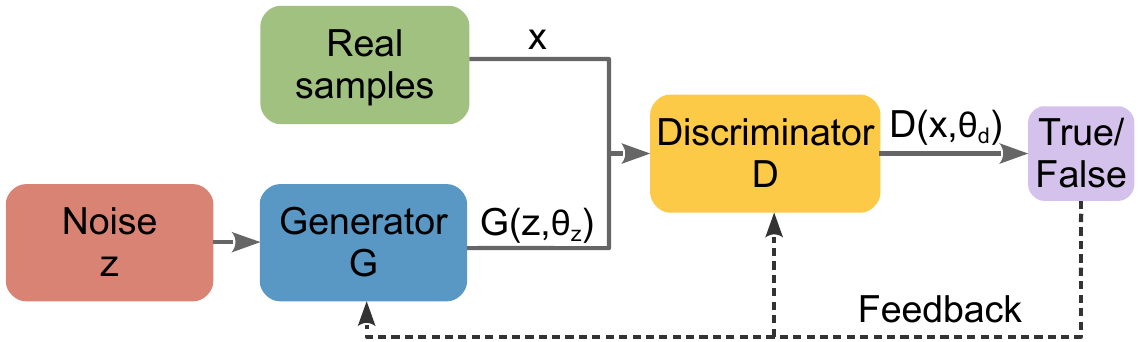}
    \caption{Typical schematic of a GAN structure.}
    \label{fig:GANs}
\end{figure}

As shown in Fig. \ref{fig:GANs}, the framework of GAN consists of two neural network models: a generative model called a generator that captures the data distribution, and a discriminative model that distinguishes whether a sample came from the real dataset or from a generated (``fake'') one. In a nutshell, the generator aims to learn the distribution of real data,  while the discriminator aims to correctly determine whether the input data is from the real data or from the generator.  In order to win the game,  the two participants need to continuously optimize themselves to improve the generation ability and the discrimination ability, respectively. Therefore, during the training procedure, the two models compete with each other. The generator is designed to generate data as realistically as possible so that it is difficult to distinguish them from the truth, while the discriminator as a binary classifier aims to identify real and fake data as accurately as possible. The generator and discriminator are optimized alternately until the augmented data is indistinguishable from the actual data
\cite{wang2017generative, gui2021review,wang2021artificial}. In other words, drawing upon the principles of game theory, GANs utilize a minimax strategy, where the generator and discriminator networks strive to optimize their respective objectives, resulting in a dynamic equilibrium. This competitive nature of GAN training resembles a zero-sum game, in which the gains made by one network come at the expense of the other. However, it is important to note that GANs, in their formal definition, may not strictly adhere to the conditions of a zero-sum game. In a zero-sum game, the total utility or payoff remains constant, and any gain by one player directly corresponds to a loss for the other player. In contrast, the training process of GANs does not necessarily maintain a fixed overall utility.

In a more mathematical formulation, to learn the generator distribution $p_g$ over data $x$ with distribution $p_x(x)$, a prior to input noise variables is defined as $P_z(z)$, where $z$ is the noise variable. Then, the  generator  represents a  mapping  from  noise  space  to data space as $G(z,\theta_g)$, where $G$ is a differentiable function represented  by  a NN  with  parameters $\theta_g$.  The other NN, $D(x,\theta_d)$, is also defined with parameters $\theta_d$,  but  the  output  of $D(x)$ is a  single  scalar. $D(x)$ denotes  the  probability  that $x$ comes  from  the  data  rather than from the generator $G$. The discriminator $D$ is trained to  maximize  the  probability  of  assigning  a  correct  label  to both  real  training  data  and  fake  examples  generated  by the generator $G$. Simultaneously, $G$ is trained to minimize $\log(1- D(G(z)))$. Therefore, the optimization of a GAN can be formulated as a minimax problem:
\begin{equation}
    \min_G \max_D \{E_{x}[\log D(x)]+E_{z}[\log(1- D(G(z)))]\},
\end{equation}
where $E[\ldots]$ represents  the  expectation value.
 
In practice, the training for such structures is inherently unstable, such that an alternative training  method  is used.  In a nutshell, this alternative training occurs in two stages:  freeze the $\theta_g$ parameters  and  optimize $\theta_d$ to  maximize  the  discrimination  accuracy  of $D$; froze the $\theta_d$ parameters  and  optimize $\theta_g$ to minimize the discrimination accuracy of $D$.  This  process  alternates  and  we  could  achieve  the  global optimal  solution  if  and  only  if $p_x=p_g$. 

Finally, it is important to highlight some of the best practices when using this type of structure\cite{radford2016unsupervised}:
\begin{itemize}
    \item Scale properly the real data $x$ and the Generator output $G(z,\theta_g)$. A problem at this step can cause sample oscillation and model instability. So, it is recommended to avoid applying batchnorm to the generator output layer and the discriminator input layer.
    \item The data fed into this merged model can either be a mix of real and fake data (from the generator), or it can be purely real and purely fake. The latter is a better approach, since having the data separated into fake and real improved the GAN performance.
    \item It is recommended to use the leaky ReLU activation unit in all layers of the GAN except the output of the generator, where we should use tanh.
    \item In Ref.~\cite{radford2016unsupervised}, the authors initialize all weights using a zero-centered Gaussian distribution with a standard deviation of 0.02.
    \item Use techniques to stabilize training: There are several techniques that can be used to stabilize the training of GANs, such as using batch normalization, using a history of generated samples in the discriminator, or using a two-time scale update rule for the generator and discriminator.
    \item Use a stable optimizer: GANs can be sensitive to the choice of an optimizer. Using a stable optimizer such as Adam can help to improve the training process.
    \item Monitor the training process carefully: It is important to monitor the training process carefully and track metrics such as the generator and discriminator loss. This can help to identify issues such as mode collapse, where the generator generates only a few types of samples, or the discriminator becomes too strong, and the generator is unable to improve.
\end{itemize}
As an example, in optical applications, GANs have been used for the end-to-end model for geometric constellation shaping applicable for any nonlinearity-limited optical communication channel\cite{cohen2021generative}.

\section{How to Choose your NN Architecture: The Hyperparameter Search}\label{sec:choose}

One of the most important steps in the NNs is the design of the NN architecture. Indeed, the hyperparameters of the NN model (e.g., number of layers, number of neurons, type of activation function, learning rate, etc.) affect the speed and accuracy of the learning process of the NN models and ultimately define its functioning.  However, due to the lack of analytical approaches to calculating such hyperparameters, only a limited number of options (e.g., exhaustive and random search) have been typically used. In this section, we will describe one of the efficient techniques known as  Bayesian optimization (BO), explaining how it can help us to design our NN architecture with the aim to maximize NN's performance~\cite{nguyen2019bayesian,cho2020basic,wu2019hyperparameter}. We will also briefly introduce the concept of reinforcement learning for hyperparameter search, as it is gaining popularity in academic and industrial circles as a replacement for BO and other searching techniques.

\begin{figure}[ht!]
    \centering
    \includegraphics[width=0.9\linewidth]{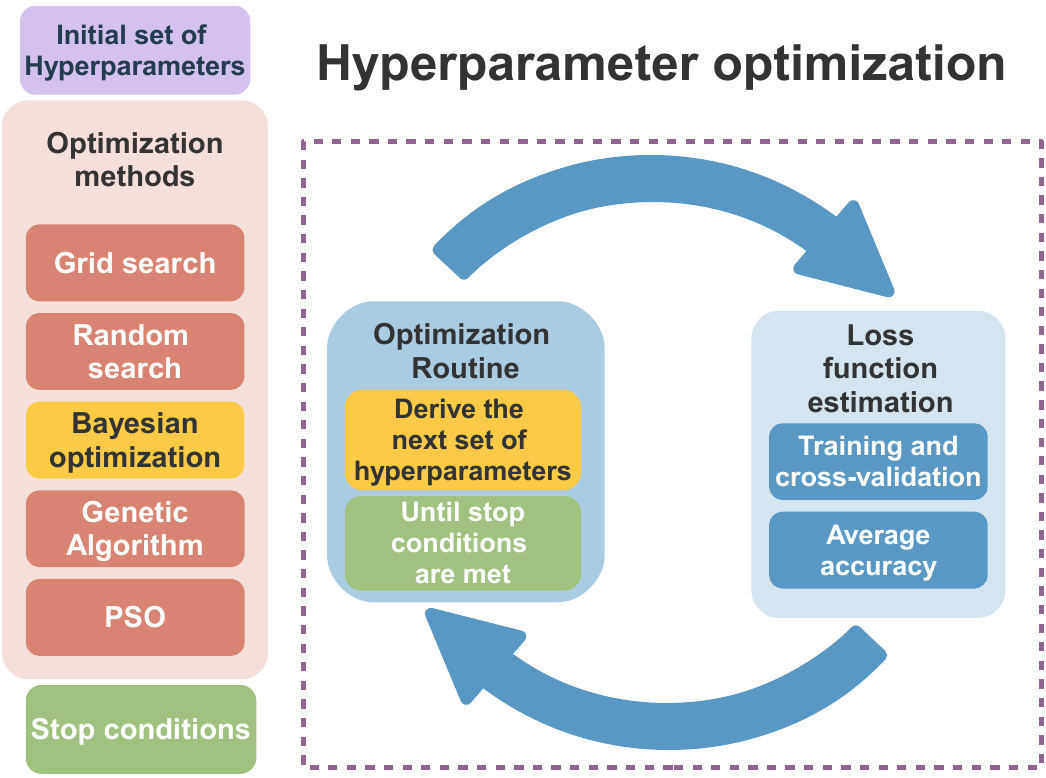}
    \caption{Hyperparameter tuning routine using optimization techniques}
    \label{fig:hyperparameter}
\end{figure}

\subsection{The problem of hyperparameter tuning}
Given a certain NN model that solves a problem under investigation, in which a given arbitrary input \(x\) yields a response \(y\), the
model accuracy can be evaluated through an objective function \(f\). A
hyperparameter set \(\theta\) fully represents the architecture of the NN, such that the objective function is described as \(f = f(\theta,s,r)\),
which for simplicity can be written as\(\ f = f\left( \theta \right)\). In order to estimate the
optimal model accuracy, \(f\) must be subject to an optimization process
with respect to \(\theta\). However, in most cases, this optimization of
\(f\) is bounded by two important restrictions; they are \cite{sena2021bayesian}:
\begin{quote}
1) \emph{Computational complexity} -- The number of evaluations
performed on \(f\) is limited, typically in the range of a few hundred. This condition frequently arises because each evaluation takes a substantial amount of time.

2) \emph{Non-differentiability} -- First- and second-order derivatives
of \(f\) with respect to \(\theta\), are not easy to obtain, thus, preventing the application
of methods like gradient descent, Newton's or quasi-Newton
methods.
\end{quote}

There are a few possible search methods that suppress some of these aforementioned restrictions: Grid search, Random search, Genetic algorithm, Particle Swarm Optimization, and the Bayesian optimization  (BO) \footnote{This list is not exhaustive, and new alternatives and methods' variants emerge constantly\cite{smithson2016neural,talbi2021automated,pinos2022evolutionary}.}. However, from our experience, the BO is the most promising among them because it needs relatively few evaluations of \(f\), it is a derivative-free method, and it is fairly robust to noisy objective function evaluations.

In summary, all the search methods for NN hyperparameter tuning have the same core, which we illustrate schematically in Fig.~\ref{fig:hyperparameter}. First, we define which hyperparameters we want to optimize (e.g., the number of filters), their initial values, which search method we will use (e.g., the BO) as the seed, and what is the search space for each hyperparameter (e.g., we wish that the number of filters ranges from 5 to 350, etc.). Next, using this hyperparameter set, we proceed to the training validation phase. To estimate the accuracy of the NN model efficiently, we can use the cross-validation method\footnote{If enough data is available, instead of cross-validation we can have independent datasets for training, validation, and testing as well.}, which divides the dataset into $k$ sections, training with $k-1$ sections, and testing with the remaining one to get the model accuracy. This process is repeated until all sections have been used for testing, and the average accuracy is calculated. This average accuracy is assigned to the set of hyperparameters, and this is the feedback to the search model that uses it, to suggest the next set of hyperparameters (or to decline the following iteration).  This search cycle ends when the whole space is searched in the case of the grid search, when a certain number of interactions were done in the case of random search, or when the model converged in the case of genetic algorithm/particle swarm optimization/BO. Finally, when the cycle is finished, the hyperparameters with the best average accuracy are taken as the ones that will be used to design the NN model.

Next, we will detail further how the BO algorithm functions, also pointing out its drawbacks.

\subsection{Bayesian optimization algorithm}

The BO algorithm is based on two core principles. First, it builds a basic
\emph{surrogate function} \(f^{*}\) to ``fit'' the objective \(f\) and
estimate its response to unknown entries \(\theta\). Second, it bypasses
the impossibility of using gradient descent methods on \(f\) by
introducing an \emph{acquisition function}, i.e., a statistical operator
that orients the optimum search.

\begin{figure}[ht!]
    \centering
    \includegraphics[width=0.75\linewidth]{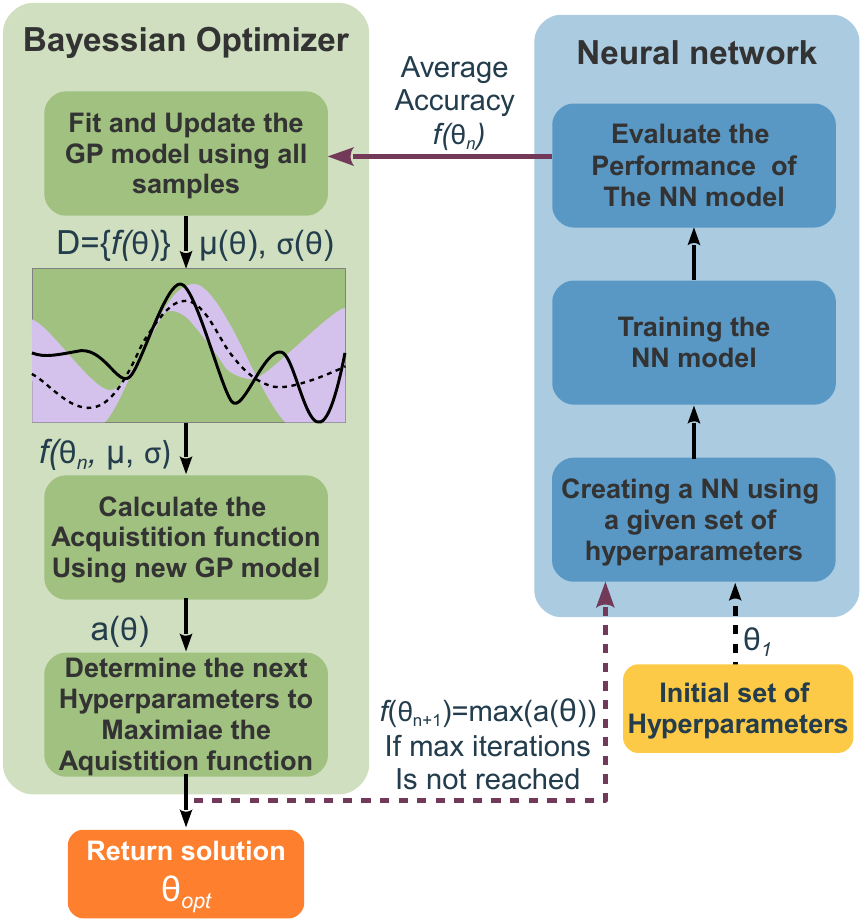}
    \caption{Bayesian optimization flowchart.}
    \label{fig:bo}
\end{figure}

Regarding the idea behind the surrogate function, it can be understood as a function
\(f^{*} = p\left( f \middle| \mathcal{D} \right)\) that estimates the
value of the objective function \(f\) for arbitrary \(\theta\), i.e.,
\(f(\theta)\), conditioned on a limited sub-set of \emph{n}-observed
data points
(\(\mathcal{D = \{}f(\theta_{1}),f(\theta_{2}),\ldots,{f(\theta}_{n})\)\}).
To build \(f^{*}\), the BO algorithm models \(p(f|\mathcal{D)}\) as a
Gaussian process (GP), which permits to represent the posterior distribution
\(p\left( f \middle| \mathcal{D} \right)\) by the normal distribution
\(\mathcal{N}(\mu,\sigma)\), with the mean value $\mu$ and dispersion $\sigma$.

Acquisition functions are crucial to the BO scheme: they are used to  choose the next vector of hyperparameters as the one which has the highest probability of improvement over the current state. In a nutshell,  the acquisition function \(a\) can be evaluated for any arbitrary hyperparameter input
\(\theta\), and it quantifies how promising the next sampling decision
\(\theta_{n + 1}\) is to indicate the location of the global optimum. By
maximizing the acquisition function, i.e., $\theta_{n + 1} = \max a(\theta)$
to select the next numerical evaluation \(f(\theta_{n + 1})\), we merely
substitute our initial optimization problem with another
optimization, but now with a cheaper function. A common choice for the
acquisition function is the expected improvement (EI), computed as\cite{sena2021bayesian}:
\begin{equation}
    a(\theta) = [\mu(\theta) -f^{+}]\phi(Z)+ \sigma(\theta)\phi(Z),
\end{equation}
where \(f^{+} = \mathrm{max}\mathcal{(D)}\) and
\(Z = \ \frac{\mu\left( \theta \right) - f^{+}}{\sigma\left( \theta \right)}\ \)
if \(\sigma\left( \theta \right) > 0\) or \(Z = \ 0\ \) if
\(\sigma\left( \theta \right) = 0\). The functions \(\Phi\) and \(\phi\)
correspond to the cumulative and probability density functions of the
standard normal distribution \(\mathcal{N}(0,1)\), respectively. Since
\(a\left( \theta \right)\) can be analytically expressed as a function
of \(\mu\left( \theta \right)\), \(\sigma\left( \theta \right)\) and
\(f^{+}\), which are directly obtained from the surrogate function
\(f^{*}\), the sampling point \(\theta_{n + 1}\) is easily found by
numerically evaluating \(a\left( \theta \right)\) for all
\(\theta\) in the searching space.

To summarize, the BO algorithm can be defined by the
scheme in Fig. \ref{fig:bo}. First, the set
\(\mathcal{\text{\ D}}\) is initialized by sampling \(f\) with an initial hyperparameter set \(\theta_1\). It should be noted that this sampling can be
performed either randomly when no previous information is known about
\(f\), or deterministically, when there is some indication about the
optimum of \(f\). $f$ is defined by the same training and evaluation phases described previously. Then, the BO is programmed to run until a
maximum number of iterations is reached.
For each \(i\)-th iteration loop, the \emph{surrogate} function
\(f^{*}\) is computed, i.e., \(\mu\) and \(\sigma^{2}\) are
calculated, and these are used to maximize an \emph{acquisition} function \(a\), which, provides a new sampling decision \(\theta_{n + i}\).
Finally, the sampling decision is evaluated
\(f\left( \theta_{n + i} \right)\) and incorporated to
\(\mathcal{D}\) before a new cycle starts. When this iterative
process ends, the hyperparameter \(\theta\) that yields the
maximum \(f(\theta)\) in \(\mathcal{D}\) is selected as the optimal
solution \(\theta_{\text{opt}}\).

Finally, it is important to highlight some drawbacks of the BO. The Bayesian optimization is restricted to problems of moderate dimension. This is a difficult problem: to ensure that a global optimum is found, we require good coverage of searching space of $\theta$, but as the dimensionality increases, the number of evaluations needed to cover searching space of $\theta$ increases exponentially\cite{shahriari2015taking}. In this sense, we recommend that the number of hyperparameters should be less than 20, even though other works in the literature show that the BO still produces some advantages depending on the problem tuning up to 76 parameters \cite{hutter2011sequential}.In this sense, unless cost function evaluation is rather costly and the dimensionality of the problem is somewhat small, BO will tend to produce the same performance as the random search \cite{joyce2018review}.

\subsection{Reinforcement Learning}

Reinforcement learning (RL) has emerged as a promising technique for optimizing the hyperparameters of NN structures in various domains. RL leverages an agent-environment interaction paradigm to learn an optimal policy that maximizes a cumulative reward signal.
The utilization of RL in finding the best hyperparameters of a NN structure involves the formulation of the problem as a Markov decision process (MDP). In this setting, the NN structure is considered the agent, while the selection of hyperparameters constitutes the action space. The environment provides feedback to the agent in the form of rewards, typically based on performance metrics such as accuracy, loss, or other domain-specific objectives. One popular approach is using deep Q-networks (DQNs), which combine deep learning with Q-learning. In this domain, Ref.~\cite{baker2016designing} is a seminal paper, which presents a meta-modeling approach utilizing reinforcement learning to generate customized CNN designs for various image classification tasks. In this approach, a common set of hyperparameters is employed to train all network topologies during the Q-learning phase. Subsequently, the hyperparameters are fine-tuned for the top models selected by the Meta \egor{Q-agent}. Another example can be found in Ref.~\cite{9420739}, where they have built upon the aforementioned work by employing Q-learning to define learning agents per layer. 

\begin{figure}[ht!]
    \centering
    \includegraphics[width=0.9\linewidth]{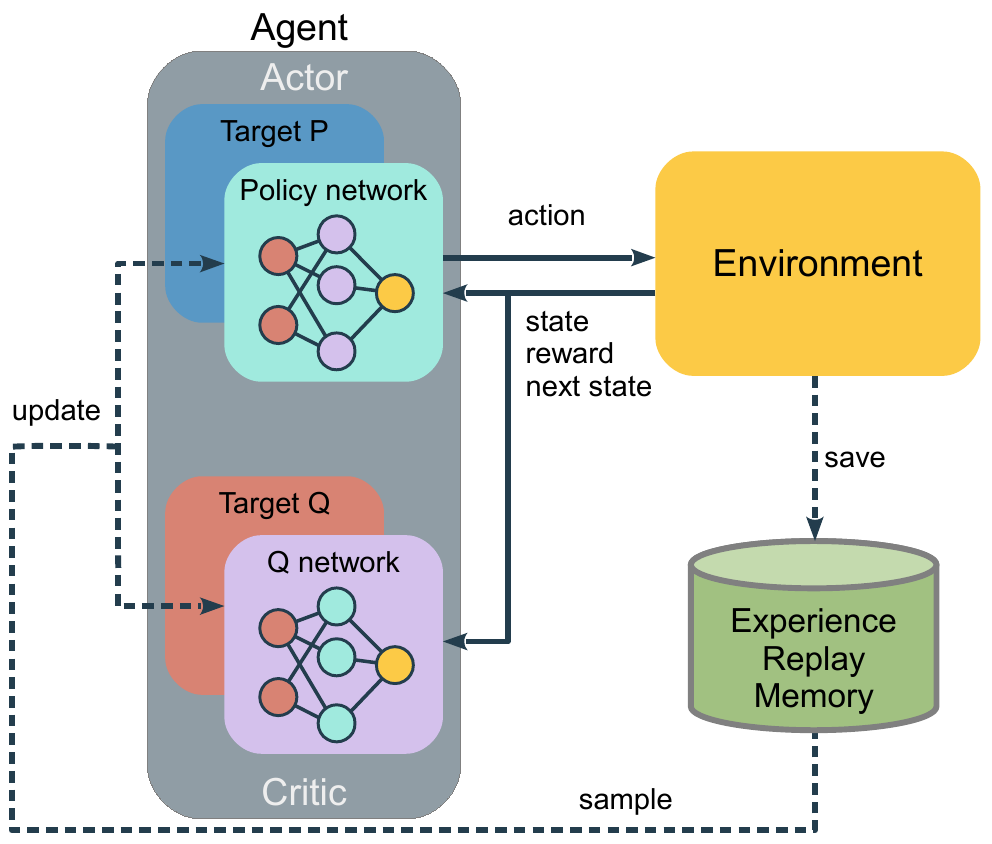}
    \caption{Reinforcement learning framework for Volterra nonlinear equalizer. Schematics reproduced from ~\cite{xu2022automatic}.}
    \label{fig: reinforcement learning}
\end{figure}

This approach partitions the design space into independent, smaller design subspaces, wherein each agent fine-tunes the hyperparameters of the assigned layer based on a global reward. This methodology aims to expedite the design space search while maintaining accuracy. Finally, moving to the optics field,  Ref.~\cite{xu2022automatic} proposed a similar application of reinforcement learning. However, instead of optimizing the hyperparameters of a specific neural network architecture, the study employs reinforcement learning to design an optimum Volterra nonlinear equalizer. The schematic of the architecture used is shown in Fig.~\ref{fig: reinforcement learning}. This approach utilizes deep deterministic policy gradient (DDPG) agents to interact with the environment (the equalizer) and learn an effective search policy. The DDPG agent's output actions represent the structural parameters of the Volterra equalizer, including memory length for each order, feedback memory length, and pruning rate. The reward from the environment is defined as a function of the bit error rate (BER) after equalization and the complexity of the equalizer.

\section{Applications of neural networks in different photonics areas}\label{sec:optcom}
\begin{figure}[ht!]
    \centering
    \includegraphics[width=\linewidth]{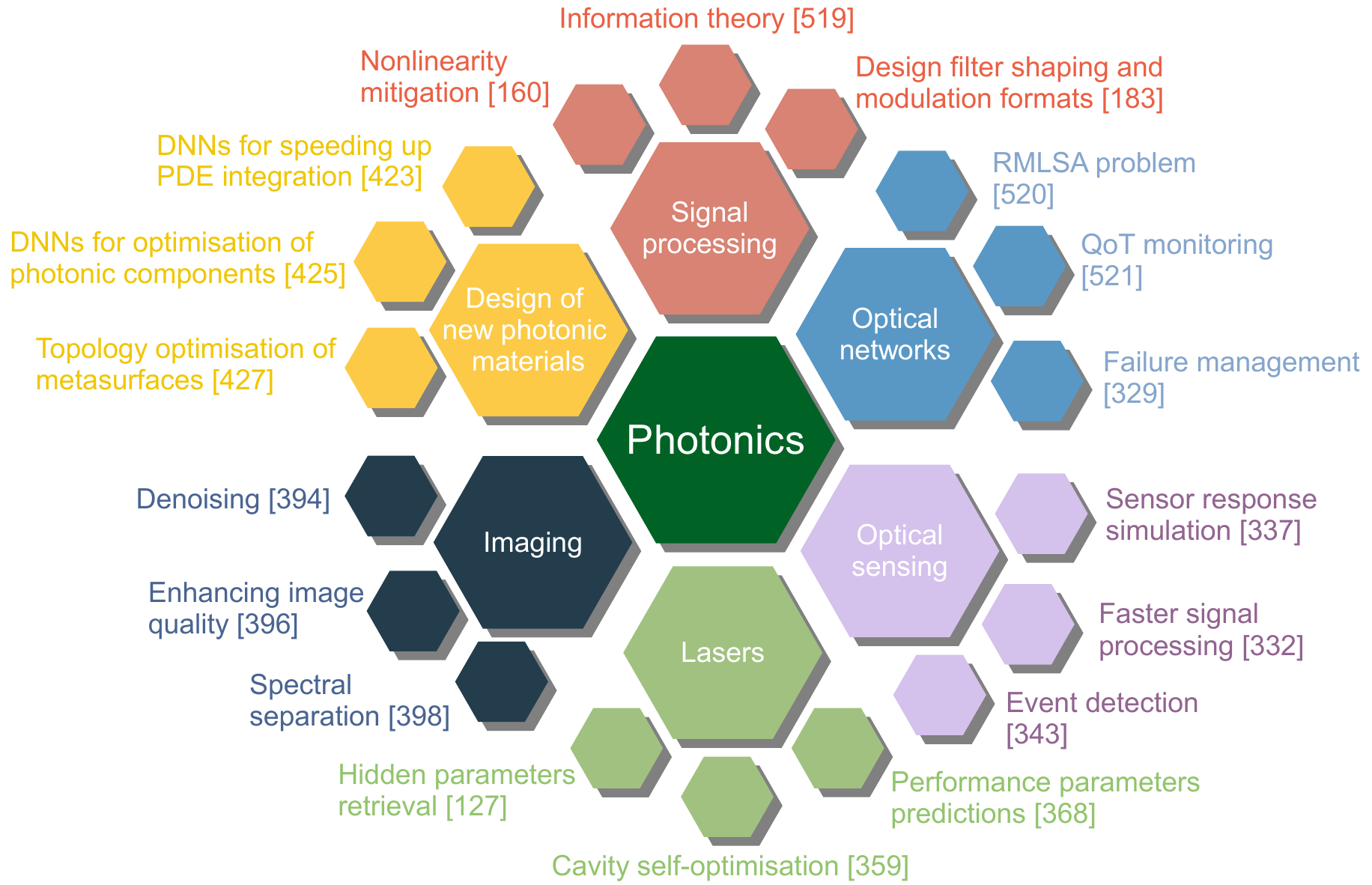}
    \caption{Graphical depiction of the NN applications discussed in this tutorial. \egor{We have updated the references of this figure after removing the duplicated references}}
    \label{fig: ML in photonics}
\end{figure}

In this section, we discuss some photonic applications of NN structures introduced above. 
We would like to reiterate that we do not aim here to present a comprehensive overview of all numerous important applications of artificial NNs in photonics, as illustrated by Fig.~\ref{fig: ML in photonics} (many of the missing areas can be found in recent review papers  \cite{R01,R02,R03Darko,R04,R05,R06,R07,R08,R09,R10,R11,NNMaterial01,R13,R14}). Instead, we often use several specific examples to illustrate how the NNs are used in these fields. 
Where appropriate, we try to identify when the complexity of NNs can be an issue in these particular applications and stress the key point of our work -- the reduction of the complexity of NNs used in photonics. Though we will use optical communications as an example for the illustration of the complexity reduction methods, we point out that the majority of our results can be applied in various other areas.

\subsection{Optical communications: channel modeling}\label{subsec:modeling}
First, we address one of the NN applications in optical communications, which is acquiring more and more popularity today: the use of different NN structures for the simulations of signal propagation down the fiber-optic channel. Of course, most of the simulations related to the fiber systems analysis are still carried out using the well-elaborated, efficient, and accurate split-step Fourier method (SSFM)\cite{agrawal21}. A more advanced version of this algorithm, with a  memory filter within the nonlinear step,  is proposed in  Ref.~\cite{rafique2011compensation}\footnote{The filtered version typically provides better performance at the low spatial resolution, with an almost negligible computational complexity increase.}, see also \cite{napoli2014reduced}. Nonetheless, sometimes the simulation of the optical fiber system's functioning is a bottleneck, consuming too much time. We also note that, at present,  the new (say NN-based) techniques for signal propagation modeling in fiber-optic systems are typically compared with an ``ordinary'' SSFM without memory filters. Thus, perhaps, it is too early to state that the NN methods are superior to the ``traditional'' approach, and more work in this direction is still required to make a fair comparison.  

When dealing with the NN-based optical channel modeling, we naturally aim at obtaining a good quality result (the output signal that has passed through the communication system and experienced the respective distortions) at a lower ``complexity'' cost, i.e., the NN can simply render the desired result faster. The latter becomes a significant bottleneck in, e.g., the simulations of a wideband signal propagation\cite{musetti2018accuracy,serena2019numerical}, or at high powers, when the spatial step of the SSFM has to be very small to guarantee a satisfactory modeling accuracy\cite{jaworski2008step,schmauss2012recent}. However,  the existing NN-based modeling propositions do not address the ultra-wide-band systems, mostly because of the novelty of the subject itself and, perhaps, because the data collection for wideband/extra-high powers is a truly time-consuming process. Thus, the advanced NNs application for the wideband simulations can be rated as an interesting and practically important open problem.

\begin{figure}[ht!]
    \centering
    \includegraphics[width=0.6\linewidth]{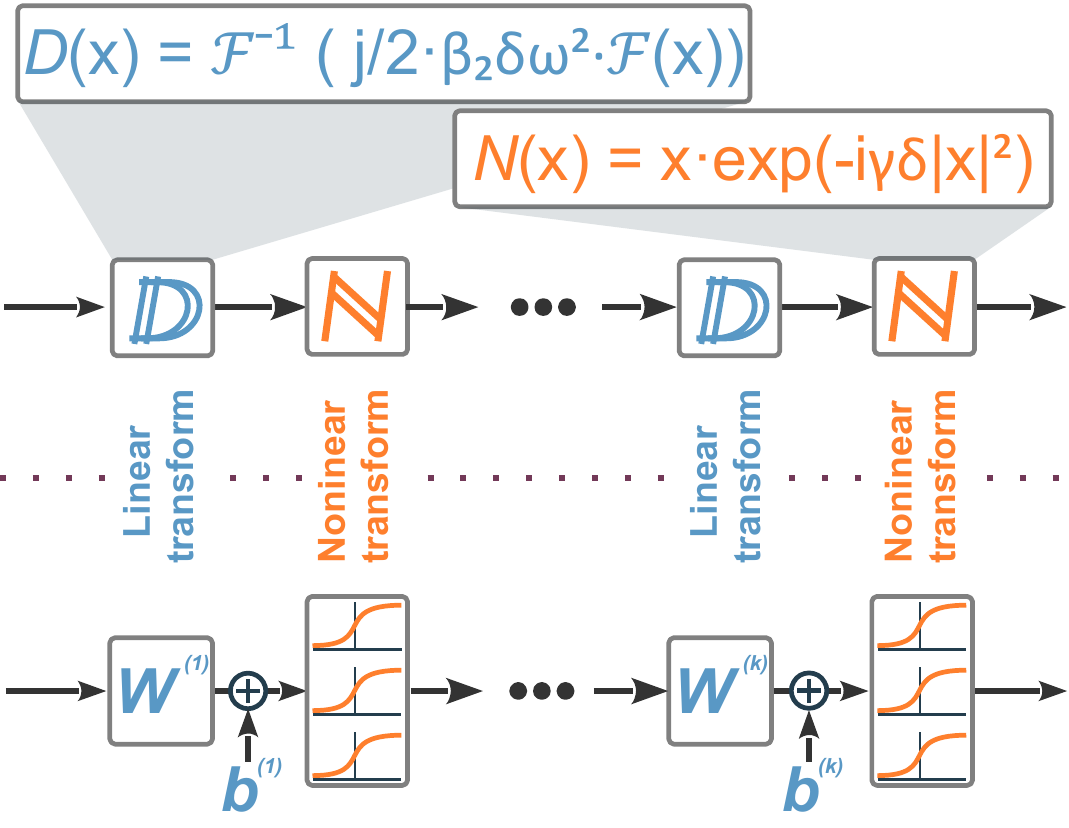}
    \caption{Correspondence between linear dispersion $D$ and non-linearity $N$ operators, used within the SSFM to approximate the light evolution down the fiber, and the linear (vector-matrix convolution) and non-linear (activation) transformation steps in a feed-forward NN structure.}
    \label{fig: NLE and NN correspondence}
\end{figure}

The first and somewhat obvious replacement of the channel function for our modeling task is, as mentioned above, to recast SSFM as a learnable NN-type framework\cite{hager2018nonlinear}. The linear SSFM step, i.e., the Fourier transform, is a vector-matrix convolution, while the nonlinear step amounts to the use of the operator $N$, see Fig.~\ref{fig: NLE and NN correspondence}, applied to the result of the preceding convolution. Now, if we allow the elements of the matrices (the linear step) to be optimized by some training procedure, we arrive at the NN-type structure, where the network parameters can be optimized with some standard procedure used in NN training.
The SSFM sequence of operations is virtually identical to the NN functioning, but
with the important differences: i) for the NN-type implementation, the weights are now considered as the parameters to optimize through training, and ii) the nonlinear activation functions are now rendered by the mathematics behind the approximation approach, but not taken from some ``standard'' deep NN set (say ReLU, etc.). We notice that ii) can potentially be a source of problems along the NN training, as the ``non-standard'' activation functions often result in the exploding/vanishing gradients' problem, such that for employing this method, we typically need to initialize the trainable weights using the (assumed known) fiber propagation parameters participating in the SSFM. However, the neural approximation of the SSFM has so far been used for the so-called learned digital back-propagation concept\cite{hager2020physics,ZHANG202243}, i.e., for the channel equalization purpose, and has not been tested specifically for the channel modeling. This approach can be ascribed to the so-called model- or physics-driven approaches, meaning that we utilize the mathematical model of the channel (in the form of its SSFM approximation) and recast it as an NN structure with the trainable weights.

Another bevy of methods that can be used for modeling the signal propagation down the fiber refers to the so-called physics-informed NN (PINN) concept\cite{NNPDE03,karniadakis2021physics}; see also the comprehensive review of the method's applications with some analysis in Ref.~\cite{cuomo2022scientific}. PINNs are the NN structures encoding the problem governing equation, i.e., the Manakov equation for the fiber-optic transmission systems, as a part of the NN, or, in other words, the scheme that adopts the physical laws of the true channel model to parameterize its solution via the NN. PINNs approximate the equation solutions by training an NN to minimize a specific loss function that includes terms corresponding to the initial/boundary conditions and the equation's residual at selected points in the space-time domain (called collocation point). PINNs, given an input point in the integration domain, produce an estimated solution in that point of a differential equation after training. The PINN concept's application to channel modeling was studied by two groups\cite{jiang2021solving,zang2021principle}. Both studies agree that the PINN can be used to accurately model pulse evolution down the fibers with less complexity as compared to the SSFM-based modeling, also underlining the universality of the approach. An important observation is that while the SSFM step has to be reduced when we simulate the high-power signals, the PINN method is insensitive to that and, thus, utilizes the same complexity as we have for low-power signals. A tutorial on the application of PINNs in optical communications can be found in \cite{wang2022applications}.

An interesting approach for the NN-based channel transfer function modeling, introduced recently, uses GANs\cite{yang2020fast}. The authors of the above Ref. claim that the GAN-based method can indeed learn the accurate transfer function of the fiber channel well, and the approach can be extended to model the signal propagation to arbitrary distances. Importantly, the GANs show noticeable generalization capabilities, such that we can model the propagation with different optical launch powers, signal modulation formats, and input
signal distributions. Comparing the complexity of GAN-based method to the SSFM modeling, the total multiplication number for the GAN modeling was found to be around 2\% compared to that of a ``standard'' SSFM, which means a considerable reduction in the simulations' complexity. 

One more recently introduced approach for channel modeling is based on the concept of Fourier Neural Operator (FNO)\cite{li2020fourier}. The latter method belongs to the supervised operator learning methods family\cite{lu2021learning}, a machine learning framework proven to be efficient in modeling the evolution of spatiotemporal dynamical systems and approximating
general black-box relationships between functional
data\cite{wang2021learning}. The feature of the FSO, which makes it attractive for channel modeling, is that the FSO is mesh-independent, which is similar to the PINN but different from the standard deep learning methods such as CNN-type SSMF mentioned above. Thus, the FNO network can be trained on one mesh and evaluated on another: by parameterizing the model in function space, it learns the continuous transfer function instead of discretized vectors, which is, of course, a highly desirable ingredient for the optical channel modeling, where we operate with randomly-modulated signals with different characteristics. The test regarding the FSO utilization for the channel modeling was carried out in Ref.~\cite{he2022fourier}. It was shown that the effective SNRs differences between the
proposed FNO and SSFM are all within 1~dB at 1200~km of the range of launch powers, and, importantly, the results rendered by the FSO modeling are also close to that obtained in the experiment. At the same time, the authors report the improvement in the complexity against standard SSFM simulations. It is interesting to note that the FNO method implies the increase in the dimensionality for the internal NN representation of the signal's evolution, such that the whole structure for the NN channel looks like a so-called over-complete autoencoder. Notably, it is opposite to another interesting method used for the nonlinear Schrödinger equation modeling\cite{gin2021deep}, which is quite similar to the optical channel modeling task: in the latter method, the parsimony principle is used, such that the channel model becomes a traditional autoencoder. This demonstrates that the universal recommendation on which particular modeling method we should use cannot be made at the moment. Perhaps, the only feature that the authors of the aforementioned works require from their NN optical channel analogs is that the resulting structure's inference takes fewer operations than the modeling with SSFM. It would be, then, interesting to compare the existing approaches in different conditions and describe each method's benefits and shortcomings.

Yet another method for channel modeling refers to the use of transformers\cite{zhang2022transformer}: in that Ref. the authors specifically addressed the case of orthogonal frequency division multiplexing (OFDM) transmission. It was noted that the model-driven approaches could suffer in balancing accuracy and efficiency, especially for complex and long-haul transmission. The authors proposed a simplified transformer, combining it with a feature-decoupled distributed scheme for fast and accurate fiber channel modeling.  The decoder part of the transformer was removed, and the self-attention was dropped out, as the latter contributes significantly to the inference complexity.  The modeling performance was investigated, taking into account the generalization ability, while the method demonstrated the high precision and robustness of the model. Furthermore, the modeling was studied for different transmission rates and was proven reliable over a wide bandwidth. Compared to the bidirectional LSTM (Bi-LSTM), the transformer performed better in accuracy and had lower computational and memory costs. For models under the same conditions, the required running time of the transformer was about 60\% of Bi-LSTM, and less than 1\% of that corresponding to the SSMF in the same scenario.

Finally, we turn to a very recent modeling approach, where the complexity reduction problem is posed in general but does not refer to the multiplications' reduction compared to the SSFM technique: OptiDistillNet \cite{gautam2022optidistillnet}. The approach uses a deep CNN to solve the nonlinear Schrödinger equation. Then, the so-called knowledge distillation (KD) based framework for compressing a CNN is considered, which involves the original complex model as a teacher, the knowledge of which is used to train the reduced model called a student. By using the latter, we gain faster modeling, whereas the quality of modeling is very close to that of the original model. 

Overall, as we see, while the modeling of the optical channel with the use of NN is gaining more and more attention, and we already have a plethora of different methods, the structuration of approaches and their face-to-face comparison is yet to be done; at the moment, it is difficult to distinguish a particular most promising direction in the simulations of signal propagation down the fiber. We also note that it would be interesting to investigate the existing approaches for the sake of their incorporation into the end-to-end impairments' mitigation framework, Subsec.~\ref{subsubsec:e2e}.

\subsection{Optical Communications: Signal Processing for
Impairments Equalization}\label{subsec:equalization}
It is widely accepted that we are approaching the capacity limits of the fiber-optic communications channels largely imposed by the nonlinearity-induced impairments, or, rather, by the interplay of nonlinearity with dispersion and noise
\cite{agrawal21,winzer2018fiber}. Thus, the search for efficient nonlinearity-mitigation solutions (i.e., the channel equalization tools) in optical transmission lines continues to be one of the primary research topics in the optical communication community. 
Up to now, numerous DSP algorithms have been proposed and studied for the optical fiber channel equalization problem \cite{Cartledge:17}. However, over the past few years, the ``conventional'' equalizers/soft-demappers have started to evolve towards designs incorporating machine learning techniques. In general, various machine learning-based approaches and, more specifically, the NN structures are rapidly finding their way into the telecommunication sector due to their ability to efficiently mitigate transmission- and devices-induced impairments and, also because of the considerable speed of optical transmission provides sufficiently large datasets in a short time so that we can have sufficient datasets to train our models\cite{jarajreh2014artificial,R02,R05,Hunt_2009,eriksson2017applying,hager2018nonlinear,zhang2019field,R03Darko,R08,khan2017machine,Karanov:10,Khan19,giacoumidis2018harnessing}. 

\subsubsection{Post-equalizers}\label{subsubsec:post-eq}

\begin{figure}[ht!]
    \centering
    \includegraphics[width=\linewidth]{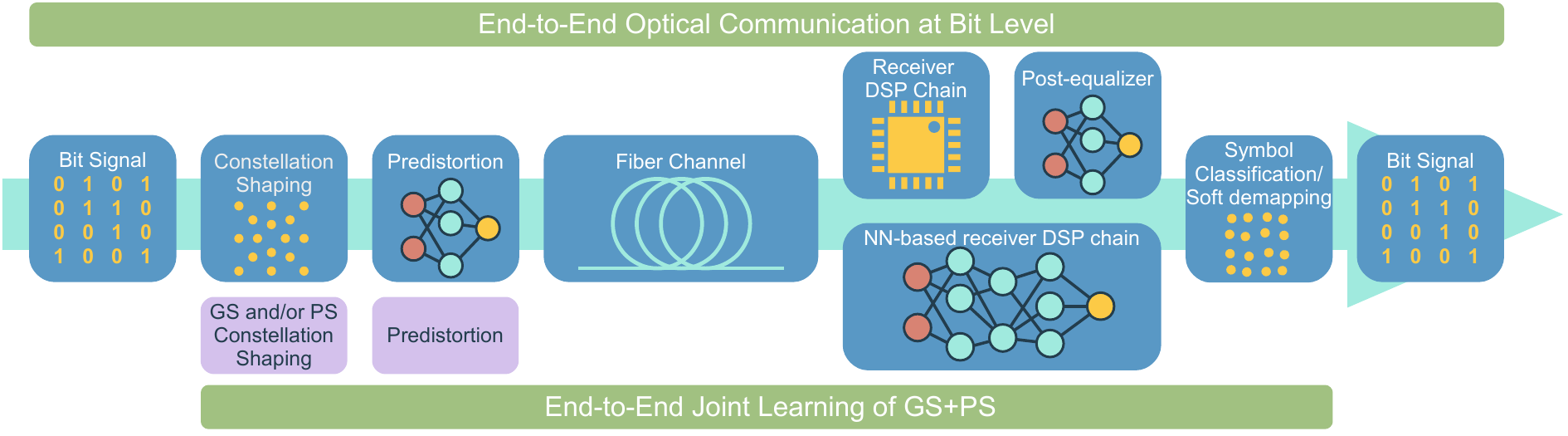}
    \caption{Flowchart depicting the variants of using the NNs in optical communications at a physical layer, including post-equalization, full NN-based DSP, pre-distortion, symbol-to-symbol, and bit-to-bit end-to-end systems. Note that the latter two require the surrogate optical channel to pass the gradients over to the receiver.}
    \label{fig:NN optcomm}
\end{figure}

Perhaps, the simplest and most straightforward NN-based concept to mitigate the signal distortions in optical fiber systems relies on the use of post-equalizers: at the receiver (Rx) side, we add the neural structure that has to revert the channel function and recover the transmitted information\cite{freire2022pitfalls}. The post-equalizer means that in Fig.~\ref{fig:NN optcomm} we use the NN only after the fiber channel at the Rx side.

\begin{figure}[ht!]
    \centering
    \includegraphics[width=0.85\linewidth]{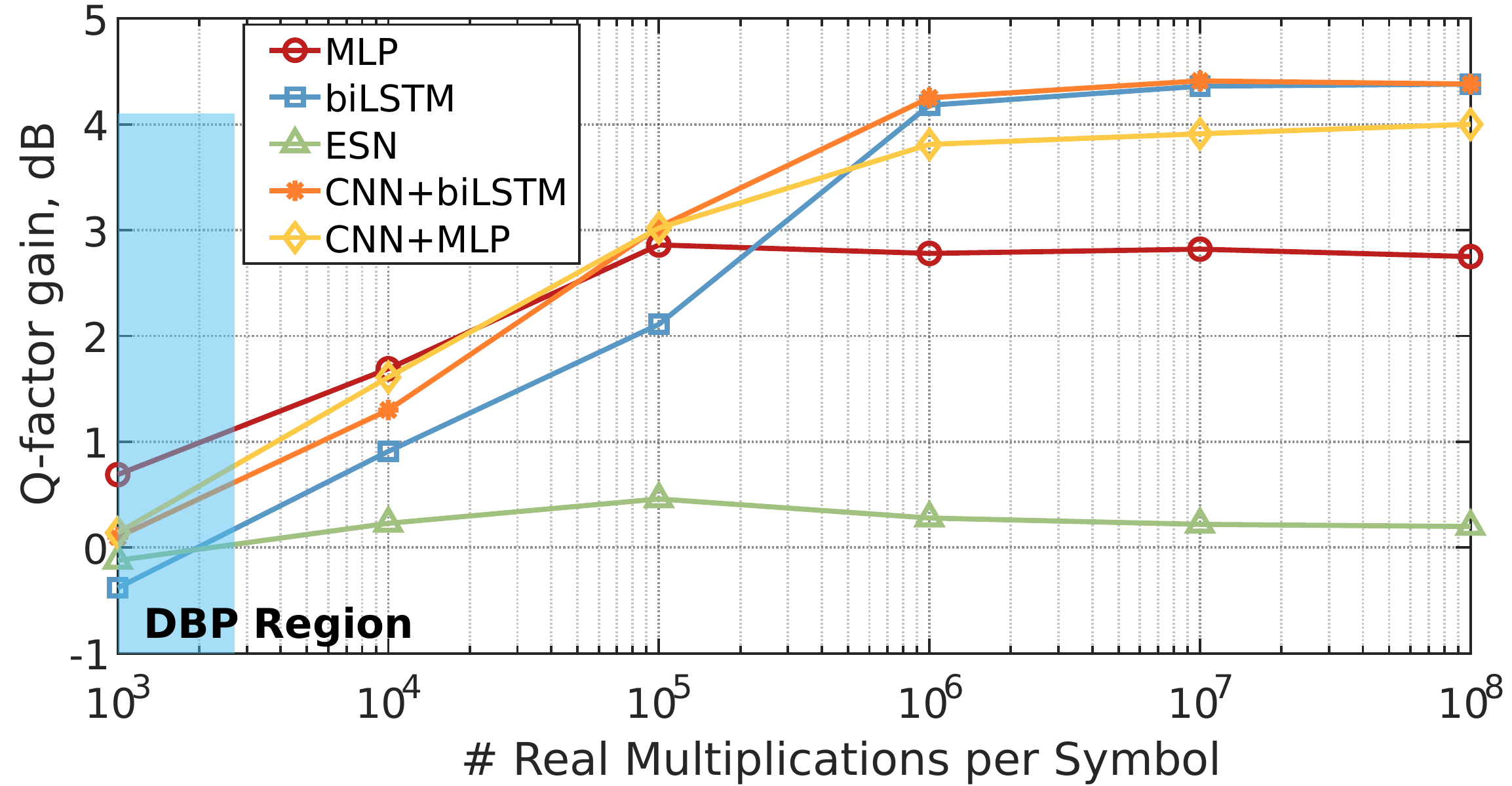}
    \caption{Comparison of performance (Q-factor gain over chromatic dispersion compensation) for different NN post-equalization topologies as a function of the number of real multiplication per symbol (a single symbol output NN). The legend identifies the line types for MLP, bidirectional LSTM, Echo State Network, and combined architectures, CNN+biLSTM and CNN+MLP. The system: dual polarization, 34~Gbd, 16 QAM, single channel TrueWave Classic fiber, 9$\times$50~km propagation distance. Simulation results (the evaluation of the experimental transmission showed a very similar trend). Redrawn from Ref.~\cite{freire2021performance}.}
    \label{fig:com_vs_perform}
\end{figure}

Even though the use of NNs for wireless transmission was considered already in  2003\cite{charalabopoulos2003frequency}, the implementation of NN-based equalizers in application to the optical transmission was first presented some 10 years later\cite{jarajreh2014artificial}. Already in that seemingly first publication on the subject, it was stated that the MLP rendered a better performance compared to the Volterra-series equalization, while the MLP itself was rated as a low-complexity method. Since then, this direction has flourished (the ``early years'' of the subject's development are reviewed in Refs.~\cite{R02,khan2017machine}), and at the moment, we have over a hundred papers addressing different aspects of the subject. First, we note that the channel equalization for the intensity-modulation direct-detection (IM-DD) systems\cite{estaran2016artificial,ye2017demonstration,sang2021multi,arnold2022spiking} represents a simpler task compared to the coherent optical systems\cite{zhang2019field,freire2020complex,freire2020complex,freire2022pitfalls}, as the dimensionality of output objects in the latter case is higher (i.e., the real numbers versus the complex ones). In particular, the reservoir computing-type approaches are relatively efficient in the IM-DD shot-reach systems \cite{da2020reservoir,da2021machine,wang2021signal,da2022reservoir}, while in the long-haul coherent communications, the capacity of the reservoir computing has been found insufficient\cite{freire2021performance}, even though this subject could benefit from some further investigations. At the same time, the latter Ref. demonstrates that the usage of the echo-state network for the post-equalization is tantamount to the MLP in terms of complexity: the interplay between complexity (for the neural equalizers' structures obtained with the BO) and performance is given in Fig.~\ref{fig:com_vs_perform}.

The next question to address when designing an NN equalizer is: which type of predictive modeling should we use to get the most from the equalizer? Ref.~\cite{freire2022deep} analyzes this question, comparing the benefits and deficiencies of each predictive modeling type in the context of coherent optical channel equalization and soft symbol demapping. The issue here is that the datasets usually used for training the NN in transmission problems contain a very low number of errors (especially when we use the NN after the Rx DSP chain). Therefore, when the NN is based on the classification, we can merely have an insufficient number of data points that would induce the training. The latter results in the infamous exploding/vanishing gradient problem, and the NN cannot train well enough. In contrast, when we use the regression task, each dataset point contributes to the training as we have a small continuous deviation of each constellation point's location compared to its initial value, and the difficulties are alleviated. The features of using the NNs described in this paragraph are a specific peculiarity of the transmission post-equalization for almost every transmission system, and these should be accounted for when designing equalization techniques. Some special loss functions that work better specifically for the optical transmission tasks have also been proposed\cite{diedolo2022nonlinear}. Together with this, we have to remember that the ultimate characteristic that we ought to improve when equalizing an optical system is the bit error rate (BER). However, the BER as a function of NN parameters is not differentiable, and, therefore, we make do with the MSE (or some other function of the difference between the true and predicted symbols) as a measure of prediction accuracy. Whence, we arrive at the mismatch between the actual ``goal'' of the equalizer and the result of the NN prediction, such that we need to check our results in terms of the achieved BER (or some other metrics, say the Q-factor, that are expressed through the BER), but the use of the MSE-type metrics, say, the effective signal-to-noise ratio (SNR), can bring about misleading conclusions and wrongly-working designs. 

Now, we need to pay attention to how to structure the output of our NN. We note that the initial equalizers' designs operated with the single-symbol recovery\cite{jarajreh2014artificial,freire2020complex,freire2021performance,sidelnikov2018equalization,giacoumidis2016experimental}, such that the NN returned the predicted value of the real and imaginary symbol parts for the coherent transmission (or just one real number for the IM-DD setups). However, the newest trend now is to use the multi-symbol equalization: it was used in Ref.~\cite{deligiannidis2021performance} to reduce the complexity of the overall post-processing and further assessed in Ref.~\cite{freire2022computational} for the coherent systems and in Refs.~\cite{sang2021multi,sang2022low} for the IM-DD short-reach systems. In particular, in the latter Refs., it was found that the multi-symbol output works well for both feed-forward and recurrent NN topologies. By increasing the number of NNs output symbols, the number of slide windows in equalization can be sharply reduced, so the complexity is also reduced. With this, more information is brought to the multi-symbol NNs in the back-propagations, resulting in a better learning capability and, therefore, better overall performance of the structure. In particular, in Ref.~\cite{sang2022low}, it was shown that the multi-symbol NN equalization in the short-reach IM-DD systems outperforms the single-symbol equalization, even though in the latter case, the task looks simpler. On the contrary, in recent Ref.~\cite{freire2022reducing}, no performance benefits were found (but even a slight degradation) when using a multi-symbol output for the long-haul coherent optical system. However, the direct detailed comparison of multi-symbol versus single-symbol equalization for the coherent long-haul system is yet to be carried out, so it can be an interesting subject for further research.

Next, when designing an equalizer, it is pertinent to think whether we wish to use some predefined structure (e.g., some black-box) solution and further optimize it\cite{freire2021performance,deligiannidis2020compensation,deligiannidis2021performance,sidelnikov2018equalization}, or we incorporate the elements of some known equalization techniques and/or recast that technique as a trainable/learnable approach. An example of the latter is the learned DBP \cite{hager2018nonlinear,hager2020physics,Bitachon20,sidelnikov2021advanced}, where we perform the back-propagation with the use of the SSFM-type architecture but allow the weights in the matrices (which used to be a Fourier convolution) to become trainable (Ref.~\cite{fan2020advancing} contains a good analysis of the method). Another popular approach is to use the perturbation theory, but we now allow the perturbation parameters to be trainable\cite{zhang2019field,luo2022nonlinear,barreiro2022data,melek2020nonlinearity,melek2021fiber,li2022convolutional,redyuk2020compensation,dzieciol2022inverse}. One more interesting direction is to base the learnable approach on the Volterra series technique\cite{castro2022novel,huang2022low}. Finally, Ref.~\cite{freire2020complex} proposes a special NN architecture combining an NN-based nonlinear step and NN additions, which also shows promising performance. No general recommendation on which of the two paths (a black-box approach or a trainable version of some existing method) to follow can be given, as both directions have their positive and negative features.

The following question addresses the particular NN architecture: whether we work with real numbers or adopt the complex-valued NNs. While the latter path is more complicated, there have been considerable advances in the development of complex-valued NNs framework\cite{complexNN}, and we can benefit from using the complex-valued NNs applied to optical channel equalization, see Refs.~\cite{liu2017multilevel,freire2020complex,wang2021optical,bogdanov2021application,he2021fiber}.

Now, let us turn to the selection of the particular NN type/topology. In general, many versatile NN-type structures can be used for equalization. The earlier studies incorporated the MLP \cite{jarajreh2014artificial,giacoumidis2016experimental,sidelnikov2018equalization}, as it is the most studied structure. At present, a great lot of other structures have been assessed in the channel equalization context. First, as a natural extension of the feedforward MLP, the CNN-type based structures have been considered for both coherent and IM-DD systems\cite{sidelnikov2021advanced,yang2021modified}. However, we note that optical transmission setups typically feature essential memory effects. And here it is the right place to recall that the RNN-type topology and its advanced modifications, like GRU, LSTM, etc., are specifically tailored to handle the memory. Therefore, recent studies have begun shifting more and more towards the equalizers incorporating various recurrent structures \cite{deligiannidis2020compensation,deligiannidis2021performance,freire2021performance,ming2022ultralow}, including advanced models with attention mechanism\cite{liu2022attention,shahkarami2021attention}, and some combinations of recurrent and feed-forward NN parts\cite{freire2021performance,shahkarami2023efficient,luo2022nonlinear}, where we can expect to have benefits rendered by both topologies. Fig.~\ref{fig:com_vs_perform} shows the performance of the BO-optimized different post-equalizing NN structures vs. the complexity, i.e., the number of multiplications required to process one symbol, for a long-haul coherent system. The complexity was upper-bounded by setting the upper limits for the BO process. It can be seen that the LSTM-based recurrent structures typically outperform their feedforward counterparts when we allow the complexity to be high. Interestingly, when we restrict the allowed complexity to lower values, a simple MLP can emerge as the most efficient solution. Thus, in general, we recommend testing several structures/topologies before deciding upon an ultimate design for the equalizer and then applying the complexity reduction procedures described in the next section. 

Finally, we note that positioning our NN after the ``traditional'' Rx DSP chain is not the only option. It seems more efficient to replace/impute the whole DSP chain at the Rx side with the NN elements. The latter design was assessed in Ref.~\cite{huang2022design}: the authors of that Ref. underline the interpretability of the resulting design, a truly important feature when we want to understand why we have some specific behavior/functioning problems of the NN setup.

To end up this section, we mention that various difficulties emerging in the design and training of NN-based post-equalizers incoherent optical systems are amply described in Ref.~\cite{freire2022pitfalls}; moreover, some pitfalls depicted in that paper are generally pertinent to the NN usage in communications, not only to the NN equalizers.

\subsubsection{Pre-distortion}
Pre-distortion, i.e. the pre-compensation
of signal's distortion at the transmitter (Tx) via special] digital pre-processing of the 
symbol sequence is another popular way of combating channel nonlinearities: in optical communications, the pre-distortion is typically based on the aforementioned Volterra series approach\cite{bajaj2022efficient}. The application of learning approaches to the digital signal pre-distortion is, actually, not a new subject in general\cite{psaltis1988multilayered,bernardini1990use}. Over the recent years, a number of digital pre-distortion techniques based on the NNs' utilization have been proposed for wireless systems\cite{gotthans2014digital,hu2021convolutional}. However, the pre-distortion techniques' demonstrations for coherent optical systems are still relatively few and far between. There are methods efficient for memoryless digital pre-distortion of a Mach-Zehnder modulator \cite{schaedler2019ai} and low-resolution digital-to-analog converter \cite{abu2019neural}. A Wiener-Hammerstein model-based approach was proposed in Ref.~\cite{sasai2020wiener}. One of the most remarkable results demonstrating the efficiency of the NN-based pre-distortion in coherent optical links was given in Ref.~\cite{bajaj2022deep}. The implementation
of the pre-distortion based on the deep MLP-like NN led to the record transmission rates for single-channel \cite{bajaj2022deep,bajaj2020single}, and multi-channel dense-wavelength-division-multiplexed (DWDM) transmission \cite{bajaj202154} over 80~km single-mode fiber systems. The recent results regarding the usage of the NN-based pre-distortion approach for IM-DD systems can be found in Ref.~\cite{minelli2022multi}, and for coherent systems using RNNs -- in Ref.~\cite{bajaj2022performance}.

\subsubsection{End-to-end equalization of optical systems}\label{subsubsec:e2e}
The methods in the two subsections above, incorporating the NN structures for the channel equalization, referred to the Rx (post-equalization) or Tx (pre-distortion) transmission system parts only. Meanwhile, a more ``omnidirectional'' approach incorporating the able-to-learn elements into a communication system can be based on the so-called end-to-end (E2E) learning concept\cite{o2017introduction}. We notice that the E2E learning concept, though fitting well the communications-related problems, is a very general multipurpose method\cite{glasmachers2017limits}, very efficient, e.g., for such a famous but seemingly irrelevant problem as autonomous car driving\cite{bojarski2016end}. The E2E learning can be formulated as the method involving training a (often very complex multi-component) learning system represented by a single model (typically an NN) that represents the \textit{complete target system}, where each NN part (that can be just a layer or a complex ensemble of layers) can specialize in performing intermediate tasks. Thus, returning to optical communications, we need that the whole optical
communication link is modeled as an NN. In this respect, we can span the modeling from initial bits entering into our system down to received identified bits\cite{karanov2018end,9195215,Karanov:10}, or, potentially, assume that we model our system from symbols to symbols\cite{9890753,Neskorniuk2022}, see Fig.~\ref{fig:NN optcomm}.  To recast our system as an NN, we now need some differentiable model of emulation of the signal propagation down the fiber (it can be, e.g., some NN structure described in Subsec.~\ref{subsec:modeling}), and the DSP elements at both Tx and Rx ends can now be represented as NNs with trainable parameters. Importantly, the parameters of both Tx and Rx NN block are now \textit{optimized simultaneously} using standard NN training, using the fact that we can efficiently pass the gradients through the optical link model towards the receiver; alternatively, some advanced gradient-free methods can be used\cite{jovanovic2021gradient}, but this direction has not so far being adopted and elaborated in optical transmission. Noticeably, the benefits of using E2E setups are evident for the systems, where the optimal DSP solutions are not known. We also note that the concept of E2E systems is conceptually similar to the contractive autoencoder architecture described above.

The initial works related to the E2E application referred to the short-haul IM/DD systems, where the fiber nonlinearity was non-essential, so the fiber propagation model was linear and simple. In Refs.~\cite{karanov2018end,Karanov:10,karanov2020concept,karanov2020end}, the E2E learning of geometric constellation shaping (GS), i.e., the optimal symbol locations for IM/DD optical communication systems were researched: it was shown that the E2E methodology resulted in the essential performance gain. Developing the approach further, in Ref.~\cite{gaiarin2020end}, the E2E learning of waveforms was addressed for a very special nonlinear frequency division multiplexing optical communication system, see subsection~\ref{subsubsec:nft} below. In~\cite{jones2018deep,jones2019end, gumucs2020end, oliari2021high, niu2022end, jovanovic2021end}, the E2E learning of single-symbol GS was considered for already a more complicated coherent communication system. In particular, in Refs.~\cite{jones2018deep,jones2019end, gumucs2020end, oliari2021high, niu2022end}, only the optical channel-related distortions were taken into account, while in Ref.~\cite{jovanovic2021end} a more realistic link model, that also included the local oscillator laser noise, was studied. In~\cite{song2021end, he2022experimental, song2022model,bajaj2020single, bajaj2022efficient, bajaj202154, bajaj2022deep}, the E2E learning of GS, signal waveform, and nonlinear pre-distortion resistant to transmitter distortions were considered. However, the distortions emanating from the signal's propagation down the fiber were neglected in these works, either completely or modeled via a simplified Gaussian noise model. Ref.~\cite{uhlemann2020deep} addressed the joint E2E learning of GS and linear pre-distorter mitigating the fiber channel distortion. However, the learned linear pre-distorter's contribution to the nonlinearity mitigation is questionable. Finally, in Ref.~\cite{Neskorniuk2022}, the E2E learning of the constellation shaping for a single-channel dual-polarized 64~GBd transmission over 170 km standard single mode fiber link, which takes into account the nonlinearities and optical channel memory, was proposed and studied. With the new method, it became possible to jointly optimize symbol locations in the constellation diagram, the symbol probabilities, and the nonlinear pre-distortion: the learned transmitted signal distribution chooses the transmitted symbol based not only on the message sent in the corresponding time slot as in the conventional constellation shaping but also on the messages sent in the neighboring time slots. The feature of the approach from that Ref. is that a relatively accurate auxiliary (surrogate) channel model based on perturbation theory was used there. With this, the training procedure for the simultaneous learning of symbol probabilities~\cite{aref2022end}, was implemented. 

Overall, the E2E learning application, especially in coherent optical systems with the account of all distortion types, is truly a nascent subject. Some initial results obtained up-to-date suggest that this direction can be really fruitful. At the same time, we ought to recognize the problems associated with the E2E system's development. First, it is the method that is very difficult to implement in experimental conditions: we need to understand which changes the loss/cost metrics alternations induce in our experimental transmission setup. The latter can be quite demanding, as we typically do not know a priory the number of runs that we have to spend to achieve some desired performance values, and we are also unaware of the ``level'' (or type) of the system's alternations that the training would result in; some representatives of the latter can be technically unfeasible at all. Together with this, we have to be accurate in designing the multi-modular E2E systems, as the high complexity of the NN structure may imply that a considerable amount of data and training runs are needed to have an acceptable result, accompanied by numerous problems characteristic to very deep NNs' training. Also, the generalizability of E2E optical communication systems has to be investigated in more detail, even though we can attain some flexibility by a specially designed training\cite{karanov2021distance}. Thus, we can think of the E2E systems as of high-complexity (potentially) high-reward direction, where the machine learning-related problems may intertwine with those pertinent to optical transmission, thus requiring a researcher to apply different mitigation methods and where a lot of difficulties are still yet to be alleviated.

\subsubsection{Free space optical systems}\label{subsubsec:fso}
One of the directions that have recently started to attract more attention, including the studies of NN-based equalization, is free space optical communications (FSO). FSO systems can provide a considerable unlicensed
bandwidth for data transmission at more than one hundred
Gigabits per second \cite{ren2016experimental}, reach extremely long distances, being secure and robust to electromagnetic interference \cite{khalighi2014survey} and atmospheric turbulence\cite{li2022enhanced}.

As for the NN applications in FSO systems, a good systematization of different results is given in Ref.~\cite{amirabadi2022low}. This reference classifies the NN methods used to compensate the FSO transmission impairments into three categories: ``classical machine learning-based methods'', which include either non-NN- or shallow NN-based approaches\cite{zheng2015svm,lohani2018turbulence},  the approaches involving the CNNs \cite{hao2020high,tian2018turbo,li2018joint,bart2022deep}, and deep NN-based methods (though CNN structures can also be classified as ``deep'')\cite{zhu2019autoencoder}. Overall, whereas the application of the NNs to mitigate the detrimental impact in FSO is gaining momentum, the studies related to the complexity reduction of the processing are relatively not widespread, aside, perhaps, from the aforementioned Ref.~\cite{amirabadi2022low}, where the authors compare the complexity of their method against some existing signal-processing solutions.

\subsubsection{Nonlinear Fourier transform-based fiber systems}\label{subsubsec:nft} The nonlinear Fourier transform  (NFT) based optical signal processing and modulation techniques, and, in particular, the nonlinear frequency division multiplexing (NFDM) as the most efficient method among the NFT-based optical transmission systems, have been intensively studied over the last years \cite{YK14,TurOpt,china-rev2021,nat2017}. 
Within the NFDM systems, the data modulation and transmission take place inside the special nonlinear Fourier (NF) domain, where the nonlinear intermodal cross-talk (arising due to the Kerr effect) between the effective ``nonlinear modes'' is virtually absent~\cite{TurOpt}. Even though, theoretically, the signal's propagation in NFDM  systems is unaffected by the fiber nonlinearity provided that the signal's evolution is well approximated by an integrable evolutionary equation, in real systems, the deviation of the channel model from the idealized integrable equation results in modes' coupling. Therefore, we arrive at the mismatch between the NFT-based processing and the channel. However, the system's performance can be improved by using highly-adaptive NNs instead of ``deterministic'' NFT operations.  

The first direction in employing the NNs for improving the functioning of NFDM systems consists in applying the additional NN-based processing unit at Rx to compensate the emerging line impairments and deviations from the ideal model \cite{gdd18,kwp19,kkp19,kpk20,kkp21,kvkppt19,chen2021two,lv2022noise}: it can be deemed as the extension of the post-equalization concept, subsection~\ref{subsubsec:post-eq}. But, despite ensuing the transmission quality improvement, this type of NN usage brings about the additional complexity of the receiver. At the same time, the complexity reduction for the NFT operations has been a subject of active research, and it is undesirable to raise it further by adding more processing units. In the more viable alternative approach, the NFT operation at the receiver is entirely replaced by the NN element\cite{jgy18}. It has been shown that this approach, indeed, results in a noticeable improvement of the NFT-based transmission system functioning \cite{ymm19,jgy18,wxz20}. At the initial stage of research, the NNs emulating the NFT operation were used in the NFDM systems operating with solitons only (the study of such systems actively develops as well~\cite{mishina2021eigenvalue,mishina2021combining,takeuchi2022eigenvalue}).  However, we notice that the most efficient NFDM systems developed so far operate with the continuous nonlinear Fourier spectrum\cite{lpt14,yangzhang2019dual,yangzhang2019experimental,gui2018nonlinear,derevyanko2021channel,zhang2021correlation,balogun2022enhancing}. In the first work related to the communication system based on the continuous nonlinear Fourier spectrum ~\cite{zhang2021direct}, a standard ``imageInputLayer'' NN (developed originally for handwritten digits' recognition) from MATLAB 2019a deep learning toolbox was adapted to process and demap the data. However, such an approach utilizes the classification task, which can bring about difficulties \cite{freire2022deep}. The problem of the NN-based nonlinear Fourier spectrum recovery using the regression task was considered by Sedov et al., in Ref.~\cite{sedov2021neural}: a special CNN-type structure coined NFT-Net was proposed there. Further, it was shown that such a structure is reversible,  i.e., it can be used for the inverse NFT computation\cite{sedov2021neural1}. This direction was extended further in \cite{zhang2022serial,zhou2022nonlinear}. In Ref.~\cite{zhang2022serial} two CNN-type structures were analyzed for directly decoding NFDM data:  a small serial network scheme was designed for small user applications, and a parallel network scheme with high speed was designed for high data rates. Importantly, the questions regarding the complexity of NN signal processing were addressed. In Ref.~\cite{zhou2022nonlinear}, a diffractive fiber-based NN was proposed to discern the NFDM symbols. That NN was composed of multiple cascaded dispersive elements and phase modulators. An all-optical back-propagation algorithm was used to optimize the phase. The fiber-based time domain NN structure acts as a powerful tool for signal conversion and recognition, and such a structure can be used to recognize the symbols all optically, which can allow us to replace the NFT processing with much simpler and even system-agnostic operations.

Finally, we mention Ref.~\cite{gaiarin2020end}, where the end-to-end optimization was used for the NFDM system based on solitons. A very efficient recent NFDM system was presented in~\cite{chen2022}: using the NN equalization, the authors experimentally demonstrated a 25-channel NFDM system with polarization division multiplexing 16QAM modulation,  transmitting over 10 Tb/sec for the 800~km distance. 

\subsection{Optical communications: Network Layer}
The idea of adding intelligence to optical networks to make network operations easier and boost network performance is rapidly gaining traction in both research and industry. The optical network is a key point in the worldwide infrastructure for communications since it bridges the gap between higher-level services and the underlying physical infrastructure by allocating resources such as links, wavelengths, spectrum slots, fiber cores, and time slots. Because of this complexity, optical networks are more difficult to operate and maintain than other types of communication networks. As optical networks grow more complex, manual optimization can take too long and lead to suboptimal results. Therefore, machine learning, and more specifically NNs, are used more and more nowadays, allowing for better, faster optimization decisions to be made. 

\begin{figure}[ht!]
    \centering
    \includegraphics[width=\linewidth]{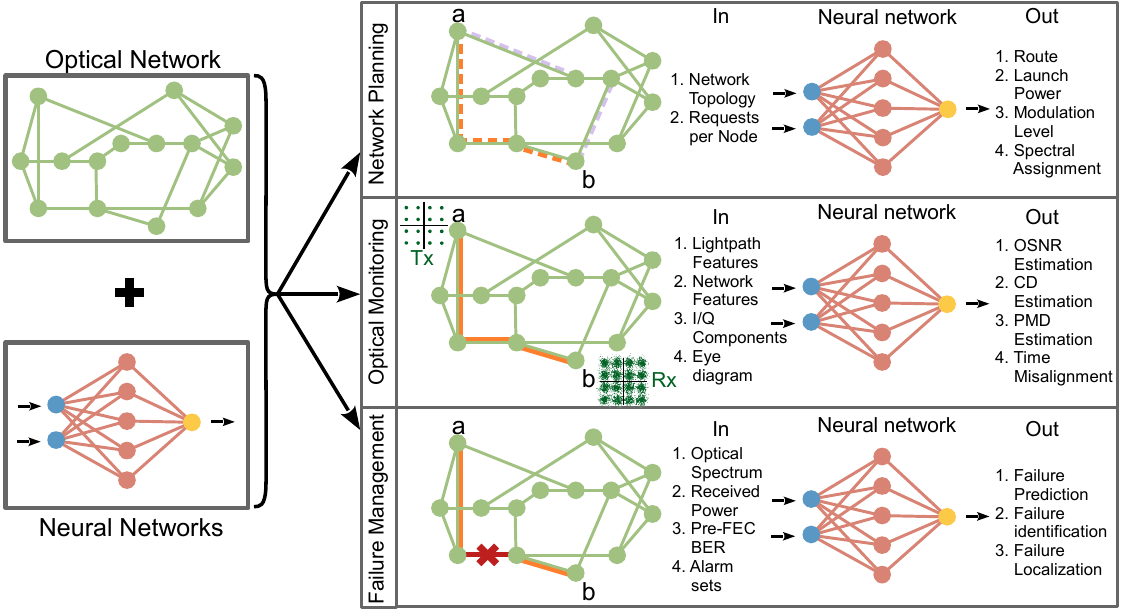}
    \caption{Three major directions of using the NNs in optical networks.}
    \label{fig:NN optnets}
\end{figure}

In Fig.~\ref{fig:NN optnets}, we have summarized the three main areas in which the NNs have been successfully applied in optical networks. Those areas are network planning, optical monitoring, and failure management\cite{gu2020machine, wang2022review, R05}.

When dealing with network planning, the NNs have been deployed for two main purposes: traffic prediction and solving the rout, modulation level, and spectra assignment (RMLSA) problem. Predicting the bandwidth requirement in the next time step based on real-time measurement of traffic characteristics is one of the key challenges in improving the efficiency of network operation. The purpose of using the NN models to predict future traffic rate variations is to do it as accurately as possible based on historical data. In this case, the NN input is the history of requests per note of the optical network (past traffic data), so the NN can forecast future traffic demands.  RNNs, such as GRU and LSTM, are among the NN structures that have been efficiently used for traffic prediction because of their ability to adaptively capture the dependencies on different time scales. In Refs.~\cite{troia2018deep,aloraifan2021deep}, the GRU is used for making the traffic matrix forecasts for both a fixed-grid WDM network and a backbone elastic optical network (EON). In addition, Refs.~\cite{hatem2019deep, zhu2020prediction} studied using LSTM models for traffic prediction in passive optical networks and core networks. For the RMLSA problem, NN has lately emerged as an alternative to standard methods such as integer linear programming, heuristics such as simulated annealing, k-shortest path routing, first-fit, and genetic algorithms. Generally speaking, the NN is capable of efficiently learning the network and physical layer aspects by having information on the network properties, fiber properties, and user requests. Then, it can provide the optimal routes, launch powers, symbol rates, modulation formats, and spectrum assignments per request to minimize network blockage and maximize spectrum utilization.

Next, we address the optical performance monitoring (OPM) category. OPM is crucial to guaranteeing a stable network, as even a momentary disruption in service due to faulty fiber or equipment can cause widespread packet loss. Parameters like OSNR (Optical Signal-to-Noise Ratio), CD (Chromatic Dispersion), PMD (Polarization Mode Dispersion), PD (Polarization Dependent Loss), OP (Optical Power), and FN (Fiber Nonlinearity) are of primary interest to OPM. In this sense, the NN is used to estimate such parameters when only the received optical signal is available in the form of eye diagrams, sampled signal, I/Q components, and other network and light-path features. Here we highlight that in Refs.~\cite{vaquero2021perturbation,wang2017convolutional,tanimura2019convolutional,tanimura2021concept,wang2019comprehensive,lohani2019dispersion,du2021forecasting}, the artificial NN was used to estimate fiber properties of an optical link. Further, in Refs.~\cite{muller2021estimating, d2020using, kashi2021neural,lonardi2020perks,lv2021joint}  the NNs were used to estimate BER  and/or OSNR. We specifically highlight recent work~\cite{muller2021estimating}, where the authors experimentally demonstrated that an NN-based quality of transmission (QoT) estimation, where the NN was trained on synthetic QoT data, could successfully estimate the SNR on a live optical network.

The last important aspect of the NN application at a network layer is failure management in optical networks. Failure management's goals are to find and fix any network problems that arise, keep everything running well, and live up to the service level agreement with customers. However, the standard approach to failure management still necessitates laborious and lengthy human involvement. Machine learning approaches have been extensively applied to the aforementioned problems in an effort to push failure management in the direction of intelligence and efficiency. First, in failure prediction, deep CNN structures \cite{inuzuka2020demonstration} were used to estimate the bend location of remote fiber by using the information of the constellation data from the receiver. An important direction concerns the prediction of failure in optical transport network (OTN) boards. Using the historical data of the operating state parameters from OTN equipment, in Ref.~\cite{zhang2020temporal} a biGRU model and in Ref.~\cite{zhang2021attention} an attention mechanism-driven LSTM model were proposed for temporal data-driven failure prediction and prognostics. For other optical equipment such as lasers, Ref.\cite{abdelli2020lifetime}, used deep NN to predict the mean time to failure of a laser by having as the input just the laser monitored parameters.  Finally, in a failures Location, the NN receives the alarm log of the system, and LSTM \cite{liu2019application}, attention/transformers \cite{jia2021transformer}, or other types of NN \cite{zhao2019accurate}, can produce an alarm root cause analysis-enabled failure location. 

\subsection{Optical sensing}
There is a great variety of optical sensors, but they all use light to detect, measure, and convert magnitudes from any domain to an optical signal, first and then to an electrical one. These domains include temperature, pressure, stress, displacement, strain, liquid level,   vibration, rotation, velocity,  acceleration, electric, magnetic and acoustic fields,  force,  pH value, chemical species,  humidity, and many others. As a result, optical sensors are used in a vast range of applications, from structural health monitoring and seismic measurement to the medicine and food industry, the oil and gas industry, power line monitoring, smart city applications, and many others. NNs are used in optical sensing for classification tasks and for improving both the accuracy and speed of raw data processing in applications varying from distributed strain sensing to biochemical optical sensing, see Refs.~\cite{Sensor05,lewis2007principal,Sensor02,Sensor03,Sensor04,suah2003applications,li2019deep,shokrekhodaei2021non,Sensor06,NNSensor09,NNSensor10} and references therein. The field of using the NNs in optical sensing is too large to discuss it in one section, so we do not aim here to cover all-optical sensing applications, but rather give several typical examples of using NNs in this field and provide references for further reading. 

The NN-based approach to analyzing optical fiber sensors' signals and the applications of NNs for fiber sensor signal interpretation are reviewed in \cite{lewis2007principal}.
The applications of NNs for pH monitoring using fiber optic sensors are discussed in \cite{suah2003applications,Sensor06}.
Deep NN for the predictions of the resonance spectra of plasmonic sensors is considered in ~\cite{li2019deep}. 
 Non-invasive glucose monitoring using optical sensors and machine learning techniques for diabetes applications are discussed in \cite{shokrekhodaei2021non}, where light sources with multiple wavelengths were used to enhance the sensitivity and selectivity of glucose detection in an aqueous solution. Machine learning techniques are employed in optical sensors to increase accuracy and noise resilience. 
 
\begin{figure}[ht!]
\includegraphics[width=.95\linewidth]{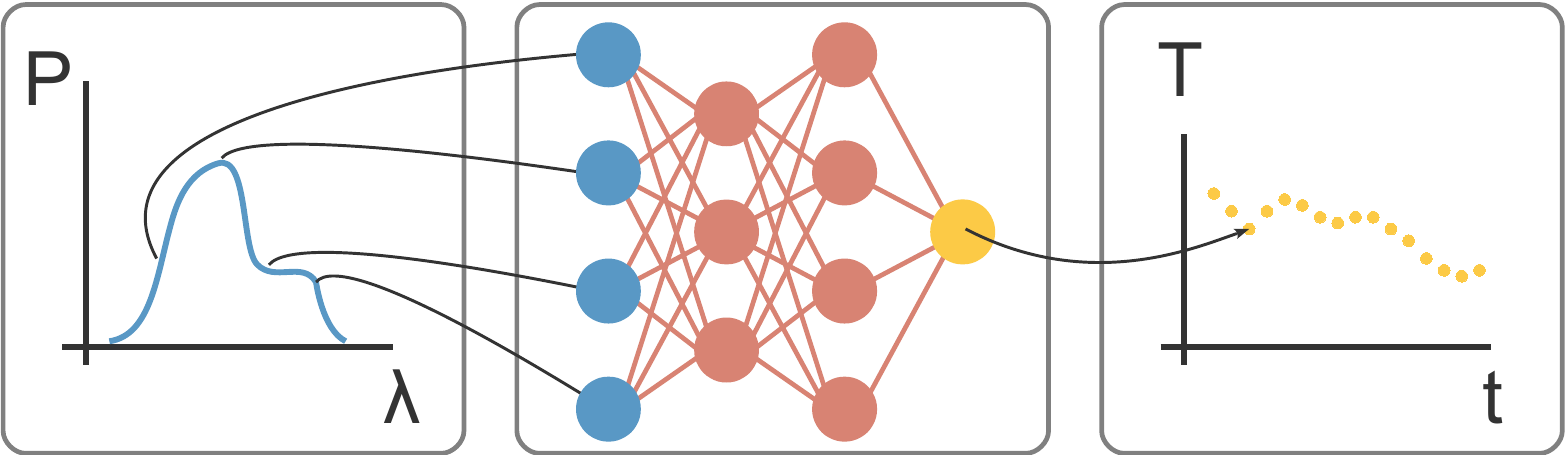}
\caption{Typical low-complexity MLP for signal processing in sensing applications. A series of optical spectra are processed to obtain a time series of the measured parameter.}
\label{fig: MLP for signal processing}
\end{figure}

 Although there are a huge number of examples of using NNs in optical sensing, not all these applications employ high-complexity NNs. In many cases (e.g., as in Refs.~\cite{lewis2007principal,suah2003applications, shokrekhodaei2021non,Sensor06}), the NNs used are simple feed-forward structures. These are often applied as a simple replacement for signal processing techniques and sensing data interpretation, as shown in Fig. \ref{fig: MLP for signal processing}. These low-complexity NNs are basically used as a black box trained to perform denoising/approximation  and consist of just a few hidden layers. As a particular example of this approach, in Ref.~\cite{Sensor06} the operation of an optical fiber pH sensor measuring the reflectance spectra of the immobilized bromophenol blue was enhanced using a relatively simple feed-forward NN. The input layer consists of six neurons, corresponding to the reflectance intensities measured at six different wavelengths from each spectrum, the output layer is a single neuron corresponding to variable pH values, and 11 neurons in the hidden layer were enough to improve the dynamic response of the pH sensor from (pH 2.00–5.00) to (pH 2.00–12.00). 
In \cite{NNSensor09}, an MLP trained via supervised learning with the Levenberg-Marquardt algorithm
was applied to reduce the errors obtained by the matrix method in optical fiber Bragg grating based on the simultaneous measurements of strain and temperature.

More complex NNs are also actively used in a number of optical sensing applications. For instance, deep NNs are used for denoising \cite{manie2020using} and event recognition \cite{shi2019event}. Deep NNs have also been employed to simulate complex sensor behavior and avoid time-consuming precise modeling of the response of the plasmonic sensor using Finite-Difference Time-Domain (FDTD) or finite element method (FEM)  \cite{li2019deep}. 
In the distributed fiber sensors, static or dynamic measurements  can be done over hundreds of kilometers  with the meter-scale spatial resolution by processing the data for Rayleigh, Brillouin, or Raman scattering. Due to the high speed of the operation of optical sensors, the analysis of the collected enormous amount of data requires advanced methods of processing, which is especially challenging for real-time raw data processing. In this situation, the reduction of the NNs computational complexity and power efficiency is the key to the development of efficient hardware for wide NNs deployment in optical sensing.

\subsection{Ultra-fast light measurements and characterization} 
Ultrafast photonics deals with high-speed  optical measurements, generation, characterization, and usage of extremely short pulses with  picosecond to attosecond scale duration, the topics which are important for a range of applications, from medical lasers and nonlinear imaging and microscopy to materials processing and 3D laser printing. One of the challenges in ultrafast photonics is that the dynamics of ultrashort pulses in many applications are highly nonlinear. Therefore, the design optimization of pulse evolution in the nonlinear medium requires time-consuming numerical modeling. NN can provide new design tools and enhance the performance of measurement and characterization techniques for ultrafast light \cite{NNUC01,R07,NNUC03,NNUC04,NNUC05,NNUC06,NNUC07,NNUC08}. For instance, the conventional approach to optimize nonlinear fiber-optic dynamics is based on numerical modeling using the generalized nonlinear Schrödinger equation. NNs can be efficiently exploited to emulate nonlinear pulse propagation and reduce computational time and memory. In \cite{NNUC01},   a recurrent NN has been applied to model and predict complex nonlinear propagation in optical fiber, using data solely from the input pulse intensity profile. The NN prediction agreed well with the experimental results for pulse compression and ultra-broadband supercontinuum generation.  Nonlinear instabilities in fiber optics, with their inherently complex light dynamics, are a good test-bed for the theoretical and computational tools required for the design and optimization of fiber devices. NN-based analysis of instabilities has been introduced in \cite{NNIns01,NNins02}. In Ref.~\cite{R07}, a supervised NN has been trained to correlate the spectral and temporal properties of modulation instability using simulations and then applied to analyze high dynamic range experimental spectra to yield the probability distribution for the highest temporal peaks in the optical field. It was also shown that unsupervised learning can be used to classify noisy modulation instability spectra into subsets associated with distinct temporal dynamic structures. 
    
The characterization of ultrashort laser pulses with femtosecond to attosecond pulse duration is yet another area where the NNs can offer new perspectives~\cite{NNUC07,NNUC08}. The characterization of the amplitude and phase of such ultra-short pulses is of critical importance for, e.g., chemical reactions and electronic phase transitions.  Employing the NN for the reconstruction of ultrashort pulses enables diagnostics of low-power pulses and/or characterization without a priori knowledge of the relations between the pulses and the measured signals \cite{NNUC07}. 
In \cite{NNUC08},  a method for the phase reconstruction of an ultrashort laser pulse based on the deep learning of the nonlinear spectral changes induced by self-phase modulation has been presented. The NNs have been trained on the simulated pulses with random initial phases and spectra and validated on experimental data produced from an ultrafast laser system, where near real-time phase reconstructions were performed. This method can be used in systems with high energy, large aperture beams, for instance, in petawatt laser systems \cite{NNUC08}.
    
The NNs can assist in retrieving the amplitude and phase of the complex electric field from the interferogram. In Ref.~\cite{NNTimestretch01}, an NN-based spectral interferometry system that utilizes a neural network to infer the magnitude and phase of femtosecond interferograms directly from the measured single-shot interference patterns has been demonstrated. A five-layer fully-connected NN was used to perform the regression, inferring the amplitude and phase spectra from the measured spectral interferogram. The input size (the size of a single input frame) of the network is defined by dividing the sampling rate of the real-time analog-to-digital converter (ADC) by the repetition of the laser \cite{NNTimestretch01}. The NN directly outputs a vector that is the concatenated magnitude and phase spectra imposed by the spectral modulator. Importantly, this method does not require a priori knowledge of the shear frequency. 
    
\subsection{Laser systems}

The NNs are well suited for dealing with non-linear problems, which makes them very useful in laser science and technology because nonlinearity plays an important role in the operation of many classes of modern lasers~\cite{siegman86,fu2018several,Laser,Laser01,Laser02,andral2016toward,Laser05}. The challenges in optimizing and control of lasers result from a large number of effective degrees of freedom (or control parameters) that need to be balanced to achieve stable operation or to achieve a specific targeted lasing regime.  Moreover, there is an increasing demand for autonomous laser operation and active self-tuning in the presence of changing environmental perturbations. In this section, we address the applications of NNs to a specific class of lasers -- the fiber lasers \cite{Laser,Laser01}.

The efficient modeling of fiber lasers can potentially lead to a breakthrough in their performance. However, laser modeling requires the accurate characterization of all elements (not always available) and efficient data analysis \cite{Laser,Laser01}. Even with the known mathematical models, the complexity of light dynamics in the laser cavity makes the design and optimization of such lasers a challenging task, namely: (i) direct numerical modeling requires the knowledge of system parameters that are not always available; (ii) even though the impact of optical noise on the evolution of radiation can be accounted for, this requires massive, time-consuming Monte Carlo simulations of the stochastic partial differential equations underlying laser operation; (iii) comprehensive design optimization requires the analysis of large amounts of data. The simultaneous presence of the three aforementioned factors is exactly why machine learning-based methods can transform the future of laser science and technology.

\begin{figure}[ht!]
\centering
\includegraphics[width=.8\linewidth]{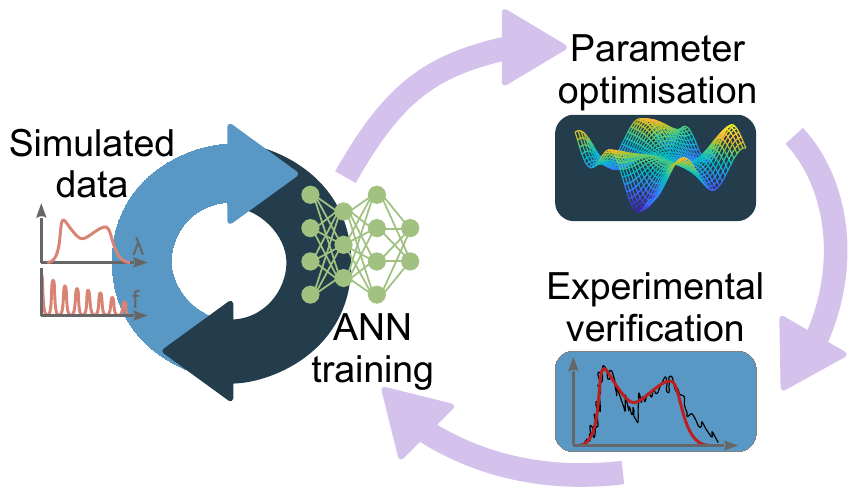}
\caption{Schematics of an ANN used for laser cavity parameter optimization. \egor{Fig. 27 and Fig. 28 are swapped to ensure that figures appear in the numerical order in the text.}} 
\label{fig: ANN for laser optimisation}
\end{figure}

It has been demonstrated recently that NN-based techniques such as deep learning can be applied to solve nonlinear stochastic partial differential equations  \cite{NNPDE01,NNPDE02,NNPDE03}. Therefore, machine learning methods, and in particular the NNs, have great potential to improve the performance of (fiber and other) lasers; such techniques can be used in the development of a new generation of ``smart'' laser systems. Lasers can quickly generate large data sets required to reach a good accuracy of NNs. Moreover, the availability of mathematical models, even without exact knowledge of all parameters, can be utilized to speed up the operation of NNs. This can be implemented by embedding available a priori information into the architectures and loss functions of the NNs or by using simulated data to train NNs as shown in Fig.~\ref{fig: ANN for laser optimisation}.

In the context of laser science and technology, machine learning can be used for:
\begin{itemize}
\item design optimization and predictions of lasing regimes; 
\item characterization of laser radiation; 
\item improving our understanding of the  physical mechanisms underlying the operation of complex laser systems; 
\item field control and self-tuning of the lasing regime. 
\end{itemize}

The machine learning-based techniques combined with the feedback loops have the potential to revolutionize the ways to design, control, and select desirable regimes in lasers, leading to self-starting, self-optimizing lasers, and robust against environmental perturbations \cite{brunton2014self,kutz2015intelligent,woodward2016towards,baumeister2018deep,winters2017NPE,woodward2017genetic,brunton2014self,kutz2015intelligent}.

\begin{figure}[ht!]
\centering
\includegraphics[width=.6\linewidth]{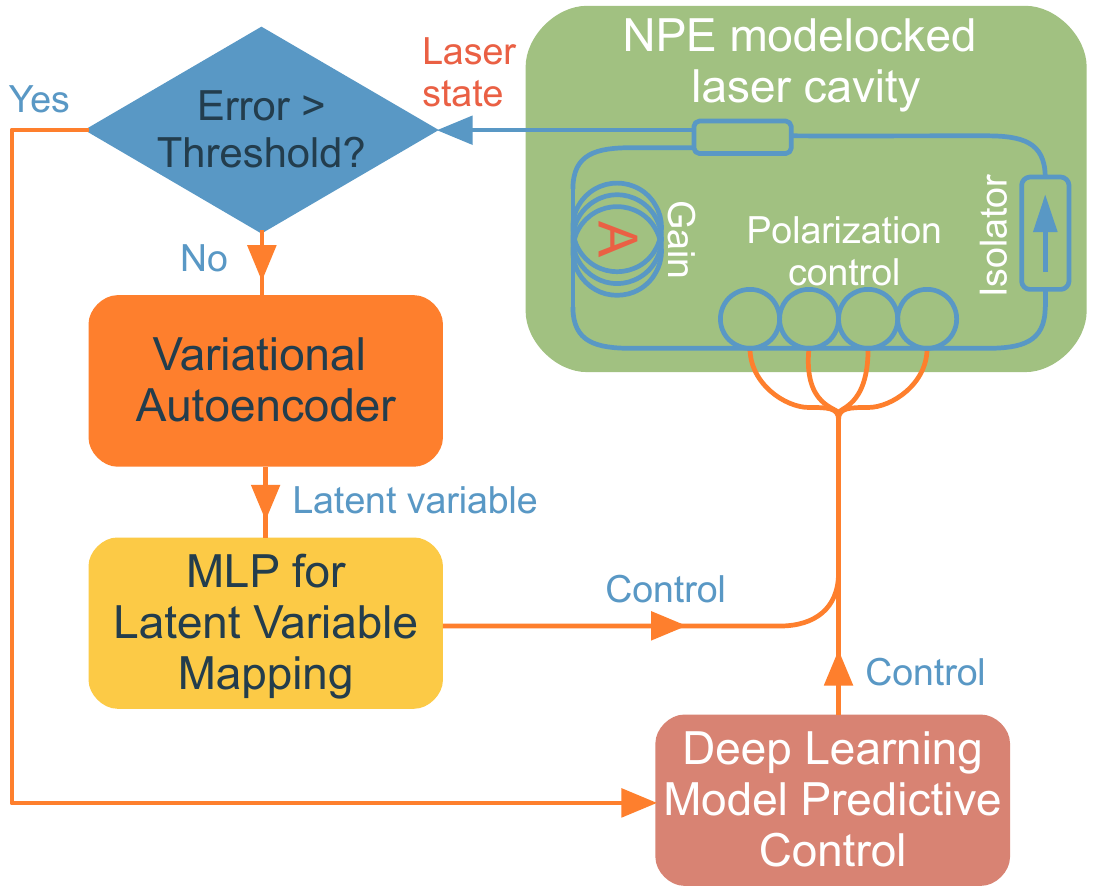}
\caption{Schematics of deep learning model predictive control circuit for hidden parameter retrieval and auto-mode-locking of a nonlinear polarization evolution fiber laser. Redrawn from \cite{baumeister2018deep}. \egor{Fig. 27 and Fig. 28 are swapped to ensure that figures appear in the numerical order in the text.}}
\label{fig: DL-MPC laser}
\end{figure}

The complexity of NNs used in laser dynamics varies from low-complexity classical MLPs for predicting performance parameters of a mode-locked laser \cite{NNLaser09} to sophisticated deep NN systems for hidden parameters' retrieval \cite{baumeister2018deep} and self-optimization of the lasing cavity \cite{andral2016toward}. Fig. \ref{fig: DL-MPC laser} shows a circuit for predictive control and hidden parameters' retrieval. There have been proposed even more complex NNs for deep reinforcement learning \cite{Sun_2020,NNLaser05,Li:22,NNLaser06}, which include both fully-connected and LSTM parts for finding various mode-locked states without any prior knowledge of the system. Dropout layers and pruning are both routinely used to reduce the computational complexity of these NNs.   In Ref.~\cite{NNLaser09}, the NN was used first to identify stable mode-locking regimes by considering the evolution of a small noise pulse. Then, the NN was trained to predict the pulse shape quickly and accurately. 

CNNs are also widely used for laser spatial distribution field analysis \cite{MD1, MD2, MD3}, for example, for measuring mode structure inside a multimode fiber or full beam characterization. These methods have become widespread, even though low-complexity algorithms \cite{MD4} are also available in this field. In the context of the multi-mode fiber lasers with complex spatio-temporal nonlinear dynamics \cite{MMLaser01,MMLaser02,MMLaser03}, the NNs are even more important and likely will play an increasing role in the design and optimization. The application of NNs for the design and optimization of an auto-setting mode-locked fiber laser cavity was discussed in \cite{andral2016toward}.

For more detailed reading, among numerous examples of the applications of NNs in laser science, we would like to mention the following most characteristic recent works:
\begin{itemize}
\item  
    \textbf{Nonlinear polarization evolution based lases:} NNs have been used for polarization control in so-called nonlinear polarization evolution (NPE) laser cavities that employ nonlinear polarization rotation as a mode-locking mechanism. This type of laser cavity is an excellent example of a complex nonlinear dynamical system governed by equations with "hidden" or non-measurable stochastic parameters such as birefringence. The pioneering works  \cite{Brunton2013ExtremumSeekingCO, brunton2014self}  demonstrated how optimization algorithms could be used to self-tune this type of lasers and make them "self-starting". More recent paper \cite{baumeister2018deep} introduced a method to extract the hidden stochastic parameter of birefringence and exploit them using deep learning and model predictive control.     Electronic initiation and optimization of the NPE evolution mode-locking in a fiber laser was studied in Ref.~\cite{winters2017NPE}. An intelligent programmable NPE-based mode-locked fiber laser with a human-like algorithm was proposed and demonstrated in Ref.~\cite{pu2019intelligent}.
\item \textbf{Double-gain, nonlinear amplifying loop mirror-based lasers:}
In Ref.~\cite{NNLaser01}, machine learning-based control was applied to 
a mode-locked fiber laser with a double-gain, nonlinear amplifying loop mirror, a modification of the figure-eight lasers. This system offers the possibility of combining simple electronic control of the nonlinearity and,
 consequently, the generated pulse properties with machine learning algorithms. 
\item  \textbf{Reduction of the number of the control measurement devices:} An important new feature that data-based approaches can bring to laser technology is the possibility of reducing the number of the control measurement devices. By combining NN-based data analysis and the dispersive Fourier transform, Ref.~\cite{NNLaser03} demonstrated the possibility of determining the temporal duration of picosecond-scale laser pulses using a nanosecond photodetector.  The trained NN makes it possible to predict the pulse duration with an average agreement of 95\%. This technique introduced in \cite{NNLaser03} paves the way to creating compact and low-cost feedback for complex laser systems.
\item  \textbf{Deep reinforcement learning in lasers:}
Application of NN-based methods generally requires arduous efforts and tuning of numerous hyperparameters. It is not obvious that the training in the laboratory environment can be smoothly transferred to the field applications or to the systems that differ from the specific laser used to develop the algorithm by design or environmental parameters. In  \cite{NNLaser05,NNLaser06}, it was demonstrated that a deep reinforcement learning (DRL) approach, based on trials and errors and sequential decisions, can be successfully used to control the generation of dissipative solitons in a mode-locked fiber laser system. It has been shown that deep Q-learning algorithms can be successfully applied to generalize knowledge about the laser system, assisting the search for conditions of  stable pulse generation. The region of stable generation was transformed by changing the pumping power of the laser cavity, while a tunable spectral filter was used as a control tool. The deep Q-learning algorithm is capable of learning the trajectory of adjusting spectral filter parameters to a stable pulsed regime, relying on the state of output radiation.  Application of the DRL for mode-locked lasers was also studied in \cite{Sun_2020} and in \cite{Li:22} with spectrum series learning control.
\end{itemize}
     
Further examples and details can be found in the recent overview of machine learning methods in ultrafast photonics \cite{R04}, which provides an insight into what are the control elements of different types of "smart" lasers, which quality measures are used to provide the feedback, and what type of algorithms can be used to control and tune such complex nonlinear dynamical systems as ultrafast lasers.
   
We anticipate that the next stage merging the data science techniques with laser technology will lead to the development of laser digital twins that will take full advantage of the available large data sets to create self-starting, self-tuned, robust laser systems.
Laser digital twins will consist of a set of adaptive models that emulate the complex light dynamics in a laser system, using real-time data to update its operation.  The laser digital twin will replicate the laser system to predict the characteristics of the output radiation and opportunities for its variation. It will also provide real-time recommendations for optimizing performance and mitigating unexpected events and situations.

\subsection{ Imaging and remote sensing}
Application of NNs in digital image processing, enhancing techniques from the statistical pattern recognition,  has a relatively long history (e.g., see \cite{Image1981,Image,Image0} and references therein).
Deep learning algorithms, in particular, deep CNNs  effectively became a methodology of choice for analyzing a stream of images, and the leading machine-learning tool for image classification and processing \cite{Image01,Image02,Image03,Image04,Image05,Image06}. 
CNNs are used for denoising \cite{yan2019fringe, lin2020optical}, enhancing image quality \cite{shi2019label, yu2020deep}, and separation of spectral channels \cite{zhang2010snapshot, qian2020single}.
Deep CNNs are capable of finding patterns and learning abstractions from raw data, in this case, images. Photonic technologies, on the other hand,  are capable to generate vast amounts of images in a short period of time. Therefore, deep learning methods are naturally well suited for merging with optical imaging and sensing techniques. 
Fast automatic and robust analysis of optical images is important for a variety of scientific and engineering applications, including medical imaging, quality control, hyperspectral analysis, nondestructive testing, image recognition for security, environmental science, bio-photonics, remote sensing, microscopy,  agri-tech, urban land-use, and many others.
Digital image processing completely transformed optical metrology making it possible to convert the results of measurements (typically displayed in the form of deformed fringe/speckle patterns) into the object characteristics, such as geometric coordinates, displacements, strain, refractive index, and so on. NN applications in image processing are so huge and developed area that we limit the discussion here to several references for more detailed reading (e.g, \cite{Imagebook01,Imagebook02,Imagebook03,Image01,Image02}) and a few particular examples.  

Deep learning algorithms are actively used in the remote sensing imagery   \cite{Sensor01,Sensor01a}. Deep learning has been applied for remote sensing image analysis tasks including image fusion, image registration, scene classification, object detection, land use and land cover classification, segmentation, and object-based image analysis \cite{Sensor01}. In particular, urban land-use mapping is one of the important though challenging problems in the field of remote sensing  \cite{Sensor01a}. Deep CNNs can substantially improve the accuracy and efficiency of the classification methods used for land-use information in urban areas. However, the conventional  CNNs uniformly decompose large images into small ones for processing, which is not ideal for land-use analysis applications. In \cite{Sensor01a}, a semi-transfer deep CNN technique was introduced to take advantage of three-channel  multispectral remote sensing images, maintaining the integrity of the land-use patterns. It is likely that remote sensing and imaging will be instrumental in developing digital twins for urban applications.  

Multi-mode fiber (MMF) has great potential to revolutionize the field of medical endoscopy and to become an optical tool of choice for endoscopic imaging due to the possibility of direct image transmission using multiple spatial modes. However, the propagating optical field is distorted by the fiber mode dispersion. Therefore, the usage of MMF as an imaging optical element, similar to a lens, requires calibration and the signal's post-processing. The NNs offer a natural approach to image reconstruction in MMF-based imaging, enhancing performance \cite{Caramazza2019,NNImage01,NNImage02,NNImage03}. 
Optical techniques allow transforming images to make processing more efficient. For instance, the high-speed all-fiber imaging combining transformation of two-dimensional spatial information into one-dimensional temporal pulsed streams by leveraging high intermodal dispersion in a multimode fiber and image reconstruction using deep learning have been demonstrated in \cite{Liu2022}.
 The fiber probe detected micron-scale objects with a high frame rate (15.4~Mfps) and large frame depth (10,000). The scheme proposed in that Ref. combines high speed with high mechanical flexibility and integration, making it attractive for various in vivo applications.

Imaging in the infrared spectral interval is of high interest for numerous applications. In recent work on the hyperspectral imaging of narrow absorption lines \cite{Voumard:20}, a machine learning technique was utilized to reconstruct images obtained with a massively parallelized infrared detection. The gathering and subsequent processing of optical images make it possible to characterize the processes occurring in systems without introducing measuring elements directly into the system. For example, remote sensing using NNs for aerial/satellite images scene classification is used for forest fire monitoring \cite{Lentile2006, DALDEGAN2019111340}. It has become increasingly used in agricultural technologies and for monitoring ecosystems. Ref.~\cite{reviewCNNvegetationRemoteSensing} provides a review of CNN-based approaches to the characterization of both spatial and temporal vegetation patterns. It is shown that CNNs are very effective in extracting a wide array of vegetation properties from remote sensing imagery. Another deep learning approach \cite{Reichstein2019} has been applied to obtain more information about global climate processes occurring on Earth, as well as to improve the accuracy of predicting seasonal fluctuations in observed atmospheric parameters and to capture long-range relationships of climate processes on various time scales.
Overall, optical imaging and remote sensing is the area where
merging of photonic techniques and NN processing is already happening, leading to the emerging field of digital optical imaging.

\subsection{Neural networks for the design of new photonic materials}
The design of new materials and structures (photonic crystals, meta-materials, and meta-surfaces, to mention a few) is one of the most important progress drivers in photonics.  
The direct problem of material design, in general, is extremely computationally intensive as it requires electromagnetic modeling based on numerical full-wave simulation methods (such as the finite-difference time-domain method, the finite-element method, and others, see Refs.~\cite{Material,Material01,Material02}) and then lots of trials together with global optimization techniques to find a good design. There exists a plethora of different advanced optimization techniques. However, combined with the direct electromagnetic solvers with high computational costs, these numerical frameworks are too time-consuming and inefficient in complex design tasks. Moreover, the performance of the conventional optimization techniques degrades with the increase of the number of additional constraints, limiting the practical applicability further. The inverse design \cite{Material03,Material04}, dealing with finding the appropriate optical material structure that can provide for the desired properties, is an even harder task insofar as it requires the search within a much larger design space. 

NNs brought computational effectiveness in this field \cite{NNMaterial01,NNMaterial02,NNMaterial04,NNMaterial05} since the NNs can be much less computationally complex than traditional approaches. The NNs have found numerous applications in material design, from predicting new materials and revealing hidden relations to the construction and optimization of nanophotonic and metamaterial structures to manipulate electromagnetic fields at the subwavelength scales.  However, a typical NN used in material design is still pretty complex, consisting of millions of trainable weights,  and complexity reduction in this field is in high demand.

\begin{figure}[ht!]
\centering
\includegraphics[width=.95\linewidth]{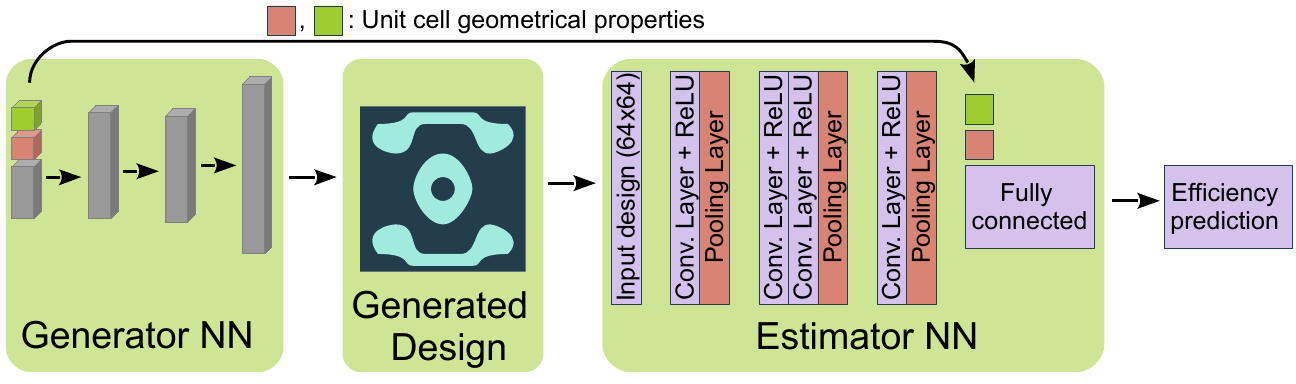}
\caption{Generative and estimating neural networks for global optimization of plasmonic nanoantennas. Redrawn from \cite{kudyshev2020machine2}.}
\label{fig: NNs for material design}
\end{figure}

In Ref.~\cite{kudyshev2020machine2}, the authors utilized generative and discriminative NNs for the efficient and fast optimization of plasmonic metamaterials. Fig.~\ref{fig: NNs for material design} shows the schematics of the NN-based optimization routine employed in the aforementioned Ref.

Several reviews \cite{InverseDesignML2021,NNMaterial01} provide a summary of machine learning techniques in material design and show how deep NNs, configured as discriminative networks, can learn from training sets and operate as high-speed surrogate electromagnetic solvers. These are often used to speed up the process of numerical simulations of the proposed layouts, which usually requires costly simulation techniques when Maxwells and/or material equations are solved using mesh methods such as FDTD or FEM.

For example, a physics-inspired loss function can be used to train a deep NN \cite{MaxwellNet2022} that is capable of solving partial differential equations without employing computationally costly mesh methods. Deep learning is often utilized for inverse design in nano-photonics where the accurate mathematical description of many problems is still challenging. A recent review \cite{Wiecha:21} of machine learning and, in particular, deep learning methods used for inverse design in nano-optics provides a good insight into how NNs have changed the field of material design and have gone beyond that with solving partial differential equations, interpreting physical properties and photonics experiments. Deep NNs and generative models are also employed for the design and optimization of photonic splitters \cite{Tahersima2019}. More examples of how deep NNs are utilized in the design of photonic materials, polymers, and other materials \cite{ma15051811} or designing new photonic structures \cite{R09} show the broad interest in applying NNs in this field.

Jiang and Fan have shown \cite{jiang2019global} how a physics-inspired conditional generative NN can be applied for global optimization of the topology of metasurfaces. Two NNs were trained on forward and adjoint electromagnetic simulations. The authors show that the topologies obtained by the developed approach have advantages over the ones obtained by the standard global optimization-based approach while being more energy efficient.

In a paper on machine learning framework for a quantum sampling of highly constrained, continuous optimization problems \cite{MLframeworkQuantumSampling2021}, the authors developed a machine learning framework that maps inverse design optimization problems into surrogate quadratic unconstrained binary optimization problems by employing a binary variational autoencoder and a factorization machine. Then, using the D-Wave advantage hybrid sampler and simulated annealing, the authors demonstrate how diffractive meta-gratings for highly efficient beam steering can be developed.

The perspectives of deep NNs in photonics and photonic materials are summarized in Ref.~\cite{R06} which describes the future opportunities for machine learning methods in the domains of photonics, nanophotonics, plasmonics, and photonic materials' discovery, including metamaterials.

\section{Reducing the complexity of neural networks}\label{sec:reduce}

In the ML model development stage, the primary challenge is often the time consumption of the NN model. From a time complexity perspective, the focus is on minimizing the computational resources required during training and inference. This entails reducing the number of operations and memory requirements, thus enabling faster and more efficient model execution. Time-critical applications such as real-time image or speech recognition often prioritize time complexity optimization to ensure timely and responsive predictions. In this sense, complex models with high parallelization can reduce the time complexity, and the use of GPUs and batch processing can further enhance efficiency. However, it's important to balance model complexity with training efficiency, as highly complex models can require a large amount of data and time to train.

In contrast, signal processing problems pose specific challenges when it comes to the implementation complexity of NN models. In these scenarios, it is crucial to carefully consider how the NN models are implemented and deployed. The inference complexity of NNs, which encompasses the computational requirements during the prediction phase, assumes a critical role in signal processing applications. One key aspect to consider in signal processing is the impact of inference complexity on power consumption. As the complexity of the model increases, the computational demands also rise, leading to higher power consumption. This is a significant concern, particularly in resource-constrained environments or energy-efficient systems. By reducing the complexity of the NN model, such as minimizing the number of layers or parameters, it becomes possible to achieve a more efficient signal processing system that consumes less power without compromising performance. Another vital consideration is the effect of inference complexity on time delay. 

In real-time signal processing applications, prompt and timely responses are imperative. Complex NN models often require longer inference times, which can introduce undesirable delays. By optimizing the model's complexity, such as utilizing efficient algorithms or reducing redundant computations, the inference time can be minimized, ensuring timely predictions and responses to signal processing tasks. Additionally, the ability to adapt rapidly is of utmost importance in signal-processing applications. The characteristics of the transmitted signals or the environment in which the processing takes place may vary over time. Therefore, the NN model must possess the capability to quickly adapt and adjust to these changes to maintain optimal performance. This adaptability can be achieved through techniques such as online learning, where the model learns and updates itself in real-time based on incoming data.

Determining the relative importance of time complexity and implementation complexity depends on the unique requirements and constraints of the application under consideration. In situations where promptness is of the uttermost importance, it may be preferable to sacrifice implementation complexity for greater temporal efficiency. In situations where power consumption and hardware dimensions are of critical importance, a harmonious balance between time complexity and optimization of implementation complexity is required. Consequently, we will now evaluate the techniques used to reduce complexity in both the training and inference stages, as well as the metrics that can be used to evaluate their positive impact on NN model design.

\subsection{Computational complexity metrics for Training and Inference Stages}

In the domain of training neural networks, the evaluation of training complexity traditionally revolves around two key metrics: the number of trainable parameters within the NN and the time required for training to attain a specific performance level. Although both metrics offer some insight into training complexity, it is important to note that two NNs with an equal number of trainable parameters can exhibit distinct training complexities, as exemplified in the work referenced as Ref.~\cite{freire2022pitfalls}.
Moreover, evaluating training time as a metric is somewhat challenging due to its dependence on hardware resources utilized for training and the size of the training dataset. To address these limitations, two additional metrics have been developed to bridge the gaps. The first metric, known as NENB (number of epochs × number of batches), combines the number of training epochs and the number of data subsets (batches) used in each epoch. Generally, an increased number of training epochs implies a more intricate and computationally demanding model. The number of batches refers to the subsets of data employed during each epoch, and is influenced proportionally by the dataset size and batch size. Consequently, NENB can serve as a suitable metric for comprehensively assessing training complexity. Evaluating the number of epochs or batches individually would fail to provide a holistic evaluation. For instance, one model may require more epochs but fewer batches, while another model might necessitate fewer epochs, but a larger number of batches.

\begin{figure}[ht!]
    \centering
    \includegraphics[width=0.9\linewidth]{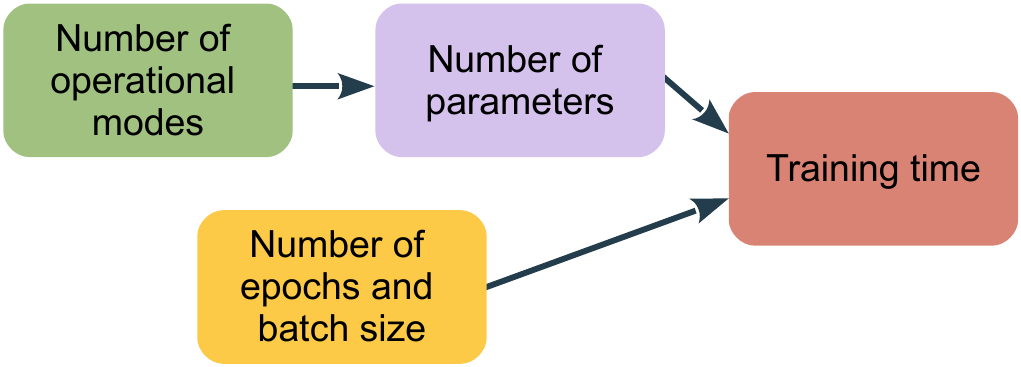}
    \caption{Training complexity metrics and their dependencies.}
    \label{fig:training complexity}
\end{figure}

The second metric, in its objective to gauge the versatility and generalization capabilities of the model, seeks to quantify the count of operational ranges in which the NN equalizer can effectively function with an acceptable level of gain. This metric provides valuable insights into the NN's ability to adapt and perform across various scenarios and contexts. When the NN is constrained to a specific task or domain, it becomes necessary to frequently retrain the model to ensure its proficiency in handling new or evolving situations. This need for frequent retraining adds to the overall complexity of the system, as it demands additional computational resources, time, and effort to maintain the desired performance levels. By considering this aspect of the NN's operational scope, the second metric offers a valuable perspective on the adaptability and complexity of the model within its intended application domain.

Moving to the computational complexity of the inference, from a computer science perspective, computational complexity analysis is almost always attributed to the Big-$O$ notation of the algorithm \cite{maass1994computational,alizadeh2020managing,wiedemann2019compact}. In general, the Big-$O$ notation is used to express an algorithm's complexity while assessing its efficiency, which means that we are interested in how effectively the algorithm scales with the size of the dataset in terms of running time \cite{amin2008analysis,kerr2005big, blondel2000survey}. However, from the engineering standpoint, the Big-$O$ is often an oversimplified measure that cannot be immediately translated into the hardware resources required to realize the algorithm (NNs) in a hardware platform \cite{gysel2016hardware}.

\begin{figure}[ht!]
\centering
\includegraphics[width=.55\linewidth]{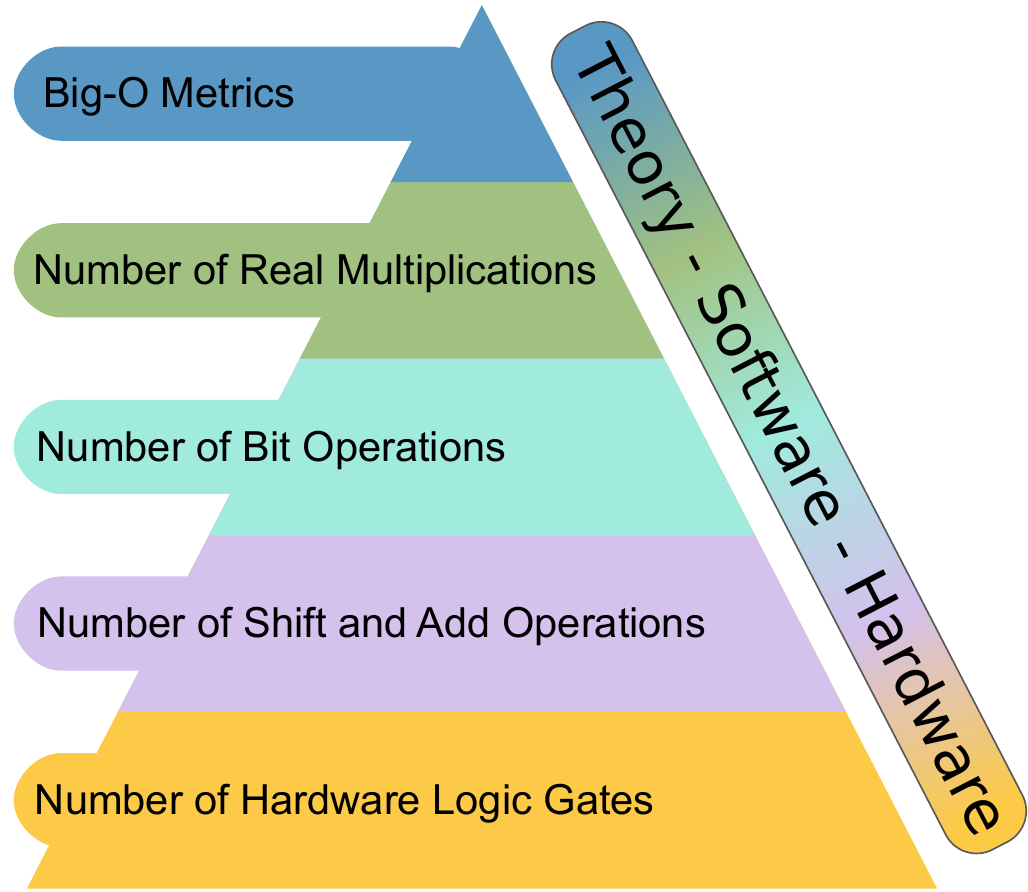}
\caption{Different levels of inference computational complexity metrics: from software notions down to hardware logic elements.}
\label{fig: computational complexity pyramid}
\end{figure}

Due to the absence of some ``universal'' complexity measure, various works started to present complexity in terms of multiply and accumulate (MAC) \cite{sze2017efficient, gysel2016hardware,li2017reducing,yang2017designing}, Kolmogorov complexity \cite{balcazar1997computational}, the number of bit-operations (BOP) \cite{van2020bayesian, baskin2021uniq}, the number of real multiplications (RM) \cite{deligiannidis2021performance, sidelnikov2018equalization, freire2021performance}, number of shift and add operations \cite{freire2022computational}, and number of hardware logic gates \cite{sahin2006neural, 5280233}. In this paper, we have summarized a sequence of useful complexity matrices going from a software level to a hardware level, which is also depicted in Fig.~\ref{fig: computational complexity pyramid}.

The first, most software-oriented, level of estimation traditionally deals only with counting the RM number of the algorithm \cite{jacobsen2007fast,spinnler2010equalizer} (quite often defined per one processed element, say a sample or a symbol).  When comparing computational complexity, the purpose of this high-level metric is to consider only the multipliers required, ignoring additions, because the implementation of the latter in hardware or software is initially considered cheap, while the multiplier is generally the slowest element in the system and consumes the largest chip area \cite{mirzaei2006fpga,jacobsen2007fast}. This ignoring of the additions can also be easily understood by looking at the Big-$O$ analysis of multiplier versus adder. When multiplying two integers with $n$ digits, the computational complexity of the multiplication instance is $O(n^2)$, whereas the addition of the same two numbers has a computational complexity of $\Theta(n)$ \cite{jahani2009zot}\footnote{The Big-$O$ notation represents the worst case or the upper bound of the time required to perform the operation, Big Omega ($\Omega$) shows the best case or the lower bound whereas the Big Theta ($\Theta$) notation defines the tight bound of the amount of time required; in other words, $f(n)$ is claimed to be $\Theta(g(n))$ if $f(n)$ is $O(g(n))$ and $f(n)$ is $\Omega(g(n))$.}. As a result, if we deal with float values with 16 decimal digits, multiplication is by far the most time-consuming part of the implementation procedure.
Therefore, when comparing solutions that use floating-point arithmetic with the same bitwidth precision, the RM metric provides an acceptable comparative estimate to qualitatively assess the complexity against some existing benchmarks (e.g., against the DSP operations for optical channel equalization tasks~\cite{spinnler2010equalizer}).

When moving to fixed-point arithmetic, the second metric, known as the number of bit-operations (BOP), must be adopted to understand the impact of changing the bitwidth precision on the complexity. The BOP metric provides a good insight into mixed-precision arithmetic performance since we can forecast the BOP needed for fundamental arithmetic operations like addition and multiplication, given the bitwidth of two operands. In a nutshell, the BOP metric aims to generalize floating-point operations (FLOPs) to heterogeneously quantized NNs, as far as the FLOPs cannot be efficiently used to evaluate integer arithmetic operations~\cite{baskin2021uniq, hawks2021ps}. For the BOP metric, we have to include the complexity contribution of both multiplications and additions, since now we evaluate the complexity in terms of the most common operations in NNs: the multiply-and-accumulate operations (MACs) \cite{baskin2021uniq, hawks2021ps,wu2018deep}. However, the BOP accounts for the scaling of the number of multipliers with the bitwidth of two operands and the scaling of the number of adders with the accumulator bitwidth. Note that since most real DSP implementations use dedicated logic macros (e.g., DSP slice in Field Programmable Gate Arrays [FPGA] or MAC in Application Specific Integrated Circuit [ASIC]), the BOP metric fits as a good complexity estimation inasmuch as the BOP also accesses the MAC taking into account the particular bitwidth of two operands.

 The progress in developing new advanced NN quantization techniques \cite{li2019additive,Koike2021,elhoushi2021deepshift,you2020shiftaddnet} allowed implementing the fixed point multiplications participating in NNs efficiently, namely with the use of a few bit-shifters and adders \cite{gentili1995efficient, evans1994efficient, lee2003frequency}.  Since the BOP cannot properly assess the effect of different quantization strategies on the complexity, a new, more sophisticated metric is required.
 We can introduce the third complexity metric that counts the number of total equivalent additions to represent the multiplication operation, called the number of additions and bit shifts (NABS)\cite{freire2022computational}. The number of shift operations can be neglected when calculating the computational complexity because, in the hardware, the shift can be performed without extra costs in constant time with the $O(1)$ complexity. Even though the cost of bit shifts can be ignored due to the aforementioned reasons, and only the total number of adders has to be accounted for to measure the computational complexity, we prefer to keep the full NABS name to highlight that the multiplication is now represented as shifts and adders.

Finally, the metric which is closest to the hardware level is the number of logic gates (NLG) that is used for our evaluating method's hardware (e.g., ASIC or FPGA) implementation. It is different from the NABS metric, as it now reflects the true cost of implementation in particular hardware. In this case, in contrast to the other complexity metrics, the cost of activation functions is also taken into account because, to achieve better complexity, they are frequently implemented using look-up tables (LUT) rather than adders and multipliers. Additionally, other metrics like the number of flip-flops (FFs) or registers, the number of logic blocks used for general logic and memory blocks, or other special functional macros used in the design are also relevant. As it is clear from this explanation, we cannot present a straightforward equation to convert the NABS to NLG, as the latter depends on the circuit design adopted by the developer: special tools like Synopsys Synthesis \cite{kurup2012logic} for ASIC implementation can provide this information. However, concerning the FPGA design, it is harder to get a correct estimate of the gate count from the report of the FPGA tools \cite{li2016efficient}.

From an analytical perspective, Table~\ref{Table:Main} presents a quantitative analysis of the primary layers utilized in machine learning, evaluated using the following complexity metrics: RM, BOP, and NABs. The equations listed in the table are derived based on the input and output dimensions of the layers, the assumed bitwidth for operations, and the design hyperparameters specific to these neural network layers. For a detailed explanation of the methodologies used to compute these metrics, refer to the work by Ref.~\cite{freire2022computational}, which also discusses the computational complexity of additional layers such as RESNET and Transformers.

\begin{table*}[htbp]
\caption{Summary of the three computational complexity metrics per layer (the number of real multiplications, the number of bit operations, the number of additions and bit shifts) for a zoo of neural network layers as a function of their designing hyper-parameters; the number of neurons ($n_n$),  the number of features in the input vector ($n_i$), the number of filters ($n_f$), the kernel size ($n_k$), the input time sequence size ($n_s$), the number of hidden units ($n_h$), the number of internal hidden neuron units of the reservoir ($N_r$), sparsity parameter ($s_p$), the number of output neurons ($n_o$), 
weight bitwidth ($b_w$), input bitwidth ($b_i$), activation bitwidth ($b_a$), and the number of adders required at most to represent the multiplication ($X_w$)} \label{tab:formulas}
\resizebox{\textwidth}{!}{
\begin{tabular}{|c|c|c|c|}
\hline
Network type & Real multiplications (RM)                & Number of  bit-operations (BOP)                                                                                                                                                                                                                                      & Number of additions and bit shifts(NABS)                                                                                                                                                                                                                                                                       \\ \hline\hline
MLP          & $n_{n}n_{i}$                             & $n_{n}n_{i}\big[b_{w} b_{i} + \text{Acc}(n_{i}, b_{w}, b_{i})\big]$                                                                                                                                                                                                  & $n_{n}n_{i}(X_{w} +1)\text{Acc}(n_{i}, b_{w}, b_{i})$                                                                                                                                                                                                                                                          \\ \hline
1D-CNN       & $n_{f} n_{i} n_{k} \cdot Output Size$    & \begin{tabular}[c]{@{}c@{}}$OutputSize \cdot n_{f}\text{Mult}(n_{i}n_{k}, b_{w}, b_{i})$\\ $+n_{f}\text{Acc}(n_{i}n_{k}, b_{w}, b_{i})$\end{tabular}                                                                                                                 & \begin{tabular}[c]{@{}c@{}}$OutputSize \cdot n_{f}\big[n_{i}n_{k}(X_{w} + 1 )-1\big]$ \\ $\cdot\text{Acc}(n_{i}n_{k}, b_{w}, b_{i})$\\ $+n_{f}\text{Acc}(n_{i}n_{k}, b_{w}, b_{i})$\end{tabular}                                                                                                               \\ \hline
Vanilla RNN  & $n_{s}n_{h}(n_{i} + n_{h})$              & \begin{tabular}[c]{@{}c@{}}$n_{s}n_{h}\text{Mult}(n_{i}, b_{w}, b_{i})$\\ $+ n_{s}n_{h}\text{Mult}(n_{h}, b_{w}, b_{a})$\\ $+2n_{s}n_{h}\text{Acc}(n_{h}, b_{w}, b_{a})$\end{tabular}                                                                                & \begin{tabular}[c]{@{}c@{}}$n_{s}n_{h}\big[ n_{i}(X_{w} + 1) -1\big]\text{Acc}(n_{i}, b_{w}, b_{i})$\\ $+ n_{s}n_{h}\big[ n_{h}(X_{w} + 1) +1\big]\text{Acc}(n_{h}, b_{w}, b_{a})$\end{tabular}                                                                                                                \\ \hline
LSTM         & $n_{s}n_{h}(4n_{i} + 4n_{h} + 3)$        & \begin{tabular}[c]{@{}c@{}}$4n_{s}n_{h}\text{Mult}(n_{i}, b_{w}, b_{i})$\\ $+ 4n_{s}n_{h}\text{Mult}(n_{h}, b_{w}, b_{a})$\\ $+ 3n_{s}n_{h}b_{a}^2$\\ $+ 9n_{s}n_{h}\text{Acc}(n_{h}, b_{w}, b_{a})$\end{tabular}                                                    & \begin{tabular}[c]{@{}c@{}}$4n_{s}n_{h}\big[n_{i}(X_{w} +1)-1\big]\text{Acc}(n_{i}, b_{w}, b_{i})$\\ $+ 4n_{s}n_{h}\big[n_{h}(X_{w} +1) +1\big]\text{Acc}(n_{h}, b_{w}, b_{a})$\\ $+ 6n_{s}n_{h}b_{a}$\end{tabular}                                                                                            \\ \hline
GRU          & $n_{s}n_{h}(3n_{i} + 3n_{h} + 3)$        & \begin{tabular}[c]{@{}c@{}}$3n_{s}n_{h}\text{Mult}(n_{i}, b_{w}, b_{i})$\\ $+ 3n_{s}n_{h}\text{Mult}(n_{h}, b_{w}, b_{a})$\\ $+3n_{s}n_{h}b_{a}^2$\\ $+ 8n_{s}n_{h}\text{Acc}(n_{h}, b_{w}, b_{a})$\end{tabular}                                                     & \begin{tabular}[c]{@{}c@{}}$3n_{s}n_{h}\big[n_{i}(X_{w} + 1)-1\big]\text{Acc}(n_{i}, b_{w}, b_{i})$\\ $+ n_{s}n_{h}\big[3n_{h}(X_{w} + 1)+5\big]\text{Acc}(n_{h}, b_{w}, b_{a})$\\ $+ 6n_{s}n_{h}b_{a}$\end{tabular}                                                                                           \\ \hline
ESN          & $n_sN_{r}(n_{i} + N_{r}s_{p}+2 + n_{o})$ & \begin{tabular}[c]{@{}c@{}}$n_{s}N_{r}\text{Mult}(n_{i}, b_{w}, b_{i})$\\ $+ n_{s}N_{r}s_{p}\text{Mult}(N_{r}, b_{w}, b_{a})$\\ $+ n_{s}N_{r}\text{Mult}(n_{o}, b_{w}, b_{a})$\\ $+2n_{s}N_{r}b_{a}^2$\\ $+ 4n_{s}N_{r}\text{Acc}(N_{r}, b_{w}, b_{a})$\end{tabular} & \begin{tabular}[c]{@{}c@{}}$n_{s}N_{r}\big[n_{i}(X_{w} + 1) -1\big]\text{Acc}(n_{i}, b_{w}, b_{i})$\\ $+ n_{s}N_{r}\big[s_{p}(N_{r}X_{w} + N_{r} -1\big] +4 )\text{Acc}(N_{r}, b_{w}, b_{a})$\\ $+ n_{s}N_{r}\big[n_{o}(X_{w} + 1) -1\big]\text{Acc}(n_{o}, b_{w}, b_{a})$\\ $+ 4n_{s}N_{r}b_{a}$\end{tabular} \\ \hline
ResNet       & Ref.~\cite{freire2022computational} Eq.~(36)                                                                                                                                & Ref.~\cite{freire2022computational} Eq.~(38)                                                                                                                                                                                                                                      & Ref.~\cite{freire2022computational} Eq.~(40)                                                                                                                                                                                                                                                                                      \\ \hline                                                                        
Transformer  & \begin{tabular}[c]{@{}c@{}}MultiHead: Ref.~\cite{freire2022computational} Eq.~(47)\\ Point-wise FFN: Ref.~\cite{freire2022computational} Eq.~(50)\\ Add\&Norm: Ref.~\cite{freire2022computational} Eq.~(53)\end{tabular}   & \begin{tabular}[c]{@{}c@{}}MultiHead: Ref.~\cite{freire2022computational} Eq.~(48)\\ Point-wise FFN: Ref.~\cite{freire2022computational} Eq.~(51)\\ Add\&Norm: Ref.~\cite{freire2022computational} Eq.~(54)\end{tabular}                                                                                                              & \begin{tabular}[c]{@{}c@{}}MultiHead: Ref.~\cite{freire2022computational} Eq.~(49)\\ Point-wise FFN: Ref.~\cite{freire2022computational} Eq.~(52)\\ Add\&Norm: Ref.~\cite{freire2022computational} Eq.~(55)\end{tabular}                                                                                                                                    		   \\ \hline
\end{tabular}}
\label{Table:Main}
\end{table*}

\begin{figure}[ht!]
    \centering
    \includegraphics[width=\linewidth]{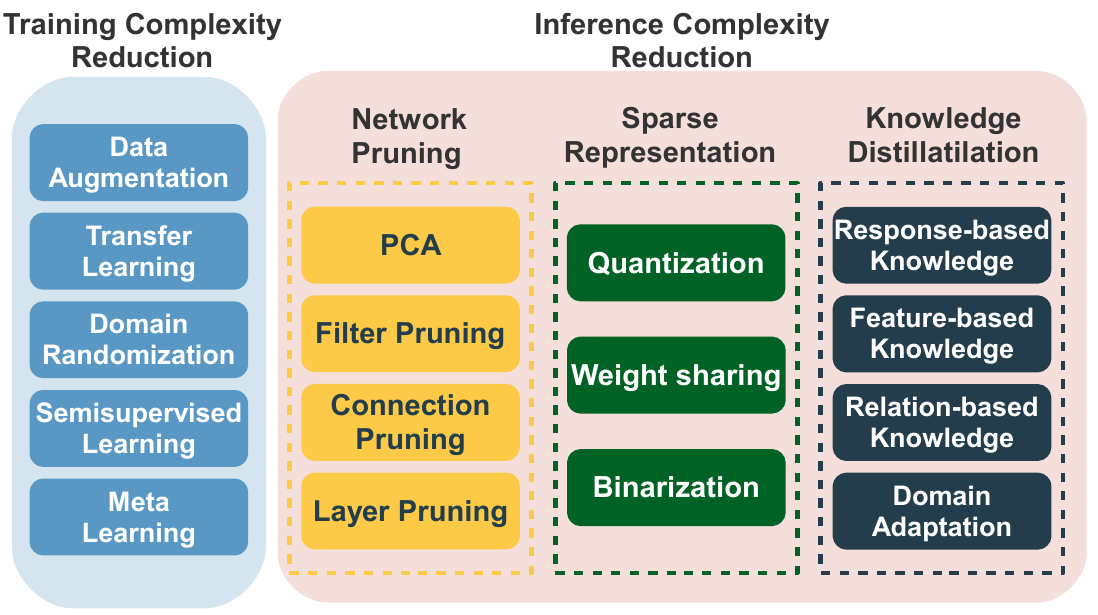}
    \caption{Overview strategies for complexity reduction in the training and inference phases when deploying NN-based solutions in Photonics.}
    \label{fig:reductioncomplexity1}
\end{figure}

\subsection{Reducing the complexity of training}
In many applications, when designing an NN structure with some particular purpose, we, first and foremost, pay attention to the performance of the respective model. Typically, we expect that this performance is better than some established benchmark: for instance, the performance of post-equalizers is gauged against the digital back propagation method with some number of steps. However, when considering the implementation aspects, the ultimate cost of the processing chain has to be taken into account, i.e., we need to assess the computational complexity of our NN. When talking about NN-based devices, we can distinguish two important factors related to the NN complexity: the complexity of training, which is often omitted as the training is assumed to be made off-line, and the complexity of inference, i.e., the complexity associated with the on-line optical signal processing for the subject considered. In this section, we devote attention to both directions insofar as the training complexity can be associated with the reconfigurability of an NN device, indicating how the device can readjust itself for some changes in the usage environment. The overview of the strategies for complexity reduction in the training and inference is shown in Fig.~\ref{fig:reductioncomplexity1}.

 \subsubsection{Data augmentation}
Data augmentation is the technique of producing additional data points from the current data obtained to artificially increase the amount of available data. Data can be augmented in a number of ways, e.g., by making small modifications or by employing machine learning models to generate new data points in the latent space of original data.  Having a large dataset is essential for the effectiveness of machine learning and deep learning models. In a nutshell, data augmentation is a technique used in machine learning to increase the size of a dataset by generating additional data points based on the existing data, reducing the complexity of the learning process by providing the model with more examples to learn from, which can improve the model's generalization and reduce the risk of overfitting.

Data augmentation is often used in image classification tasks, where the model is trained to recognize objects in images. By generating additional images that are slightly modified versions of the original images (e.g., by rotating, cropping, or adding noise), the model is able to learn more about the features that are relevant to the task, as well as how to be robust to small variations in the data.

 In optical communications, the data augmentation has been recently considered in network scenarios for predicting failures~\cite{cui2019deep,zhuang2020machine} and traffic peculiarities~\cite{li2019adaptive, li2019deep}. The aforementioned network applications suggested new object generation by GANs~\cite{zhuang2020machine, li2019adaptive, li2019deep} utilizing the heuristics~\cite{cui2019deep}. It is worth noting that training supervised learning algorithms for every particular task require a unique dataset structure and, hence, a unique data augmentation procedure. Therefore, the aforementioned data augmentation techniques from the networking layer are not applicable to signal distortion mitigation at the physical layer of optical communications. Only in Ref.~\cite{9333417}, the data augmentation technique was successfully used for improving the training of supervised learned algorithms for the compensation of nonlinear distortion compensation in fiber-optic communication systems. In this case, it was shown, both numerically and experimentally, that data transformations that account for underlying propagation equation symmetries (e.g., Manakov equation) can be used to synthetically expand the training dataset. 

\subsubsection{Transfer learning}
Transfer learning is a machine learning framework that uses a pre-trained model as a starting point to solve a new task rather than training a model from scratch. It can reduce the complexity of the learning process by leveraging the knowledge learned by the pre-trained model and adapting it to the new task.

Transfer learning is often used when there is a limited amount of labeled data available for the new task, as the pre-trained model can provide a good starting point for the new task even with a small amount of labeled data. It can also be used to reduce the number of computations required to train a new model, as the pre-trained model has already learned many of the general patterns and features that are relevant to the new task. This procedure is more likely to succeed if the features are universal or applicable to both the base and target tasks.

Transfer learning in optical networks has been mainly used for optical signal-to-noise ratio (OSNR) monitoring. In~\cite{mo2018ann}, this application was introduced using an artificial NN-based transfer learning approach to accurately predict the QoT of different optical networks without re-training NN models from scratch. In that paper, the source domain was a 4$\times$80~km (4 spans) large effective area fiber (LEAF) link using QPSK modulation. The target domain was the same system but with a different number of spans (propagation distance) and different modulations formats (4$\times$80~km LEAF with 16-QAM; 2$\times$80~km LEAF with 16-QAM; and 3$\times$80~km dispersion-shifted fiber with QPSK). The results showed that when using transfer learning, just 2\% of the original training dataset size was enough to calibrate the NN for the new target domain. More recently, in~Ref.~\cite{cheng2020transfer}, the experimental demonstration of the application of transfer learning for joint OSNR monitoring and
modulation format identification from 64-QAM signals was presented. It was shown that by implementing the transfer of learning from simulation
to experiment, the number of training samples and epochs needed for the same prediction quality was reduced by 24.5\% and 44.4\%, respectively. 
Another recent application of transfer learning was in the spectrum optimization problem for resource reservation~\cite{yao2019transductive}. To predict a spectrum defragmentation time, the pre-trained NN model for a source domain (having a 6-node topology) was transferred and trained again using the data from the target domain (the NSFNet with 14 nodes). It was shown that by using this technique, the proportion of affected services was reduced, the overall likelihood of resource reservation failure was diminished, and the spectrum resource utilization was improved.

Only a few works have addressed transfer learning for nonlinearity mitigation, and these mostly focus on short-haul IM-DD systems. In~\cite{xu2020feedforward}, the successful transfer of the knowledge for the links with different bit rates and fiber lengths was demonstrated. Both feedforward and recurrent NNs were tested for the transfer learning application: about 90\% (feed-forward) and 87.5\% (recurrent) reduction in epochs were achieved, and 62.5\% (feed-forward) and 53.8\% (recurrent) reduction in training symbols were demonstrated. Another work in direct detection, Ref.~\cite{zhang2019fast}, applied the transfer from 5~dBm launch power to other powers (ranging from -7~dBm to 9~dBm) and from one transmission distance (640~km) to other ones (from 80~km up to 800~km). The experimental results showed that the iterations with transfer learning constitute approximately one-fourth of the full NN training iterations. Additionally, the transfer learning did not result in a performance penalty in a five-channel transmission when transferring the learned features from training just the middle channel to the four other channels. 

Finally,  the transfer learning in coherent optical systems was first investigated in Ref.~\cite{zhang2020nonlinear}. In that work, the authors applied the TL for different launch powers but provided a very brief explanation of the technique. However, only in Ref.~\cite{9523752} a comprehensive description of how the transfer learning can be efficiently used to realize flexible NN equalizers for adaptation to changes in launch power, modulation format, symbol rate, and fiber setup, was presented.

 \subsubsection{Domain randomization}
Domain randomization is a technique used in machine learning to train models that are robust to changes in data distribution. It involves training a model on a wide range of simulated data that is randomly generated within a certain domain, rather than training on real-world data\cite{tobin2017domain}.
By training on a wide range of simulated data, the model is able to learn the general patterns and features that are common across the entire domain, rather than being specifically adapted to the data distribution of a particular dataset. This can make the model more robust to changes in the data distribution and less prone to overfitting\cite{chen2021understanding}.

The usage of domain randomization can reduce the complexity of the training process by decreasing the amount of real-world data required to train a model. It can also reduce the need for extensive data preprocessing, as the simulated data is generated randomly and does not need to be cleaned or normalized\cite{muratore2022robot}.

 In particular, when talking about the optical channel equalization task, by using domain randomization, we can train the model in such a way that it can successfully work for different baud rates, powers, etc.~\cite{freire2022domain}. Quite often, the randomization is coupled with domain adaptation techniques, e.g., with transfer learning.

\subsubsection{Other approaches}
In order to minimize the time and effort needed to train NN, we can use some other approaches. The meta-learning method \cite{simeone2020learning} is the first option. Meta-learning is a machine learning technique that involves learning how to learn or learning the process of improving a learning system. It aims to reduce the complexity of the learning process by learning common patterns across various tasks and using this knowledge to improve the performance of learning algorithms. One way that meta-learning can be used to reduce training complexity is by pre-training a model on a large dataset and then fine-tuning it on a specific task, rather than training a model from scratch for each task. This can reduce the amount of data and computation required to train a model for a specific task, as the model has already learned many of the general patterns that are common across the tasks. Another way that meta-learning can be used to reduce the training complexity is by learning to adapt the model's architecture or hyperparameters based on the specific task at hand. This can allow the model to automatically adjust its complexity to the needs of the task, rather than requiring manual tuning of the model's architecture or hyperparameters.

The second possibility is to use semi-supervised learning techniques \cite{ouali2020overview}. Semi-supervised learning is a machine learning method that uses both labeled and unlabeled data to train a model. It can be used to reduce the complexity of the learning process by reducing the amount of labeled data required to train a model. In supervised learning, the model is trained using a large dataset of labeled examples, where each example has a known correct output. However, collecting and labeling a large dataset can be a time-consuming and expensive process. In semi-supervised learning, a smaller amount of labeled data is used in conjunction with a larger amount of unlabeled data. The model is trained to make predictions on the labeled data, and the predictions are then used to label the unlabeled data. This process is repeated until the model acquires the ability to make accurate predictions on the entire dataset.

Using semi-supervised learning can reduce the amount of labeled data required to train a model, as the model is able to learn from both the labeled and unlabeled data. This can be particularly useful in situations where it is difficult or expensive to obtain a large amount of labeled data.

\subsection{Inference complexity reduction}

\subsubsection{Pruning neural networks}
Pruning is a technique used in machine learning to reduce the complexity of a model by removing unnecessary (low-importance) parameters or connections. It can be used to reduce the computational complexity of a model by reducing the number of parameters that the model needs to store and the number of operations required to process input. In summary, pruning is the process of reducing the size of a preexisting NN by eliminating nonessential elements. Maintaining the network's precision while increasing its productivity is the goal of this procedure. This can also reduce the CPU time required for the NN to function.

The area of NN pruning is wide and encompasses several subcategories: (a) static or dynamic; (b) one-shot or iterative; (c) structured or unstructured; (d) magnitude-based or information-based; (e) global or layer-wise. Detailed information on the different types of pruning can be found in, e.g., Refs.~\cite{blalock2020state,liang2021pruning,liu2018rethinking, augasta2013pruning, vadera2020methods,han2015deep}. The four (most promising from our viewpoint) strategies for the iterative-pruning retraining process are fine-tuning, weight rewinding, learning rate rewinding, and BO-assisted.

Fine-tuning pruning is considered the most classic not only in the machine learning field but also in the field of equalizers for optical channel nonlinearities compensation. Such pruning technique can be used in a simpler way to eliminate, e.g., the coefficients of the Volterra equalizers \cite{chuang2019sparse,huang201893,6647643} and to trim not important triplets, making the triplet feature vector more sparse; in perturbation-method-based approaches \cite{zhang2019field, melek2020nonlinearity, kumar2021deep} and in the NN-based equalizers, subsection~\ref{subsubsec:post-eq}. For such complex NN structures, several papers investigated the use of fine-tuning, mainly in short-reach transmission (IM/DD) \cite{li2021high, wan2018nonlinear, zhang2020compressed, wang2021low, ge2020compressed, reza2018nonlinear}, to reduce the complexity of the model. So far, pruning analysis in optical channel equalization was restricted to the cases when such NN models used only the feed-forward layers.

In the context of channel equalization, seemingly the first paper that applied the Weight Rewinding Approach was  Ref.~\cite{Koike2021}, where such a technique was tested in the feed-forward model called ResMLP, which could give a sparsity of 99\% when compared with an initial over-parametrized solution with 6 layers and more than $10^6$ parameters.

\subsubsection{Weight sharing}

The weight-sharing compression approach is another method that can be explored to reduce the NN model's complexity by reducing the number of effective weights used by the model. This approach takes into account that several connections may share the same weight value, and then fine-tunes those shared weights. 
One common use of weight sharing is in CNNs, where the same set of weights is used for each convolutional filter in the network. This allows the model to learn general features that apply to multiple parts of the input, rather than learning separate features for each part. Also, weight sharing can significantly reduce the number of parameters in a model, reducing the amount of memory required to store the model and the amount of computation required to train it. It can also make the model more efficient at inference, as it requires fewer operations to process our input.

In the case of feedforward structures, this strategy was already successfully employed to minimize the complexity of NN models~\cite{han2015deep,wang2020compressing,son2018clustering,wu2018deep}.  Following the selection of a centroids' initialization technique, a minimal distance from each weight to such centroids is used to determine the shared weights for each layer of a trained network so that all weights in the same cluster share the same value~\cite{han2015deep}. The weights are not shared between the layers to prevent further performance loss and because sharing weights between sequential layers does not lower the computational complexity. Using the weight-sharing approach has the advantage of reducing the number of distinct multipliers in matrix multiplication to at least the number of clusters per input element. Then, the results of the multipliers are sent to the different adders.

 \subsubsection{Quantization techniques}
Quantization is used to lower the bitwidth of the numbers participating in arithmetic operations along the signal processing, which typically helps to significantly reduce the computation complexity of the processing. This means that a quantized model can use, for example, integers instead of floating-point numbers for some/all operations. Therefore, quantization allows representing the model using less memory and doing high-performance vectorized operations on a variety of hardware platforms \cite{gholami2021survey}. 

Quantization has demonstrated excellent and consistent results when used during the training and inference in different NN models \cite{gholami2021survey,liang2021pruning,cheng2017survey, weng2021neural}. It is particularly effective during inference because it saves computing resources without significantly decreasing the accuracy. NNs benefit from quantization because they are remarkably robust to aggressive quantization and extreme discretization.  This robustness emerges from the large number of parameters involved in the NN, meaning that the NNs are typically over-parameterized. In this subsection, we present the categories of quantization in terms of their mode (post-training quantization\cite{bai2021towards} or quantization-aware training \cite{alvarez2016efficient}) and quantization approach (homogeneous \cite{duarte2018fast}, or heterogeneous \cite{coelho2021automatic}).

Many quantization strategies have been investigated for equalizing the optical channel. Regarding the post-training quantization, Ref.~\cite{kaneda2020fpga} implemented an MLP-based equalizer with two hidden layers in an FPGA (XCZU9EG FFVC900) by using the post-training quantization with the traditional uniform int8 quantization, and it was tested on an experimental setup of  50Gb/s  PON with a 30~km SSMF link. Next, this time using an RNN-based equalizer, Ref.~\cite{huang2022low} tested the equalizer in a PAM4-based 100-Gbps passive optical network (PON) signal over a 20~km SSMF fiber testbed and applied a post-training quantization changing the weight's bitwidth from 8 to 2 bits to study the BER degradation due to the quantization noise. Also,  the authors of Ref.~\cite{huang2022low} realized such an equalizer in FPGA using the Xilinx Vivado toolset for high synthesis.
For coherent transmission, paper \cite{he2021fiber} introduced a complex-valued dimension-reduced triplet input NN and experimentally tested it with a 16-QAM 80 Gbps single-polarization transmission at 1800 km, with 100 km SSMF in the loop. In this study, to validate the robustness of such an NN equalizer on the quantization errors, the authors managed to reduce the bit precision of weights down to 2 bits with some acceptable decrease in performance.  

Moving on to the quantization-aware training (QAT) strategy, an important discussion on the quantization of NN weights was held in Ref.~\cite{aoudia2019towards}, emphasizing that the equalizer's inference should be performed by a fixed point system rather than a floating-point system. In this paper, an MLP-based equalizer was used, and its weights were quantized with a powers-of-two (PoT) quantization strategy. The authors incorporated the quantization error in the training of the equalizer by using the  Learning-Compression (LC) algorithm, which characterizes a QAT strategy. Then, considering a theoretical dispersive channel with additive white Gaussian noise and inter-symbol interference, Ref.~\cite{xu2019efficient} used a deep CNN equalizer to show its proposed quantization strategy, which combines QAT and post-equalization to find the most appropriate number of bits in the uniform quantization. The CNN equalizer performs comparably to the full-precision model using just 5-bit weights.  More recently, paper \cite{Koike2021} showed that instead of using PoT, the additive powers-of-two (APoT) strategy would bring much more resilience in terms of not degrading the performance as the PoT does.  In this work, a ResMLP equalizer was tested in simulation for a  dual-polarization 64/256~QAM, 34~GBd 11CH-WDM transmissions over 22 spans of 80~km SSMF fiber, and the QAT for APoT quantized weights was used for assessing the performance limits of such quantization strategy. More recently,  Ref.~\cite{freire2022reducing} reported a complete and comprehensive description and comparison study of various quantization approaches that have been applied to feed-forward and recurrent NN designs in the context of optical channel equalization. Finally, in Ref.~\cite{ron2022experimental}, the experimental implementation of an MLP-based coherent optical channel equalizer functioning realized in Raspberry Pi and Jentson Nano using pruning and quantization was performed.

\subsubsection{Knowledge
Distillation} 

The term ''knowledge distillation'' (KD) \cite{hinton2015distilling} is used to describe the process of condensing the information contained inside a large, complex model or set of models and passing it into a more manageable, standalone model suitable for deployment in the real world applications. In other words, KD  is a technique used in machine learning to reduce the complexity of a model by transferring the knowledge learned by a larger, pre-trained model (called the teacher model) to a smaller model (called the student model), allowing the student model to achieve similar performance to the teacher model with fewer parameters and less computation. This occurs because the student model learns the precise behavior of the teacher model by attempting to mimic its outputs at each level (not just for the final loss metric). The different forms of KD are response-based KD, feature-based KD, and relation-based KD \cite{gou2021knowledge}. The response-based knowledge focuses on the final output layer of the teacher model. The hypothesis for this KD type is that the student model will learn to mimic the predictions of the teacher model. The feature-based KD focuses on what intermediate layers learn to discriminate specific features; this knowledge is further used to train a student model. Finally, the relation-based KD focuses on capturing the relationship between feature maps (e.g., graphs, similarity matrix, feature embeddings, or probabilistic distributions) to train a student model.

\subsubsection{Parallelization Aspect of NN Implementation} 
 Feed-forward neural networks have been designed to be inherently parallelizable as the computations within each layer of a feed-forward network are independent of each other, enabling them to be parallelized across multiple processors or cores. In contrast, Recurrent Neural Networks have a recurrent structure, making parallelization more challenging. Specifically, computations within each time step of an RNN depend on the computations from the previous time step, thus making it infeasible to parallelize computations across time steps. Consequently, hardware parallelization of RNNs is limited. However, to address this challenge, specialized hardware such as Graphics Processing Units (GPUs) has been developed to support the parallelization of RNNs. A common technique to parallelize RNNs on GPUs is to unroll the RNN over a fixed number of time steps \cite{chang2017hardware}, creating a feed-forward network with shared weights. This enables parallel computation across the unrolled time steps, thereby allowing for the effective use of the parallelization capabilities of GPUs.However, the choice of the number of time steps to unroll an RNN can have an impact on its performance, with longer sequences requiring more memory and potentially leading to overfitting. Therefore, the optimal number of time steps to unroll an RNN is often a topic of research and experimentation. Another approach to address this issue is to use specialized hardware that is designed to handle the recurrent nature of RNNs. An example of such specialized hardware is the Neural Processing Unit \cite{willi2019recurrent}, which is a hardware accelerator for neural networks that is optimized for both feed-forward and recurrent computation.
 I
 
 Finally, it is worth mentioning that Ref.~\cite{9319148} presented an important study that evaluated architectural variations with majorly different degrees of parallelism which produced tradeoffs between area, speed, and reliability. In fact, with the increasing complexity and size of state-of-the-art neural network topologies, it becomes impractical to deploy all the required processing elements (PEs) on FPGA devices due to limitations in available logic resources. Consequently, it is crucial to analyze and discuss the tradeoffs involving area utilization, performance, and reliability when employing different levels of parallelism for neural network accelerators. In fact, as shown in Ref.~\cite{9319148}, the tradeoffs associated with varying degrees of parallelism in neural network accelerators are crucial considerations in optimizing resource utilization while maintaining performance and reliability. Achieving maximum parallelism by incorporating a large number of PEs might yield high computational efficiency, but it comes at the cost of increased logic resource utilization and potential limitations imposed by the FPGA device's capacity. In order to determine the optimal level of parallelism that strikes a balance between resource utilization, performance, and dependability, it would be advisable to evaluate these trade-offs. This analysis will contribute to the development of efficient and robust neural network accelerator designs, taking into consideration the limitations of hardware (such as FPGA devices\footnote{Note that Refs.~\cite{9079640, 7927161,luo2020towards, li2015fpga,danopoulos2022lstm,du2020model,wang2023model} can provide further insights in regard to the parallelization of NN structures when designing them in hardware.}).

 \section{Conclusions and perspectives}

Optical systems are capable of generating large datasets in a short time, making various data-driven methods and techniques especially attractive and efficient in this field. Neural networks, in particular, can fundamentally transform the design approaches in material science and optical engineering; the methodology of optical measurements and characterization; the architecture, operation, and control process of photonic devices and systems. Data-based techniques revolutionize optical sensing and imaging,  improving resolution accuracy, speed, and power consumption. Machine learning will increasingly contribute to the data-driven discovery (via sparsity-promoting techniques) of the physical models and master equations underlying the operation of complex photonic systems.
 We anticipate that in the future, laser systems with growing complexity will evolve into digital laser twins that will allow converting acquired data into the efficient control of such systems. In optical communications, the channel model is typically presented by complicated nonlinear equations, and the NN-based approach has already proven to be utterly efficient in modeling signal propagation down the system, the inversion of optical channels (for the equalization), or for the transmission systems' control. Together with this, in optical applications, we almost always have to deal with non-negligible noise impact, a situation in which machine learning methods can really flourish.

We want to reiterate that there are a plethora of other important research areas at the interface of photonics and machine learning that have not been discussed in this review/tutorial, for instance, the all-optical implementation of neural networks that we anticipate to grow substantially in the nearest future. 

 Though many existing (and future) problems in photonics can be solved using already available and well-established machine learning approaches, we anticipate that emerging data-driven photonics will require the development of new efficient algorithms specifically designed for optical applications, making a feedback impact on data science and leading to new synergetic concepts at the interface of photonics and machine learning. Overall, one of our goals in this tutorial is to stimulate the exchange of methods and ideas between data scientists and optical researchers/engineers. We strongly believe that the mutual penetration and cross-fertilization of these two disciplines will soon lead to unexpected innovations and fundamental breakthroughs in both fields.

\egor{numbers to insert on Fig. \ref{fig: ML in photonics} from 12 o'clock, clock-wise: \cite{jarajreh2014artificial}, \cite{Karanov:10}, \cite{aloraifan2021deep}, \cite{muller2021estimating}, \cite{liu2019application}, \cite{li2019deep}, \cite{lewis2007principal}, \cite{shi2019event}, \cite{NNLaser09}, \cite{andral2016toward}, \cite{baumeister2018deep}, \cite{qian2020single}, \cite{yu2020deep}, \cite{lin2020optical}, \cite{jiang2019global}, \cite{Tahersima2019}, \cite{MaxwellNet2022}, \cite{hager2018nonlinear}}


\section*{DISCLOSURES}
The author declares that there are no conflicts of interest related to this article.

\section*{DATA AVAILABILITY}
The data used for the results presented in this work is available upon request from the authors.

\section*{Acknowledgments:} This paper was supported by the EU Horizon 2020 program under the Marie Sklodowska-Curie grant agreement 813144 (REAL-NET). EM acknowledges the support of the EPSRC project EP/W002868/1. JEP is supported by Leverhulme Trust, Grant No. RPG-2018-063. SKT acknowledges the support of the EPSRC project TRANSNET.\\

\bibliography{sample}

\begin{thebibliography}{100}
\newcommand{\enquote}[1]{``#1''}

\bibitem{TM1997}
T.~Mitchell, \emph{Machine Learning} (McGraw-Hill, New York, 1997).

\bibitem{yegnanarayana2009artificial}
B.~Yegnanarayana, \emph{Artificial neural networks} (PHI Learning Pvt. Ltd., 2009).

\bibitem{pakkim}
M.~Pak and S.~Kim, \enquote{A review of deep learning in image recognition,} in \emph{2017 4th International Conference on Computer Applications and Information Processing Technology (CAIPT),}  (2017), pp. 1--3.

\bibitem{gu2018recent}
J.~Gu, Z.~Wang, J.~Kuen, L.~Ma, A.~Shahroudy, B.~Shuai, T.~Liu, X.~Wang, G.~Wang, J.~Cai, and T.~Chen, \enquote{Recent advances in convolutional neural networks,} {\protect\JournalTitle{Pattern Recognition}} \textbf{77}, 354 -- 377 (2018).

\bibitem{karhunen2015unsupervised}
J.~Karhunen, T.~Raiko, and K.~Cho, \enquote{Unsupervised deep learning: A short review,} {\protect\JournalTitle{Advances in Independent Component Analysis and Learning Machines}} pp. 125--142 (2015).

\bibitem{van2020survey}
J.~E. Van~Engelen and H.~H. Hoos, \enquote{A survey on semi-supervised learning,} {\protect\JournalTitle{Machine Learning}} \textbf{109}, 373--440 (2020).

\bibitem{NEURIPS2018_c1fea270}
A.~Oliver, A.~Odena, C.~A. Raffel, E.~D. Cubuk, and I.~Goodfellow, \enquote{Realistic evaluation of deep semi-supervised learning algorithms,} in \emph{Advances in Neural Information Processing Systems,}  vol.~31 S.~Bengio, H.~Wallach, H.~Larochelle, K.~Grauman, N.~Cesa-Bianchi, and R.~Garnett, eds. (Curran Associates, Inc., 2018).

\bibitem{mnih2015humanlevel}
V.~Mnih, K.~Kavukcuoglu, D.~Silver, A.~A. Rusu, J.~Veness, M.~G. Bellemare, A.~Graves, M.~Riedmiller, A.~K. Fidjeland, G.~Ostrovski, S.~Petersen, C.~Beattie, A.~Sadik, I.~Antonoglou, H.~King, D.~Kumaran, D.~Wierstra, S.~Legg, and D.~Hassabis, \enquote{Human-level control through deep reinforcement learning,} {\protect\JournalTitle{Nature}} \textbf{518}, 529--533 (2015).

\bibitem{Silver_2016}
D.~Silver, A.~Huang, C.~J. Maddison, A.~Guez, L.~Sifre, G.~van~den Driessche, J.~Schrittwieser, I.~Antonoglou, V.~Panneershelvam, M.~Lanctot, S.~Dieleman, D.~Grewe, J.~Nham, N.~Kalchbrenner, I.~Sutskever, T.~Lillicrap, M.~Leach, K.~Kavukcuoglu, T.~Graepel, and D.~Hassabis, \enquote{Mastering the game of {Go} with deep neural networks and tree search,} {\protect\JournalTitle{Nature}} \textbf{529}, 484--489 (2016).

\bibitem{li2017deep}
Y.~Li, \enquote{Deep reinforcement learning: An overview,} {\protect\JournalTitle{arXiv preprint arXiv:1701.07274}}  (2017).

\bibitem{henderson2018deep}
P.~Henderson, R.~Islam, P.~Bachman, J.~Pineau, D.~Precup, and D.~Meger, \enquote{Deep reinforcement learning that matters,} in \emph{Proceedings of the AAAI conference on artificial intelligence,}  vol.~32 (2018).

\bibitem{freire2022pitfalls}
P.~J. Freire, A.~Napoli, B.~Spinnler, N.~Costa, S.~K. Turitsyn, and J.~E. Prilepsky, \enquote{Neural networks-based equalizers for coherent optical transmission: Caveats and pitfalls,} {\protect\JournalTitle{IEEE Journal of Selected Topics in Quantum Electronics}} \textbf{28}, 1--23 (2022).

\bibitem{R01}
D.~Zibar, H.~Wymeersch, and I.~Lyubomirsky, \enquote{Machine learning under the spotlight,} {\protect\JournalTitle{Nature Photonics}} \textbf{11} (2017).

\bibitem{R02}
D.~Zibar, M.~Piels, R.~Jones, and C.~G. Schäeffer, \enquote{Machine learning techniques in optical communication,} {\protect\JournalTitle{Journal of Lightwave Technology}} \textbf{34}, 1442--1452 (2016).

\bibitem{R03Darko}
F.~Musumeci, C.~Rottondi, A.~Nag, I.~Macaluso, D.~Zibar, M.~Ruffini, and M.~Tornatore, \enquote{An overview on application of machine learning techniques in optical networks,} {\protect\JournalTitle{IEEE Communications Surveys and Tutorials}} \textbf{21}, 1383--1408 (2019).

\bibitem{R04}
G.~Genty, L.~Salmela, J.~M. Dudley, D.~Brunner, A.~Kokhanovskiy, S.~M. Kobtsev, and S.~K. Turitsyn, \enquote{Machine learning and applications in ultrafast photonics,} {\protect\JournalTitle{Nature Photonics}} \textbf{15}, 91 (2021).

\bibitem{R05}
J.~W. Nevin, S.~Nallaperuma, N.~A. Shevchenko, X.~Li, M.~S. Faruk, and S.~J. Savory, \enquote{Machine learning for optical fiber communication systems: An introduction and overview,} {\protect\JournalTitle{APL Photonics}} \textbf{6}, 121101 (2021).

\bibitem{R06}
D.~Piccinotti, K.~F. MacDonald, S.~A. Gregory, I.~Youngs, and N.~I. Zheludev, \enquote{Artificial intelligence for photonics and photonic materials,} {\protect\JournalTitle{Reports on Progress in Physics}} \textbf{84}, 012401 (2020).

\bibitem{R07}
M.~N{\"a}rhi, L.~Salmela, J.~Toivonen, C.~Billet, J.~M. Dudley, and G.~Genty, \enquote{Machine learning analysis of extreme events in optical fibre modulation instability,} {\protect\JournalTitle{Nature Communications}} \textbf{9}, 4923 (2018). Doi: 10.1038/s41467-018-07355-y.

\bibitem{R08}
F.~N. Khan, C.~Lu, and A.~P.~T. Lau, \enquote{Machine learning methods for optical communication systems,} in \emph{Advanced Photonics 2017 (IPR, NOMA, Sensors, Networks, SPPCom, PS),}  (Optica Publishing Group, 2017), p. SpW2F.3.

\bibitem{R09}
W.~Ma, Z.~Liu, Z.~Kudyshev, A.~Boltasseva, W.~Cai, and Y.~Liu, \enquote{Deep learning for the design of photonic structures,} {\protect\JournalTitle{Nature Photonics}} \textbf{15}, 77 (2021).

\bibitem{R10}
L.~Pilozzi, F.~A. Farrelly, G.~Marcucci, and C.~Conti, \enquote{Machine learning inverse problem for topological photonics,} {\protect\JournalTitle{Communications Physics}} \textbf{1}, 1--7 (2018).

\bibitem{R11}
L.~Pilozzi, F.~A. Farrelly, G.~Marcucci, and C.~Conti, \enquote{Topological nanophotonics and artificial neural networks,} {\protect\JournalTitle{Nanotechnology}} \textbf{32}, 142001 (2021).

\bibitem{NNMaterial01}
J.~Jiang, M.~Chen, and J.~A. Fan, \enquote{Deep neural networks for the evaluation and design of photonic devices,} {\protect\JournalTitle{Nature Reviews Materials}} \textbf{6}, 679 --700 (2021).

\bibitem{R13}
Y.~Xu, X.~Zhang, Y.~Fu, and Y.~Liu, \enquote{Interfacing photonics with artificial intelligence: an innovative design strategy for photonic structures and devices based on artificial neural networks,} {\protect\JournalTitle{Photon. Res.}} \textbf{9}, B135--B152 (2021).

\bibitem{R14}
F.~Vernuccio, A.~Bresci, V.~Cimini, A.~Giuseppi, G.~Cerullo, D.~Polli, and C.~M. Valensise, \enquote{Artificial intelligence in classical and quantum photonics,} {\protect\JournalTitle{Laser \& Photonics Reviews}} \textbf{16}, 2100399 (2022).

\bibitem{Saad02}
Y.~P. Raykov and D.~Saad, \enquote{Principled machine learning,} {\protect\JournalTitle{IEEE Journal of Selected Topics in Quantum Electronics}} \textbf{28}, 1--19 (2022).

\bibitem{wettlin2020complexity}
T.~Wettlin, S.~Pachnicke, T.~Rahman, J.~Wei, S.~Calabro, and N.~Stojanovic, \enquote{Complexity reduction of volterra nonlinear equalization for optical short-reach im/dd systems,} in \emph{Photonic Networks; 21th ITG-Symposium,}  (VDE, 2020), pp. 1--6.

\bibitem{wei2020special}
J.~Wei, L.~Yi, E.~Giacoumidis, Q.~Cheng, and A.~Lau, \enquote{Special issue on “optics for ai and ai for optics”,} {\protect\JournalTitle{Applied Sciences}} \textbf{10}, 3262 (2020).

\bibitem{shen2017deep}
Y.~Shen, N.~C. Harris, S.~Skirlo, M.~Prabhu, T.~Baehr-Jones, M.~Hochberg, X.~Sun, S.~Zhao, H.~Larochelle, D.~Englund \emph{et~al.}, \enquote{Deep learning with coherent nanophotonic circuits,} {\protect\JournalTitle{Nature photonics}} \textbf{11}, 441--446 (2017).

\bibitem{sunada2021photonic}
S.~Sunada and A.~Uchida, \enquote{Photonic neural field on a silicon chip: large-scale, high-speed neuro-inspired computing and sensing,} {\protect\JournalTitle{Optica}} \textbf{8}, 1388--1396 (2021).

\bibitem{huang2021silicon}
C.~Huang, S.~Fujisawa, T.~F. de~Lima, A.~N. Tait, E.~C. Blow, Y.~Tian, S.~Bilodeau, A.~Jha, F.~Yaman, H.-T. Peng \emph{et~al.}, \enquote{A silicon photonic--electronic neural network for fibre nonlinearity compensation,} {\protect\JournalTitle{Nature Electronics}} \textbf{4}, 837--844 (2021).

\bibitem{shastri2022silicon}
B.~J. Shastri, C.~Huang, A.~N. Tait, T.~F. de~Lima, and P.~R. Prucnal, \enquote{Silicon photonic neural network applications and prospects,} in \emph{AI and Optical Data Sciences III,}  vol. 12019 (SPIE, 2022), pp. 135--144.

\bibitem{huang2022prospects}
C.~Huang, V.~J. Sorger, M.~Miscuglio, M.~Al-Qadasi, A.~Mukherjee, L.~Lampe, M.~Nichols, A.~N. Tait, T.~Ferreira~de Lima, B.~A. Marquez \emph{et~al.}, \enquote{Prospects and applications of photonic neural networks,} {\protect\JournalTitle{Advances in Physics: X}} \textbf{7}, 1981155 (2022).

\bibitem{Prucnal02}
T.~F. de~Lima, H.-T. Peng, A.~N. Tait, M.~A. Nahmias, H.~B. Miller, B.~J. Shastri, and P.~R. Prucnal, \enquote{Machine learning with neuromorphic photonics,} {\protect\JournalTitle{Journal of Lightwave Technology}} \textbf{37}, 1515--1534 (2019).

\bibitem{Prucnal03}
T.~F. de~Lima, B.~J. Shastri, A.~N. Tait, M.~A. Nahmias, and P.~R. Prucnal, \enquote{Progress in neuromorphic photonics,} {\protect\JournalTitle{Nanophotonics}} \textbf{6}, 577--599 (2017).

\bibitem{Prucnal05}
K.~Berggren, Q.~Xia, K.~K. Likharev, D.~B. Strukov, H.~Jiang, T.~Mikolajick, D.~Querlioz, M.~Salinga, J.~R. Erickson, S.~Pi \emph{et~al.}, \enquote{Roadmap on emerging hardware and technology for machine learning,} {\protect\JournalTitle{Nanotechnology}} \textbf{32}, 012002 (2020).

\bibitem{Fan01}
G.~Wetzstein, A.~Ozcan, S.~Gigan, and et~al., \enquote{Inference in artificial intelligence with deep optics and photonics,} {\protect\JournalTitle{Nature}} \textbf{588}, 39--47 (2020).

\bibitem{Fan05}
D.~Brunner, M.~C. Soriano, and S.~Fan, \enquote{Neural network learning with photonics and for photonic circuit design,} {\protect\JournalTitle{Nanophotonics}} \textbf{12}, 773--775 (2023).

\bibitem{pai2023experimentally}
S.~Pai, Z.~Sun, T.~W. Hughes, T.~Park, B.~Bartlett, I.~A. Williamson, M.~Minkov, M.~Milanizadeh, N.~Abebe, F.~Morichetti \emph{et~al.}, \enquote{Experimentally realized in situ backpropagation for deep learning in photonic neural networks,} {\protect\JournalTitle{Science}} \textbf{380}, 398--404 (2023).

\bibitem{Volker01}
M.~Miscuglio, Z.~Hu, S.~Li, J.~K. George, R.~Capanna, H.~Dalir, P.~M. Bardet, P.~Gupta, and V.~J. Sorger, \enquote{Massively parallel amplitude-only fourier neural network,} {\protect\JournalTitle{Optica}} \textbf{7}, 1812--1819 (2020).

\bibitem{Volker02}
M.~Miscuglio and V.~Sorger, \enquote{Photonic tensor cores for machine learning,} {\protect\JournalTitle{Applied Physics Reviews}} \textbf{7}, 031404 (2020).

\bibitem{Volker03}
M.~Miscuglio, A.~Mehrabian, Z.~Hu, S.~I. Azzam, J.~George, A.~V. Kildishev, M.~Pelton, and V.~J. Sorger, \enquote{All-optical nonlinear activation function for photonic neural networks,} {\protect\JournalTitle{Opt. Mater. Express}} \textbf{8}, 3851--3863 (2018).

\bibitem{VOlker04}
N.~Peserico, B.~J. Shastri, and V.~J. Sorger, \enquote{Integrated photonic tensor processing unit for a matrix multiply: A review,} {\protect\JournalTitle{Journal of Lightwave Technology}} pp. 1--14 (2023).

\bibitem{christensen20222022}
D.~V. Christensen, R.~Dittmann, B.~Linares-Barranco, A.~Sebastian, M.~Le~Gallo, A.~Redaelli, S.~Slesazeck, T.~Mikolajick, S.~Spiga, S.~Menzel \emph{et~al.}, \enquote{2022 roadmap on neuromorphic computing and engineering,} {\protect\JournalTitle{Neuromorphic Computing and Engineering}} \textbf{2}, 022501 (2022).

\bibitem{peserico2022emerging}
N.~Peserico, T.~F. de~Lima, P.~Prucnal, and V.~J. Sorger, \enquote{Emerging devices and packaging strategies for electronic-photonic ai accelerators: opinion,} {\protect\JournalTitle{Optical Materials Express}} \textbf{12}, 1347--1351 (2022).

\bibitem{mehonic2022brain}
A.~Mehonic and A.~J. Kenyon, \enquote{Brain-inspired computing needs a master plan,} {\protect\JournalTitle{Nature}} \textbf{604}, 255--260 (2022).

\bibitem{mcculloch1943logical}
W.~S. McCulloch and W.~Pitts, \enquote{A logical calculus of the ideas immanent in nervous activity,} {\protect\JournalTitle{The Bulletin of Mathematical Biophysics}} \textbf{5}, 115--133 (1943).

\bibitem{rosenblatt1958perceptron}
F.~Rosenblatt, \enquote{The perceptron: a probabilistic model for information storage and organization in the brain.} {\protect\JournalTitle{Psychological Review}} \textbf{65}, 386 (1958).

\bibitem{hong2020low}
S.~Hong, H.~Kang, J.~Kim, and K.~Cho, \enquote{Low voltage time-based matrix multiplier-and-accumulator for neural computing system,} {\protect\JournalTitle{Electronics}} \textbf{9}, 2138 (2020).

\bibitem{heidari2019analog}
M.~Heidari and H.~Shamsi, \enquote{Analog programmable neuron and case study on vlsi implementation of multi-layer perceptron (mlp),} {\protect\JournalTitle{Microelectronics Journal}} \textbf{84}, 36--47 (2019).

\bibitem{geng2020analog}
C.~Geng, Q.~Sun, and S.~Nakatake, \enquote{An analog cmos implementation for multi-layer perceptron with relu activation,} in \emph{2020 9th International conference on modern circuits and systems technologies (MOCAST),}  (IEEE, 2020), pp. 1--6.

\bibitem{abden2020multilayer}
S.~Abden and E.~Azab, \enquote{Multilayer perceptron analog hardware implementation using low power operational transconductance amplifier,} in \emph{2020 32nd International Conference on Microelectronics (ICM),}  (IEEE, 2020), pp. 1--4.

\bibitem{sarpeshkar1998analog}
R.~Sarpeshkar, \enquote{Analog versus digital: extrapolating from electronics to neurobiology,} {\protect\JournalTitle{Neural computation}} \textbf{10}, 1601--1638 (1998).

\bibitem{zhou2022photonic}
H.~Zhou, J.~Dong, J.~Cheng, W.~Dong, C.~Huang, Y.~Shen, Q.~Zhang, M.~Gu, C.~Qian, H.~Chen \emph{et~al.}, \enquote{Photonic matrix multiplication lights up photonic accelerator and beyond,} {\protect\JournalTitle{Light: Science \& Applications}} \textbf{11}, 30 (2022).

\bibitem{harris2018linear}
N.~C. Harris, J.~Carolan, D.~Bunandar, M.~Prabhu, M.~Hochberg, T.~Baehr-Jones, M.~L. Fanto, A.~M. Smith, C.~C. Tison, P.~M. Alsing \emph{et~al.}, \enquote{Linear programmable nanophotonic processors,} {\protect\JournalTitle{Optica}} \textbf{5}, 1623--1631 (2018).

\bibitem{bogaerts2020programmable}
W.~Bogaerts, D.~P{\'e}rez, J.~Capmany, D.~A. Miller, J.~Poon, D.~Englund, F.~Morichetti, and A.~Melloni, \enquote{Programmable photonic circuits,} {\protect\JournalTitle{Nature}} \textbf{586}, 207--216 (2020).

\bibitem{Fan02}
T.~W. Hughes, M.~Minkov, Y.~Shi, and S.~Fan, \enquote{Training of photonic neural networks through in situ backpropagation and gradient measurement,} {\protect\JournalTitle{Optica}} \textbf{5}, 864--871 (2018).

\bibitem{roques2023learning}
C.~Roques-Carmes, \enquote{Learning photons go backward,} {\protect\JournalTitle{Science}} \textbf{380}, 341--342 (2023).

\bibitem{cheng2021photonic}
J.~Cheng, H.~Zhou, and J.~Dong, \enquote{Photonic matrix computing: from fundamentals to applications,} {\protect\JournalTitle{Nanomaterials}} \textbf{11}, 1683 (2021).

\bibitem{zhang2021optical}
H.~Zhang, M.~Gu, X.~Jiang, J.~Thompson, H.~Cai, S.~Paesani, R.~Santagati, A.~Laing, Y.~Zhang, M.~Yung \emph{et~al.}, \enquote{An optical neural chip for implementing complex-valued neural network,} {\protect\JournalTitle{Nature communications}} \textbf{12}, 1--11 (2021).

\bibitem{peserico2022photonic}
N.~Peserico, X.~Ma, B.~J. Shastri, and V.~J. Sorger, \enquote{Photonic tensor core for machine learning: a review,} {\protect\JournalTitle{Emerging Topics in Artificial Intelligence (ETAI) 2022}} \textbf{12204}, 53--60 (2022).

\bibitem{thomaschewski2023high}
M.~Thomaschewski, Z.~Hu, B.~M. Nouri, Y.~Gui, H.~Wang, S.~Altaleb, H.~Dalir, and V.~J. Sorger, \enquote{High-performance optoelectronics for integrated photonic neural networks,} in \emph{AI and Optical Data Sciences IV,}  vol. 12438 (SPIE, 2023), pp. 262--271.

\bibitem{kiranyaz20211d}
S.~Kiranyaz, O.~Avci, O.~Abdeljaber, T.~Ince, M.~Gabbouj, and D.~J. Inman, \enquote{1d convolutional neural networks and applications: A survey,} {\protect\JournalTitle{Mechanical systems and signal processing}} \textbf{151}, 107398 (2021).

\bibitem{woods2017fpga}
R.~Woods, J.~McAllister, G.~Lightbody, and Y.~Yi, \enquote{Fpga-based implementation of signal processing systems,}  (2017).

\bibitem{chang2018hybrid}
J.~Chang, V.~Sitzmann, X.~Dun, W.~Heidrich, and G.~Wetzstein, \enquote{Hybrid optical-electronic convolutional neural networks with optimized diffractive optics for image classification,} {\protect\JournalTitle{Scientific reports}} \textbf{8}, 1--10 (2018).

\bibitem{elman1990finding}
J.~L. Elman, \enquote{Finding structure in time,} {\protect\JournalTitle{Cognitive science}} \textbf{14}, 179--211 (1990).

\bibitem{lipton2015critical}
Z.~C. Lipton, J.~Berkowitz, and C.~Elkan, \enquote{A critical review of recurrent neural networks for sequence learning,} {\protect\JournalTitle{arXiv preprint arXiv:1506.00019}}  (2015).

\bibitem{bengio1994vanishgrad}
Y.~Bengio, P.~Simard, and P.~Frasconi, \enquote{Learning long-term dependencies with gradient descent is difficult,} {\protect\JournalTitle{IEEE Transactions on Neural Networks}} \textbf{5}, 157--166 (1994).

\bibitem{sanchez2020exploiting}
A.~Sanchez-Caballero, D.~Fuentes-Jimenez, and C.~Losada-Guti{\'e}rrez, \enquote{Exploiting the convlstm: Human action recognition using raw depth video-based recurrent neural networks,} {\protect\JournalTitle{arXiv preprint arXiv:2006.07744}}  (2020).

\bibitem{saha2022comprehensive}
S.~Saha, N.~Majumder, D.~Sangani, and A.~Das~Bhattacharjee, \enquote{Comprehensive forecasting-based analysis using hybrid and stacked stateful/stateless models,} in \emph{Advances in Distributed Computing and Machine Learning,}  (Springer, 2022), pp. 567--579.

\bibitem{pham2019recurrent}
T.-T. Pham, M.~Pister, and P.~Couv{\'e}e, \enquote{Recurrent neural network for classifying of hpc applications,} in \emph{2019 Spring Simulation Conference (SpringSim),}  (IEEE, 2019), pp. 1--12.

\bibitem{van2001reference}
M.~E. Van~Valkenburg, \emph{Reference data for engineers: radio, electronics, computers and communications} (Newnes, 2001).

\bibitem{storn1996differential}
R.~Storn, \enquote{Differential evolution design of an iir-filter,} in \emph{Proceedings of IEEE international conference on evolutionary computation,}  (IEEE, 1996), pp. 268--273.

\bibitem{kalman12}
R.~G. Brown and P.~Y.~C. Hwang, \emph{Introduction to Random Signals and Applied Kalman Filtering: with {M}atlab exercises} (John Wiley \& Sons, Inc., 2012), 4th ed.

\bibitem{brookner1998tracking}
E.~Brookner, \emph{Tracking and Kalman filtering made easy} (Wiley New York, 1998).

\bibitem{cruse2006neural}
H.~Cruse, \enquote{Neural networks as cybernetic systems,} {\protect\JournalTitle{Brains, Minds \& Media, Bielefeld, Germany}}  (2006).

\bibitem{juang1993estimation}
J.-N. Juang, C.-W. Chen, and M.~Phan, \enquote{Estimation of kalman filter gain from output residuals,} {\protect\JournalTitle{Journal of Guidance, Control, and Dynamics}} \textbf{16}, 903--908 (1993).

\bibitem{decruyenaere1992comparison}
J.~DeCruyenaere and H.~Hafez, \enquote{A comparison between kalman filters and recurrent neural networks,} in \emph{[Proceedings 1992] IJCNN International Joint Conference on Neural Networks,}  vol.~4 (IEEE, 1992), pp. 247--251.

\bibitem{jospin2022bayesian}
L.~V. Jospin, H.~Laga, F.~Boussaid, W.~Buntine, and M.~Bennamoun, \enquote{Hands-on bayesian neural networks—a tutorial for deep learning users,} {\protect\JournalTitle{IEEE Computational Intelligence Magazine}} \textbf{17}, 29--48 (2022).

\bibitem{chung1991thesis}
C.-W. Chen, \enquote{Integrated system identification and adaptive state estimation for control of flexible space structures,} Ph.D. thesis, Old Dominion University (1991).

\bibitem{chenna04}
S.~K. Chenna, Y.~K. Jain, H.~Kapoor, R.~S. Bapi, N.~Yadaiah, A.~Negi, V.~S. Rao, and B.~L. Deekshatulu, \enquote{State estimation and tracking problems: A comparison between kalman filter and recurrent neural networks,} in \emph{Neural Information Processing,}  N.~R. Pal, N.~Kasabov, R.~K. Mudi, S.~Pal, and S.~K. Parui, eds. (Springer Berlin Heidelberg, Berlin, Heidelberg, 2004), pp. 275--281.

\bibitem{parlos01}
A.~Parlos, S.~Menon, and A.~Atiya, \enquote{An algorithmic approach to adaptive state filtering using recurrent neural networks,} {\protect\JournalTitle{IEEE Transactions on Neural Networks}} \textbf{12}, 1411--1432 (2001).

\bibitem{haykin2001kalman}
S.~S. Haykin, ed., \emph{Kalman filtering and neural networks}, vol. 284 (Wiley Online Library, 2001).

\bibitem{mandic2009complex}
D.~P. Mandic and V.~S.~L. Goh, \emph{Complex valued nonlinear adaptive filters: noncircularity, widely linear and neural models} (John Wiley \& Sons, 2009).

\bibitem{shao2021training}
Y.~Shao, F.~M. Dietrich, C.~Nettelblad, and C.~Zhang, \enquote{Training algorithm matters for the performance of neural network potential: A case study of adam and the kalman filter optimizers,} {\protect\JournalTitle{The Journal of Chemical Physics}} \textbf{155}, 204108 (2021).

\bibitem{tait2017neuromorphic}
A.~N. Tait, T.~F. De~Lima, E.~Zhou, A.~X. Wu, M.~A. Nahmias, B.~J. Shastri, and P.~R. Prucnal, \enquote{Neuromorphic photonic networks using silicon photonic weight banks,} {\protect\JournalTitle{Scientific reports}} \textbf{7}, 1--10 (2017).

\bibitem{bueno2018reinforcement}
J.~Bueno, S.~Maktoobi, L.~Froehly, I.~Fischer, M.~Jacquot, L.~Larger, and D.~Brunner, \enquote{Reinforcement learning in a large-scale photonic recurrent neural network,} {\protect\JournalTitle{Optica}} \textbf{5}, 756--760 (2018).

\bibitem{shastri2021photonics}
B.~J. Shastri, A.~N. Tait, T.~Ferreira~de Lima, W.~H. Pernice, H.~Bhaskaran, C.~D. Wright, and P.~R. Prucnal, \enquote{Photonics for artificial intelligence and neuromorphic computing,} {\protect\JournalTitle{Nature Photonics}} \textbf{15}, 102--114 (2021).

\bibitem{Prucnal04}
H.-T. Peng, J.~C. Lederman, L.~Xu, T.~F. de~Lima, C.~Huang, B.~J. Shastri, D.~Rosenbluth, and P.~R. Prucnal, \enquote{A photonics-inspired compact network: Toward real-time ai processing in communication systems,} {\protect\JournalTitle{IEEE Journal of Selected Topics in Quantum Electronics}} \textbf{28}, 1--17 (2022).

\bibitem{Fan03}
T.~W. Hughes, I.~A.~D. Williamson, M.~Minkov, and S.~Fan, \enquote{Wave physics as an analog recurrent neural network,} {\protect\JournalTitle{Science Advances}} \textbf{5}, eaay6946 (2019).

\bibitem{hochreiter1997long}
S.~Hochreiter and J.~Schmidhuber, \enquote{Long short-term memory,} {\protect\JournalTitle{Neural Computation}} \textbf{9}, 1735--1780 (1997).

\bibitem{gers2000learning}
F.~A. Gers, J.~Schmidhuber, and F.~Cummins, \enquote{Learning to forget: Continual prediction with lstm,} {\protect\JournalTitle{Neural computation}} \textbf{12}, 2451--2471 (2000).

\bibitem{cho2014learning}
K.~Cho, B.~Van~Merri{\"e}nboer, C.~Gulcehre, D.~Bahdanau, F.~Bougares, H.~Schwenk, and Y.~Bengio, \enquote{Learning phrase representations using rnn encoder-decoder for statistical machine translation,} {\protect\JournalTitle{arXiv preprint arXiv:1406.1078}}  (2014).

\bibitem{chung2014empirical}
J.~Chung, C.~Gulcehre, K.~Cho, and Y.~Bengio, \enquote{Empirical evaluation of gated recurrent neural networks on sequence modeling,} in \emph{NIPS 2014 Workshop on Deep Learning,}  (2014).

\bibitem{dey2017gate}
R.~Dey and F.~M. Salem, \enquote{Gate-variants of gated recurrent unit (gru) neural networks,} in \emph{2017 IEEE 60th international midwest symposium on circuits and systems (MWSCAS),}  (IEEE, 2017), pp. 1597--1600.

\bibitem{heck2017simplified}
J.~C. Heck and F.~M. Salem, \enquote{Simplified minimal gated unit variations for recurrent neural networks,} in \emph{2017 IEEE 60th International Midwest Symposium on Circuits and Systems (MWSCAS),}  (IEEE, 2017), pp. 1593--1596.

\bibitem{jaeger2004harnessing}
H.~Jaeger and H.~Haas, \enquote{Harnessing nonlinearity: Predicting chaotic systems and saving energy in wireless communication,} {\protect\JournalTitle{Science}} \textbf{304}, 78--80 (2004).

\bibitem{wu2018statistical}
Q.~Wu, E.~Fokoue, and D.~Kudithipudi, \enquote{On the statistical challenges of echo state networks and some potential remedies,} {\protect\JournalTitle{arXiv preprint arXiv:1802.07369}}  (2018).

\bibitem{sorokina2019fiber}
M.~Sorokina, S.~Sergeyev, and S.~Turitsyn, \enquote{Fiber echo state network analogue for high-bandwidth dual-quadrature signal processing,} {\protect\JournalTitle{Optics Express}} \textbf{27}, 2387--2395 (2019).

\bibitem{brain-inspired}
S.~S. {Mosleh}, L.~{Liu}, C.~{Sahin}, Y.~R. {Zheng}, and Y.~{Yi}, \enquote{Brain-inspired wireless communications: Where reservoir computing meets mimo-ofdm,} {\protect\JournalTitle{IEEE Transactions on Neural Networks and Learning Systems}} \textbf{29}, 4694--4708 (2018).

\bibitem{sun2020review}
C.~Sun, M.~Song, S.~Hong, and H.~Li, \enquote{A review of designs and applications of echo state networks,} {\protect\JournalTitle{arXiv preprint arXiv:2012.02974}}  (2020).

\bibitem{jaeger2007optimization}
H.~Jaeger, M.~Luko{\v{s}}evi{\v{c}}ius, D.~Popovici, and U.~Siewert, \enquote{Optimization and applications of echo state networks with leaky-integrator neurons,} {\protect\JournalTitle{Neural networks}} \textbf{20}, 335--352 (2007).

\bibitem{van2017advances}
G.~Van~der Sande, D.~Brunner, and M.~C. Soriano, \enquote{Advances in photonic reservoir computing,} {\protect\JournalTitle{Nanophotonics}} \textbf{6}, 561--576 (2017).

\bibitem{bahdanau2014neural}
D.~Bahdanau, K.~Cho, and Y.~Bengio, \enquote{Neural machine translation by jointly learning to align and translate,} {\protect\JournalTitle{arXiv preprint arXiv:1409.0473}}  (2014).

\bibitem{quinn2019dive}
J.~Quinn, J.~McEachen, M.~Fullan, M.~Gardner, and M.~Drummy, \emph{Dive into deep learning: Tools for engagement} (Corwin Press, 2019).

\bibitem{luong2015effective}
M.-T. Luong, H.~Pham, and C.~D. Manning, \enquote{Effective approaches to attention-based neural machine translation,} {\protect\JournalTitle{arXiv preprint arXiv:1508.04025}}  (2015).

\bibitem{kim2017structured}
Y.~Kim, C.~Denton, L.~Hoang, and A.~M. Rush, \enquote{Structured attention networks,} {\protect\JournalTitle{arXiv preprint arXiv:1702.00887}}  (2017).

\bibitem{vaswani2017attention}
A.~Vaswani, N.~Shazeer, N.~Parmar, J.~Uszkoreit, L.~Jones, A.~N. Gomez, {\L}.~Kaiser, and I.~Polosukhin, \enquote{Attention is all you need,} {\protect\JournalTitle{Advances in neural information processing systems}} \textbf{30} (2017).

\bibitem{he2016deep}
K.~He, X.~Zhang, S.~Ren, and J.~Sun, \enquote{Deep residual learning for image recognition,} in \emph{Proceedings of the IEEE conference on computer vision and pattern recognition,}  (2016), pp. 770--778.

\bibitem{ba2016layer}
J.~L. Ba, J.~R. Kiros, and G.~E. Hinton, \enquote{Layer normalization,} {\protect\JournalTitle{arXiv preprint arXiv:1607.06450}}  (2016).

\bibitem{hamgini2023application}
B.~B. Hamgini, H.~Najafi, A.~Bakhshali, and Z.~Zhang, \enquote{Application of transformers for nonlinear channel compensation in optical systems,} {\protect\JournalTitle{arXiv preprint arXiv:2304.13119}}  (2023).

\bibitem{Srivastava2015}
R.~K. Srivastava, K.~Greff, and J.~Schmidhuber, \enquote{Highway networks,}  (2015).

\bibitem{huang2017densely}
G.~Huang, Z.~Liu, L.~Van Der~Maaten, and K.~Q. Weinberger, \enquote{Densely connected convolutional networks,} in \emph{Proceedings of the IEEE conference on computer vision and pattern recognition,}  (2017), pp. 4700--4708.

\bibitem{dou2020residual}
H.~Dou, Y.~Deng, T.~Yan, H.~Wu, X.~Lin, and Q.~Dai, \enquote{Residual d 2 nn: training diffractive deep neural networks via learnable light shortcuts,} {\protect\JournalTitle{Optics Letters}} \textbf{45}, 2688--2691 (2020).

\bibitem{gin2021deep}
C.~Gin, B.~Lusch, S.~L. Brunton, and J.~N. Kutz, \enquote{Deep learning models for global coordinate transformations that linearise pdes,} {\protect\JournalTitle{European Journal of Applied Mathematics}} \textbf{32}, 515--539 (2021).

\bibitem{broomhead1988radial}
D.~S. Broomhead and D.~Lowe, \enquote{Radial basis functions, multi-variable functional interpolation and adaptive networks,} Tech. rep., Royal Signals and Radar Establishment Malvern (United Kingdom) (1988).

\bibitem{que2016back}
Q.~Que and M.~Belkin, \enquote{Back to the future: Radial basis function networks revisited,} in \emph{Artificial intelligence and statistics,}  (PMLR, 2016), pp. 1375--1383.

\bibitem{beheim2004new}
L.~Beheim, A.~Zitouni, F.~Belloir, and C.~d.~M. de~la Housse, \enquote{New rbf neural network classifier with optimized hidden neurons number,} {\protect\JournalTitle{WSEAS Transactions on Systems}} \textbf{2}, 467--472 (2004).

\bibitem{de2000mahalanobis}
R.~De~Maesschalck, D.~Jouan-Rimbaud, and D.~L. Massart, \enquote{The mahalanobis distance,} {\protect\JournalTitle{Chemometrics and intelligent laboratory systems}} \textbf{50}, 1--18 (2000).

\bibitem{park1991universal}
J.~Park and I.~W. Sandberg, \enquote{Universal approximation using radial-basis-function networks,} {\protect\JournalTitle{Neural computation}} \textbf{3}, 246--257 (1991).

\bibitem{Georg2021book}
G.~B{\"o}cherer, \enquote{Lecture notes on machine learning for communications,}  (2021).

\bibitem{bank2020autoencoders}
D.~Bank, N.~Koenigstein, and R.~Giryes, \enquote{Autoencoders,} {\protect\JournalTitle{arXiv preprint arXiv:2003.05991}}  (2020).

\bibitem{Sensor05}
A.~Venketeswaran, N.~Lalam, J.~Wuenschell, P.~R. Ohodnicki~Jr., M.~Badar, K.~P. Chen, P.~Lu, Y.~Duan, B.~Chorpening, and M.~Buric, \enquote{Recent advances in machine learning for fiber optic sensor applications,} {\protect\JournalTitle{Advanced Intelligent Systems}} \textbf{4}, 2100067 (2022).

\bibitem{doersch2016tutorial}
C.~Doersch, \enquote{Tutorial on variational autoencoders,} {\protect\JournalTitle{arXiv preprint arXiv:1606.05908}}  (2016).

\bibitem{kingma2013auto}
D.~P. Kingma and M.~Welling, \enquote{Auto-encoding variational bayes,} {\protect\JournalTitle{arXiv preprint arXiv:1312.6114}}  (2013).

\bibitem{baumeister2018deep}
T.~Baumeister, S.~L. Brunton, and J.~N. Kutz, \enquote{Deep learning and model predictive control for self-tuning mode-locked lasers,} {\protect\JournalTitle{JOSA B}} \textbf{35}, 617--626 (2018).

\bibitem{chen2023photonic}
Y.~Chen, T.~Zhou, J.~Wu, H.~Qiao, X.~Lin, L.~Fang, and Q.~Dai, \enquote{Photonic unsupervised learning variational autoencoder for high-throughput and low-latency image transmission,} {\protect\JournalTitle{Science Advances}} \textbf{9}, eadf8437 (2023).

\bibitem{makhzani2015adversarial}
A.~Makhzani, J.~Shlens, N.~Jaitly, I.~Goodfellow, and B.~Frey, \enquote{Adversarial autoencoders,} {\protect\JournalTitle{arXiv preprint arXiv:1511.05644}}  (2015).

\bibitem{kudyshev2020machine}
Z.~A. Kudyshev, A.~V. Kildishev, V.~M. Shalaev, and A.~Boltasseva, \enquote{Machine-learning-assisted metasurface design for high-efficiency thermal emitter optimization,} {\protect\JournalTitle{Applied Physics Reviews}} \textbf{7}, 021407 (2020).

\bibitem{kudyshev2020machine2}
Z.~A. Kudyshev, A.~V. Kildishev, V.~M. Shalaev, and A.~Boltasseva, \enquote{Machine learning--assisted global optimization of photonic devices,} {\protect\JournalTitle{Nanophotonics}} \textbf{10}, 371--383 (2020).

\bibitem{creswell2018denoising}
A.~Creswell and A.~A. Bharath, \enquote{Denoising adversarial autoencoders,} {\protect\JournalTitle{IEEE transactions on neural networks and learning systems}} \textbf{30}, 968--984 (2018).

\bibitem{xie2020autoencoder}
W.~Xie, B.~Liu, Y.~Li, J.~Lei, and Q.~Du, \enquote{Autoencoder and adversarial-learning-based semisupervised background estimation for hyperspectral anomaly detection,} {\protect\JournalTitle{IEEE Transactions on Geoscience and Remote Sensing}} \textbf{58}, 5416--5427 (2020).

\bibitem{goodfellow2020generative}
I.~Goodfellow, J.~Pouget-Abadie, M.~Mirza, B.~Xu, D.~Warde-Farley, S.~Ozair, A.~Courville, and Y.~Bengio, \enquote{Generative adversarial networks,} {\protect\JournalTitle{Communications of the ACM}} \textbf{63}, 139--144 (2020).

\bibitem{wang2017generative}
K.~Wang, C.~Gou, Y.~Duan, Y.~Lin, X.~Zheng, and F.-Y. Wang, \enquote{Generative adversarial networks: introduction and outlook,} {\protect\JournalTitle{IEEE/CAA Journal of Automatica Sinica}} \textbf{4}, 588--598 (2017).

\bibitem{gui2021review}
J.~Gui, Z.~Sun, Y.~Wen, D.~Tao, and J.~Ye, \enquote{A review on generative adversarial networks: Algorithms, theory, and applications,} {\protect\JournalTitle{IEEE Transactions on Knowledge and Data Engineering}}  (2021).

\bibitem{wang2021artificial}
D.~Wang and M.~Zhang, \enquote{Artificial intelligence in optical communications: From machine learning to deep learning,} {\protect\JournalTitle{Frontiers in Communications and Networks}} \textbf{2}, 656786 (2021).

\bibitem{radford2016unsupervised}
A.~Radford, L.~Metz, and S.~Chintala, \enquote{Unsupervised representation learning with deep convolutional generative adversarial networks,}  (2016).

\bibitem{cohen2021generative}
A.~Cohen and S.~Derevyanko, \enquote{Generative adversarial network and end-to-end learning for optical fiber communication systems limited by the nonlinear phase noise,} in \emph{2021 IEEE International Conference on Microwaves, Antennas, Communications and Electronic Systems (COMCAS),}  (IEEE, 2021), pp. 241--246.

\bibitem{nguyen2019bayesian}
V.~Nguyen, \enquote{Bayesian optimization for accelerating hyper-parameter tuning,} in \emph{2019 IEEE second international conference on artificial intelligence and knowledge engineering (AIKE),}  (IEEE, 2019), pp. 302--305.

\bibitem{cho2020basic}
H.~Cho, Y.~Kim, E.~Lee, D.~Choi, Y.~Lee, and W.~Rhee, \enquote{Basic enhancement strategies when using bayesian optimization for hyperparameter tuning of deep neural networks,} {\protect\JournalTitle{IEEE Access}} \textbf{8}, 52588--52608 (2020).

\bibitem{wu2019hyperparameter}
J.~Wu, X.-Y. Chen, H.~Zhang, L.-D. Xiong, H.~Lei, and S.-H. Deng, \enquote{Hyperparameter optimization for machine learning models based on bayesian optimization,} {\protect\JournalTitle{Journal of Electronic Science and Technology}} \textbf{17}, 26--40 (2019).

\bibitem{sena2021bayesian}
M.~Sena, M.~S. Erkilinc, T.~Dippon, B.~Shariati, R.~Emmerich, J.~K. Fischer, and R.~Freund, \enquote{Bayesian optimization for nonlinear system identification and pre-distortion in cognitive transmitters,} {\protect\JournalTitle{Journal of Lightwave Technology}} \textbf{39}, 5008--5020 (2021).

\bibitem{smithson2016neural}
S.~C. Smithson, G.~Yang, W.~J. Gross, and B.~H. Meyer, \enquote{Neural networks designing neural networks: multi-objective hyper-parameter optimization,} in \emph{2016 IEEE/ACM International Conference on Computer-Aided Design (ICCAD),}  (IEEE, 2016), pp. 1--8.

\bibitem{talbi2021automated}
E.-G. Talbi, \enquote{Automated design of deep neural networks: A survey and unified taxonomy,} {\protect\JournalTitle{ACM Computing Surveys (CSUR)}} \textbf{54}, 1--37 (2021).

\bibitem{pinos2022evolutionary}
M.~Pinos, V.~Mrazek, and L.~Sekanina, \enquote{Evolutionary approximation and neural architecture search,} {\protect\JournalTitle{Genetic Programming and Evolvable Machines}} \textbf{23}, 351--374 (2022).

\bibitem{shahriari2015taking}
B.~Shahriari, K.~Swersky, Z.~Wang, R.~P. Adams, and N.~De~Freitas, \enquote{Taking the human out of the loop: A review of bayesian optimization,} {\protect\JournalTitle{Proceedings of the IEEE}} \textbf{104}, 148--175 (2015).

\bibitem{hutter2011sequential}
F.~Hutter, H.~H. Hoos, and K.~Leyton-Brown, \enquote{Sequential model-based optimization for general algorithm configuration,} in \emph{International conference on learning and intelligent optimization,}  (Springer, 2011), pp. 507--523.

\bibitem{joyce2018review}
T.~Joyce and J.~M. Herrmann, \enquote{A review of no free lunch theorems, and their implications for metaheuristic optimisation,} {\protect\JournalTitle{Nature-inspired algorithms and applied optimization}} pp. 27--51 (2018).

\bibitem{baker2016designing}
B.~Baker, O.~Gupta, N.~Naik, and R.~Raskar, \enquote{Designing neural network architectures using reinforcement learning,} {\protect\JournalTitle{arXiv preprint arXiv:1611.02167}}  (2016).

\bibitem{9420739}
A.~Iranfar, M.~Zapater, and D.~Atienza, \enquote{Multiagent reinforcement learning for hyperparameter optimization of convolutional neural networks,} {\protect\JournalTitle{IEEE Transactions on Computer-Aided Design of Integrated Circuits and Systems}} \textbf{41}, 1034--1047 (2022).

\bibitem{xu2022automatic}
Y.~Xu, L.~Huang, W.~Jiang, L.~Xue, W.~Hu, and L.~Yi, \enquote{Automatic optimization of volterra equalizer with deep reinforcement learning for intensity-modulated direct-detection optical communications,} {\protect\JournalTitle{Journal of Lightwave Technology}} \textbf{40}, 5395--5406 (2022).

\bibitem{agrawal21}
G.~P. Agrawal, \emph{Fiber-Optic Communication Systems} (Wiley, 2021), 5th ed.

\bibitem{rafique2011compensation}
D.~Rafique, M.~Mussolin, M.~Forzati, J.~M{\aa}rtensson, M.~N. Chugtai, and A.~D. Ellis, \enquote{Compensation of intra-channel nonlinear fibre impairments using simplified digital back-propagation algorithm,} {\protect\JournalTitle{Optics express}} \textbf{19}, 9453--9460 (2011).

\bibitem{napoli2014reduced}
A.~Napoli, Z.~Maalej, V.~A. Sleiffer, M.~Kuschnerov, D.~Rafique, E.~Timmers, B.~Spinnler, T.~Rahman, L.~D. Coelho, and N.~Hanik, \enquote{Reduced complexity digital back-propagation methods for optical communication systems,} {\protect\JournalTitle{Journal of Lightwave Technology}} \textbf{32}, 1351--1362 (2014).

\bibitem{musetti2018accuracy}
S.~Musetti, P.~Serena, and A.~Bononi, \enquote{On the accuracy of split-step fourier simulations for wideband nonlinear optical communications,} {\protect\JournalTitle{Journal of Lightwave Technology}} \textbf{36}, 5669--5677 (2018).

\bibitem{serena2019numerical}
P.~Serena, C.~Lasagni, S.~Musetti, and A.~Bononi, \enquote{On numerical simulations of ultra-wideband long-haul optical communication systems,} {\protect\JournalTitle{Journal of Lightwave Technology}} \textbf{38}, 1019--1031 (2019).

\bibitem{jaworski2008step}
M.~Jaworski, \enquote{Step-size distribution strategies in ssfm simulation of dwdm links,} in \emph{2008 2nd ICTON Mediterranean Winter,}  (IEEE, 2008), pp. 1--6.

\bibitem{schmauss2012recent}
B.~Schmauss, R.~Asif, and C.-Y. Lin, \enquote{Recent advances in digital backward propagation algorithm for coherent transmission systems with higher order modulation formats,} {\protect\JournalTitle{Next-Generation Optical Communication: Components, Sub-Systems, and Systems}} \textbf{8284}, 151--165 (2012).

\bibitem{hager2018nonlinear}
C.~H{\"a}ger and H.~D. Pfister, \enquote{Nonlinear interference mitigation via deep neural networks,} in \emph{2018 Optical Fiber Communications Conference and Exposition (OFC),}  (IEEE, 2018), pp. 1--3.

\bibitem{hager2020physics}
C.~H{\"a}ger and H.~D. Pfister, \enquote{Physics-based deep learning for fiber-optic communication systems,} {\protect\JournalTitle{IEEE Journal on Selected Areas in Communications}} \textbf{39}, 280--294 (2020).

\bibitem{ZHANG202243}
S.~Zhang and C.~Häger, \enquote{Chapter two - machine learning for long-haul optical systems,} in \emph{Machine Learning for Future Fiber-Optic Communication Systems,}  A.~P.~T. Lau and F.~N. Khan, eds. (Academic Press, 2022), pp. 43--64.

\bibitem{NNPDE03}
M.~Raissi, P.~Perdikaris, and G.~Karniadakis, \enquote{Physics-informed neural networks: A deep learning framework for solving forward and inverse problems involving nonlinear partial differential equations,} {\protect\JournalTitle{Journal of Computational Physics}} \textbf{378}, 686--707 (2019).

\bibitem{karniadakis2021physics}
G.~E. Karniadakis, I.~G. Kevrekidis, L.~Lu, P.~Perdikaris, S.~Wang, and L.~Yang, \enquote{Physics-informed machine learning,} {\protect\JournalTitle{Nature Reviews Physics}} \textbf{3}, 422--440 (2021).

\bibitem{cuomo2022scientific}
S.~Cuomo, V.~S. Di~Cola, F.~Giampaolo, G.~Rozza, M.~Raissi, and F.~Piccialli, \enquote{Scientific machine learning through physics-informed neural networks: Where we are and what's next,} {\protect\JournalTitle{arXiv preprint arXiv:2201.05624}}  (2022).

\bibitem{jiang2021solving}
X.~Jiang, D.~Wang, Q.~Fan, M.~Zhang, C.~Lu, and A.~P.~T. Lau, \enquote{Solving the nonlinear schr{\"o}dinger equation in optical fibers using physics-informed neural network,} in \emph{Optical Fiber Communication Conference,}  (Optica Publishing Group, 2021), pp. M3H--8.

\bibitem{zang2021principle}
Y.~Zang, Z.~Yu, K.~Xu, X.~Lan, M.~Chen, S.~Yang, and H.~Chen, \enquote{Principle-driven fiber transmission model based on pinn neural network,} {\protect\JournalTitle{Journal of Lightwave Technology}} \textbf{40}, 404--414 (2021).

\bibitem{wang2022applications}
D.~Wang, X.~Jiang, Y.~Song, M.~Fu, Z.~Zhang, X.~Chen, and M.~Zhang, \enquote{Applications of physics-informed neural network for optical fiber communications,} {\protect\JournalTitle{IEEE Communications Magazine}} \textbf{60}, 32--37 (2022).

\bibitem{yang2020fast}
H.~Yang, Z.~Niu, S.~Xiao, J.~Fang, Z.~Liu, D.~Fainsin, and L.~Yi, \enquote{Fast and accurate optical fiber channel modeling using generative adversarial network,} {\protect\JournalTitle{Journal of Lightwave Technology}} \textbf{39}, 1322--1333 (2020).

\bibitem{li2020fourier}
Z.~Li, N.~B. Kovachki, K.~Azizzadenesheli, K.~Bhattacharya, A.~Stuart, A.~Anandkumar \emph{et~al.}, \enquote{Fourier neural operator for parametric partial differential equations,} in \emph{International Conference on Learning Representations,}  (2020).

\bibitem{lu2021learning}
L.~Lu, P.~Jin, G.~Pang, Z.~Zhang, and G.~E. Karniadakis, \enquote{Learning nonlinear operators via deeponet based on the universal approximation theorem of operators,} {\protect\JournalTitle{Nature Machine Intelligence}} \textbf{3}, 218--229 (2021).

\bibitem{wang2021learning}
S.~Wang, H.~Wang, and P.~Perdikaris, \enquote{Learning the solution operator of parametric partial differential equations with physics-informed deeponets,} {\protect\JournalTitle{Science advances}} \textbf{7}, eabi8605 (2021).

\bibitem{he2022fourier}
X.~He, L.~Yan, L.~Jiang, A.~Yi, Z.~Pu, Y.~Yu, H.~Chen, W.~Pan, and B.~Luo, \enquote{Fourier neural operator for accurate optical fiber modeling with low complexity,} {\protect\JournalTitle{Journal of Lightwave Technology}} \textbf{in press} (2022).

\bibitem{zhang2022transformer}
N.~Zhang, H.~Yang, Z.~Niu, L.~Zheng, C.~Chen, S.~Xiao, and L.~Yi, \enquote{Transformer-based long distance fiber channel modeling for optical ofdm systems,} {\protect\JournalTitle{Journal of Lightwave Technology}}  (2022).

\bibitem{gautam2022optidistillnet}
N.~Gautam, V.~Kaushik, A.~Choudhary, and B.~Lall, \enquote{Optidistillnet: Learning nonlinear pulse propagation using the student-teacher model,} {\protect\JournalTitle{Optics Express}} \textbf{30}, 42430--42439 (2022).

\bibitem{winzer2018fiber}
P.~J. Winzer, D.~T. Neilson, and A.~R. Chraplyvy, \enquote{Fiber-optic transmission and networking: the previous 20 and the next 20 years,} {\protect\JournalTitle{Optics Express}} \textbf{26}, 24190--24239 (2018).

\bibitem{Cartledge:17}
J.~C. Cartledge, F.~P. Guiomar, F.~R. Kschischang, G.~Liga, and M.~P. Yankov, \enquote{Digital signal processing for fiber nonlinearities [invited],} {\protect\JournalTitle{Optics Express}} \textbf{25}, 1916--1936 (2017).

\bibitem{jarajreh2014artificial}
M.~A. Jarajreh, E.~Giacoumidis, I.~Aldaya, S.~T. Le, A.~Tsokanos, Z.~Ghassemlooy, and N.~J. Doran, \enquote{Artificial neural network nonlinear equalizer for coherent optical ofdm,} {\protect\JournalTitle{IEEE Photonics Technology Letters}} \textbf{27}, 387--390 (2014).

\bibitem{Hunt_2009}
S.~Hunt, Y.~Sun, A.~Shafarenko, R.~Adams, N.~Davey, B.~Slater, R.~Bhamber, S.~Boscolo, and S.~K. Turitsyn, \enquote{Adaptive electrical signal post-processing with varying representations in optical communication systems,} in \emph{Engineering Applications of Neural Networks,}  (Springer Berlin Heidelberg, 2009), pp. 235--245.

\bibitem{eriksson2017applying}
T.~A. Eriksson, H.~B{\"u}low, and A.~Leven, \enquote{Applying neural networks in optical communication systems: possible pitfalls,} {\protect\JournalTitle{IEEE Photonics Technology Letters}} \textbf{29}, 2091--2094 (2017).

\bibitem{zhang2019field}
S.~Zhang, F.~Yaman, K.~Nakamura, T.~Inoue, V.~Kamalov, L.~Jovanovski, V.~Vusirikala, E.~Mateo, Y.~Inada, and T.~Wang, \enquote{Field and lab experimental demonstration of nonlinear impairment compensation using neural networks,} {\protect\JournalTitle{Nature Communications}} \textbf{10}, 3033 (2019).

\bibitem{khan2017machine}
F.~N. Khan, C.~Lu, and A.~P.~T. Lau, \enquote{Machine learning methods for optical communication systems,} in \emph{Signal Processing in Photonic Communications,}  (Optical Society of America, 2017), pp. SpW2F--3.

\bibitem{Karanov:10}
B.~Karanov, D.~Lavery, P.~Bayvel, and L.~Schmalen, \enquote{End-to-end optimized transmission over dispersive intensity-modulated channels using bidirectional recurrent neural networks,} {\protect\JournalTitle{Optics Express}} \textbf{27}, 19650--19663 (2019).

\bibitem{Khan19}
F.~N. Khan, Q.~Fan, C.~Lu, and A.~P.~T. Lau, \enquote{An optical communication's perspective on machine learning and its applications,} {\protect\JournalTitle{J. Lightw. Technol.}} \textbf{37}, 493--516 (2019).

\bibitem{giacoumidis2018harnessing}
E.~Giacoumidis, Y.~Lin, J.~Wei, I.~Aldaya, A.~Tsokanos, and L.~P. Barry, \enquote{Harnessing machine learning for fiber-induced nonlinearity mitigation in long-haul coherent optical ofdm,} {\protect\JournalTitle{Future internet}} \textbf{11}, 2 (2018).

\bibitem{charalabopoulos2003frequency}
G.~Charalabopoulos, P.~Stavroulakis, and A.~H. Aghvami, \enquote{A frequency-domain neural network equalizer for ofdm,} in \emph{GLOBECOM'03. IEEE Global Telecommunications Conference (IEEE Cat. No. 03CH37489),}  vol.~2 (IEEE, 2003), pp. 571--575.

\bibitem{estaran2016artificial}
J.~Estaran, R.~Rios-M{\"u}ller, M.~Mestre, F.~Jorge, H.~Mardoyan, A.~Konczykowska, J.-Y. Dupuy, and S.~Bigo, \enquote{Artificial neural networks for linear and non-linear impairment mitigation in high-baudrate im/dd systems,} in \emph{ECOC 2016; 42nd European Conference on Optical Communication,}  (VDE, 2016), pp. 1--3.

\bibitem{ye2017demonstration}
C.~Ye, D.~Zhang, X.~Huang, H.~Feng, and K.~Zhang, \enquote{Demonstration of 50gbps im/dd pam4 pon over 10ghz class optics using neural network based nonlinear equalization,} in \emph{2017 European Conference on Optical Communication (ECOC),}  (IEEE, 2017), pp. 1--3.

\bibitem{sang2021multi}
B.~Sang, J.~Zhang, C.~Wang, M.~Kong, Y.~Tan, L.~Zhao, W.~Zhou, D.~Shang, Y.~Zhu, H.~Yi \emph{et~al.}, \enquote{Multi-symbol output long short-term memory neural network equalizer for 200+ gbps im/dd system,} in \emph{2021 European Conference on Optical Communication (ECOC),}  (IEEE, 2021), pp. 1--4.

\bibitem{arnold2022spiking}
E.~Arnold, G.~B{\"o}cherer, E.~M{\"u}ller, P.~Spilger, J.~Schemmel, S.~Calabr{\`o}, and M.~Kuschnerov, \enquote{Spiking neural network equalization for im/dd optical communication,} {\protect\JournalTitle{arXiv preprint arXiv:2205.04263}}  (2022).

\bibitem{freire2020complex}
P.~J. Freire, V.~Neskornuik, A.~Napoli, B.~Spinnler, N.~Costa, G.~Khanna, E.~Riccardi, J.~E. Prilepsky, and S.~K. Turitsyn, \enquote{Complex-valued neural network design for mitigation of signal distortions in optical links,} {\protect\JournalTitle{Journal of Lightwave Technology}} \textbf{39}, 1696--1705 (2021).

\bibitem{da2020reservoir}
F.~Da~Ros, S.~M. Ranzini, H.~B{\"u}low, and D.~Zibar, \enquote{Reservoir-computing based equalization with optical pre-processing for short-reach optical transmission,} {\protect\JournalTitle{IEEE Journal of Selected Topics in Quantum Electronics}} \textbf{26}, 1--12 (2020).

\bibitem{da2021machine}
F.~Da~Ros, S.~M. Ranzini, R.~Dischler, A.~Cem, V.~Aref, H.~B{\"u}low, and D.~Zibar, \enquote{Machine-learning-based equalization for short-reach transmission: neural networks and reservoir computing,} in \emph{Metro and Data Center Optical Networks and Short-Reach Links IV,}  vol. 11712 (SPIE, 2021), p. 1171205.

\bibitem{wang2021signal}
S.~Wang, N.~Fang, and L.~Wang, \enquote{Signal recovery based on optoelectronic reservoir computing for high speed optical fiber communication system,} {\protect\JournalTitle{Optics Communications}} \textbf{495}, 127082 (2021).

\bibitem{da2022reservoir}
F.~Da~Ros, S.~M. Ranzini, Y.~Osadchuk, A.~Cem, B.~J.~G. Castro, and D.~Zibar, \enquote{Reservoir-computing and neural-network-based equalization for short reach communication,} in \emph{Signal Processing in Photonic Communications,}  (Optica Publishing Group, 2022), pp. SpTu1J--1.

\bibitem{freire2021performance}
P.~J. Freire, Y.~Osadchuk, B.~Spinnler, A.~Napoli, W.~Schairer, N.~Costa, J.~E. Prilepsky, and S.~K. Turitsyn, \enquote{Performance versus complexity study of neural network equalizers in coherent optical systems,} {\protect\JournalTitle{Journal of Lightwave Technology}} \textbf{39}, 6085--6096 (2021).

\bibitem{freire2022deep}
P.~J. Freire, J.~E. Prilepsky, Y.~Osadchuk, S.~K. Turitsyn, and V.~Aref, \enquote{Deep neural network-aided soft-demapping in coherent optical systems: Regression versus classification,} {\protect\JournalTitle{IEEE Transactions on Communications}} \textbf{70}, 7973--7988 (2022).

\bibitem{diedolo2022nonlinear}
F.~Diedolo, G.~B{\"o}cherer, M.~Sch{\"a}dler, and S.~Calabr{\'o}, \enquote{Nonlinear equalization for optical communications based on entropy-regularized mean square error,} {\protect\JournalTitle{arXiv preprint arXiv:2206.01004}}  (2022).

\bibitem{sidelnikov2018equalization}
O.~Sidelnikov, A.~Redyuk, and S.~Sygletos, \enquote{Equalization performance and complexity analysis of dynamic deep neural networks in long haul transmission systems,} {\protect\JournalTitle{Optics Express}} \textbf{26}, 32765--32776 (2018).

\bibitem{giacoumidis2016experimental}
E.~Giacoumidis, S.~T. Le, I.~Aldaya, J.~Wei, M.~McCarthy, N.~Doran, and B.~J. Eggleton, \enquote{Experimental comparison of artificial neural network and volterra based nonlinear equalization for co-ofdm,} in \emph{Optical Fiber Communication Conference,}  (Optical Society of America, 2016), pp. W3A--4.

\bibitem{deligiannidis2021performance}
S.~Deligiannidis, C.~Mesaritakis, and A.~Bogris, \enquote{Performance and complexity analysis of bi-directional recurrent neural network models versus volterra nonlinear equalizers in digital coherent systems,} {\protect\JournalTitle{Journal of Lightwave Technology}} \textbf{39}, 5791--5798 (2021).

\bibitem{freire2022computational}
P.~J. Freire, S.~Srivallapanondh, A.~Napoli, J.~E. Prilepsky, and S.~K. Turitsyn, \enquote{Computational complexity evaluation of neural network applications in signal processing,} {\protect\JournalTitle{arXiv preprint arXiv:2206.12191}}  (2022).

\bibitem{sang2022low}
B.~Sang, W.~Zhou, Y.~Tan, M.~Kong, C.~Wang, M.~Wang, L.~Zhao, J.~Zhang, and J.~Yu, \enquote{Low complexity neural network equalization based on multi-symbol output technique for 200+ gbps im/dd short reach optical system,} {\protect\JournalTitle{Journal of Lightwave Technology}} \textbf{40}, 2890--2900 (2022).

\bibitem{freire2022reducing}
P.~J. Freire, A.~Napoli, D.~A. Ron, B.~Spinnler, M.~Anderson, W.~Schairer, T.~Bex, N.~Costa, S.~K. Turitsyn, and J.~E. Prilepsky, \enquote{Reducing computational complexity of neural networks in optical channel equalization: From concepts to implementation,} {\protect\JournalTitle{arXiv preprint arXiv:2208.12866}}  (2022).

\bibitem{deligiannidis2020compensation}
S.~{Deligiannidis}, A.~{Bogris}, C.~{Mesaritakis}, and Y.~{Kopsinis}, \enquote{Compensation of fiber nonlinearities in digital coherent systems leveraging long short-term memory neural networks,} {\protect\JournalTitle{Journal of Lightwave Technology}} \textbf{38}, 5991--5999 (2020).

\bibitem{Bitachon20}
B.~I. Bitachon, A.~Ghazisaeidi, M.~Eppenberger, B.~Baeuerle, M.~Ayata, and J.~Leuthold, \enquote{Deep learning based digital backpropagation demonstrating snr gain at low complexity in a 1200 km transmission link,} {\protect\JournalTitle{Optics Express}} \textbf{28}, 29318--29334 (2020).

\bibitem{sidelnikov2021advanced}
O.~Sidelnikov, A.~Redyuk, S.~Sygletos, M.~Fedoruk, and S.~Turitsyn, \enquote{Advanced convolutional neural networks for nonlinearity mitigation in long-haul wdm transmission systems,} {\protect\JournalTitle{Journal of Lightwave Technology}} \textbf{39}, 2397--2406 (2021).

\bibitem{fan2020advancing}
Q.~Fan, G.~Zhou, T.~Gui, C.~Lu, and A.~P.~T. Lau, \enquote{Advancing theoretical understanding and practical performance of signal processing for nonlinear optical communications through machine learning,} {\protect\JournalTitle{Nature Communications}} \textbf{11}, 1--11 (2020).

\bibitem{luo2022nonlinear}
X.~Luo, C.~Bai, X.~Chi, H.~Xu, Y.~Fan, L.~Yang, P.~Qin, Z.~Wang, and X.~Lv, \enquote{Nonlinear impairment compensation using transfer learning-assisted convolutional bidirectional long short-term memory neural network for coherent optical communication systems,} {\protect\JournalTitle{Photonics}} \textbf{9}, 919 (2022).

\bibitem{barreiro2022data}
A.~Barreiro, G.~Liga, and A.~Alvarado, \enquote{Data-driven enhancement of the time-domain first-order regular perturbation model,} {\protect\JournalTitle{arXiv preprint arXiv:2210.05340}}  (2022).

\bibitem{melek2020nonlinearity}
M.~M. Melek and D.~Yevick, \enquote{Nonlinearity mitigation with a perturbation based neural network receiver,} {\protect\JournalTitle{Optical and Quantum Electronics}} \textbf{52}, 1--10 (2020).

\bibitem{melek2021fiber}
M.~M. Melek and D.~Yevick, \enquote{Fiber nonlinearity mitigation with a perturbation based siamese neural network receiver,} {\protect\JournalTitle{Optical Fiber Technology}} \textbf{66}, 102641 (2021).

\bibitem{li2022convolutional}
C.~Li, Y.~Wang, J.~Wang, H.~Yao, X.~Liu, R.~Gao, L.~Yang, H.~Xu, Q.~Zhang, P.~Ma \emph{et~al.}, \enquote{Convolutional neural network-aided dp-64 qam coherent optical communication systems,} {\protect\JournalTitle{Journal of Lightwave Technology}} \textbf{40}, 2880--2889 (2022).

\bibitem{redyuk2020compensation}
A.~Redyuk, E.~Averyanov, O.~Sidelnikov, M.~Fedoruk, and S.~Turitsyn, \enquote{Compensation of nonlinear impairments using inverse perturbation theory with reduced complexity,} {\protect\JournalTitle{Journal of Lightwave Technology}} \textbf{38}, 1250--1257 (2020).

\bibitem{dzieciol2022inverse}
H.~Dzieciol, T.~Koike-Akino, Y.~Wang, and K.~Parsons, \enquote{Inverse regular perturbation with ml-assisted phasor correction for fiber nonlinearity compensation,} {\protect\JournalTitle{Optics Letters}} \textbf{47}, 3471--3474 (2022).

\bibitem{castro2022novel}
N.~Castro and S.~Sygletos, \enquote{A novel learned volterra-based scheme for time-domain nonlinear equalization,} in \emph{CLEO: Science and Innovations,}  (Optica Publishing Group, 2022), pp. SF3M--1.

\bibitem{huang2022low}
X.~Huang, D.~Zhang, X.~Hu, C.~Ye, and K.~Zhang, \enquote{Low-complexity recurrent neural network based equalizer with embedded parallelization for 100-gbit/s/$\lambda$ pon,} {\protect\JournalTitle{Journal of Lightwave Technology}} \textbf{40}, 1353--1359 (2022).

\bibitem{complexNN}
A.~A. Cruz, K.~S. Mayer, and D.~S. Arantes, \enquote{Rosenpy: An open source python framework for complex-valued neural networks,} \url{https://dx.doi.org/10.2139/ssrn.4252610}. Accessed: 2022-12-01.

\bibitem{liu2017multilevel}
S.~Liu, M.~Xu, J.~Wang, F.~Lu, W.~Zhang, H.~Tian, and G.-K. Chang, \enquote{A multilevel artificial neural network nonlinear equalizer for millimeter-wave mobile fronthaul systems,} {\protect\JournalTitle{Journal of Lightwave Technology}} \textbf{35}, 4406--4417 (2017).

\bibitem{wang2021optical}
L.~Wang, M.~Gao, Y.~Zhang, F.~Cao, and H.~Huang, \enquote{Optical phase conjugation with complex-valued deep neural network for wdm 64-qam coherent optical systems,} {\protect\JournalTitle{IEEE Photonics Journal}} \textbf{13}, 1--8 (2021).

\bibitem{bogdanov2021application}
S.~A. Bogdanov, O.~S. Sidelnikov, and A.~A. Redyuk, \enquote{Application of complex fully connected neural networks to compensate for nonlinearity in fibre-optic communication lines with polarisation division multiplexing,} {\protect\JournalTitle{Quantum Electronics}} \textbf{51}, 1076 (2021).

\bibitem{he2021fiber}
P.~He, F.~Wu, M.~Yang, A.~Yang, P.~Guo, Y.~Qiao, and X.~Xin, \enquote{A fiber nonlinearity compensation scheme with complex-valued dimension-reduced neural network,} {\protect\JournalTitle{IEEE Photonics Journal}} \textbf{13}, 1--7 (2021).

\bibitem{yang2021modified}
H.~Yang, X.~Zhang, A.~Yi, R.~Wang, B.~Lin, H.~Xing, and B.~Sha, \enquote{A modified convolutional neural network-based signal demodulation method for direct detection ofdm/oqam-pon,} {\protect\JournalTitle{Optics Communications}} \textbf{489}, 126843 (2021).

\bibitem{ming2022ultralow}
H.~Ming, X.~Chen, X.~Fang, L.~Zhang, C.~Li, and F.~Zhang, \enquote{Ultralow complexity long short-term memory network for fiber nonlinearity mitigation in coherent optical communication systems,} {\protect\JournalTitle{Journal of Lightwave Technology}} \textbf{40}, 2427--2434 (2022).

\bibitem{liu2022attention}
Y.~Liu, V.~Sanchez, P.~J. Freire, J.~E. Prilepsky, M.~J. Koshkouei, and M.~D. Higgins, \enquote{Attention-aided partial bidirectional rnn-based nonlinear equalizer in coherent optical systems,} {\protect\JournalTitle{Optics Express}} \textbf{30}, 32908--32923 (2022).

\bibitem{shahkarami2021attention}
A.~Shahkarami, M.~I. Yousefi, and Y.~Jaou{\"e}n, \enquote{Attention-based neural network equalization in fiber-optic communications,} in \emph{Asia Communications and Photonics Conference,}  (Optical Society of America, 2021), pp. M5H--3.

\bibitem{shahkarami2023efficient}
A.~Shahkarami, M.~I. Yousefi, and Y.~Jaouen, \enquote{Efficient deep learning of kerr nonlinearity in fiber-optic channels using a convolutional recurrent neural network,} in \emph{Deep Learning Applications, Volume 4,}  (Springer, 2023), pp. 317--338.

\bibitem{huang2022design}
X.~Huang, W.~Jiang, X.~Yi, J.~Zhang, T.~Jin, Q.~Zhang, B.~Xu, and K.~Qiu, \enquote{Design of fully interpretable neural networks for digital coherent demodulation,} {\protect\JournalTitle{Optics Express}} \textbf{30}, 35526--35538 (2022).

\bibitem{bajaj2022efficient}
V.~Bajaj, M.~Chagnon, S.~Wahls, and V.~Aref, \enquote{Efficient training of volterra series-based pre-distortion filter using neural networks,} in \emph{2022 Optical Fiber Communications Conference and Exhibition (OFC),}  (IEEE, 2022), pp. 1--3.

\bibitem{psaltis1988multilayered}
D.~Psaltis, A.~Sideris, and A.~A. Yamamura, \enquote{A multilayered neural network controller,} {\protect\JournalTitle{IEEE Control Systems Magazine}} \textbf{8}, 17--21 (1988).

\bibitem{bernardini1990use}
A.~Bernardini, M.~Carrarini, and S.~De~Fina, \enquote{The use of a neural net for copeing with nonlinear distortions,} in \emph{1990 20th European Microwave Conference,}  vol.~2 (IEEE, 1990), pp. 1718--1723.

\bibitem{gotthans2014digital}
T.~Gotthans, G.~Baudoin, and A.~Mbaye, \enquote{Digital predistortion with advance/delay neural network and comparison with volterra derived models,} in \emph{2014 IEEE 25th Annual International Symposium on Personal, Indoor, and Mobile Radio Communication (PIMRC),}  (IEEE, 2014), pp. 811--815.

\bibitem{hu2021convolutional}
X.~Hu, Z.~Liu, X.~Yu, Y.~Zhao, W.~Chen, B.~Hu, X.~Du, X.~Li, M.~Helaoui, W.~Wang \emph{et~al.}, \enquote{Convolutional neural network for behavioral modeling and predistortion of wideband power amplifiers,} {\protect\JournalTitle{IEEE Transactions on Neural Networks and Learning Systems}}  (2021).

\bibitem{schaedler2019ai}
M.~Schaedler, M.~Kuschnerov, S.~Calabr{\`o}, F.~Pittal{\`a}, C.~Bluemm, and S.~Pachnicke, \enquote{Ai-based digital predistortion for iq mach-zehnder modulators,} in \emph{2019 Asia Communications and Photonics Conference (ACP),}  (IEEE, 2019), pp. 1--3.

\bibitem{abu2019neural}
M.~Abu-Romoh, S.~Sygletos, I.~D. Phillips, and W.~Forysiak, \enquote{Neural-network-based pre-distortion method to compensate for low resolution dac nonlinearity,} in \emph{45th European Conference on Optical Communication (ECOC 2019),}  (IET, 2019), pp. 1--4.

\bibitem{sasai2020wiener}
T.~Sasai, M.~Nakamura, E.~Yamazaki, A.~Matsushita, S.~Okamoto, K.~Horikoshi, and Y.~Kisaka, \enquote{Wiener-hammerstein model and its learning for nonlinear digital pre-distortion of optical transmitters,} {\protect\JournalTitle{Optics Express}} \textbf{28}, 30952--30963 (2020).

\bibitem{bajaj2022deep}
V.~Bajaj, F.~Buchali, M.~Chagnon, S.~Wahls, and V.~Aref, \enquote{Deep neural network-based digital pre-distortion for high baudrate optical coherent transmission,} {\protect\JournalTitle{Journal of Lightwave Technology}} \textbf{40}, 597--606 (2022).

\bibitem{bajaj2020single}
V.~Bajaj, F.~Buchali, M.~Chagnon, S.~Wahls, and V.~Aref, \enquote{Single-channel 1.61 tb/s optical coherent transmission enabled by neural network-based digital pre-distortion,} in \emph{2020 European Conference on Optical Communications (ECOC),}  (IEEE, 2020), pp. 1--4.

\bibitem{bajaj202154}
V.~Bajaj, F.~Buchali, M.~Chagnon, S.~Wahls, and V.~Aref, \enquote{54.5 tb/s wdm transmission over field deployed fiber enabled by neural network-based digital pre-distortion,} in \emph{Optical Fiber Communication Conference,}  (Optica Publishing Group, 2021), pp. M5F--2.

\bibitem{minelli2022multi}
L.~Minelli, F.~Forghieri, A.~Nespola, S.~Straullu, and R.~Gaudino, \enquote{A multi-rate approach for nonlinear pre-distortion using end-to-end deep learning in im-dd systems,} {\protect\JournalTitle{Journal of Lightwave Technology}}  (2022).

\bibitem{bajaj2022performance}
V.~Bajaj, V.~Aref, and S.~Wahls, \enquote{Performance analysis of recurrent neural network-based digital pre-distortion for optical coherent transmission,} in \emph{2022 European Conference on Optical Communication (ECOC),}  (IEEE, 2022), pp. 1--4.

\bibitem{o2017introduction}
T.~O’shea and J.~Hoydis, \enquote{An introduction to deep learning for the physical layer,} {\protect\JournalTitle{IEEE Transactions on Cognitive Communications and Networking}} \textbf{3}, 563--575 (2017).

\bibitem{glasmachers2017limits}
T.~Glasmachers, \enquote{Limits of end-to-end learning,} in \emph{Asian conference on machine learning,}  (PMLR, 2017), pp. 17--32.

\bibitem{bojarski2016end}
M.~Bojarski, D.~Del~Testa, D.~Dworakowski, B.~Firner, B.~Flepp, P.~Goyal, L.~D. Jackel, M.~Monfort, U.~Muller, J.~Zhang \emph{et~al.}, \enquote{End to end learning for self-driving cars,} {\protect\JournalTitle{arXiv preprint arXiv:1604.07316}}  (2016).

\bibitem{karanov2018end}
B.~Karanov, M.~Chagnon, F.~Thouin, T.~A. Eriksson, H.~B{\"u}low, D.~Lavery, P.~Bayvel, and L.~Schmalen, \enquote{End-to-end deep learning of optical fiber communications,} {\protect\JournalTitle{Journal of Lightwave Technology}} \textbf{36}, 4843--4855 (2018).

\bibitem{9195215}
B.~{Karanov}, M.~{Chagnon}, V.~{Aref}, F.~{Ferreira}, D.~{Lavery}, P.~{Bayvel}, and L.~{Schmalen}, \enquote{Experimental investigation of deep learning for digital signal processing in short reach optical fiber communications,} in \emph{2020 IEEE Workshop on Signal Processing Systems (SiPS),}  (2020), pp. 1--6.

\bibitem{9890753}
V.~Neskorniuk, A.~Carnio, D.~Marsella, S.~K. Turitsyn, J.~E. Prilepsky, and V.~Aref, \enquote{Model-based deep learning of joint probabilistic and geometric shaping for optical communication,} in \emph{2022 Conference on Lasers and Electro-Optics (CLEO),}  (2022), pp. 1--2.

\bibitem{Neskorniuk2022}
V.~Neskorniuk, A.~Carnio, D.~Marsella, S.~K. Turitsyn, J.~E. Prilepsky, and V.~Aref, \enquote{Memory-aware end-to-end learning of channel distortions in optical coherent communications,} {\protect\JournalTitle{Optics Express}} \textbf{31}, 1--20 (2023).

\bibitem{jovanovic2021gradient}
O.~Jovanovic, M.~P. Yankov, F.~Da~Ros, and D.~Zibar, \enquote{Gradient-free training of autoencoders for non-differentiable communication channels,} {\protect\JournalTitle{Journal of Lightwave Technology}} \textbf{39}, 6381--6391 (2021).

\bibitem{karanov2020concept}
B.~Karanov, M.~Chagnon, V.~Aref, D.~Lavery, P.~Bayvel, and L.~Schmalen, \enquote{Concept and experimental demonstration of optical im/dd end-to-end system optimization using a generative model,} in \emph{Optical Fiber Communication Conference,}  (Optica Publishing Group, 2020), pp. Th2A--48.

\bibitem{karanov2020end}
B.~Karanov, V.~Oliari, M.~Chagnon, G.~Liga, A.~Alvarado, V.~Aref, D.~Lavery, P.~Bayvel, and L.~Schmalen, \enquote{End-to-end learning in optical fiber communications: Experimental demonstration and future trends,} in \emph{2020 European Conference on Optical Communications (ECOC),}  (IEEE, 2020), pp. 1--4.

\bibitem{gaiarin2020end}
S.~Gaiarin, F.~Da~Ros, R.~T. Jones, and D.~Zibar, \enquote{End-to-end optimization of coherent optical communications over the split-step fourier method guided by the nonlinear fourier transform theory,} {\protect\JournalTitle{Journal of Lightwave Technology}} \textbf{39}, 418--428 (2020).

\bibitem{jones2018deep}
R.~T. Jones, T.~A. Eriksson, M.~P. Yankov, and D.~Zibar, \enquote{Deep learning of geometric constellation shaping including fiber nonlinearities,} in \emph{2018 European Conference on Optical Communication (ECOC),}  (IEEE, 2018), pp. 1--3.

\bibitem{jones2019end}
R.~T. Jones, M.~P. Yankov, and D.~Zibar, \enquote{End-to-end learning for gmi optimized geometric constellation shape,} in \emph{2019 European Conference on Optical Communication (ECOC),}  (IET, 2019), pp. 1--3.

\bibitem{gumucs2020end}
K.~G{\"u}m{\"u}{\c{s}}, A.~Alvarado, B.~Chen, C.~H{\"a}ger, and E.~Agrell, \enquote{End-to-end learning of geometrical shaping maximizing generalized mutual information,} in \emph{2020 Optical Fiber Communications Conference and Exhibition (OFC),}  (IEEE, 2020), pp. 1--3.

\bibitem{oliari2021high}
V.~Oliari, B.~Karanov, S.~Goossens, G.~Liga, O.~Vassilieva, I.~Kim, P.~Palacharla, C.~Okonkwo, and A.~Alvarado, \enquote{High-cardinality hybrid shaping for 4d modulation formats in optical communications optimized via end-to-end learning,} {\protect\JournalTitle{arXiv preprint arXiv:2112.10471}}  (2021).

\bibitem{niu2022end}
Z.~Niu, H.~Yang, H.~Zhao, C.~Dai, W.~Hu, and L.~Yi, \enquote{End-to-end deep learning for long-haul fiber transmission using differentiable surrogate channel,} {\protect\JournalTitle{Journal of Lightwave Technology}} \textbf{40}, 2807--2822 (2022).

\bibitem{jovanovic2021end}
O.~Jovanovic, M.~P. Yankov, F.~Da~Ros, and D.~Zibar, \enquote{End-to-end learning of a constellation shape robust to variations in snr and laser linewidth,} in \emph{2021 European Conference on Optical Communication (ECOC),}  (IEEE, 2021), pp. 1--4.

\bibitem{song2021end}
J.~Song, C.~H{\"a}ger, J.~Schr{\"o}der, A.~G. i~Amat, and H.~Wymeersch, \enquote{End-to-end autoencoder for superchannel transceivers with hardware impairment,} in \emph{Optical Fiber Communication Conference,}  (Optica Publishing Group, 2021), pp. F4D--6.

\bibitem{he2022experimental}
Z.~He, J.~Song, C.~H{\"a}ger, A.~G. i~Amat, H.~Wymeersch, P.~A. Andrekson, M.~Karlsson, and J.~Schr{\"o}der, \enquote{Experimental demonstration of learned pulse shaping filter for superchannels,} in \emph{Optical Fiber Communication Conference,}  (Optica Publishing Group, 2022), pp. W2A--33.

\bibitem{song2022model}
J.~Song, C.~H{\"a}ger, J.~Schr{\"o}der, A.~G.~I. Amat, and H.~Wymeersch, \enquote{Model-based end-to-end learning for wdm systems with transceiver hardware impairments,} {\protect\JournalTitle{IEEE Journal of Selected Topics in Quantum Electronics}} \textbf{28}, 1--14 (2022).

\bibitem{uhlemann2020deep}
T.~Uhlemann, S.~Cammerer, A.~Span, S.~D{\"o}rner, and S.~ten Brink, \enquote{Deep-learning autoencoder for coherent and nonlinear optical communication,} in \emph{Photonic Networks; 21th ITG-Symposium,}  (VDE, 2020), pp. 1--8.

\bibitem{aref2022end}
V.~Aref and M.~Chagnon, \enquote{End-to-end learning of joint geometric and probabilistic constellation shaping,} in \emph{2022 Optical Fiber Communications Conference and Exhibition (OFC),}  (IEEE, 2022), pp. 1--3.

\bibitem{karanov2021distance}
B.~Karanov, L.~Schmalen, and A.~Alvarado, \enquote{Distance-agnostic auto-encoders for short reach fiber communications,} in \emph{2021 Optical Fiber Communications Conference and Exhibition (OFC),}  (IEEE, 2021), pp. 1--3.

\bibitem{ren2016experimental}
Y.~Ren, Z.~Wang, P.~Liao, L.~Li, G.~Xie, H.~Huang, Z.~Zhao, Y.~Yan, N.~Ahmed, A.~Willner \emph{et~al.}, \enquote{Experimental characterization of a 400 gbit/s orbital angular momentum multiplexed free-space optical link over 120 m,} {\protect\JournalTitle{Optics letters}} \textbf{41}, 622--625 (2016).

\bibitem{khalighi2014survey}
M.~A. Khalighi and M.~Uysal, \enquote{Survey on free space optical communication: A communication theory perspective,} {\protect\JournalTitle{IEEE communications surveys \& tutorials}} \textbf{16}, 2231--2258 (2014).

\bibitem{li2022enhanced}
Y.~Li, Z.~Chen, Z.~Hu, D.~M. Benton, A.~A. Ali, M.~Patel, M.~P. Lavery, and A.~D. Ellis, \enquote{Enhanced atmospheric turbulence resiliency with successive interference cancellation dsp in mode division multiplexing free-space optical links,} {\protect\JournalTitle{Journal of Lightwave Technology}}  (2022).

\bibitem{amirabadi2022low}
M.~A. Amirabadi, M.~H. Kahaei, and S.~A. Nezamalhosseni, \enquote{Low complexity deep learning algorithms for compensating atmospheric turbulence in the free space optical communication system,} {\protect\JournalTitle{IET Optoelectronics}} \textbf{16}, 93--105 (2022).

\bibitem{zheng2015svm}
C.~Zheng, S.~Yu, and W.~Gu, \enquote{A svm-based processor for free-space optical communication,} in \emph{2015 IEEE 5th International Conference on Electronics Information and Emergency Communication,}  (IEEE, 2015), pp. 30--33.

\bibitem{lohani2018turbulence}
S.~Lohani and R.~T. Glasser, \enquote{Turbulence correction with artificial neural networks,} {\protect\JournalTitle{Optics Letters}} \textbf{43}, 2611--2614 (2018).

\bibitem{hao2020high}
Y.~Hao, L.~Zhao, T.~Huang, Y.~Wu, T.~Jiang, Z.~Wei, D.~Deng, A.-P. Luo, and H.~Liu, \enquote{High-accuracy recognition of orbital angular momentum modes propagated in atmospheric turbulences based on deep learning,} {\protect\JournalTitle{IEEE Access}} \textbf{8}, 159542--159551 (2020).

\bibitem{tian2018turbo}
Q.~Tian, Z.~Li, K.~Hu, L.~Zhu, X.~Pan, Q.~Zhang, Y.~Wang, F.~Tian, X.~Yin, and X.~Xin, \enquote{Turbo-coded 16-ary oam shift keying fso communication system combining the cnn-based adaptive demodulator,} {\protect\JournalTitle{Optics Express}} \textbf{26}, 27849--27864 (2018).

\bibitem{li2018joint}
J.~Li, M.~Zhang, D.~Wang, S.~Wu, and Y.~Zhan, \enquote{Joint atmospheric turbulence detection and adaptive demodulation technique using the cnn for the oam-fso communication,} {\protect\JournalTitle{Optics Express}} \textbf{26}, 10494--10508 (2018).

\bibitem{bart2022deep}
M.~P. Bart, N.~J. Savino, P.~Regmi, L.~Cohen, H.~Safavi, H.~C. Shaw, S.~Lohani, T.~A. Searles, B.~T. Kirby, H.~Lee \emph{et~al.}, \enquote{Deep learning for enhanced free-space optical communications,} {\protect\JournalTitle{arXiv preprint arXiv:2208.07712}}  (2022).

\bibitem{zhu2019autoencoder}
Z.-R. Zhu, J.~Zhang, R.-H. Chen, and H.-Y. Yu, \enquote{Autoencoder-based transceiver design for owc systems in log-normal fading channel,} {\protect\JournalTitle{IEEE Photonics Journal}} \textbf{11}, 1--12 (2019).

\bibitem{YK14}
M.~Yousefi and F.~Kschischang, \enquote{Information transmission using the nonlinear {F}ourier transform, part i--iii,} {\protect\JournalTitle{IEEE Transactions on Information Theory}} \textbf{60}, 4312--4369 (2014).

\bibitem{TurOpt}
S.~K. Turitsyn, J.~E. Prilepsky, S.~T. Le, S.~Wahls, L.~L. Frumin, M.~Kamalian, and S.~A. Derevyanko, \enquote{{Nonlinear Fourier transform for optical data processing and transmission: advances and perspectives},} {\protect\JournalTitle{Optica}} \textbf{4}, 307 (2017).

\bibitem{china-rev2021}
L.~Xi, J.~Wei, and W.~Zhang, \enquote{Applications of machine learning on nonlinear frequency division multiplexing optic-fiber communication systems,} in \emph{2021 IEEE 9th International Conference on Information, Communication and Networks (ICICN),}  (2021), pp. 190--194.

\bibitem{nat2017}
S.~Le, V.~Aref, and H.~Buelow, \enquote{Nonlinear signal multiplexing for communication beyond the {K}err nonlinearity limit,} {\protect\JournalTitle{Nature Photonics}} \textbf{11}, 570 (2017).

\bibitem{gdd18}
S.~Gaiarin, F.~Da~Ros, N.~De~Renzis, E.~P. da~Silva, and D.~Zibar, \enquote{Dual-polarization nfdm transmission using distributed raman amplification and nft-domain equalization,} {\protect\JournalTitle{IEEE Photonics Technology Letters}} \textbf{30}, 1983--1986 (2018).

\bibitem{kwp19}
J.~Koch, R.~Weixer, and S.~Pachnicke, \enquote{Equalization of soliton transmission based on nonlinear fourier transform using neural networks,} in \emph{45th European Conference on Optical Communication (ECOC),}  (2019), pp. 1--3.

\bibitem{kkp19}
O.~Kotlyar, M.~K. Kopae, J.~E. Prilepsky, M.~Pankratova, and S.~K. Turitsyn, \enquote{Machine learning for performance improvement of periodic nft-based communication system,} in \emph{2019 European Conference on Optical Communications,}  (2019), pp. 1--3.

\bibitem{kpk20}
O.~Kotlyar, M.~Pankratova, M.~Kamalian-Kopae, A.~Vasylchenkova, J.~E. Prilepsky, and S.~K. Turitsyn, \enquote{Combining nonlinear fourier transform and neural network-based processing in optical communications,} {\protect\JournalTitle{Optics Letters}} \textbf{45}, 3462--3465 (2020).

\bibitem{kkp21}
O.~Kotlyar, M.~Kamalian-Kopae, M.~Pankratova, A.~Vasylchenkova, J.~E. Prilepsky, and S.~K. Turitsyn, \enquote{Convolutional long short-term memory neural network equalizer for nonlinear fourier transform-based optical transmission systems,} {\protect\JournalTitle{Optics Express}} \textbf{29}, 11254--11267 (2021).

\bibitem{kvkppt19}
M.~{Kamalian-Kopae}, A.~{Vasylchenkova}, O.~{Kotlyar}, M.~{Pankratova}, J.~{Prilepsky}, and S.~{Turitsyn}, \enquote{Artificial neural network-based equaliser in the nonlinear {F}ourier domain for fibre-optic communication applications,} in \emph{2019 Conference on Lasers and Electro-Optics Europe European Quantum Electronics Conference (CLEO/Europe-EQEC),}  (2019).

\bibitem{chen2021two}
X.~Chen, H.~Ming, C.~Li, G.~He, and F.~Zhang, \enquote{Two-stage artificial neural network-based burst-subcarrier joint equalization in nonlinear frequency division multiplexing systems,} {\protect\JournalTitle{Optics Letters}} \textbf{46}, 1700--1703 (2021).

\bibitem{lv2022noise}
X.~Lv, C.~Bai, Q.~Qi, H.~Xu, X.~Luo, X.~Chi, L.~Yang, and L.~Xi, \enquote{Noise equalization scheme based on complex-valued ann for multiple-eigenvalue modulated nonlinear frequency division multiplexing systems,} {\protect\JournalTitle{Applied Optics}} \textbf{61}, 10755--10765 (2022).

\bibitem{jgy18}
R.~T. Jones, S.~Gaiarin, M.~P. Yankov, and D.~Zibar, \enquote{Time-domain neural network receiver for nonlinear frequency division multiplexed systems,} {\protect\JournalTitle{IEEE Photonics Technology Letters}} \textbf{30}, 1079--1082 (2018).

\bibitem{ymm19}
S.~Yamamoto, K.~Mishina, and A.~Maruta, \enquote{Demodulation of optical eigenvalue modulated signal using neural network,} {\protect\JournalTitle{IEICE Communications Express}} \textbf{8}, 507--512 (2019).

\bibitem{wxz20}
Y.~Wu, L.~Xi, X.~Zhang, Z.~Zheng, J.~Wei, S.~Du, W.~Zhang, and X.~Zhang, \enquote{Robust neural network receiver for multiple-eigenvalue modulated nonlinear frequency division multiplexing system,} {\protect\JournalTitle{Optics Express}} \textbf{28}, 18304--18316 (2020).

\bibitem{mishina2021eigenvalue}
K.~Mishina, S.~Sato, Y.~Yoshida, D.~Hisano, and A.~Maruta, \enquote{Eigenvalue-domain neural network demodulator for eigenvalue-modulated signal,} {\protect\JournalTitle{Journal of Lightwave Technology}} \textbf{39}, 4307--4317 (2021).

\bibitem{mishina2021combining}
K.~Mishina, T.~Maeda, D.~Hisano, Y.~Yoshida, and A.~Maruta, \enquote{Combining ist-based cfo compensation and neural network-based demodulation for eigenvalue-modulated signal,} {\protect\JournalTitle{Journal of Lightwave Technology}} \textbf{39}, 7370--7382 (2021).

\bibitem{takeuchi2022eigenvalue}
H.~Takeuchi, K.~Mishina, Y.~Terashi, D.~Hisano, Y.~Yoshida, and A.~Maruta, \enquote{Eigenvalue-domain neural network receiver for 4096-ary eigenvalue-modulated signal,} in \emph{2022 Optical Fiber Communications Conference and Exhibition (OFC),}  (IEEE, 2022), pp. 01--03.

\bibitem{lpt14}
S.~Le, J.~E. Prilepsky, and S.~K. Turitsyn, \enquote{Nonlinear inverse synthesis for high spectral efficiency transmission in optical fibers,} {\protect\JournalTitle{Optics Express}} \textbf{22}, 26720--26741 (2014).

\bibitem{yangzhang2019dual}
X.~Yangzhang, V.~Aref, S.~T. Le, H.~Buelow, D.~Lavery, and P.~Bayvel, \enquote{Dual-polarization non-linear frequency-division multiplexed transmission with $ b $-modulation,} {\protect\JournalTitle{Journal of Lightwave Technology}} \textbf{37}, 1570--1578 (2019).

\bibitem{yangzhang2019experimental}
X.~Yangzhang, S.~T. Le, V.~Aref, H.~Buelow, D.~Lavery, and P.~Bayvel, \enquote{Experimental demonstration of dual-polarization nfdm transmission with $ b $-modulation,} {\protect\JournalTitle{IEEE Photonics Technology Letters}} \textbf{31}, 885--888 (2019).

\bibitem{gui2018nonlinear}
T.~Gui, G.~Zhou, C.~Lu, A.~P.~T. Lau, and S.~Wahls, \enquote{Nonlinear frequency division multiplexing with b-modulation: shifting the energy barrier,} {\protect\JournalTitle{Optics express}} \textbf{26}, 27978--27990 (2018).

\bibitem{derevyanko2021channel}
S.~Derevyanko, M.~Balogun, O.~Aluf, D.~Shepelsky, and J.~E. Prilepsky, \enquote{Channel model and the achievable information rates of the optical nonlinear frequency division-multiplexed systems employing continuous b-modulation,} {\protect\JournalTitle{Optics Express}} \textbf{29}, 6384--6406 (2021).

\bibitem{zhang2021correlation}
Q.~Zhang and F.~R. Kschischang, \enquote{Correlation-aided nonlinear spectrum detection,} {\protect\JournalTitle{Journal of Lightwave Technology}} \textbf{39}, 4923--4931 (2021).

\bibitem{balogun2022enhancing}
M.~Balogun and S.~Derevyanko, \enquote{Enhancing the spectral efficiency of nonlinear frequency division multiplexing systems via hermite-gaussian subcarriers,} {\protect\JournalTitle{Journal of Lightwave Technology}} \textbf{40}, 6071--6077 (2022).

\bibitem{zhang2021direct}
W.~Q. Zhang, T.~H. Chan, and S.~Afshar, \enquote{Direct decoding of nonlinear ofdm-qam signals using convolutional neural network,} {\protect\JournalTitle{Optics Express}} \textbf{29}, 11591--11604 (2021).

\bibitem{sedov2021neural}
E.~V. Sedov, P.~J. Freire, V.~V. Seredin, V.~A. Kolbasin, M.~Kamalian-Kopae, I.~S. Chekhovskoy, S.~K. Turitsyn, and J.~E. Prilepsky, \enquote{Neural networks for computing and denoising the continuous nonlinear fourier spectrum in focusing nonlinear schr{\"o}dinger equation,} {\protect\JournalTitle{Scientific Reports}} \textbf{11}, 22857 (2021).

\bibitem{sedov2021neural1}
E.~V. Sedov, I.~S. Chekhovskoy, and J.~E. Prilepsky, \enquote{Neural network for calculating direct and inverse nonlinear fourier transform,} {\protect\JournalTitle{Quantum Electronics}} \textbf{51}, 1118--1121 (2021).

\bibitem{zhang2022serial}
W.~Q. Zhang, T.~H. Chan, and S.~A. Vahid, \enquote{Serial and parallel convolutional neural network schemes for nfdm signals,} {\protect\JournalTitle{Scientific reports}} \textbf{12}, 1--12 (2022).

\bibitem{zhou2022nonlinear}
J.~Zhou, Q.~Hu, and H.~Pu, \enquote{Nonlinear fourier transform receiver based on a time domain diffractive deep neural network,} {\protect\JournalTitle{Optics Express}} \textbf{30}, 38576--38586 (2022).

\bibitem{chen2022}
X.~Chen, X.~Fang, F.~Yang, and F.~Zhang, \enquote{10.83 tb/s over 800 km nonlinear frequency division multiplexing wdm transmission,} {\protect\JournalTitle{Journal of Lightwave Technology}} \textbf{40}, 5385--5394 (2022).

\bibitem{gu2020machine}
R.~Gu, Z.~Yang, and Y.~Ji, \enquote{Machine learning for intelligent optical networks: A comprehensive survey,} {\protect\JournalTitle{Journal of Network and Computer Applications}} \textbf{157}, 102576 (2020).

\bibitem{wang2022review}
D.~Wang, C.~Zhang, W.~Chen, H.~Yang, M.~Zhang, and A.~P.~T. Lau, \enquote{A review of machine learning-based failure management in optical networks,} {\protect\JournalTitle{Science China Information Sciences}} \textbf{65}, 1--19 (2022).

\bibitem{troia2018deep}
S.~Troia, R.~Alvizu, Y.~Zhou, G.~Maier, and A.~Pattavina, \enquote{Deep learning-based traffic prediction for network optimization,} in \emph{2018 20th International Conference on Transparent Optical Networks (ICTON),}  (IEEE, 2018), pp. 1--4.

\bibitem{aloraifan2021deep}
D.~Aloraifan, I.~Ahmad, and E.~Alrashed, \enquote{Deep learning based network traffic matrix prediction,} {\protect\JournalTitle{International Journal of Intelligent Networks}} \textbf{2}, 46--56 (2021).

\bibitem{hatem2019deep}
J.~A. Hatem, A.~R. Dhaini, and S.~Elbassuoni, \enquote{Deep learning-based dynamic bandwidth allocation for future optical access networks,} {\protect\JournalTitle{IEEE Access}} \textbf{7}, 97307--97318 (2019).

\bibitem{zhu2020prediction}
X.~Zhu, O.~Xu, and G.~Li, \enquote{Prediction accuracy improvement of passive optical network traffic by a lstm model with a new activation function,} in \emph{2020 7th International Conference on Information, Cybernetics, and Computational Social Systems (ICCSS),}  (IEEE, 2020), pp. 662--666.

\bibitem{vaquero2021perturbation}
F.~J. Vaquero-Caballero, D.~J. Ives, and S.~J. Savory, \enquote{Perturbation-based frequency domain linear and nonlinear noise estimation,} {\protect\JournalTitle{Journal of Lightwave Technology}} \textbf{40}, 6055--6063 (2022).

\bibitem{wang2017convolutional}
D.~Wang, M.~Zhang, Z.~Li, J.~Li, C.~Song, J.~Li, and M.~Wang, \enquote{Convolutional neural network-based deep learning for intelligent osnr estimation on eye diagrams,} in \emph{2017 European Conference on Optical Communication (ECOC),}  (IEEE, 2017), pp. 1--3.

\bibitem{tanimura2019convolutional}
T.~Tanimura, T.~Hoshida, T.~Kato, S.~Watanabe, and H.~Morikawa, \enquote{Convolutional neural network-based optical performance monitoring for optical transport networks,} {\protect\JournalTitle{Journal of Optical Communications and Networking}} \textbf{11}, A52--A59 (2019).

\bibitem{tanimura2021concept}
T.~Tanimura, S.~Yoshida, K.~Tajima, S.~Oda, and T.~Hoshida, \enquote{Concept and implementation study of advanced dsp-based fiber-longitudinal optical power profile monitoring toward optical network tomography,} {\protect\JournalTitle{Journal of Optical Communications and Networking}} \textbf{13}, E132--E141 (2021).

\bibitem{wang2019comprehensive}
D.~Wang, Y.~Xu, J.~Li, M.~Zhang, J.~Li, J.~Qin, C.~Ju, Z.~Zhang, and X.~Chen, \enquote{Comprehensive eye diagram analysis: a transfer learning approach,} {\protect\JournalTitle{IEEE Photonics Journal}} \textbf{11}, 1--19 (2019).

\bibitem{lohani2019dispersion}
S.~Lohani, E.~M. Knutson, W.~Zhang, and R.~T. Glasser, \enquote{Dispersion characterization and pulse prediction with machine learning,} {\protect\JournalTitle{OSA Continuum}} \textbf{2}, 3438--3445 (2019).

\bibitem{du2021forecasting}
W.~Du, D.~C{\^o}t{\'e}, C.~Barber, and Y.~Liu, \enquote{Forecasting loss of signal in optical networks with machine learning,} {\protect\JournalTitle{Journal of Optical Communications and Networking}} \textbf{13}, E109--E121 (2021).

\bibitem{muller2021estimating}
J.~M{\"u}ller, T.~Fehenberger, S.~K. Patri, K.~Kaeval, H.~Griesser, M.~Tikas, and J.-P. Elbers, \enquote{Estimating quality of transmission in a live production network using machine learning,} in \emph{Optical Fiber Communication Conference,}  (Optica Publishing Group, 2021), pp. Tu1G--2.

\bibitem{d2020using}
A.~D’Amico, S.~Straullu, A.~Nespola, I.~Khan, E.~London, E.~Virgillito, S.~Piciaccia, A.~Tanzi, G.~Galimberti, and V.~Curri, \enquote{Using machine learning in an open optical line system controller,} {\protect\JournalTitle{Journal of Optical Communications and Networking}} \textbf{12}, C1--C11 (2020).

\bibitem{kashi2021neural}
A.~S. Kashi, J.~C. Cartledge, and W.-Y. Chan, \enquote{Neural network training framework for nonlinear signal-to-noise ratio estimation in heterogeneous optical networks,} in \emph{2021 Optical Fiber Communications Conference and Exhibition (OFC),}  (IEEE, 2021), pp. 1--3.

\bibitem{lonardi2020perks}
M.~Lonardi, J.~Pesic, T.~Zami, and N.~Rossi, \enquote{The perks of using machine learning for qot estimation with uncertain network parameters,} in \emph{Photonic Networks and Devices,}  (Optica Publishing Group, 2020), pp. NeM3B--2.

\bibitem{lv2021joint}
H.~Lv, X.~Zhou, J.~Huo, and J.~Yuan, \enquote{Joint osnr monitoring and modulation format identification on signal amplitude histograms using convolutional neural network,} {\protect\JournalTitle{Optical Fiber Technology}} \textbf{61}, 102455 (2021).

\bibitem{inuzuka2020demonstration}
F.~Inuzuka, T.~Oda, T.~Tanaka, K.~Kitamura, S.~Kuwabara, A.~Hirano, and M.~Tomizawa, \enquote{Demonstration of a novel framework for proactive maintenance using failure prediction and bit lossless protection with autonomous network diagnosis system,} {\protect\JournalTitle{Journal of Lightwave Technology}} \textbf{38}, 2695--2702 (2020).

\bibitem{zhang2020temporal}
C.~Zhang, D.~Wang, L.~Wang, J.~Song, S.~Liu, J.~Li, L.~Guan, Z.~Liu, and M.~Zhang, \enquote{Temporal data-driven failure prognostics using bigru for optical networks,} {\protect\JournalTitle{Journal of Optical Communications and Networking}} \textbf{12}, 277--287 (2020).

\bibitem{zhang2021attention}
C.~Zhang, D.~Wang, J.~Jia, L.~Wang, S.~Liu, L.~Guan, and M.~Zhang, \enquote{Attention mechanism-driven potential fault cause identification in optical networks,} in \emph{2021 Optical Fiber Communications Conference and Exhibition (OFC),}  (IEEE, 2021), pp. 1--3.

\bibitem{abdelli2020lifetime}
K.~Abdelli, D.~Rafique, H.~Grie{\ss}er, and S.~Pachnicke, \enquote{Lifetime prediction of 1550 nm dfb laser using machine learning techniques,} in \emph{Optical Fiber Communication Conference,}  (Optica Publishing Group, 2020), pp. Th2A--3.

\bibitem{liu2019application}
T.~Liu, H.~Mei, Q.~Sun, and H.~Zhou, \enquote{Application of neural network in fault location of optical transport network,} {\protect\JournalTitle{China Communications}} \textbf{16}, 214--225 (2019).

\bibitem{jia2021transformer}
J.~Jia, D.~Wang, C.~Zhang, H.~Yang, L.~Guan, X.~Chen, and M.~Zhang, \enquote{Transformer-based alarm context-vectorization representation for reliable alarm root cause identification in optical networks,} in \emph{2021 European Conference on Optical Communication (ECOC),}  (IEEE, 2021), pp. 1--4.

\bibitem{zhao2019accurate}
X.~Zhao, H.~Yang, H.~Guo, T.~Peng, and J.~Zhang, \enquote{Accurate fault location based on deep neural evolution network in optical networks for 5g and beyond,} in \emph{Optical Fiber Communication Conference,}  (Optical Society of America, 2019), pp. M3J--5.

\bibitem{lewis2007principal}
E.~Lewis, C.~Sheridan, M.~O’Farrell, D.~King, C.~Flanagan, W.~Lyons, and C.~Fitzpatrick, \enquote{Principal component analysis and artificial neural network based approach to analysing optical fibre sensors signals,} {\protect\JournalTitle{Sensors and Actuators A: Physical}} \textbf{136}, 28--38 (2007).

\bibitem{Sensor02}
S.~Kowarik, M.-T. Hussels, S.~Chruscicki, S.~Münzenberger, A.~Lämmerhirt, P.~Pohl, and M.~Schubert, \enquote{Fiber optic train monitoring with distributed acoustic sensing: Conventional and neural network data analysis,} {\protect\JournalTitle{Sensors}} \textbf{20} (2020).

\bibitem{Sensor03}
S.~Liehr, L.~A. J\"{a}ger, C.~Karapanagiotis, S.~M\"{u}nzenberger, and S.~Kowarik, \enquote{Real-time dynamic strain sensing in optical fibers using artificial neural networks,} {\protect\JournalTitle{Optics Express}} \textbf{27}, 7405--7425 (2019).

\bibitem{Sensor04}
S.~Liehr, \enquote{Artificial neural networks for distributed optical fiber sensing (invited),} in \emph{Optical Fiber Communication Conference (OFC) 2021,}  (Optica Publishing Group, 2021), p. Th4F.2.

\bibitem{suah2003applications}
F.~B.~M. Suah, M.~Ahmad, and M.~N. Taib, \enquote{Applications of artificial neural network on signal processing of optical fibre ph sensor based on bromophenol blue doped with sol--gel film,} {\protect\JournalTitle{Sensors and Actuators B: Chemical}} \textbf{90}, 182--188 (2003).

\bibitem{li2019deep}
X.~Li, J.~Shu, W.~Gu, and L.~Gao, \enquote{Deep neural network for plasmonic sensor modeling,} {\protect\JournalTitle{Optical Materials Express}} \textbf{9}, 3857--3862 (2019).

\bibitem{shokrekhodaei2021non}
M.~Shokrekhodaei, D.~P. Cistola, R.~C. Roberts, and S.~Quinones, \enquote{Non-invasive glucose monitoring using optical sensor and machine learning techniques for diabetes applications,} {\protect\JournalTitle{IEEE Access}} \textbf{9}, 73029--73045 (2021).

\bibitem{Sensor06}
F.~B.~M. Suah, M.~Ahmad, and M.~N. Taib, \enquote{Optimisation of the range of an optical fibre ph sensor using feed-forward artificial neural network,} {\protect\JournalTitle{Sensors and Actuators B: Chemical}} \textbf{90}, 175--181 (2003).

\bibitem{NNSensor09}
I.~Dias, R.~Oliveira, and O.~Fraz{\~a}o, \enquote{Intelligent optical sensors using artificial neural network approach,} in \emph{Innovation in Manufacturing Networks,}  A.~Azevedo, ed. (Springer US, Boston, MA, 2008), pp. 289--294.

\bibitem{NNSensor10}
L.~Zhao, J.~Wang, and X.~Chen, \enquote{Bp neural network with regularization and sensor array for prediction of component concentration of mixed gas,} in \emph{Advances in Neural Networks -- ISNN 2018,}  T.~Huang, J.~Lv, C.~Sun, and A.~V. Tuzikov, eds. (Springer International Publishing, Cham, 2018), pp. 541--548.

\bibitem{manie2020using}
Y.~C. Manie, J.-W. Li, P.-C. Peng, R.-K. Shiu, Y.-Y. Chen, and Y.-T. Hsu, \enquote{Using a machine learning algorithm integrated with data de-noising techniques to optimize the multipoint sensor network,} {\protect\JournalTitle{Sensors}} \textbf{20}, 1070 (2020).

\bibitem{shi2019event}
Y.~Shi, Y.~Wang, L.~Zhao, and Z.~Fan, \enquote{An event recognition method for $\phi$-otdr sensing system based on deep learning,} {\protect\JournalTitle{Sensors}} \textbf{19}, 3421 (2019).

\bibitem{NNUC01}
L.~Salmela, N.~Tsipinakis, A.~Foi, C.~Billet, J.~Dudley, and G.~Genty, \enquote{Predicting ultrafast nonlinear dynamics in fibre optics with a recurrent neural network,} {\protect\JournalTitle{Nature Machine Intelligence}} \textbf{3}, 344–354 (2021).

\bibitem{NNUC03}
L.~Salmela, C.~Lapre, J.~M. Dudley, and G.~Genty, \enquote{Machine learning analysis of rogue solitons in supercontinuum generation,} {\protect\JournalTitle{Scientific reports}} \textbf{10}, 1--8 (2020).

\bibitem{NNUC04}
A.~Ermolaev, A.~Sheveleva, G.~Genty, C.~Finot, and J.~Dudley, \enquote{Data-driven model discovery of ideal four-wave mixing in nonlinear fibre optics,} {\protect\JournalTitle{Scientific Reports}} \textbf{12}, 12711 (2022).

\bibitem{NNUC05}
X.~Jiang, D.~Wang, Q.~Fan, M.~Zhang, C.~Lu, and A.~P.~T. Lau, \enquote{Physics-informed neural network for nonlinear dynamics in fiber optics,} {\protect\JournalTitle{Laser \& Photonics Reviews}} \textbf{16}, 2100483 (2022).

\bibitem{NNUC06}
M.~Soltani, F.~Da~Ros, A.~Carena, and D.~Zibar, \enquote{Spectral and spatial power evolution design with machine learning-enabled raman amplification,} {\protect\JournalTitle{Journal of Lightwave Technology}}  (2022).

\bibitem{NNUC07}
T.~Zahavy, A.~Dikopoltsev, D.~Moss, G.~I. Haham, O.~Cohen, S.~Mannor, and M.~Segev, \enquote{Deep learning reconstruction of ultrashort pulses,} {\protect\JournalTitle{Optica}} \textbf{5}, 666--673 (2018).

\bibitem{NNUC08}
M.~Stanfield, J.~Ott, C.~Gardner, N.~F. Beier, D.~M. Farinella, C.~A. Mancuso, P.~Baldi, and F.~Dollar, \enquote{Real-time reconstruction of high energy, ultrafast laser pulses using deep learning,} {\protect\JournalTitle{Scientific reports}} \textbf{12(1)}, 5299 (2022).

\bibitem{NNIns01}
M.~Mabed, F.~Meng, L.~Salmela, C.~Finot, G.~Genty, and J.~M. Dudley, \enquote{Machine learning analysis of instabilities in noise-like pulse lasers,} {\protect\JournalTitle{Optics Express}} \textbf{30}, 15060--15072 (2022).

\bibitem{NNins02}
L.~Salmela, M.~Hary, M.~Mabed, A.~Foi, J.~M. Dudley, and G.~Genty, \enquote{Feed-forward neural network as nonlinear dynamics integrator for supercontinuum generation: erratum,} {\protect\JournalTitle{Optics Letters}} \textbf{47}, 1741--1741 (2022).

\bibitem{NNTimestretch01}
G.~Pu and B.~Jalali, \enquote{Neural network enabled time stretch spectral regression,} {\protect\JournalTitle{Optics Express}} \textbf{29}, 20786--20794 (2021).

\bibitem{siegman86}
A.~E. Siegman, \emph{Lasers} (University Science Books, 1986).

\bibitem{fu2018several}
W.~Fu, L.~G. Wright, P.~Sidorenko, S.~Backus, and F.~W. Wise, \enquote{Several new directions for ultrafast fiber lasers,} {\protect\JournalTitle{Optics Express}} \textbf{26}, 9432--9463 (2018). Doi: 10.1364/OE.26.009432.

\bibitem{Laser}
O.~G. Okhotnikov, ed., \emph{Fiber Lasers} (John Wiley \& Sons, Ltd, 2012).

\bibitem{Laser01}
S.~K. Turitsyn, B.~G. Bale, and M.~P. Fedoruk, \enquote{Dispersion-managed solitons in fiber systems and lasers,} {\protect\JournalTitle{Physics Reports}} \textbf{521}, 135--203 (2012).

\bibitem{Laser02}
S.~K. Turitsyn, S.~A. Babin, D.~V. Churkin, I.~D. Vatnik, M.~Nikulin, and E.~V. Podivilov, \enquote{Random distributed feedback fibre lasers,} {\protect\JournalTitle{Physics Reports}} \textbf{542}, 133--193 (2014).

\bibitem{andral2016toward}
U.~Andral, J.~Buguet, R.~S. Fodil, F.~Amrani, F.~Billard, E.~Hertz, and P.~Grelu, \enquote{Toward an autosetting mode-locked fiber laser cavity,} {\protect\JournalTitle{Journal of Optical Society of America B}} \textbf{33}, 825--833 (2016).

\bibitem{Laser05}
J.~N. Kutz, \enquote{Mode‐locked soliton lasers,} {\protect\JournalTitle{SIAM Review}} \textbf{48}, 629--678 (2006).

\bibitem{NNPDE01}
J.~N. Kutz, \enquote{Deep learning in fluid dynamics,} {\protect\JournalTitle{Journal of Fluid Mechanics}} \textbf{814}, 1–4 (2017).

\bibitem{NNPDE02}
M.~Raissi, \enquote{Deep hidden physics models: Deep learning of nonlinear partial differential equations,} {\protect\JournalTitle{J. Mach. Learn. Res.}} \textbf{19}, 932–955 (2018).

\bibitem{brunton2014self}
S.~L. Brunton, X.~Fu, and J.~N. Kutz, \enquote{Self-tuning fiber lasers,} {\protect\JournalTitle{IEEE Journal of Selected Topics in Quantum Electronics}} \textbf{20}, 464--471 (2014).

\bibitem{kutz2015intelligent}
J.~N. Kutz and S.~L. Brunton, \enquote{Intelligent systems for stabilizing mode-locked lasers and frequency combs: machine learning and equation-free control paradigms for self-tuning optics,} {\protect\JournalTitle{Nanophotonics}} \textbf{4}, 459--471 (2015).

\bibitem{woodward2016towards}
R.~Woodward and E.~J. Kelleher, \enquote{Towards a smart lasers: self-optimisation of an ultrafast pulse source using a genetic algorithm,} {\protect\JournalTitle{Scientific Reports}} \textbf{6}, 37616 (2016).

\bibitem{winters2017NPE}
D.~G. Winters, M.~S. Kirchner, S.~J. Backus, and H.~C. Kapteyn, \enquote{Electronic initiation and optimization of nonlinear polarization evolution mode-locking in a fiber laser,} {\protect\JournalTitle{Optics Express}} \textbf{25}, 33216--33225 (2017).

\bibitem{woodward2017genetic}
R.~Woodward and E.~Kelleher, \enquote{Genetic algorithm-based control of birefringent filtering for self-tuning, self-pulsing fiber lasers,} {\protect\JournalTitle{Optics Letters}} \textbf{42}, 2952--2955 (2017).

\bibitem{NNLaser09}
X.~Ma, J.~Lin, C.~Dai, J.~Lv, P.~Yao, L.~Xu, and C.~Gu, \enquote{Machine learning method for calculating mode-locking performance of linear cavity fiber lasers,} {\protect\JournalTitle{Optics and Laser Technology}} \textbf{149}, 107883 (2022).

\bibitem{Sun_2020}
C.~Sun, E.~Kaiser, S.~L. Brunton, and J.~N. Kutz, \enquote{Deep reinforcement learning for optical systems: A case study of mode-locked lasers,} {\protect\JournalTitle{Machine Learning: Science and Technology}} \textbf{1}, 045013 (2020).

\bibitem{NNLaser05}
E.~Kuprikov, A.~Kokhanovskiy, K.~Serebrennikov, and S.~Turitsyn, \enquote{Deep reinforcement learning for self-tuning laser source of dissipative solitons,} {\protect\JournalTitle{Scientific Reports}} \textbf{12}, 7185 (2022).

\bibitem{Li:22}
Z.~Li, S.~Yang, Q.~Xiao, T.~Zhang, Y.~Li, L.~Han, D.~Liu, X.~Ouyang, and J.~Zhu, \enquote{Deep reinforcement with spectrum series learning control for a mode-locked fiber laser,} {\protect\JournalTitle{Photon. Res.}} \textbf{10}, 1491--1500 (2022).

\bibitem{NNLaser06}
A.~Kokhanovskiy, A.~Shevelev, K.~Serebrennikov, E.~Kuprikov, and S.~Turitsyn, \enquote{A deep reinforcement learning algorithm for smart control of hysteresis phenomena in a mode-locked fiber laser,} {\protect\JournalTitle{Photonics}} \textbf{9} (2022).

\bibitem{MD1}
A.~Liu, T.~Lin, H.~Han, X.~Zhang, Z.~Chen, F.~Gan, H.~Lv, and X.~Liu, \enquote{Analyzing modal power in multi-mode waveguide via machine learning,} {\protect\JournalTitle{Optics Express}} \textbf{26}, 22100--22109 (2018).

\bibitem{MD2}
Y.~An, L.~Huang, J.~Li, J.~Leng, L.~Yang, and P.~Zhou, \enquote{Deep learning-based real-time mode decomposition for multimode fibers,} {\protect\JournalTitle{IEEE Journal of Selected Topics in Quantum Electronics}} \textbf{26}, 1--6 (2020).

\bibitem{MD3}
Y.~An, L.~Huang, J.~Li, J.~Leng, L.~Yang, and P.~Zhou, \enquote{Learning to decompose the modes in few-mode fibers with deep convolutional neural network,} {\protect\JournalTitle{Optics Express}} \textbf{27}, 10127--10137 (2019).

\bibitem{MD4}
E.~Manuylovich, A.~Donodin, and S.~Turitsyn, \enquote{Intensity-only-measurement mode decomposition in few-mode fibers,} {\protect\JournalTitle{Optics Express}} \textbf{29}, 36769--36783 (2021).

\bibitem{MMLaser01}
L.~G. Wright, W.~H. Renninger, D.~N. Christodoulides, and F.~W. Wise, \enquote{Nonlinear multimode photonics: nonlinear optics with many degrees of freedom,} {\protect\JournalTitle{Optica}} \textbf{9}, 824--841 (2022).

\bibitem{MMLaser02}
L.~Wright, P.~Sidorenko, H.~Pourbeyram, Z.~H. Ziegler, A.~Isichenko, B.~A. Malomed, C.~R. Menyuk, D.~N. Christodoulides, and F.~W. Wise, \enquote{Mechanisms of spatiotemporal mode-locking,} {\protect\JournalTitle{Nature Physics}} \textbf{6}, 565–570 (2020).

\bibitem{MMLaser03}
H.~Haig, P.~Sidorenko, A.~Dhar, N.~Choudhury, R.~Sen, D.~Christodoulides, and F.~Wise, \enquote{Multimode mamyshev oscillator,} {\protect\JournalTitle{Optics Letters}} \textbf{47}, 46--49 (2022).

\bibitem{Brunton2013ExtremumSeekingCO}
S.~L. Brunton, X.~Fu, and J.~N. Kutz, \enquote{Extremum-seeking control of a mode-locked laser,} {\protect\JournalTitle{IEEE Journal of Quantum Electronics}} \textbf{49}, 852--861 (2013).

\bibitem{pu2019intelligent}
G.~Pu, L.~Yi, L.~Zhang, and W.~Hu, \enquote{Intelligent programmable mode-locked fiber laser with a human-like algorithm,} {\protect\JournalTitle{Optica}} \textbf{6}, 362--369 (2019). Doi: 10.1038/421805a.

\bibitem{NNLaser01}
A.~Kokhanovskiy, A.~Ivanenko, S.~Kobtsev, S.~Smirnov, and S.~Turitsyn, \enquote{Machine learning methods for control of fibre lasers with double gain nonlinear loop mirror,} {\protect\JournalTitle{Scientific Reports}} \textbf{9}, 2916 (2019).

\bibitem{NNLaser03}
A.~Kokhanovskiy, E.~Kuprikov, A.~Bednyakova, I.~Popkov, S.~Smirnov, and S.~Turitsyn, \enquote{Inverse design of mode-locked fiber laser by particle swarm optimization algorithm,} {\protect\JournalTitle{Scientific Reports}} \textbf{11(1)}, 1--9 (2021).

\bibitem{Image1981}
B.~K. Horn and B.~G. Schunck, \enquote{Determining optical flow,} {\protect\JournalTitle{Artificial Intelligence}} \textbf{17}, 185--203 (1981).

\bibitem{Image}
J.~A. Marshall, \enquote{Self-organizing neural networks for perception of visual motion,} {\protect\JournalTitle{Neural Networks}} \textbf{3}, 45--74 (1990).

\bibitem{Image0}
M.~Egmont-Petersen, D.~{de Ridder}, and H.~Handels, \enquote{Image processing with neural networks—a review,} {\protect\JournalTitle{Pattern Recognition}} \textbf{35}, 2279--2301 (2002).

\bibitem{Image01}
C.~Zuo, J.~Qian, S.~Feng, W.~Yin, Y.~Li, P.~Fan, J.~Han, K.~Qian, and Q.~Chen, \enquote{Deep learning in optical metrology: a review,} {\protect\JournalTitle{Light: Science and Applications}} \textbf{11}, 39 (2022).

\bibitem{Image02}
H.~Greenspan, B.~van Ginneken, and R.~M. Summers, \enquote{Guest editorial deep learning in medical imaging: Overview and future promise of an exciting new technique,} {\protect\JournalTitle{IEEE Transactions on Medical Imaging}} \textbf{35}, 1153--1159 (2016).

\bibitem{Image03}
M.~T. McCann, K.~H. Jin, and M.~Unser, \enquote{Convolutional neural networks for inverse problems in imaging: A review,} {\protect\JournalTitle{IEEE Signal Processing Magazine}} \textbf{34}, 85--95 (2017).

\bibitem{Image04}
U.~S. Kamilov, I.~N. Papadopoulos, M.~H. Shoreh, A.~Goy, C.~Vonesch, M.~Unser, and D.~Psaltis, \enquote{Learning approach to optical tomography,} {\protect\JournalTitle{Optica}} \textbf{2}, 517--522 (2015).

\bibitem{Image05}
Z.~Zhang, J.~Liu, D.~Yang, U.~S. Kamilov, and G.~D. Hugo, \enquote{Deep learning-based motion compensation for four-dimensional cone-beam computed tomography (4d-cbct) reconstruction,} {\protect\JournalTitle{Medical Physics}} \textbf{n/a} (2022).

\bibitem{Image06}
Z.~Wu, Y.~Sun, A.~Matlock, J.~Liu, L.~Tian, and U.~S. Kamilov, \enquote{Simba: Scalable inversion in optical tomography using deep denoising priors,} {\protect\JournalTitle{IEEE Journal of Selected Topics in Signal Processing}} \textbf{14}, 1163--1175 (2020).

\bibitem{yan2019fringe}
K.~Yan, Y.~Yu, C.~Huang, L.~Sui, K.~Qian, and A.~Asundi, \enquote{Fringe pattern denoising based on deep learning,} {\protect\JournalTitle{Optics Communications}} \textbf{437}, 148--152 (2019).

\bibitem{lin2020optical}
B.~Lin, S.~Fu, C.~Zhang, F.~Wang, and Y.~Li, \enquote{Optical fringe patterns filtering based on multi-stage convolution neural network,} {\protect\JournalTitle{Optics and Lasers in Engineering}} \textbf{126}, 105853 (2020).

\bibitem{shi2019label}
J.~Shi, X.~Zhu, H.~Wang, L.~Song, and Q.~Guo, \enquote{Label enhanced and patch based deep learning for phase retrieval from single frame fringe pattern in fringe projection 3d measurement,} {\protect\JournalTitle{Optics Express}} \textbf{27}, 28929--28943 (2019).

\bibitem{yu2020deep}
H.~Yu, D.~Zheng, J.~Fu, Y.~Zhang, C.~Zuo, and J.~Han, \enquote{Deep learning-based fringe modulation-enhancing method for accurate fringe projection profilometry,} {\protect\JournalTitle{Optics Express}} \textbf{28}, 21692--21703 (2020).

\bibitem{zhang2010snapshot}
Z.~Zhang, D.~P. Towers, and C.~E. Towers, \enquote{Snapshot color fringe projection for absolute three-dimensional metrology of video sequences,} {\protect\JournalTitle{Applied optics}} \textbf{49}, 5947--5953 (2010).

\bibitem{qian2020single}
J.~Qian, S.~Feng, Y.~Li, T.~Tao, J.~Han, Q.~Chen, and C.~Zuo, \enquote{Single-shot absolute 3d shape measurement with deep-learning-based color fringe projection profilometry,} {\protect\JournalTitle{Optics Letters}} \textbf{45}, 1842--1845 (2020).

\bibitem{Imagebook01}
L.~Lu, Y.~Zheng, G.~Carneiro, and L.~Yang, eds., \emph{Deep Learning and Convolutional Neural Networks for Medical Image Computing}, Advances in Computer Vision and Pattern Recognition (Springer, Cham, 2017).

\bibitem{Imagebook02}
G.~Wang, Y.~Zhang, X.~Ye, and X.~Mou, \emph{Machine Learning for Tomographic Imaging}, 2053-2563 (IOP Publishing, 2019).

\bibitem{Imagebook03}
D.~Hemanth and V.~Estrela, \emph{Deep Learning for Image Processing Applications}, Advances in Parallel Computing (IOS Press, 2017).

\bibitem{Sensor01}
L.~Ma, Y.~Liu, X.~Zhang, Y.~Ye, G.~Yin, and B.~A. Johnson, \enquote{Deep learning in remote sensing applications: A meta-analysis and review,} {\protect\JournalTitle{ISPRS Journal of Photogrammetry and Remote Sensing}} \textbf{152}, 166--177 (2019).

\bibitem{Sensor01a}
B.~Huang, B.~Zhao, and Y.~Song, \enquote{Urban land-use mapping using a deep convolutional neural network with high spatial resolution multispectral remote sensing imagery,} {\protect\JournalTitle{Remote Sensing of Environment}} \textbf{214}, 73--86 (2018).

\bibitem{Caramazza2019}
P.~Caramazza, O.~Moran, R.~Murray-Smith, and D.~Faccio, \enquote{Transmission of natural scene images through a multimode fibre,} {\protect\JournalTitle{Nature Communications}} \textbf{10}, 2029 (2019).

\bibitem{NNImage01}
S.~Aisawa, K.~Noguchi, and T.~Matsumoto, \enquote{Remote image classification through multimode optical fiber using a neural network,} {\protect\JournalTitle{Optics Letters}} \textbf{16}, 645--647 (1991).

\bibitem{NNImage02}
N.~Shabairou, E.~Cohen, O.~Wagner, D.~Malka, and Z.~Zalevsky, \enquote{Color image identification and reconstruction using artificial neural networks on multimode fiber images: towards an all-optical design,} {\protect\JournalTitle{Optics Letters}} \textbf{43}, 5603--5606 (2018).

\bibitem{NNImage03}
B.~Rahmani, I.~Oguz, U.~Tegin, J.~liang Hsieh, D.~Psaltis, and C.~Moser, \enquote{Learning to image and compute with multimode optical fibers,} {\protect\JournalTitle{Nanophotonics}} \textbf{11}, 1071--1082 (2022).

\bibitem{Liu2022}
Z.~Liu, L.~Wang, Y.~Meng, T.~He, S.~He, Y.~Yang, L.~Wang, J.~Tian, D.~Li, P.~Yan, M.~Gong, Q.~Liu, and Q.~Xiao, \enquote{All-fiber high-speed image detection enabled by deep learning,} {\protect\JournalTitle{Nature Communications}} \textbf{13}, 1433 (2022).

\bibitem{Voumard:20}
T.~Voumard, T.~Wildi, V.~Brasch, R.~G. Alvarez, G.~V. Ogando, and T.~Herr, \enquote{Dual-frequency comb hyperspectral imaging by massively parallelized infrared detection and machine learning,} in \emph{Optical Sensors and Sensing Congress,}  (Optica Publishing Group, 2020), p. EM1C.1.

\bibitem{Lentile2006}
L.~B. Lentile, Z.~A. Holden, A.~M.~S. Smith, M.~J. Falkowski, A.~T. Hudak, P.~Morgan, S.~A. Lewis, P.~E. Gessler, and N.~C. Benson, \enquote{Remote sensing techniques to assess active fire characteristics and post-fire effects,} {\protect\JournalTitle{International Journal of Wildland Fire}} \textbf{15}, 319--345 (2006).

\bibitem{DALDEGAN2019111340}
G.~A. Daldegan, D.~A. Roberts, and F.~de~Figueiredo~Ribeiro, \enquote{Spectral mixture analysis in google earth engine to model and delineate fire scars over a large extent and a long time-series in a rainforest-savanna transition zone,} {\protect\JournalTitle{Remote Sensing of Environment}} \textbf{232}, 111340 (2019).

\bibitem{reviewCNNvegetationRemoteSensing}
T.~Kattenborn, J.~Leitloff, F.~Schiefer, and S.~Hinz, \enquote{Review on convolutional neural networks (cnn) in vegetation remote sensing,} {\protect\JournalTitle{ISPRS Journal of Photogrammetry and Remote Sensing}} \textbf{173}, 24--49 (2021).

\bibitem{Reichstein2019}
M.~Reichstein, G.~Camps-Valls, B.~Stevens, M.~Jung, J.~Denzler, N.~Carvalhais, and {Prabhat}, \enquote{Deep learning and process understanding for data-driven earth system science,} {\protect\JournalTitle{Nature}} \textbf{566}, 195--204 (2019).

\bibitem{Material}
W.~Chew, E.~Michielssen, J.~M. Song, and J.~M. Jin, \emph{Fast and Efficient Algorithms in Computational Electromagnetics} (Artech House, Inc., USA, 2001).

\bibitem{Material01}
M.~N.~O. Sadiku, \emph{{Numerical Techniques in Electromagnetics}} (CRC press, 1992).

\bibitem{Material02}
B.~Gallinet, J.~Butet, and O.~J.~F. Martin, \enquote{Numerical methods for nanophotonics: standard problems and future challenges,} {\protect\JournalTitle{Laser \& Photonics Reviews}} \textbf{9}, 577--603 (2015).

\bibitem{Material03}
S.~Molesky, Z.~Lin, A.~Y. Piggott, W.~Jin, J.~Vuckovi{\'c}, and A.~W. Rodriguez, \enquote{Inverse design in nanophotonics,} {\protect\JournalTitle{Nature Photonics}} \textbf{12}, 659--670 (2018).

\bibitem{Material04}
S.~D. Campbell, D.~Sell, R.~P. Jenkins, E.~B. Whiting, J.~A. Fan, and D.~H. Werner, \enquote{Review of numerical optimization techniques for meta-device design,} {\protect\JournalTitle{Optical Materials Express}} \textbf{9}, 1842--1863 (2019).

\bibitem{NNMaterial02}
M.~Zandehshahvar, M.~H. Javani, M.~Chen, T.~Brown, Y.~Kiarashi, and A.~Adibi, \enquote{Machine learning for efficient inverse design of nanophotonics structures,} in \emph{Photonic and Phononic Properties of Engineered Nanostructures XII,}  (SPIE, 2022), p. PC120100W.

\bibitem{NNMaterial04}
R.~S. Hegde, \enquote{Deep learning: a new tool for photonic nanostructure design,} {\protect\JournalTitle{Nanoscale Advances}} \textbf{2}, 1007--1023 (2020).

\bibitem{NNMaterial05}
W.~Ma, F.~Cheng, and Y.~Liu, \enquote{Deep-learning-enabled on-demand design of chiral metamaterials,} {\protect\JournalTitle{ACS nano}} \textbf{12}, 6326--6334 (2018).

\bibitem{InverseDesignML2021}
Z.~Liu, D.~Zhu, L.~Raju, and W.~Cai, \enquote{Tackling photonic inverse design with machine learning,} {\protect\JournalTitle{Advanced Science}} \textbf{8}, 2002923 (2021).

\bibitem{MaxwellNet2022}
J.~Lim and D.~Psaltis, \enquote{Maxwellnet: Physics-driven deep neural network training based on maxwell’s equations,} {\protect\JournalTitle{APL Photonics}} \textbf{7}, 011301 (2022).

\bibitem{Wiecha:21}
P.~R. Wiecha, A.~Arbouet, C.~Girard, and O.~L. Muskens, \enquote{Deep learning in nano-photonics: inverse design and beyond,} {\protect\JournalTitle{Photon. Res.}} \textbf{9}, B182--B200 (2021).

\bibitem{Tahersima2019}
M.~H. Tahersima, K.~Kojima, T.~Koike-Akino, D.~Jha, B.~Wang, C.~Lin, and K.~Parsons, \enquote{Deep neural network inverse design of integrated photonic power splitters,} {\protect\JournalTitle{Scientific Reports}} \textbf{9}, 1368 (2019).

\bibitem{ma15051811}
J.~Wang, Y.~Wang, and Y.~Chen, \enquote{Inverse design of materials by machine learning,} {\protect\JournalTitle{Materials}} \textbf{15} (2022).

\bibitem{jiang2019global}
J.~Jiang and J.~A. Fan, \enquote{Global optimization of dielectric metasurfaces using a physics-driven neural network,} {\protect\JournalTitle{Nano letters}} \textbf{19}, 5366--5372 (2019).

\bibitem{MLframeworkQuantumSampling2021}
B.~A. Wilson, Z.~A. Kudyshev, A.~V. Kildishev, S.~Kais, V.~M. Shalaev, and A.~Boltasseva, \enquote{Machine learning framework for quantum sampling of highly constrained, continuous optimization problems,} {\protect\JournalTitle{Applied Physics Reviews}} \textbf{8}, 041418 (2021).

\bibitem{maass1994computational}
W.~Maass, \enquote{On the computational complexity of networks of spiking neurons,} {\protect\JournalTitle{Advances in neural information processing systems}} \textbf{7} (1994).

\bibitem{alizadeh2020managing}
R.~Alizadeh, J.~K. Allen, and F.~Mistree, \enquote{Managing computational complexity using surrogate models: a critical review,} {\protect\JournalTitle{Research in Engineering Design}} \textbf{31}, 275--298 (2020).

\bibitem{wiedemann2019compact}
S.~Wiedemann, K.-R. M{\"u}ller, and W.~Samek, \enquote{Compact and computationally efficient representation of deep neural networks,} {\protect\JournalTitle{IEEE transactions on neural networks and learning systems}} \textbf{31}, 772--785 (2019).

\bibitem{amin2008analysis}
A.~M. Amin, R.~R. Mahmood, and A.~I. Khan, \enquote{Analysis of pattern recognition algorithms using associative memory approach: a comparative study between the hopfield network and distributed hierarchical graph neuron (dhgn),} in \emph{2008 IEEE 8th International Conference on Computer and Information Technology Workshops,}  (IEEE, 2008), pp. 153--158.

\bibitem{kerr2005big}
S.~N. Kerr, \enquote{A big-o experiment: which function is it?} in \emph{Proceedings of the 43rd annual Southeast regional conference-Volume 1,}  (2005), pp. 317--318.

\bibitem{blondel2000survey}
V.~D. Blondel and J.~N. Tsitsiklis, \enquote{A survey of computational complexity results in systems and control,} {\protect\JournalTitle{Automatica}} \textbf{36}, 1249--1274 (2000).

\bibitem{gysel2016hardware}
P.~Gysel, M.~Motamedi, and S.~Ghiasi, \enquote{Hardware-oriented approximation of convolutional neural networks,} {\protect\JournalTitle{arXiv preprint arXiv:1604.03168}}  (2016).

\bibitem{sze2017efficient}
V.~Sze, Y.-H. Chen, T.-J. Yang, and J.~S. Emer, \enquote{Efficient processing of deep neural networks: A tutorial and survey,} {\protect\JournalTitle{Proceedings of the IEEE}} \textbf{105}, 2295--2329 (2017).

\bibitem{li2017reducing}
B.~Li and T.~N. Sainath, \enquote{Reducing the computational complexity of two-dimensional lstms.} in \emph{INTERSPEECH,}  (2017), pp. 964--968.

\bibitem{yang2017designing}
T.-J. Yang, Y.-H. Chen, and V.~Sze, \enquote{Designing energy-efficient convolutional neural networks using energy-aware pruning,} in \emph{Proceedings of the IEEE conference on computer vision and pattern recognition,}  (2017), pp. 5687--5695.

\bibitem{balcazar1997computational}
J.~L. Balc{\'a}zar, R.~Gavalda, and H.~T. Siegelmann, \enquote{Computational power of neural networks: A characterization in terms of kolmogorov complexity,} {\protect\JournalTitle{IEEE Transactions on Information Theory}} \textbf{43}, 1175--1183 (1997).

\bibitem{van2020bayesian}
M.~Van~Baalen, C.~Louizos, M.~Nagel, R.~A. Amjad, Y.~Wang, T.~Blankevoort, and M.~Welling, \enquote{Bayesian bits: Unifying quantization and pruning,} {\protect\JournalTitle{Advances in neural information processing systems}} \textbf{33}, 5741--5752 (2020).

\bibitem{baskin2021uniq}
C.~Baskin, et~al., \enquote{Uniq: Uniform noise injection for non-uniform quantization of neural networks,} {\protect\JournalTitle{ACM Transactions on Computer Systems (TOCS)}} \textbf{37}, 1--15 (2021).

\bibitem{sahin2006neural}
S.~Sahin, Y.~Becerikli, and S.~Yazici, \enquote{Neural network implementation in hardware using fpgas,} in \emph{International conference on neural information processing,}  (Springer, 2006), pp. 1105--1112.

\bibitem{5280233}
A.~Dinu, M.~N. Cirstea, and S.~E. Cirstea, \enquote{Direct neural-network hardware-implementation algorithm,} {\protect\JournalTitle{IEEE Transactions on Industrial Electronics}} \textbf{57}, 1845--1848 (2010).

\bibitem{jacobsen2007fast}
E.~Jacobsen and P.~Kootsookos, \enquote{Fast, accurate frequency estimators [dsp tips \& tricks],} {\protect\JournalTitle{IEEE Signal Processing Magazine}} \textbf{24}, 123--125 (2007).

\bibitem{spinnler2010equalizer}
B.~Spinnler, \enquote{Equalizer design and complexity for digital coherent receivers,} {\protect\JournalTitle{IEEE Journal of Selected Topics in Quantum Electronics}} \textbf{16}, 1180--1192 (2010).

\bibitem{mirzaei2006fpga}
S.~Mirzaei, A.~Hosangadi, and R.~Kastner, \enquote{Fpga implementation of high speed fir filters using add and shift method,} in \emph{2006 International Conference on Computer Design,}  (IEEE, 2006), pp. 308--313.

\bibitem{jahani2009zot}
S.~Jahani, \enquote{Zot-mk: a new algorithm for big integer multiplication,} {\protect\JournalTitle{MSc MSc, Department of Computer Science, Universiti Sains Malaysia, Penang}}  (2009).

\bibitem{hawks2021ps}
B.~Hawks, J.~Duarte, N.~J. Fraser, A.~Pappalardo, N.~Tran, and Y.~Umuroglu, \enquote{Ps and qs: Quantization-aware pruning for efficient low latency neural network inference,} {\protect\JournalTitle{arXiv preprint arXiv:2102.11289}}  (2021).

\bibitem{wu2018deep}
J.~Wu, Y.~Wang, Z.~Wu, Z.~Wang, A.~Veeraraghavan, and Y.~Lin, \enquote{Deep k-means: Re-training and parameter sharing with harder cluster assignments for compressing deep convolutions,} in \emph{International Conference on Machine Learning,}  (PMLR, 2018), pp. 5363--5372.

\bibitem{li2019additive}
Y.~Li, X.~Dong, and W.~Wang, \enquote{Additive powers-of-two quantization: An efficient non-uniform discretization for neural networks,} {\protect\JournalTitle{arXiv preprint arXiv:1909.13144}}  (2019).

\bibitem{Koike2021}
T.~Koike-Akino, Y.~Wang, K.~Kojima, K.~Parsons, and T.~Yoshida, \enquote{Zero-multiplier sparse dnn equalization for fiber-optic qam systems with probabilistic amplitude shaping,} in \emph{2021 European Conference on Optical Communications (ECOC),}  (IEEE, 2021), pp. 1--4.

\bibitem{elhoushi2021deepshift}
M.~Elhoushi, Z.~Chen, F.~Shafiq, Y.~H. Tian, and J.~Y. Li, \enquote{Deepshift: Towards multiplication-less neural networks,} in \emph{Proceedings of the IEEE/CVF Conference on Computer Vision and Pattern Recognition,}  (2021), pp. 2359--2368.

\bibitem{you2020shiftaddnet}
H.~You, X.~Chen, Y.~Zhang, C.~Li, S.~Li, Z.~Liu, Z.~Wang, and Y.~Lin, \enquote{Shiftaddnet: A hardware-inspired deep network,} {\protect\JournalTitle{arXiv preprint arXiv:2010.12785}}  (2020).

\bibitem{gentili1995efficient}
P.~Gentili, F.~Piazza, and A.~Uncini, \enquote{Efficient genetic algorithm design for power-of-two fir filters,} in \emph{1995 International conference on acoustics, speech, and signal processing,}  vol.~2 (IEEE, 1995), pp. 1268--1271.

\bibitem{evans1994efficient}
J.~B. Evans, \enquote{Efficient fir filter architectures suitable for fpga implementation,} {\protect\JournalTitle{IEEE Transactions on Circuits and Systems II: Analog and Digital Signal Processing}} \textbf{41}, 490--493 (1994).

\bibitem{lee2003frequency}
W.~R. Lee, V.~Rehbock, K.~L. Teo, and L.~Caccetta, \enquote{Frequency-response masking based fir filter design with power-of-two coefficients and suboptimum pwr,} {\protect\JournalTitle{Journal of Circuits, Systems, and Computers}} \textbf{12}, 591--599 (2003).

\bibitem{kurup2012logic}
P.~Kurup and T.~Abbasi, \emph{Logic synthesis using Synopsys{\textregistered}} (Springer Science \& Business Media, 2012).

\bibitem{li2016efficient}
H.~Li and W.~Ye, \enquote{Efficient implementation of fpga based on vivado high level synthesis,} in \emph{2016 2nd IEEE International Conference on Computer and Communications (ICCC),}  (2016), pp. 2810--2813.

\bibitem{cui2019deep}
L.~Cui, Y.~Zhao, B.~Yan, D.~Liu, and J.~Zhang, \enquote{Deep-learning-based failure prediction with data augmentation in optical transport networks,} in \emph{17th International Conference on Optical Communications and Networks (ICOCN 2018),}  vol. 11048 (International Society for Optics and Photonics, 2019), p. 110482I.

\bibitem{zhuang2020machine}
H.~Zhuang, Y.~Zhao, X.~Yu, Y.~Li, Y.~Wang, and J.~Zhang, \enquote{Machine-learning-based alarm prediction with gans-based self-optimizing data augmentation in large-scale optical transport networks,} in \emph{2020 International Conference on Computing, Networking and Communications (ICNC),}  (IEEE, 2020), pp. 294--298.

\bibitem{li2019adaptive}
S.~Li, J.~Li, M.~Zhang, D.~Wang, C.~Song, and X.~Zhen, \enquote{Adaptive traffic data augmentation using generative adversarial networks for optical networks,} in \emph{2019 Optical Fiber Communications Conference and Exhibition (OFC),}  (IEEE, 2019), pp. 1--3.

\bibitem{9333417}
V.~Neskorniuk, P.~J. Freire, A.~Napoli, B.~Spinnler, W.~Schairer, J.~E. Prilepsky, N.~Costa, and S.~K. Turitsyn, \enquote{Simplifying the supervised learning of kerr nonlinearity compensation algorithms by data augmentation,} in \emph{2020 European Conference on Optical Communications (ECOC),}  (2020), pp. 1--4.

\bibitem{mo2018ann}
W.~Mo, Y.-K. Huang, S.~Zhang, E.~Ip, D.~C. Kilper, Y.~Aono, and T.~Tajima, \enquote{Ann-based transfer learning for qot prediction in real-time mixed line-rate systems,} in \emph{2018 Optical Fiber Communications Conference and Exposition (OFC),}  (IEEE, 2018), pp. 1--3.

\bibitem{cheng2020transfer}
Y.~Cheng, W.~Zhang, S.~Fu, M.~Tang, and D.~Liu, \enquote{Transfer learning simplified multi-task deep neural network for pdm-64qam optical performance monitoring,} {\protect\JournalTitle{Optics Express}} \textbf{28}, 7607--7617 (2020).

\bibitem{yao2019transductive}
Q.~Yao, H.~Yang, A.~Yu, and J.~Zhang, \enquote{Transductive transfer learning-based spectrum optimization for resource reservation in seven-core elastic optical networks,} {\protect\JournalTitle{Journal of Lightwave Technology}} \textbf{37}, 4164--4172 (2019).

\bibitem{xu2020feedforward}
Z.~Xu, C.~Sun, T.~Ji, J.~H. Manton, and W.~Shieh, \enquote{Feedforward and recurrent neural network-based transfer learning for nonlinear equalization in short-reach optical links,} {\protect\JournalTitle{Journal of Lightwave Technology}} \textbf{39}, 475--480 (2020).

\bibitem{zhang2019fast}
J.~Zhang, L.~Xia, M.~Zhu, S.~Hu, B.~Xu, and K.~Qiu, \enquote{Fast remodeling for nonlinear distortion mitigation based on transfer learning,} {\protect\JournalTitle{Optics letters}} \textbf{44}, 4243--4246 (2019).

\bibitem{zhang2020nonlinear}
W.~Zhang, T.~Jin, T.~Xu, J.~Zhang, and K.~Qiu, \enquote{Nonlinear mitigation with tl-nn-nlc in coherent optical fiber communications,} in \emph{Asia Communications and Photonics Conference,}  (Optical Society of America, 2020), pp. M4A--321.

\bibitem{9523752}
P.~J. Freire, D.~Abode, J.~E. Prilepsky, N.~Costa, B.~Spinnler, A.~Napoli, and S.~K. Turitsyn, \enquote{Transfer learning for neural networks-based equalizers in coherent optical systems,} {\protect\JournalTitle{Journal of Lightwave Technology}} \textbf{39}, 6733--6745 (2021).

\bibitem{tobin2017domain}
J.~Tobin, R.~Fong, A.~Ray, J.~Schneider, W.~Zaremba, and P.~Abbeel, \enquote{Domain randomization for transferring deep neural networks from simulation to the real world,} in \emph{2017 IEEE/RSJ international conference on intelligent robots and systems (IROS),}  (IEEE, 2017), pp. 23--30.

\bibitem{chen2021understanding}
X.~Chen, J.~Hu, C.~Jin, L.~Li, and L.~Wang, \enquote{Understanding domain randomization for sim-to-real transfer,} {\protect\JournalTitle{arXiv preprint arXiv:2110.03239}}  (2021).

\bibitem{muratore2022robot}
F.~Muratore, F.~Ramos, G.~Turk, W.~Yu, M.~Gienger, and J.~Peters, \enquote{Robot learning from randomized simulations: A review,} {\protect\JournalTitle{Frontiers in Robotics and AI}} \textbf{9} (2022).

\bibitem{freire2022domain}
P.~J. Freire, B.~Spinnler, D.~Abode, J.~E. Prilepsky, A.~Ali, N.~Costa, W.~Schairer, A.~Napoli, A.~D. Ellis, and S.~K. Turitsyn, \enquote{Domain adaptation: the key enabler of neural network equalizers in coherent optical systems,} in \emph{2022 Optical Fiber Communications Conference and Exhibition (OFC),}  (IEEE, 2022), pp. 1--3.

\bibitem{simeone2020learning}
O.~Simeone, S.~Park, and J.~Kang, \enquote{From learning to meta-learning: Reduced training overhead and complexity for communication systems,} in \emph{2020 2nd 6G Wireless Summit (6G SUMMIT),}  (IEEE, 2020), pp. 1--5.

\bibitem{ouali2020overview}
Y.~Ouali, C.~Hudelot, and M.~Tami, \enquote{An overview of deep semi-supervised learning,} {\protect\JournalTitle{arXiv preprint arXiv:2006.05278}}  (2020).

\bibitem{blalock2020state}
D.~Blalock, J.~J.~G. Ortiz, J.~Frankle, and J.~Guttag, \enquote{What is the state of neural network pruning?} {\protect\JournalTitle{arXiv preprint arXiv:2003.03033}}  (2020).

\bibitem{liang2021pruning}
T.~Liang, J.~Glossner, L.~Wang, S.~Shi, and X.~Zhang, \enquote{Pruning and quantization for deep neural network acceleration: A survey,} {\protect\JournalTitle{Neurocomputing}} \textbf{461}, 370--403 (2021).

\bibitem{liu2018rethinking}
Z.~Liu, M.~Sun, T.~Zhou, G.~Huang, and T.~Darrell, \enquote{Rethinking the value of network pruning,} {\protect\JournalTitle{arXiv preprint arXiv:1810.05270}}  (2018).

\bibitem{augasta2013pruning}
M.~Augasta and T.~Kathirvalavakumar, \enquote{Pruning algorithms of neural networks—a comparative study,} {\protect\JournalTitle{Open Computer Science}} \textbf{3}, 105--115 (2013).

\bibitem{vadera2020methods}
S.~Vadera and S.~Ameen, \enquote{Methods for pruning deep neural networks,} {\protect\JournalTitle{arXiv preprint arXiv:2011.00241}}  (2020).

\bibitem{han2015deep}
S.~Han, H.~Mao, and W.~J. Dally, \enquote{Deep compression: Compressing deep neural networks with pruning, trained quantization and huffman coding,} {\protect\JournalTitle{arXiv preprint arXiv:1510.00149}}  (2015).

\bibitem{chuang2019sparse}
C.-Y. Chuang, W.-F. Chang, C.-C. Wei, C.-J. Ho, C.-Y. Huang, J.-W. Shi, L.~Henrickson, Y.-K. Chen, and J.~Chen, \enquote{Sparse volterra nonlinear equalizer by employing pruning algorithm for high-speed pam-4 850-nm vcsel optical interconnect,} in \emph{Optical Fiber Communication Conference,}  (Optical Society of America, 2019), pp. M1F--2.

\bibitem{huang201893}
W.-J. Huang, W.-F. Chang, C.-C. Wei, J.-J. Liu, Y.-C. Chen, K.-L. Chi, C.-L. Wang, J.-W. Shi, and J.~Chen, \enquote{93\% complexity reduction of volterra nonlinear equalizer by l1-regularization for 112-gbps pam-4 850-nm vcsel optical interconnect,} in \emph{2018 Optical Fiber Communications Conference and Exposition (OFC),}  (IEEE, 2018), pp. 1--3.

\bibitem{6647643}
F.~P. Guiomar, S.~B. Amado, N.~J. Muga, J.~D. Reis, A.~L. Teixeira, and A.~N. Pinto, \enquote{Simplified volterra series nonlinear equalizer by intra-channel cross-phase modulation oriented pruning,} in \emph{39th European Conference and Exhibition on Optical Communication (ECOC 2013),}  (2013), pp. 1--3.

\bibitem{kumar2021deep}
O.~S. Kumar, L.~Lampe, S.~Luo, M.~Jana, J.~Mitra, and C.~Li, \enquote{Deep neural network assisted second-order perturbation-based nonlinearity compensation,} in \emph{Signal Processing in Photonic Communications,}  (Optical Society of America, 2021), pp. SpF2E--2.

\bibitem{li2021high}
M.~Li, W.~Zhang, Q.~Chen, and Z.~He, \enquote{High-throughput hardware deployment of pruned neural network based nonlinear equalization for 100-gbps short-reach optical interconnect,} {\protect\JournalTitle{Optics Letters}} \textbf{46}, 4980--4983 (2021).

\bibitem{wan2018nonlinear}
Z.~Wan, J.~Li, L.~Shu, M.~Luo, X.~Li, S.~Fu, and K.~Xu, \enquote{Nonlinear equalization based on pruned artificial neural networks for 112-gb/s ssb-pam4 transmission over 80-km ssmf,} {\protect\JournalTitle{Optics Express}} \textbf{26}, 10631--10642 (2018).

\bibitem{zhang2020compressed}
W.~Zhang, L.~Ge, Y.~Zhang, C.~Liang, and Z.~He, \enquote{Compressed nonlinear equalizers for 112-gbps optical interconnects: Efficiency and stability,} {\protect\JournalTitle{Sensors}} \textbf{20}, 4680 (2020).

\bibitem{wang2021low}
L.~Wang, X.~Zeng, J.~Wang, D.~Gao, and M.~Bai, \enquote{Low-complexity nonlinear equalizer based on artificial neural network for 112 gbit/s pam-4 transmission using dml,} {\protect\JournalTitle{Optical Fiber Technology}} \textbf{67}, 102724 (2021).

\bibitem{ge2020compressed}
L.~Ge, W.~Zhang, C.~Liang, and Z.~He, \enquote{Compressed neural network equalization based on iterative pruning algorithm for 112-gbps vcsel-enabled optical interconnects,} {\protect\JournalTitle{Journal of Lightwave Technology}} \textbf{38}, 1323--1329 (2020).

\bibitem{reza2018nonlinear}
A.~G. Reza and J.-K.~K. Rhee, \enquote{Nonlinear equalizer based on neural networks for pam-4 signal transmission using dml,} {\protect\JournalTitle{IEEE Photonics Technology Letters}} \textbf{30}, 1416--1419 (2018).

\bibitem{wang2020compressing}
L.-N. Wang, W.~Liu, X.~Liu, G.~Zhong, P.~P. Roy, J.~Dong, and K.~Huang, \enquote{Compressing deep networks by neuron agglomerative clustering,} {\protect\JournalTitle{Sensors}} \textbf{20}, 6033 (2020).

\bibitem{son2018clustering}
S.~Son, S.~Nah, and K.~M. Lee, \enquote{Clustering convolutional kernels to compress deep neural networks,} in \emph{Proceedings of the European Conference on Computer Vision (ECCV),}  (2018), pp. 216--232.

\bibitem{gholami2021survey}
A.~Gholami, S.~Kim, Z.~Dong, Z.~Yao, M.~W. Mahoney, and K.~Keutzer, \enquote{A survey of quantization methods for efficient neural network inference,} {\protect\JournalTitle{arXiv preprint arXiv:2103.13630}}  (2021).

\bibitem{cheng2017survey}
Y.~Cheng, D.~Wang, P.~Zhou, and T.~Zhang, \enquote{A survey of model compression and acceleration for deep neural networks,} {\protect\JournalTitle{arXiv preprint arXiv:1710.09282}}  (2017).

\bibitem{weng2021neural}
O.~Weng, \enquote{Neural network quantization for efficient inference: A survey,} {\protect\JournalTitle{arXiv preprint arXiv:2112.06126}}  (2021).

\bibitem{bai2021towards}
H.~Bai, L.~Hou, L.~Shang, X.~Jiang, I.~King, and M.~R. Lyu, \enquote{Towards efficient post-training quantization of pre-trained language models,} {\protect\JournalTitle{arXiv preprint arXiv:2109.15082}}  (2021).

\bibitem{alvarez2016efficient}
R.~Alvarez, R.~Prabhavalkar, and A.~Bakhtin, \enquote{On the efficient representation and execution of deep acoustic models,} {\protect\JournalTitle{arXiv preprint arXiv:1607.04683}}  (2016).

\bibitem{duarte2018fast}
J.~Duarte, S.~Han, P.~Harris, S.~Jindariani, E.~Kreinar, B.~Kreis, J.~Ngadiuba, M.~Pierini, R.~Rivera, N.~Tran \emph{et~al.}, \enquote{Fast inference of deep neural networks in fpgas for particle physics,} {\protect\JournalTitle{Journal of Instrumentation}} \textbf{13}, P07027 (2018).

\bibitem{coelho2021automatic}
C.~N. Coelho, A.~Kuusela, S.~Li, H.~Zhuang, J.~Ngadiuba, T.~K. Aarrestad, V.~Loncar, M.~Pierini, A.~A. Pol, and S.~Summers, \enquote{Automatic heterogeneous quantization of deep neural networks for low-latency inference on the edge for particle detectors,} {\protect\JournalTitle{Nature Machine Intelligence}} pp. 1--12 (2021).

\bibitem{kaneda2020fpga}
N.~Kaneda, Z.~Zhu, C.-Y. Chuang, A.~Mahadevan, B.~Farah, K.~Bergman, D.~Van~Veen, and V.~Houtsma, \enquote{Fpga implementation of deep neural network based equalizers for high-speed pon,} in \emph{Optical Fiber Communication Conference,}  (Optical Society of America, 2020), pp. T4D--2.

\bibitem{aoudia2019towards}
F.~A. Aoudia and J.~Hoydis, \enquote{Towards hardware implementation of neural network-based communication algorithms,} in \emph{2019 IEEE 20th International Workshop on Signal Processing Advances in Wireless Communications (SPAWC),}  (IEEE, 2019), pp. 1--5.

\bibitem{xu2019efficient}
W.~Xu, X.~Tan, Y.~Lin, X.~You, C.~Zhang, and Y.~Be’ery, \enquote{On the efficient design of neural networks in communication systems,} in \emph{2019 53rd Asilomar Conference on Signals, Systems, and Computers,}  (IEEE, 2019), pp. 522--526.

\bibitem{ron2022experimental}
D.~A. Ron, P.~J. Freire, J.~E. Prilepsky, M.~Kamalian-Kopae, A.~Napoli, and S.~K. Turitsyn, \enquote{Experimental implementation of a neural network optical channel equalizer in restricted hardware using pruning and quantization,} {\protect\JournalTitle{Scientific Reports}} \textbf{12}, 8713 (2022).

\bibitem{hinton2015distilling}
G.~Hinton, O.~Vinyals, J.~Dean \emph{et~al.}, \enquote{Distilling the knowledge in a neural network,} {\protect\JournalTitle{arXiv preprint arXiv:1503.02531}} \textbf{2} (2015).

\bibitem{gou2021knowledge}
J.~Gou, B.~Yu, S.~J. Maybank, and D.~Tao, \enquote{Knowledge distillation: A survey,} {\protect\JournalTitle{International Journal of Computer Vision}} \textbf{129}, 1789--1819 (2021).

\bibitem{chang2017hardware}
A.~X.~M. Chang and E.~Culurciello, \enquote{Hardware accelerators for recurrent neural networks on fpga,} in \emph{2017 IEEE International symposium on circuits and systems (ISCAS),}  (IEEE, 2017), pp. 1--4.

\bibitem{willi2019recurrent}
T.~Willi, J.~Masci, J.~Schmidhuber, and C.~Osendorfer, \enquote{Recurrent neural processes,} {\protect\JournalTitle{arXiv preprint arXiv:1906.05915}}  (2019).

\bibitem{9319148}
F.~Libano, P.~Rech, B.~Neuman, J.~Leavitt, M.~Wirthlin, and J.~Brunhaver, \enquote{How reduced data precision and degree of parallelism impact the reliability of convolutional neural networks on fpgas,} {\protect\JournalTitle{IEEE Transactions on Nuclear Science}} \textbf{68}, 865--872 (2021).

\bibitem{9079640}
C.~Wang, L.~Gong, X.~Li, and X.~Zhou, \enquote{A ubiquitous machine learning accelerator with automatic parallelization on fpga,} {\protect\JournalTitle{IEEE Transactions on Parallel and Distributed Systems}} \textbf{31}, 2346--2359 (2020).

\bibitem{7927161}
G.~Zhong, A.~Prakash, S.~Wang, Y.~Liang, T.~Mitra, and S.~Niar, \enquote{Design space exploration of fpga-based accelerators with multi-level parallelism,} in \emph{Design, Automation \& Test in Europe Conference \& Exhibition (DATE), 2017,}  (2017), pp. 1141--1146.

\bibitem{luo2020towards}
C.~Luo, M.-K. Sit, H.~Fan, S.~Liu, W.~Luk, and C.~Guo, \enquote{Towards efficient deep neural network training by fpga-based batch-level parallelism,} {\protect\JournalTitle{Journal of Semiconductors}} \textbf{41}, 022403 (2020).

\bibitem{li2015fpga}
S.~Li, C.~Wu, H.~Li, B.~Li, Y.~Wang, and Q.~Qiu, \enquote{Fpga acceleration of recurrent neural network based language model,} in \emph{2015 IEEE 23rd Annual International Symposium on Field-Programmable Custom Computing Machines,}  (IEEE, 2015), pp. 111--118.

\bibitem{danopoulos2022lstm}
D.~Danopoulos, I.~Stamoulias, G.~Lentaris, D.~Masouros, I.~Kanaropoulos, A.~K. Kakolyris, and D.~Soudris, \enquote{Lstm acceleration with fpga and gpu devices for edge computing applications in b5g mec,} in \emph{Embedded Computer Systems: Architectures, Modeling, and Simulation: 22nd International Conference, SAMOS 2022, Samos, Greece, July 3--7, 2022, Proceedings,}  (Springer, 2022), pp. 406--419.

\bibitem{du2020model}
J.~Du, X.~Zhu, M.~Shen, Y.~Du, Y.~Lu, N.~Xiao, and X.~Liao, \enquote{Model parallelism optimization for distributed inference via decoupled cnn structure,} {\protect\JournalTitle{IEEE Transactions on Parallel and Distributed Systems}} \textbf{32}, 1665--1676 (2020).

\bibitem{wang2023model}
J.~Wang, W.~Tong, and X.~Zhi, \enquote{Model parallelism optimization for cnn fpga accelerator,} {\protect\JournalTitle{Algorithms}} \textbf{16}, 110 (2023).

\end{thebibliography}






\end{document}